\documentclass[10pt]{article} 
\usepackage[preprint]{tmlr}

\usepackage{hyperref}
\usepackage{breakurl}
\usepackage{graphicx}
\usepackage{amsmath}
\usepackage{amsfonts}
\usepackage{amsthm}
\usepackage{amssymb}
\usepackage{breqn}
\usepackage{ascmac}
\usepackage{subcaption}
\usepackage{wrapfig}
\usepackage{authblk}
\usepackage{algorithmicx}
\usepackage{algorithm}
\usepackage[noend]{algpseudocode}
\definecolor{uscarlet}{rgb}{1.0, 0.13, 0.0}
\definecolor{uviolet}{rgb}{0.56, 0.0, 1.0}

\title{Efficient Sparse-Reward Goal-Conditioned Reinforcement Learning with a High Replay Ratio and Regularization}

\author{\name Takuya Hiraoka \email takuya-h1@nec.com \\
      \addr NEC Corporation\\
      \addr National Institute of Advanced Industrial Science and Technology
      }

\def\month{MM}  
\def\year{YYYY} 
\def\openreview{\url{https://openreview.net/forum?id=XXXX}} 

\begin{document}

\maketitle

\begin{abstract}
Reinforcement learning (RL) methods with a high replay ratio (RR) and regularization have gained interest due to their superior sample efficiency. 
However, these methods have mainly been developed for dense-reward tasks. 
In this paper, we aim to extend these RL methods to sparse-reward goal-conditioned tasks. 
We use Randomized Ensemble Double Q-learning (REDQ)~\citep{chen2021randomized}, an RL method with a high RR and regularization. 
To apply REDQ to sparse-reward goal-conditioned tasks, we make the following modifications to it: 
(i) using hindsight experience replay and (ii) bounding target Q-values. 
We evaluate REDQ with these modifications on 12 sparse-reward goal-conditioned tasks of Robotics~\citep{plappert2018multi}, and show that it achieves about $2 \times$ better sample efficiency than previous state-of-the-art (SoTA) RL methods. 
Furthermore, we reconsider the necessity of specific components of REDQ and simplify it by removing unnecessary ones. 
The simplified REDQ with our modifications achieves $\sim 8 \times$ better sample efficiency than the SoTA methods in 4 Fetch tasks of Robotics. 
\end{abstract}

\section{Introduction}\label{sec:introduction}
In the reinforcement learning (RL) community, improving the sample efficiency of RL methods has been important. 
Traditional RL methods have been promising for solving complex control tasks, including dexterous in-hand manipulation~\citep{andrychowicz2020learning}, quadrupedal/bipedal locomotion~\citep{lee2020learning,haarnoja2023learning}, and car/drone racing~\citep{wurman2022outracing,kaufmann2023champion}. 
However, traditional RL methods are generally data-hungry and require large amounts of training samples to solve tasks~\citep{mendonca2019guided}. 
Motivated by this problem, various sample-efficient RL methods have been proposed~\citep{haarnoja2018soft,lillicrap2015continuous,schulman2017proximal,fujimoto2018addressing}. 

In recent years, RL methods using a high replay ratio (RR) and regularization have attracted attention as sample-efficient methods~\citep{janner2019trust,chen2021randomized,hiraoka2022dropout,pmlr-v162-nikishin22a,li2022efficient,d'oro2023sampleefficient,smith2023learning,sokar2023dormant,pmlr-v202-schwarzer23a}. 
RR is the ratio of components (e.g., policy and Q-functions) updates to the actual interactions with an environment. 
A high RR facilitates sufficient training of the components within a few interactions but exacerbates the components' overfitting. 
Regularization techniques (e.g., ensemble~\citep{chen2021randomized} or dropout~\citep{hiraoka2022dropout}) are employed to prevent the overfitting. 
The RL methods equipped with them have exhibited high sample efficiency and enabled training agents within mere tens of minutes in real-world tasks, such as quadrupedal robot locomotion~\citep{smith2022walk,smith2023grow} and image-based vehicle driving~\citep{stachowicz2023fastrlap}. 

However, these methods have been developed mainly on dense-reward tasks rather than sparse-reward tasks. 
Many RL tasks require RL methods to learn with a sparse reward due to the difficulty of designing dense rewards~\citep{andrychowicz2017hindsight,trott19keeping,pmlr-v164-agrawal22a,knox2023reward,booth2023perils}. 
%
A typical example of such tasks is \textbf{sparse-reward goal-conditioned tasks}~\citep{plappert2018multi}, where a positive reward is provided only upon successful goal attainment. 
RL methods that can efficiently learn in these tasks hold substantial value in numerous application scenarios, such as (i) developing versatile agents capable of achieving diverse goals~\citep{vithayathil2020survey,beck2023survey}, or (ii) constructing low-level agents to execute goals provided by high-level agents in a hierarchical framework~\citep{pateria2021hierarchical,brohan2023rt,yu2023language}. 
Therefore, it is valuable to investigate whether RL methods with a high RR and regularization efficiently work in sparse-reward goal-conditioned tasks and what additional modifications are needed if the methods do not work efficiently. 

\begin{figure}[t!]
\begin{minipage}{1.0\hsize}
\centering
\includegraphics[clip, width=0.41\hsize]{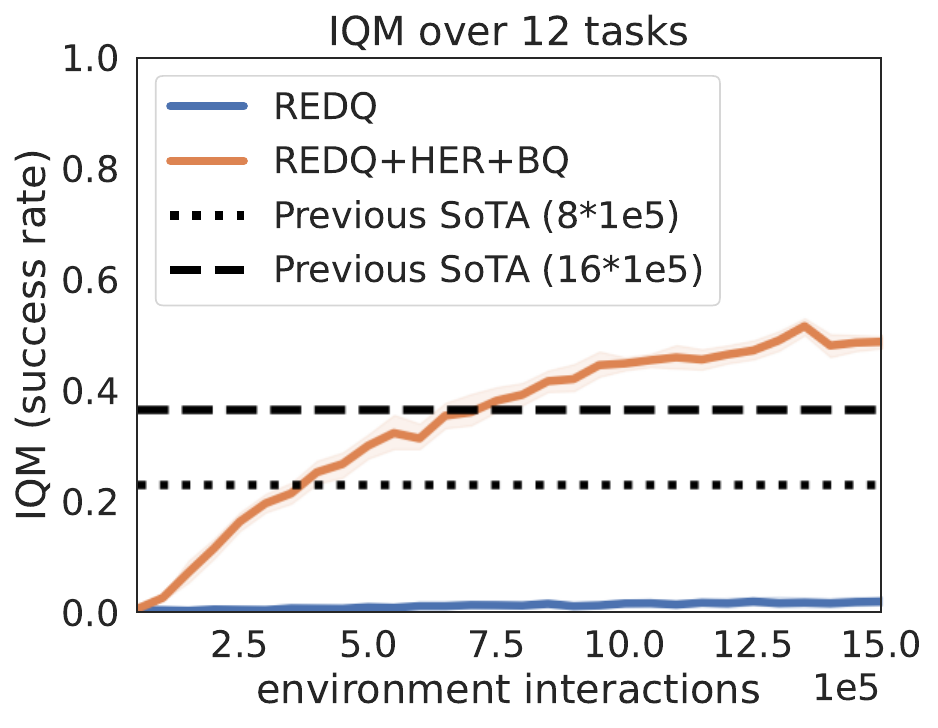}
\includegraphics[clip, width=0.41\hsize]{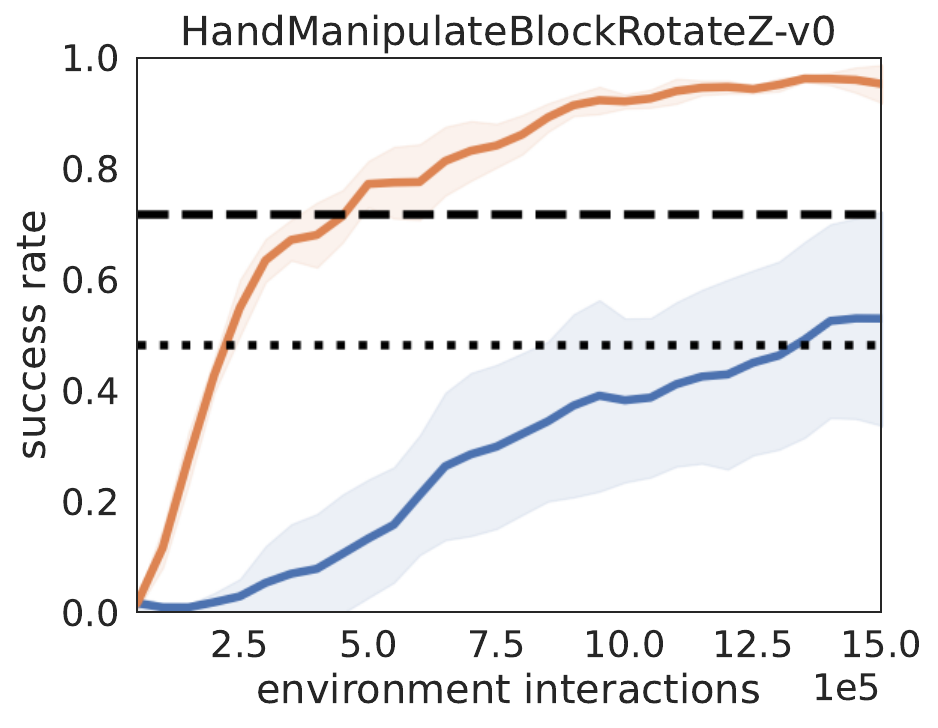}
\end{minipage}
\vspace{-0.7\baselineskip}
\caption{
The task success rate of vanilla REDQ and our modified REDQ (REDQ+HER+BQ). 
The left-hand side figure shows the interquartile mean (IQM) with a 95\% confidence interval~\citep{agarwal2021deep} for the success rate over $12$ Robotics tasks. 
The right-hand side figure shows the average scores with one standard deviation in the HandManipulateBlockRotateZ task (one of the Robotics tasks). 
We also present scores from previous SoTA methods with $8 \cdot 10^5$ and $16 \cdot 10^5$ samples (number of environment interactions). 
For context, $10^5$ samples correspond to approximately one hour of real-world experience.
The left-hand side figure shows that our modified REDQ achieves approximately $2 \times$ better sample efficiency than previous SoTA methods. 
Examples of policies learnt by our modified REDQ can be found at 
\textcolor{blue}{\url{https://drive.google.com/file/d/1UHd7JVPCwFLNFhy1QcycQfwU_nll_yII/view?usp=drive_link}}
}
\label{fig:intro-remarks}
\vspace{-0.5\baselineskip}
\end{figure}

In this paper, we apply an RL method with a high RR and regularization to sparse-reward goal-conditioned tasks. 
As our sparse-reward goal-conditioned tasks, we consider Robotics~\citep{plappert2018multi} (Section~\ref{sec:prelim}). 
As an RL method with a high RR and regularization, we employ Randomized Ensemble Double Q-learning (REDQ)~\citep{chen2021randomized} (Section~\ref{sec:base_method}). 
To adapt REDQ for the Robotics tasks, we introduce the following modifications to REDQ: 
(i) using hindsight experience replay (HER; Section~\ref{sec:her}) and (ii) bounding target Q-value (BQ; Section~\ref{sec:boundq}). 
We experimentally demonstrate that REDQ with these modifications can achieve better sample efficiency than previous state-of-the-art (SoTA) RL methods (Fig.~\ref{fig:intro-remarks}. 
See Sections~\ref{sec:experiment} and \ref{sec:simplification} for more comprehensive details). 

While our main contribution is \textbf{successful application of the RL method with a high RR and regularization to sparse-reward goal-conditioned tasks}, we make two
additional significant contributions:\\ 
\textbf{1. Illustration of the importance of stabilizing Q-value estimation in sparse-reward tasks:} 
We show that (i) the introduction of HER into REDQ makes the Q-value estimation unstable and that (ii) BQ significantly suppresses the instability (Fig.~\ref{fig:method-boundq-qvals} in Section \ref{sec:boundq}) and improves overall sample efficiency (Fig.~\ref{fig:experiment-return-sr-summary} in Section~\ref{sec:experiment}). 
\textit{Surprisingly, REDQ already uses sophisticated regularization techniques to stabilize the Q-value estimation (Section~\ref{sec:base_method}), but it is made unstable by the additional component (HER) and requires treatment (Sections~\ref{sec:boundq}).} 
Exploration ability has been considered to be the overwhelmingly important component in sparse-reward tasks (e.g., \citet{pathak2017curiosity,tang2017exploration,ecoffet2019go}). 
However, our results indicate that the stability of Q-value estimation is also non-negligibly important and suggest practitioners should pay attention to it when applying an RL method to similar tasks.\\
\textbf{2. Simplification of REDQ in sparse-reward goal-conditioned tasks:} 
REDQ uses clipped double Q-learning and an entropy term in its target Q-value calculation. 
We find that REDQ can be simplified by removing them (Figs.~\ref{fig:simplication-qvals} and \ref{fig:simplification-return-sr-summary} in Section~\ref{sec:simplification}). 
Remarkably, the simplified REDQ with our modifications achieves $\sim 8 \times$ better sample efficiency than SoTA methods in the Fetch tasks of Robotics (Fig.~\ref{fig:simplification-return-sr-summary}). 
Our findings may be valuable in maintaining the simplicity of REDQ, which improves reproducibility and reduces human effort in debugging and engineering.

\section{Preliminary: Sparse-Reward Goal-Conditioned RL}\label{sec:prelim}
We focus on sparse-reward goal-conditioned RL. 
This is typically modeled as goal-augmented Markov decision processes $\langle \mathcal{S}, \mathcal{A}, \mathcal{G}, \gamma, p_{s_0}, p_g,  \mathcal{T}, \mathcal{R} \rangle$~\citep{liu2022goal}. 
Here, $\mathcal{S}$, $\mathcal{A}$, $\mathcal{G}$, and $\gamma$ are the state space, the action space, the goal space, and the discount factor, respectively. 
$p_{s_0}: \mathcal{S} \rightarrow [0, 1]$ is the initial state distribution. 
$p_g: \mathcal{G} \rightarrow [0, 1]$ is the goal distribution. 
$\mathcal{T}: \mathcal{S} \times \mathcal{A} \times \mathcal{S} \rightarrow [0, 1]$ is the dynamics transition function. 
$\mathcal{R}: \mathcal{S} \times \mathcal{A} \times \mathcal{G} \rightarrow \mathbb{R}$ is the reward function, which is sparsely structured in our setting. 
At the beginning of an episode, an agent receives the desired goal $g \sim p_g(\cdot)$. 
At each discrete time step $t$, an environment provides the agent with a state $s_t \in \mathcal{S}$, the agent responds by selecting an action $a_t \in \mathcal{A}$, and then the environment provides the next reward $r_t \leftarrow \mathcal{R}(s_t, a_t, g)$ and state $s_{t+1} \in \mathcal{S}$. 
For convenience, as needed, we use the simpler notations of $r$, $s$, $a$, $s'$, and $a'$ to refer to a reward, state, action, next state, and next action, respectively. 
In addition, as needed, we use the notation of $g_t' \in \mathcal{G}$ and $r_t' \leftarrow \mathcal{R}(s_t, a_t, g_t')$ to refer to the goal and reward at $t$, respectively. 
The objective of sparse-reward goal-conditioned RL is to learn a goal-conditioned policy $\pi: \mathcal{S} \times \mathcal{G} \times \mathcal{A} \rightarrow [0, 1]$ that maximizes the expected cumulative rewards: 
\begin{eqnarray}
    \mathbb{E}_{a_t, g, s_{t+1}, s_0} \left[ \sum_{t=0}^{\infty} \gamma^{t} r_t \right], ~~~ a_t \sim \pi(\cdot | s_t, g), ~~~ g \sim p_g(\cdot), \nonumber
     s_{t+1} \sim \mathcal{T}(\cdot | s_t, a_t), ~~~ s_0 \sim p_{s_0}(\cdot). 
\end{eqnarray}

\begin{figure}[t!]
\begin{minipage}[t]{.48\hsize}
\centering
\includegraphics[clip, width=0.32\hsize]{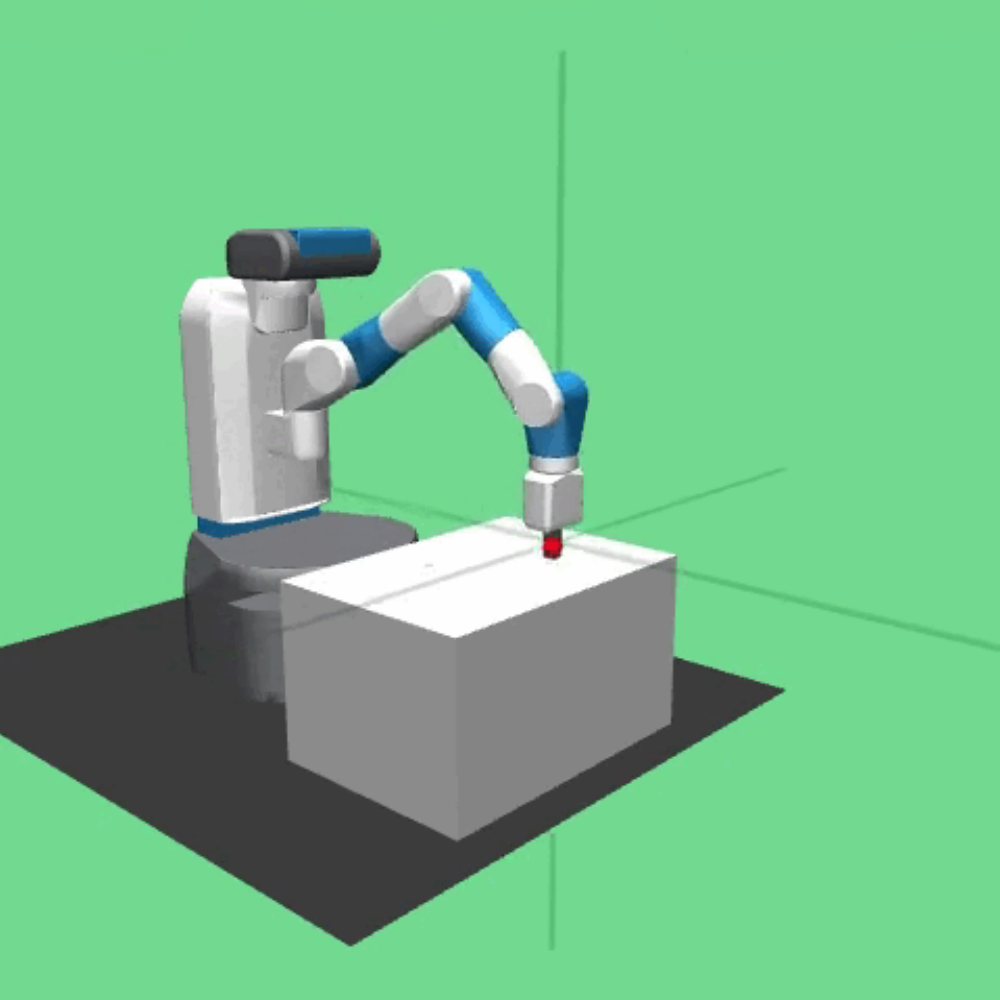}
\includegraphics[clip, width=0.32\hsize]{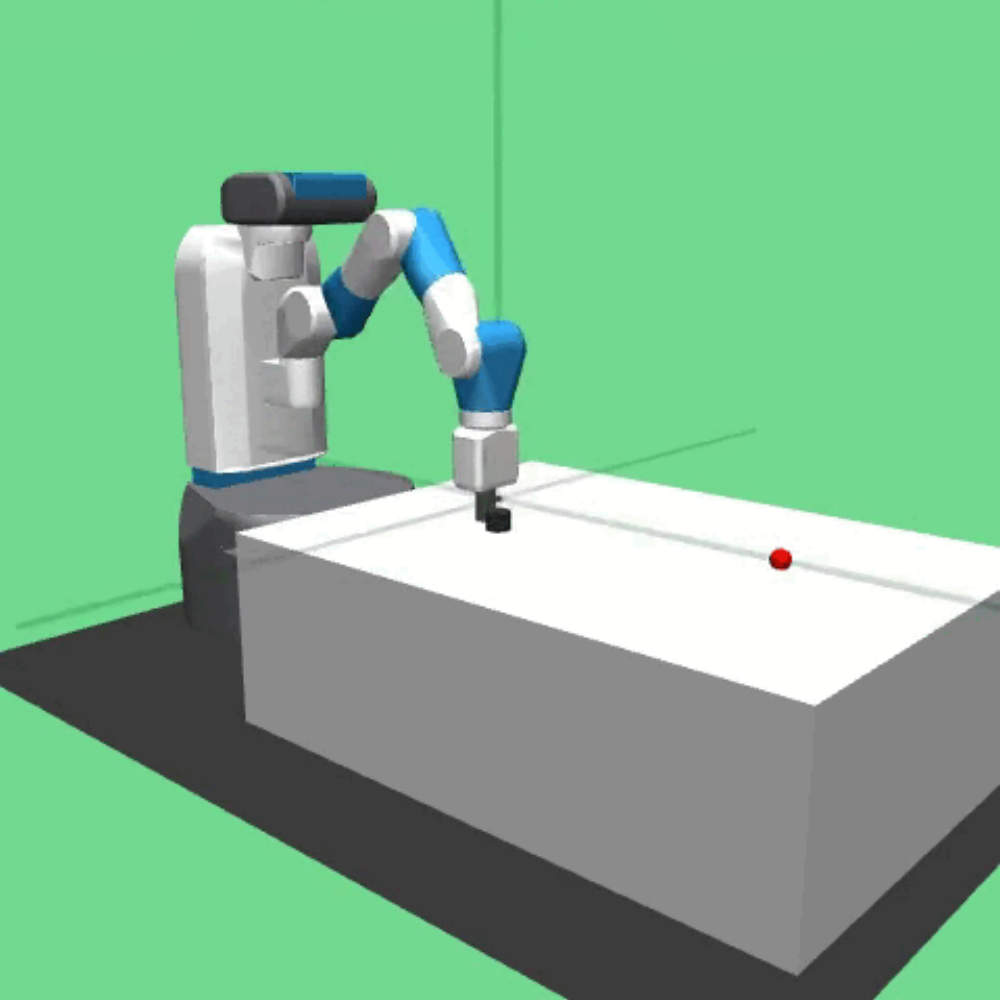}
\includegraphics[clip, width=0.32\hsize]{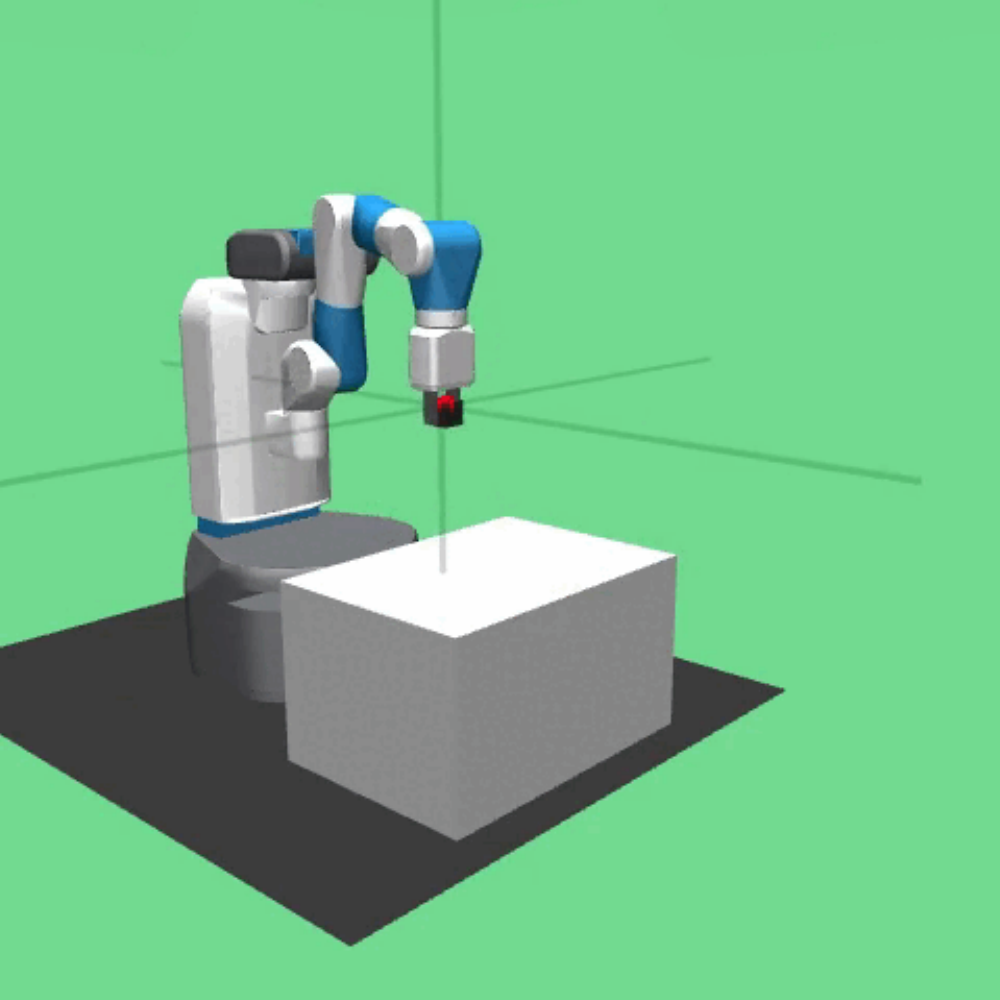}
\subcaption{Fetch tasks}
\end{minipage}\hspace{0.8\baselineskip}\begin{minipage}[t]{.48\hsize}
\centering
\includegraphics[clip, width=0.32\hsize]{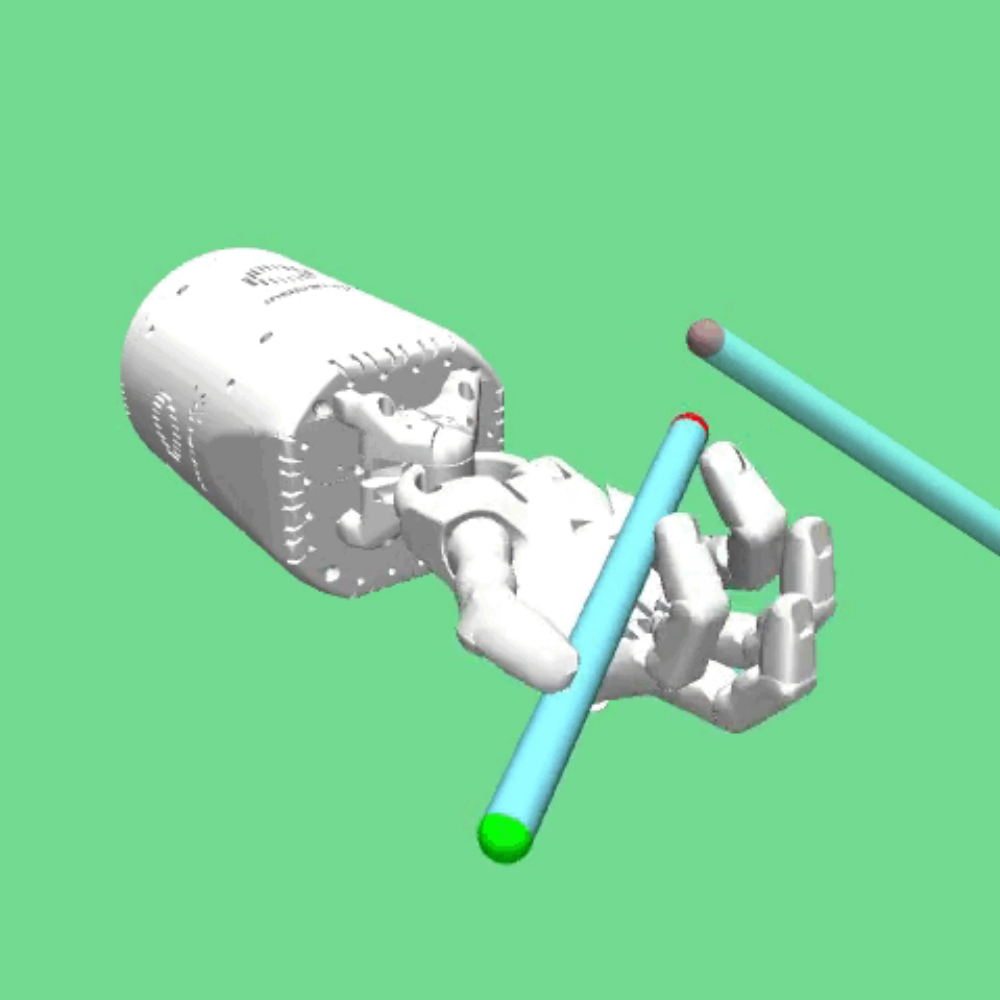}
\includegraphics[clip, width=0.32\hsize]{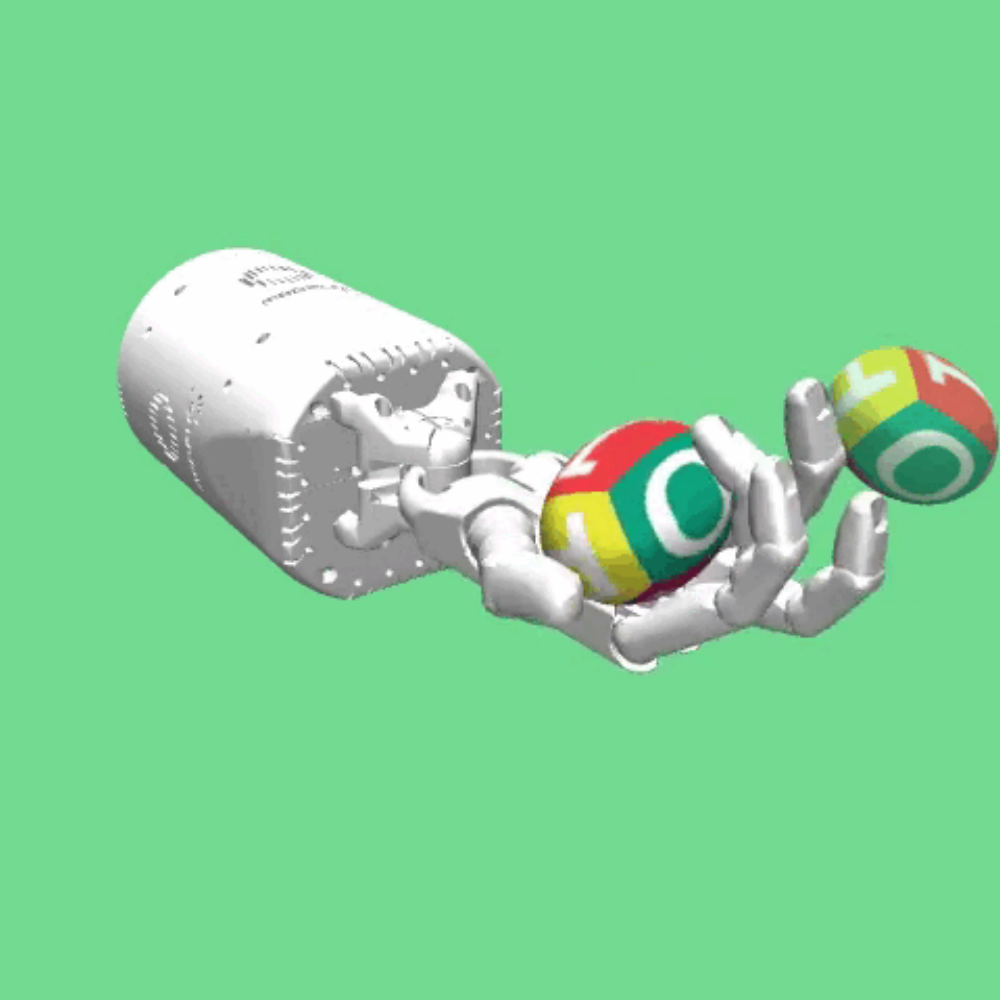}
\includegraphics[clip, width=0.32\hsize]{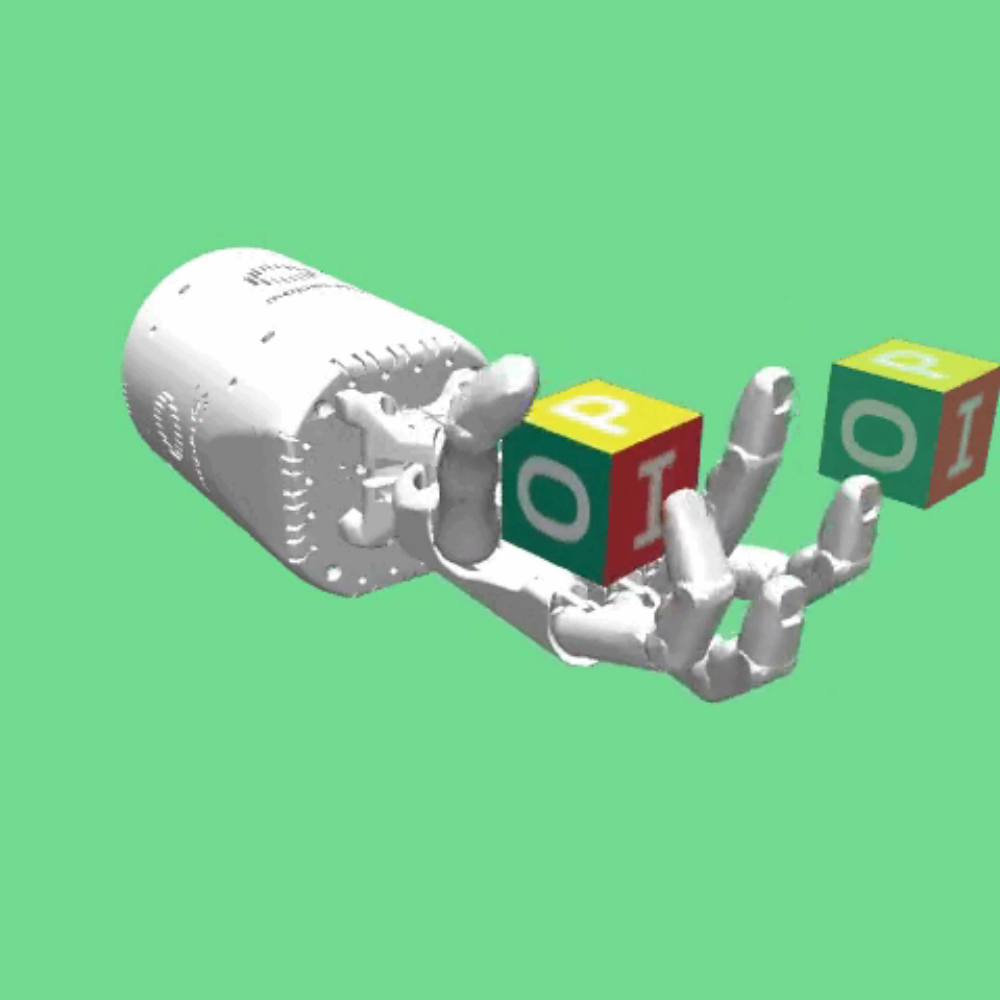}
\subcaption{HandManipulate tasks}
\end{minipage}
\vspace{-0.7\baselineskip}
\caption{Robotics~\citep{plappert2018multi} tasks.}
\label{fig:environments}
\vspace{-0.5\baselineskip}
\end{figure}

As benchmark tasks for sparse-reward goal-conditioned RL, we employ the Robotics~\citep{plappert2018multi,gymnasium_robotics2023github} tasks (Fig.~\ref{fig:environments}). 
In these tasks, an RL agent aims to learn control policies for moving and reorienting objects (e.g., a block or an egg) to target positions and orientations. 
The reward is sparsely structured: 
The agent receives a positive reward of $0$ if the distance between the positions (and orientations) of the object and the target is within a small threshold, and a negative reward of $-1$ otherwise~\footnote{A more detailed task description can be found at \url{https://robotics.farama.org/}.}. 
We use 12 Robotics tasks for our experiments: 
FetchReach, 
FetchPush, 
FetchSlide, 
FetchPickAndPlace, 
HandManipulatePenRotate, 
HandManipulateEggRotate, 
HandManipulatePenFull, 
HandManipulateEggFull, 
HandManipulateBlockFull, 
HandManipulateBlockRotateZ, 
HandManipulateBlockRotateXYZ, 
and HandManipulateBlockRotateParallel. 

\section{Our RL Method}\label{sec:method}
In this section, we introduce our method for sparse-reward goal-conditioned RL. 
The algorithmic description of our method is summarized in Algorithm~\ref{alg1:REDQwithOurDesigneDecision}. 
We use REDQ for our base method (Section~\ref{sec:base_method}). 
To apply REDQ to sparse-reward goal-conditioned tasks, we make two modifications to REDQ: 
(i) using hindsight experience replay (HER; Section~\ref{sec:her}) 
and (ii) bounding target Q-values (BQ; Section~\ref{sec:boundq}). 
\begin{algorithm*}[t]
\caption{REDQ with our modifications (\textcolor{uscarlet}{HER} and \textcolor{uviolet}{BQ})}
\label{alg1:REDQwithOurDesigneDecision}
\begin{algorithmic}[1]
\Statex Initialize policy parameters $\theta$, $N$ Q-function parameters $\phi_i$, empty replay buffer $\mathcal{D}$, and episode length $T$. Set target parameters $\bar{\phi}_i \leftarrow \phi_i$, for $i = 1, ...., N$.
\State Sample goal $g \sim p_g(\cdot)$ and initial state $s_0 \sim p_{s_0}(\cdot)$
\For{$t=0, .., T$}
\State Take action $a_t \sim \pi_\theta(\cdot | s_t)$; Observe reward $r_t$ and next state $s_{t+1}$.
\If{$t = T$}
    \State $\mathcal{D} \leftarrow \mathcal{D} \bigcup \{(s_t, a_t, r_t, s_{t+1}, g)\}_{t=0}^{T}$; \textcolor{uscarlet}{Select new goal $g_t'$; Calculate new reward $r_t' \leftarrow \mathcal{R}(s_t, a_t, g_t')$; $\mathcal{D} \leftarrow \mathcal{D} \bigcup \{(s_t, a_t, r_t', s_{t+1}, g_t')\}_{t=0}^{T}$}
\EndIf
\For{$G$ updates}
    \State Sample a mini-batch $\mathcal{B} = \{ (s, a, r, s', g) \}$ from $\mathcal{D}$. 
    \State Sample a set $\mathcal{M}$ of $M$ distinct indices from $\{1, 2, . . . , N\}$. 
    \State Compute the target Q-value $y$ (same for all $N$ Q-functions): 
    \begin{equation}
        y = r + \gamma \textcolor{uviolet}{\min \left( \max \left( \textcolor{black}{\min_{i \in \mathcal{M}} Q_{\bar{\phi}_i}(s', a', g) - \alpha \log \pi_\theta(a' | s', g)}, Q_\min \right), Q_\max \right)}, ~~ a' \sim \pi_\theta(\cdot | s', g) \nonumber
    \end{equation}
    \For{$i=1, ..., N$}
        \State Update $\phi_i$ with gradient descent using
        \begin{equation}
            \nabla_\phi \frac{1}{|B|} \sum_{(s, a, r, s', g) \in \mathcal{B}} \left( Q_{\phi_i}(s, a, g) - y \right)^2 \nonumber
        \end{equation}
    \State Update target networks with $\bar{\phi}_i \leftarrow \rho \bar{\phi}_i + (1-\rho) \phi_i $.
    \EndFor
\EndFor
\State Update $\theta$ with gradient ascent using 
\begin{equation}
    \nabla_\theta \frac{1}{|B|} \sum_{s \in \mathcal{B}} \left( \frac{1}{N} \sum_{i=1}^{N} Q_{\phi_i}(s, a, g) - \alpha \log \pi_\theta(a | s, g) \right), ~~~ a \sim \pi_\theta(\cdot | s, g) \nonumber
\end{equation}
\EndFor
\end{algorithmic}
\end{algorithm*}

\subsection{Base Method: RL Method with a High RR and Regularization}\label{sec:base_method}
Our base method is REDQ~\citep{chen2021randomized}, an RL method with a high RR and regularization:\\
\textbf{High RR.} 
REDQ uses a high RR $G$ (typically $G > 1$), which is the number of Q-function updates (lines 6--12 in Algorithm~\ref{alg1:REDQwithOurDesigneDecision}) relative to the number of actual interactions with the environment (line 3). 
A high RR promotes sufficient training of Q-functions within a few interactions. 
However, it may cause overfitting of Q-functions and degrade sample efficiency.\\
\textbf{Regularization.} To mitigate overfitting, our REDQ uses (i) ensemble and (ii) layer normalization. 
(i) Ensemble of $N$ Q-functions is used as a regularization technique (lines 8--9). 
Specifically, a random subset $\mathcal{M}$ of the ensemble is selected (line 8) and used for target calculation (line 9). 
Each Q-function in the ensemble is randomly and independently initialized but updated with the same target (lines 10--11). 
(ii) Layer normalization~\citep{ba2016layer} is applied after the weight layer in each Q-function. 
Layer normalization is not used in the original REDQ paper~\citep{chen2021randomized}, but its subsequent works~\citep{hiraoka2022dropout,ball2023efficient} show that it further suppresses the overfitting and improves sample efficiency of REDQ. 
Following these subsequent works, we use layer normalization for our REDQ. 

REDQ has demonstrated high sample efficiency in dense-reward continuous-control tasks~\citep{brockman2016openai,fu2020d4rl} based on MuJoCo~\citep{todorov2012mujoco} (see e.g., \citet{chen2021randomized}). 
However, when applied to sparse-reward goal-conditioned tasks, it performs worse than previous SoTA methods (Fig.~\ref{fig:intro-remarks}). 
In the following sections, we will make modifications to improve REDQ's performance in sparse-reward goal-conditioned tasks. 

\subsection{Modification 1: Using Hindsight Experience Replay (\textcolor{uscarlet}{HER})}\label{sec:her}
Numerous technical innovations have been developed for sparse-reward goal-conditioned RL (see Section~\ref{sec:related_work} for details), and many of these innovations can be applied to REDQ. 
We want to keep our method simple and flexible to allow its users to introduce complex innovations as needed. 
Thus, we begin our modification of REDQ by introducing the fundamental component commonly used in previous innovations. 

\begin{figure}[t!]
\centering
\begin{minipage}{.5\hsize}
\includegraphics[clip, width=0.49\hsize]{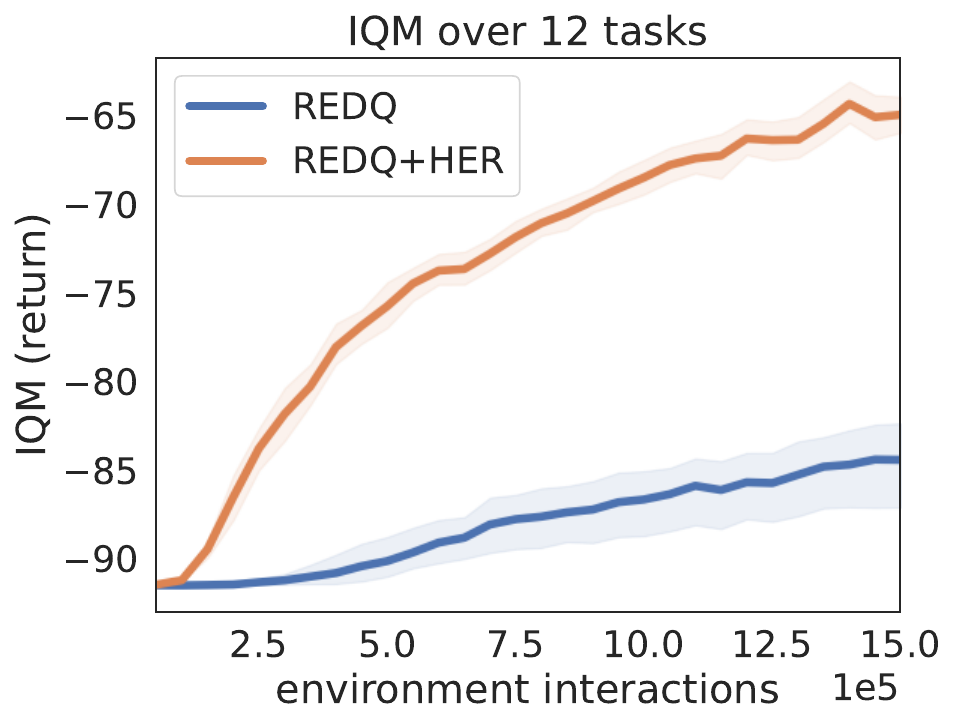}
\includegraphics[clip, width=0.49\hsize]{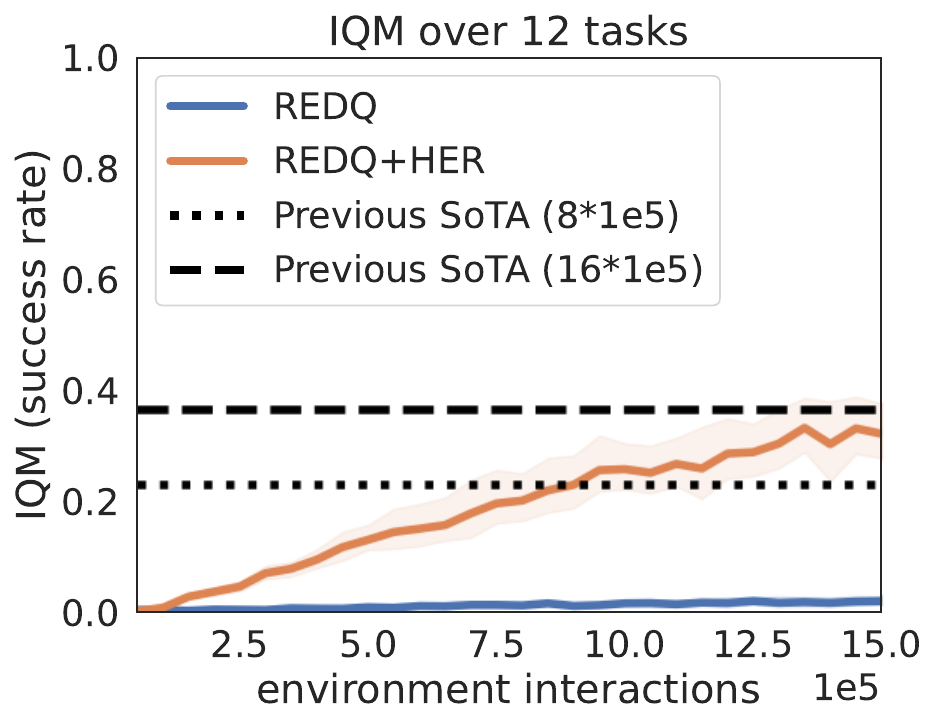}
\end{minipage}
\vspace{-0.7\baselineskip}
\caption{
Effect of HER on REDQ's performance. 
The left figure: the learning curve for a return. 
The right figure: the curve for task success rate. 
These figures indicate that the use of HER significantly improves performance. 
We can see that REDQ with HER (REDQ+HER) exhibits superior returns and success rates to vanilla REDQ. 
}
\label{fig:method-her-influence}
\vspace{-0.5\baselineskip}
\end{figure}

We introduce HER~\citep{andrychowicz2017hindsight} with a future strategy into REDQ to improve its performance. 
HER with the future strategy is commonly used in previous works for sparse-reward goal-conditioned RL methods~\citep{andrychowicz2017hindsight,plappert2018multi,zhao2018energy,zhao2019maximum,xu2023efficient}. 
HER replaces a goal $g$ of a past transition with a new goal $g_t'$ to obtain positive rewards (line 5 in Algorithm~\ref{alg1:REDQwithOurDesigneDecision}). 
For selecting the new goal $g_t'$, our HER follows the future strategy. 
In the future strategy, for each transition $(s_t, a_t, r_t', s_{t+1}, g) \in \{(s_t, a_t, r_t', s_{t+1}, g)\}_{t=0}^{T}$,  $g$ is replaced with $g_t'$, which is the achieved goal that comes from the same trajectories as the transition and was observed after it. 
HER with the future strategy significantly improves REDQ's performance (Fig.~\ref{fig:method-her-influence}). 

\subsection{Modification 2: Bounding Target Q-Value (\textcolor{uviolet}{BQ})}\label{sec:boundq}
REDQ (Section~\ref{sec:base_method}) employs (i) off-policy learning, (ii) approximation of the value function, and (iii) bootstrapping (i.e. the deadly triad~\citep{sutton2018reinforcement}). 
This deadly triad often leads to Q-value estimate divergence and consequently degrades performance~\citep{van2018deep}. 

\begin{figure}[t!]
\centering
\begin{minipage}{1.0\hsize}
\includegraphics[clip, width=0.245\hsize]{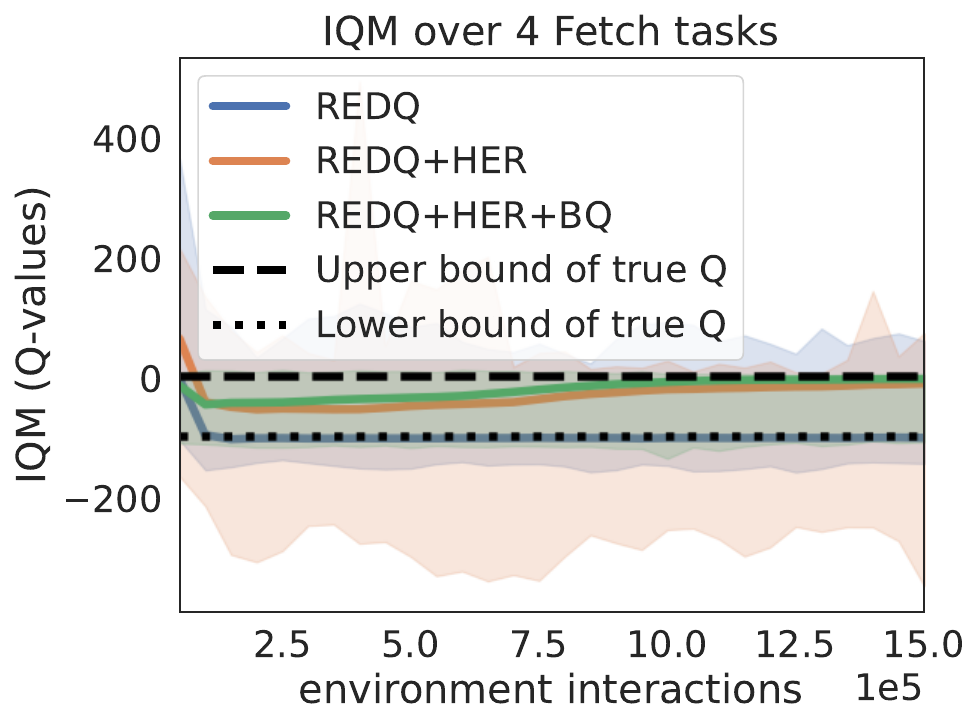}
\includegraphics[clip, width=0.245\hsize]{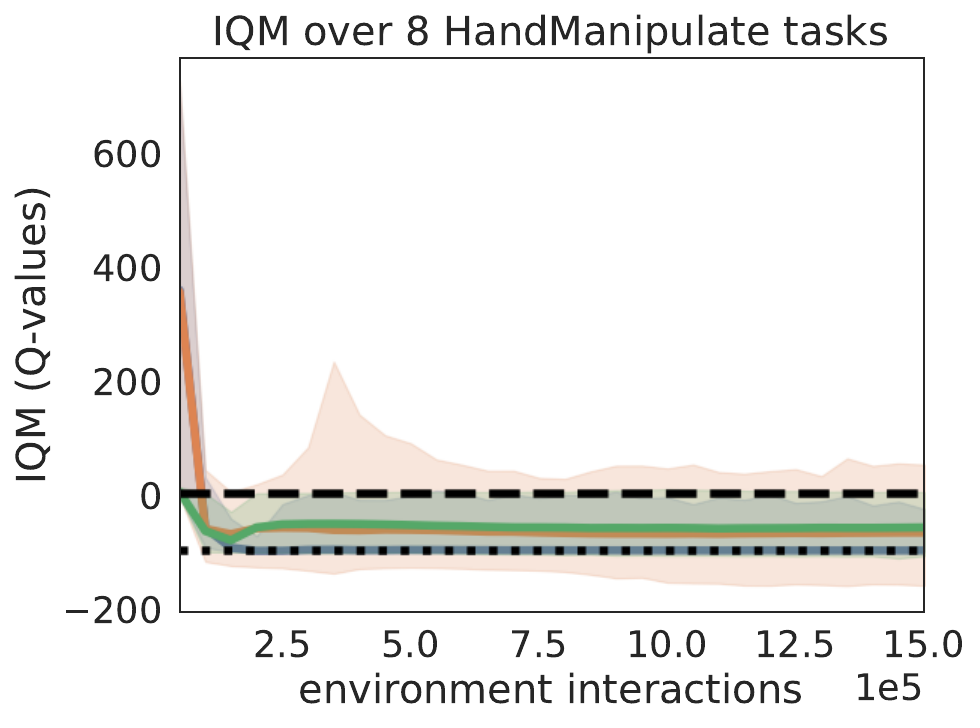}
\includegraphics[clip, width=0.245\hsize]{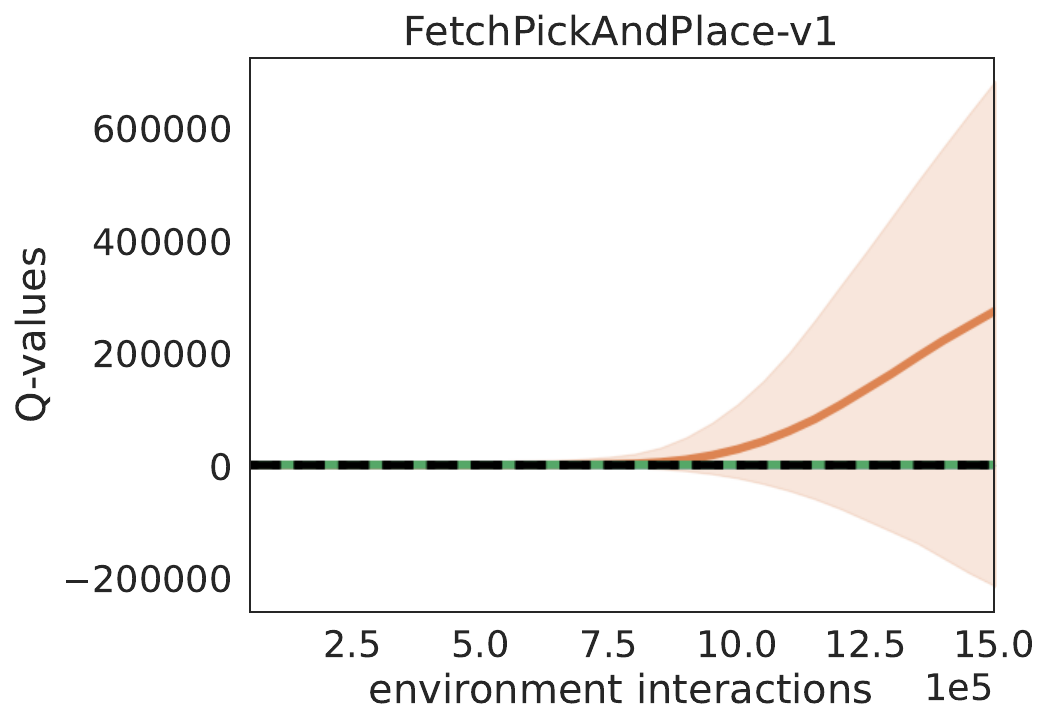}
\includegraphics[clip, width=0.245\hsize]{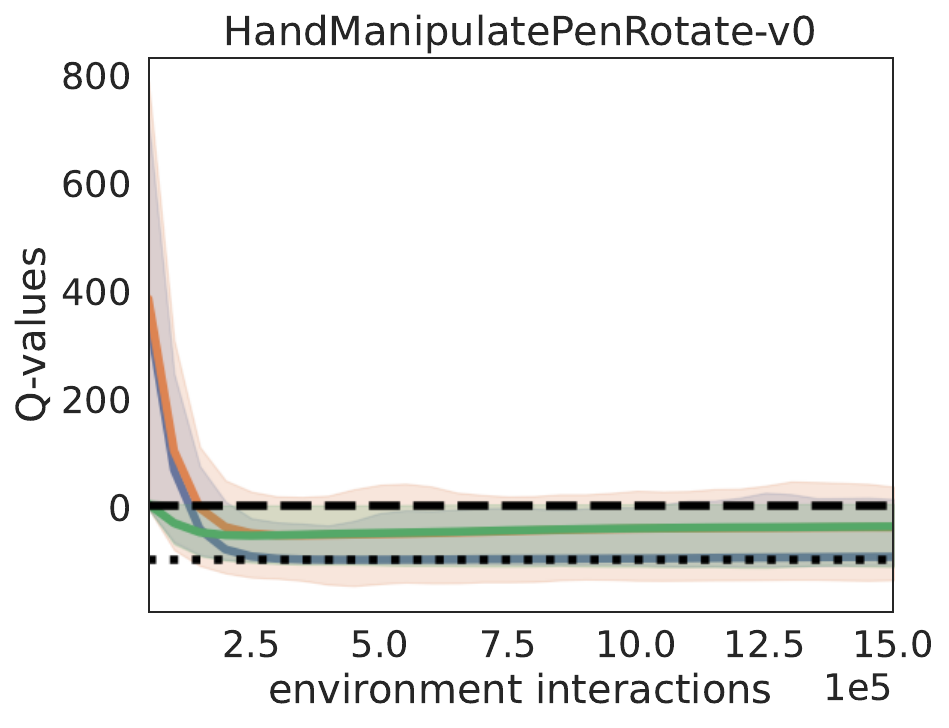}
\end{minipage}
\vspace{-0.7\baselineskip}
\caption{
The effect of BQ on Q-value divergence. 
The solid line represents the average Q-value estimate, while the shaded area represents the range of Q-value estimates.
Dashed and dotted lines represent the theoretical upper bound ($Q_\max$) and lower bound ($Q_\min$) of Q-value, respectively. 
Two figures on the left-hand side: a summary (IQM) of the scores over 4 Fetch tasks and 8 HandManipulate tasks. 
Two figures on the right-hand side: examples of scores in individual tasks (FetchPickAndPlace and HandmanipulatePenRotate): 
From the figures, we can see that (i) Q-value estimates of REDQ with HER (REDQ+HER) significantly exceed the bound range and (ii) estimates of the method using bounded target Q-value (REDQ+HER+BQ) are kept almost within the range. 
The results for all tasks are shown in Fig.~\ref{fig:app-method-boundq-qvals} in the appendix. 
}
\label{fig:method-boundq-qvals}
\vspace{-0.5\baselineskip}
\end{figure}

We observe that introducing HER to REDQ induces a divergence in its Q-value estimation. 
We assess the extent to which Q-value estimates exceed the theoretical upper bound $Q_\max$ and lower bound $Q_\min$. 
Here, $Q_\max$ is the discounted future return in the best-case scenario, where an agent consistently receives a positive reward, while $Q_\min$ is the return in the worst-case scenario with consistent negative rewards. 
In Robotics~\citep{plappert2018multi,gymnasium_robotics2023github} tasks, the positive reward is 0, and the negative reward is -1 ~\footnote{See the second paragraph in Section~\ref{sec:prelim} for a reminder of the reward structure of Robotics tasks.}. 
Thus, we estimate $Q_\max$ and $Q_\min$ as: 
For any time step $t$, $Q_\max = \sum_{t'=t}^{\infty} \gamma^{(t' - t)} \cdot 0 = 0$, and $Q_\min = \sum_{t'=t}^{\infty} \gamma^{(t'-t)} \cdot -1 = -1 / (1 - \gamma)$. 
The result (Fig.~\ref{fig:method-boundq-qvals}) shows that HER induces a divergence in Q-value estimation.
We can see that the Q-value estimates of REDQ with HER (REDQ+HER) significantly surpass theoretical bounds compared with those of REDQ. 

We bound the target Q-value to mitigate the Q-value estimate divergence. 
Specifically, we bound the target Q-value using $Q_\max$ and $Q_\min$ (line 9 in Algorithm~\ref{alg1:REDQwithOurDesigneDecision}): 
\begin{dmath}
   y = r + \gamma \min \left( \max \left( \min_{i \in \mathcal{M}} Q_{\bar{\phi}_i}(s', a', g) - \alpha \log \pi_\theta(a' | s', g), Q_\min \right), Q_\max \right). \label{eq:bounded_target}
\end{dmath}
Here, $Q_\max$ and $Q_\min$ are the same as the ones introduced in the preceding paragraph. 
This bounding effectively suppresses the Q-value estimate divergence (Fig.~\ref{fig:method-boundq-qvals}). 
We will experimentally show that this modification substantially enhances overall performance in the next section. 

\section{Experiment}\label{sec:experiment}
In the previous section, we introduced HER (Section~\ref{sec:her}) and BQ (Section~\ref{sec:boundq}) into REDQ (Section~\ref{sec:base_method}). 
In this section, we conduct experiments to answer two questions:\\ 
\textbf{Q1.} Are both HER and BQ necessary to improve the performance of REDQ?\\ 
\textbf{Q2.} Does REDQ with HER and BQ perform as well as or better than previous SoTA methods?\\
\begin{figure}[t!]
\centering
\begin{minipage}{.5\hsize}
\includegraphics[clip, width=0.49\hsize]{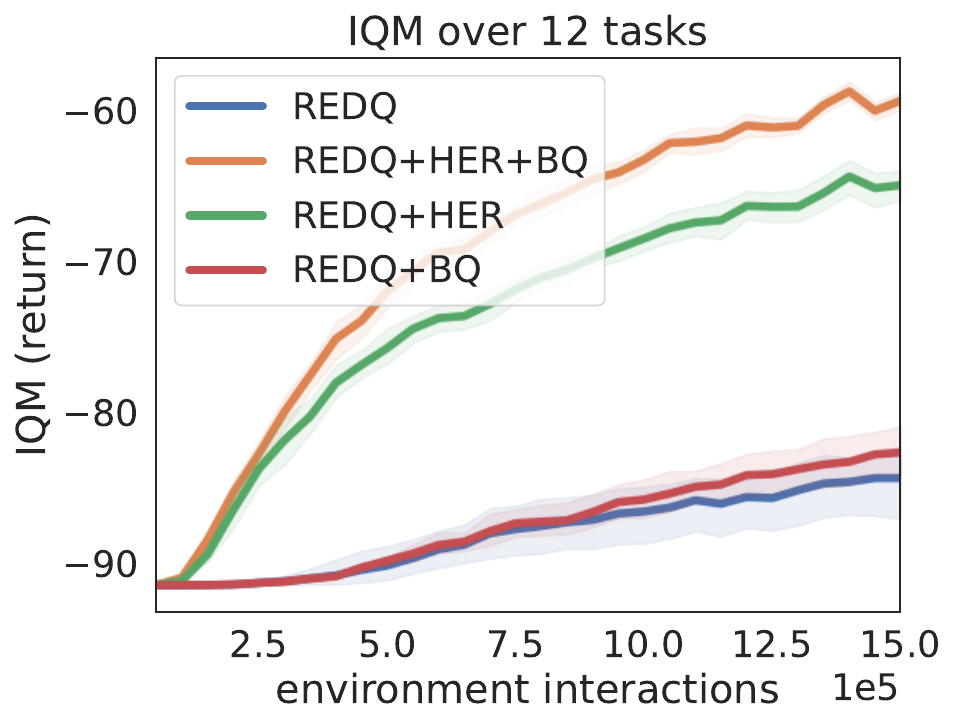}
\includegraphics[clip, width=0.49\hsize]{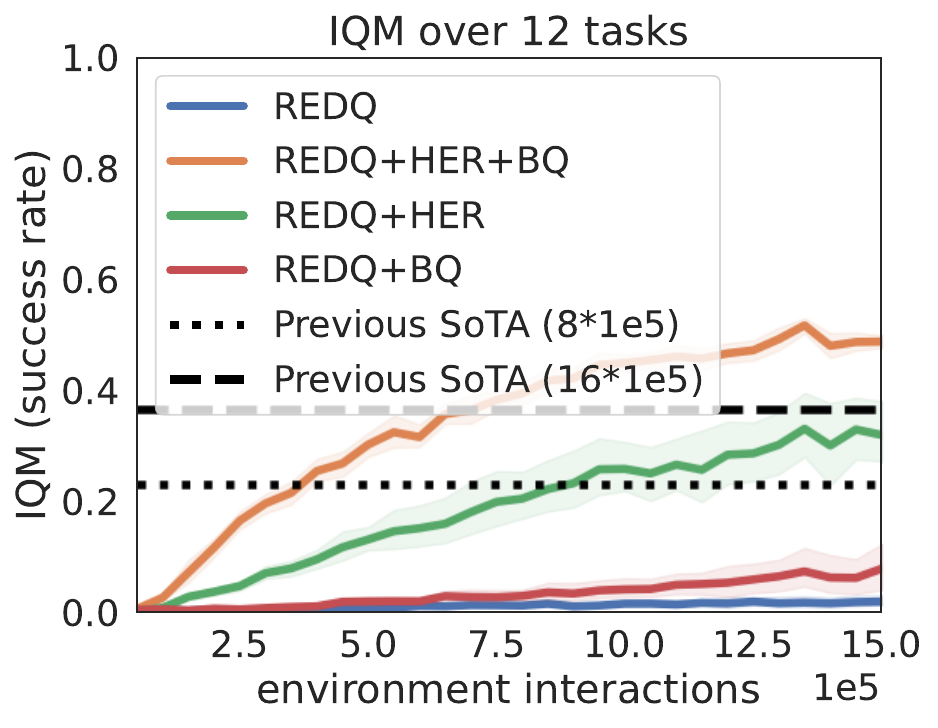}
\end{minipage}
\vspace{-0.7\baselineskip}
\caption{
IQM of performance (return and success rate) over 12 Robotics tasks. 
The left-hand side figure: the IQM of return. 
The right-hand side figure: the IQM of task success rate. 
The left-hand side figure shows that our modifications (HER and BQ) significantly contribute to score improvement over the 12 tasks. 
The right-hand figure shows that REDQ+HER+BQ achieves approximately $2 \times$ better sample efficiency than previous SoTA methods.
}
\label{fig:experiment-return-sr-summary}
\vspace{-0.5\baselineskip}
\end{figure}

\begin{figure*}[t!]
\begin{minipage}{1.0\hsize}
\includegraphics[clip, width=0.24\hsize]{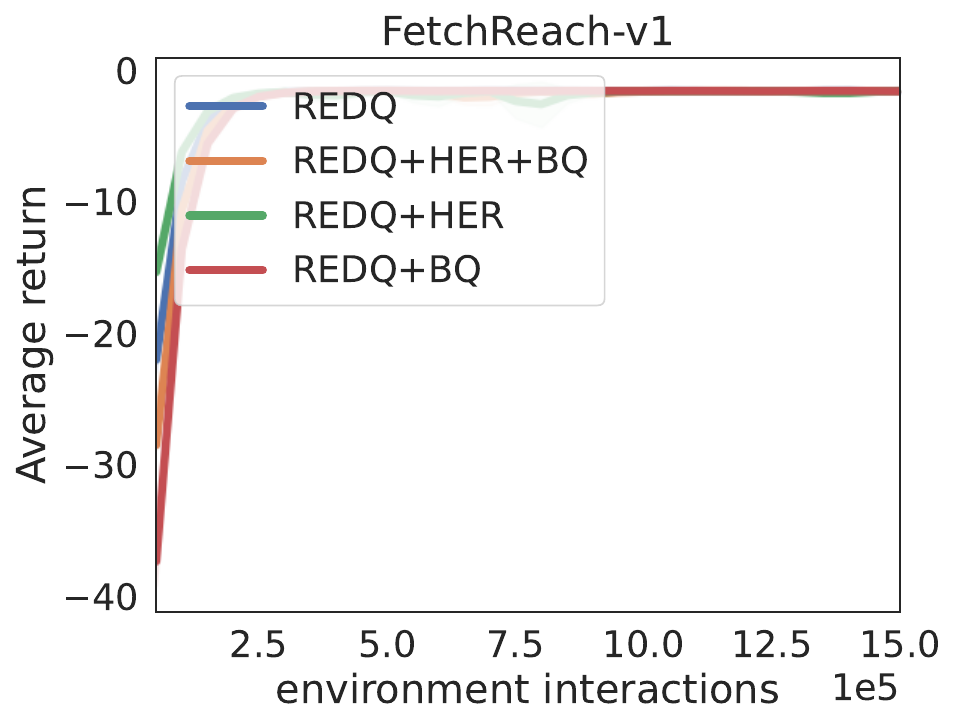}
\includegraphics[clip, width=0.24\hsize]{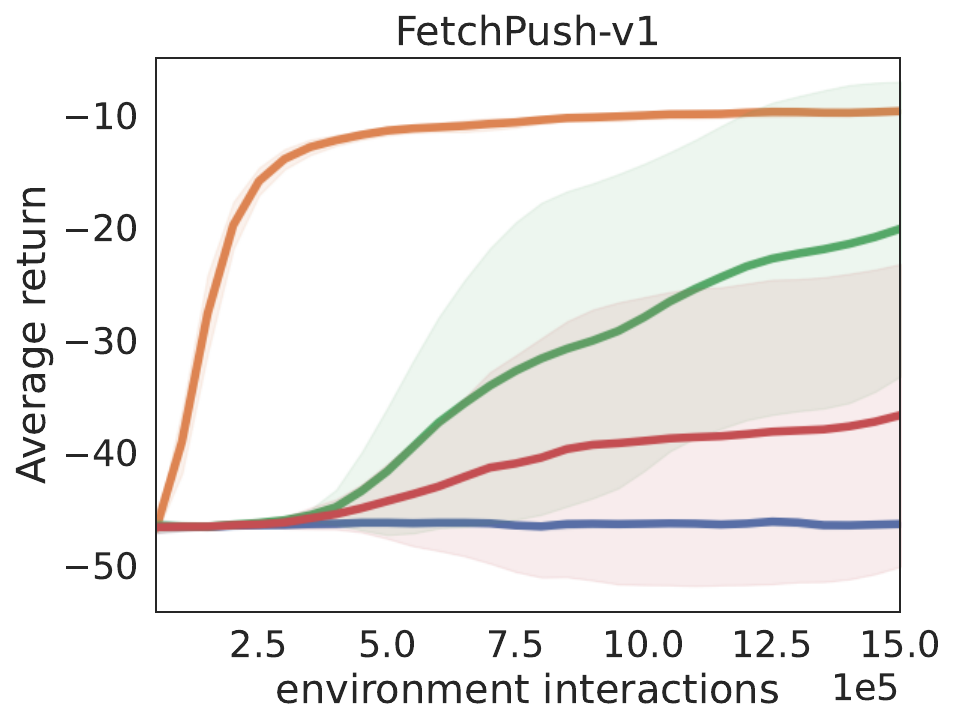}
\includegraphics[clip, width=0.24\hsize]{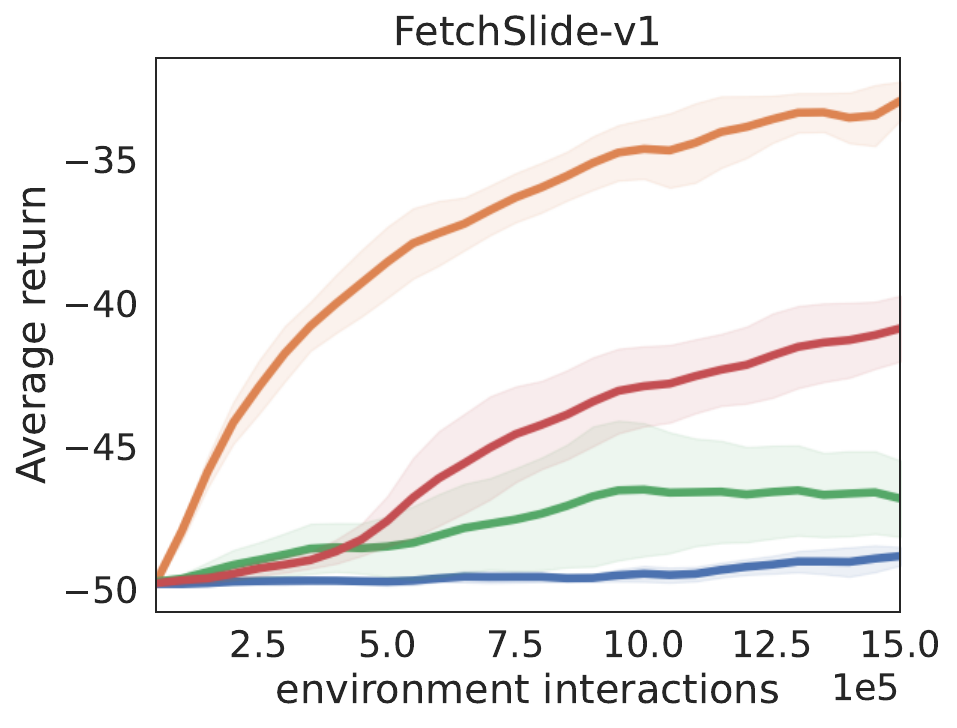}
\includegraphics[clip, width=0.24\hsize]{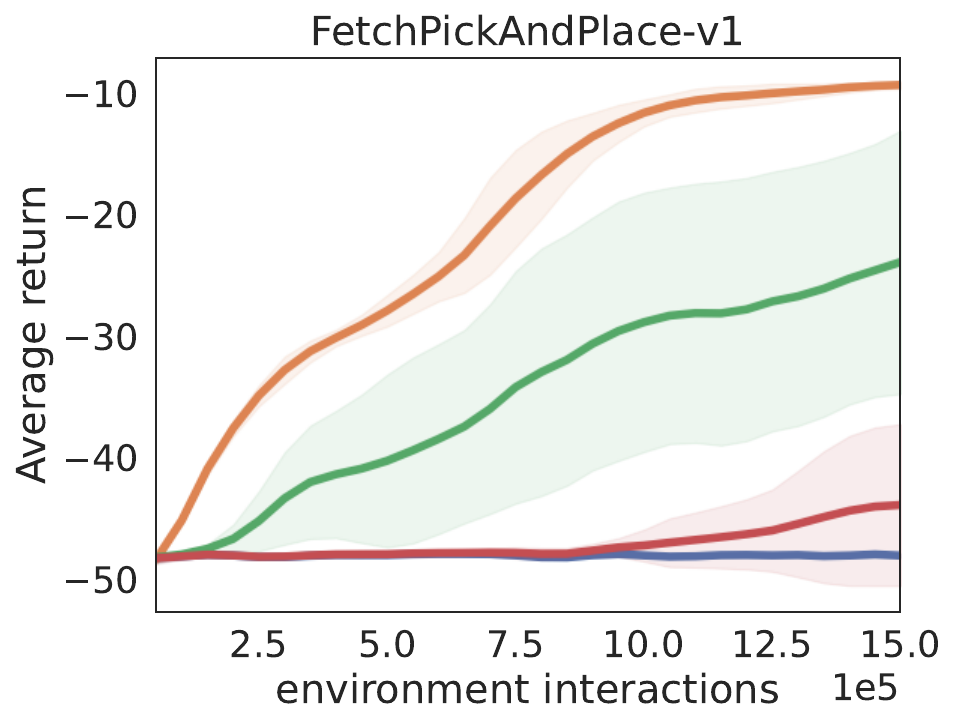}
\end{minipage}
\begin{minipage}{1.0\hsize}
\includegraphics[clip, width=0.24\hsize]{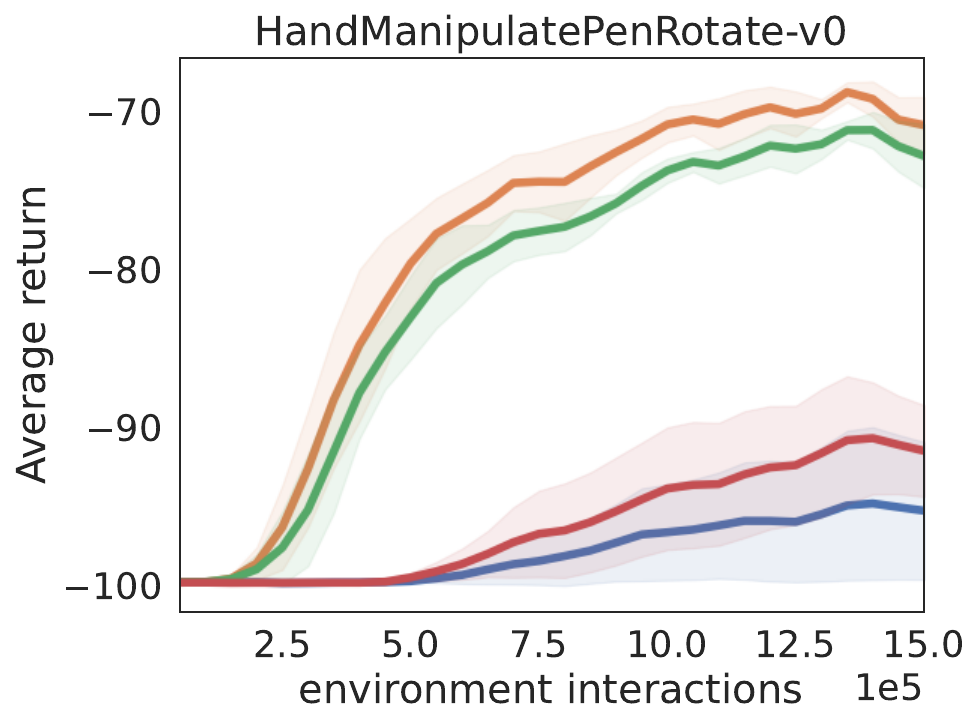}
\includegraphics[clip, width=0.24\hsize]{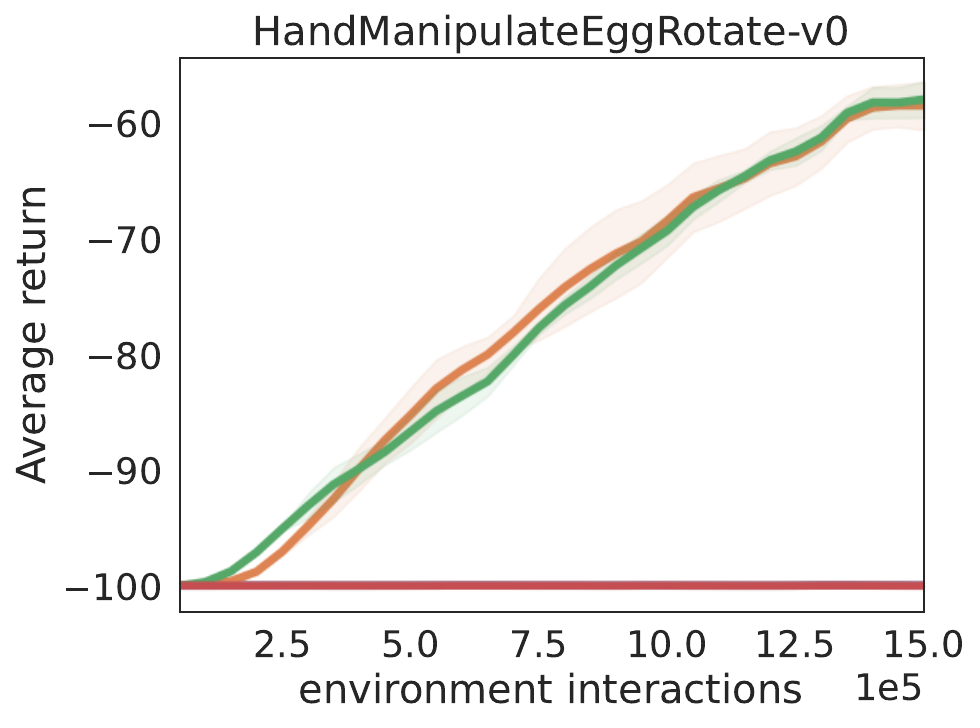}
\includegraphics[clip, width=0.24\hsize]{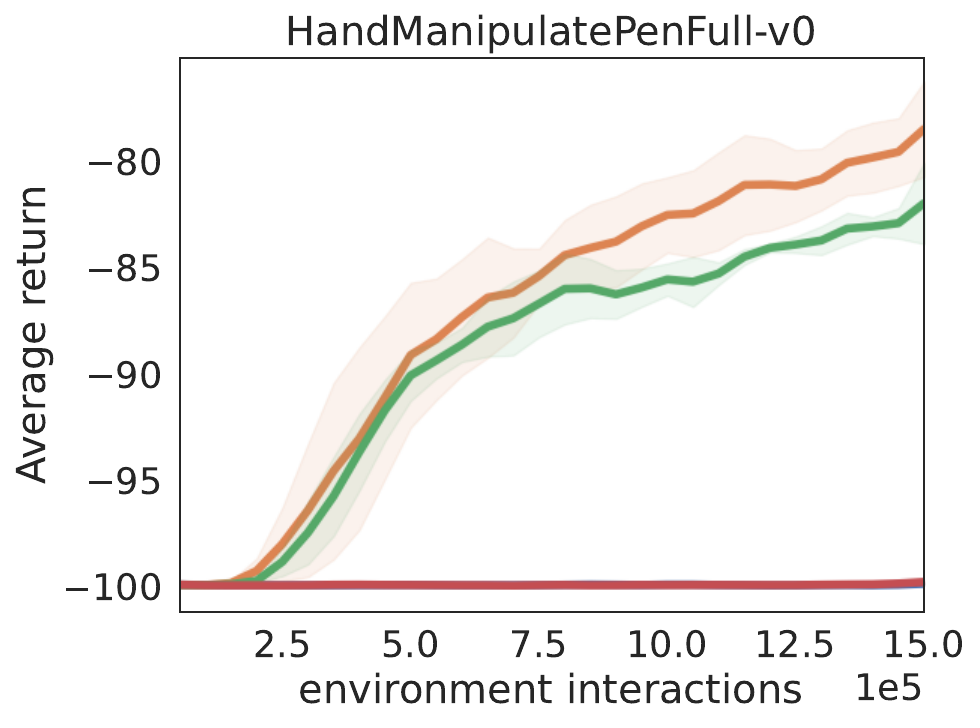}
\includegraphics[clip, width=0.24\hsize]{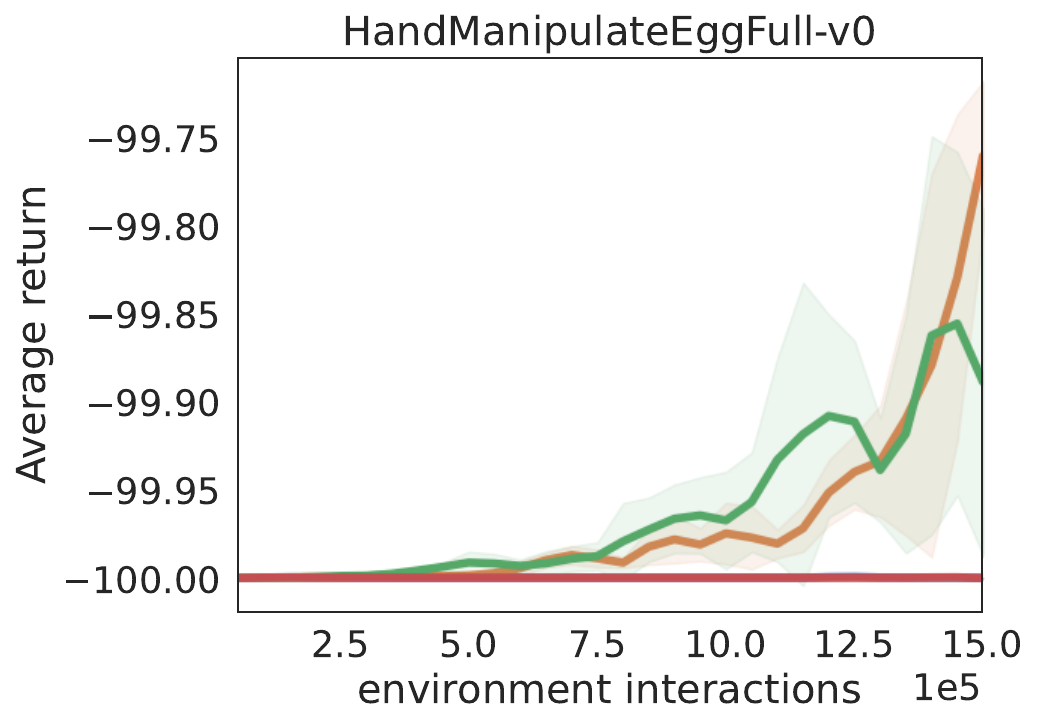}
\end{minipage}
\begin{minipage}{1.0\hsize}
\includegraphics[clip, width=0.24\hsize]{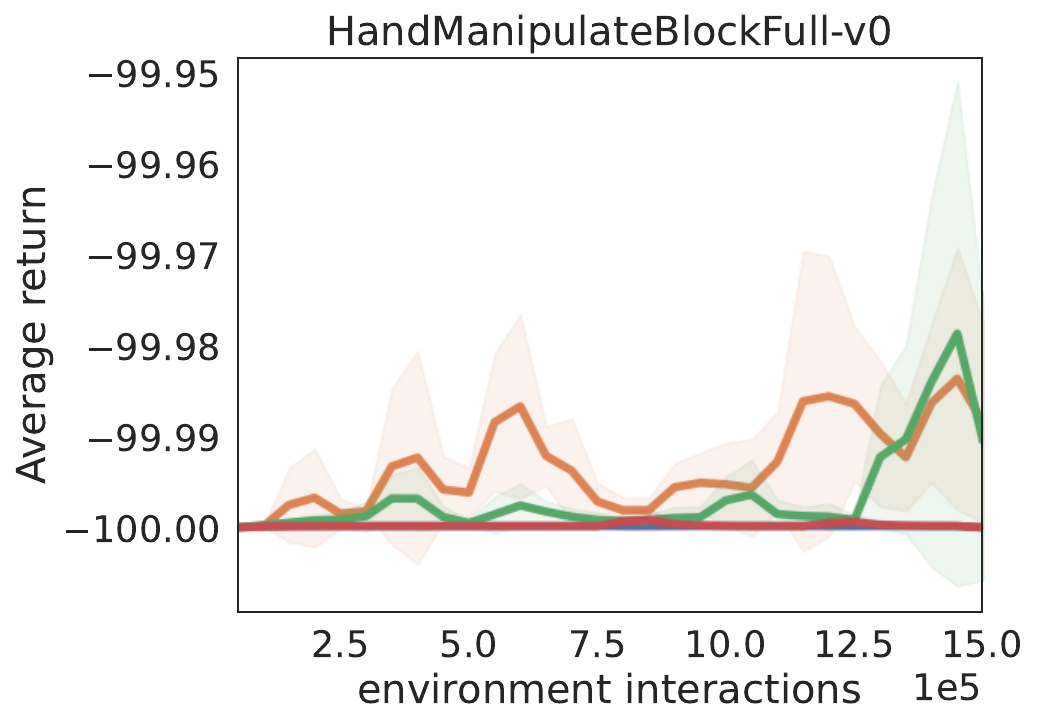}
\includegraphics[clip, width=0.24\hsize]{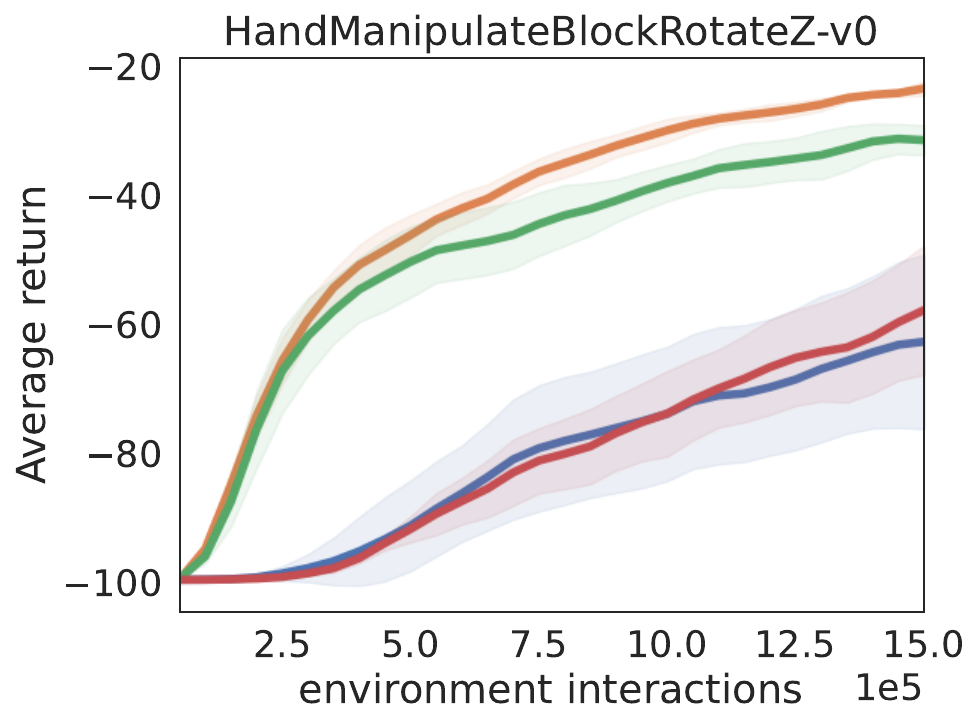}
\includegraphics[clip, width=0.24\hsize]{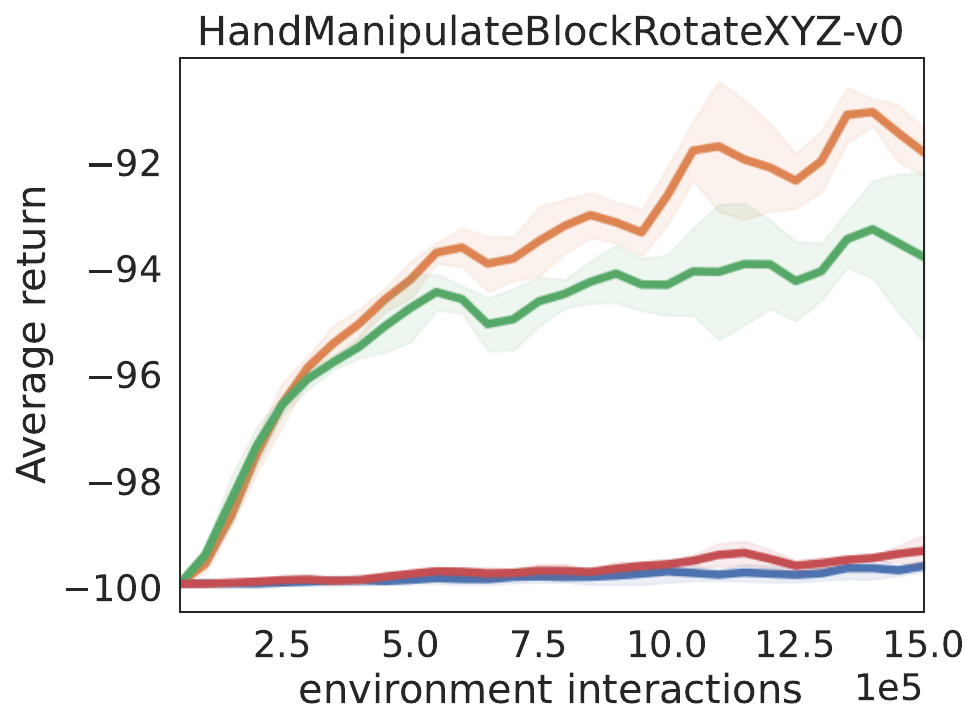}
\includegraphics[clip, width=0.24\hsize]{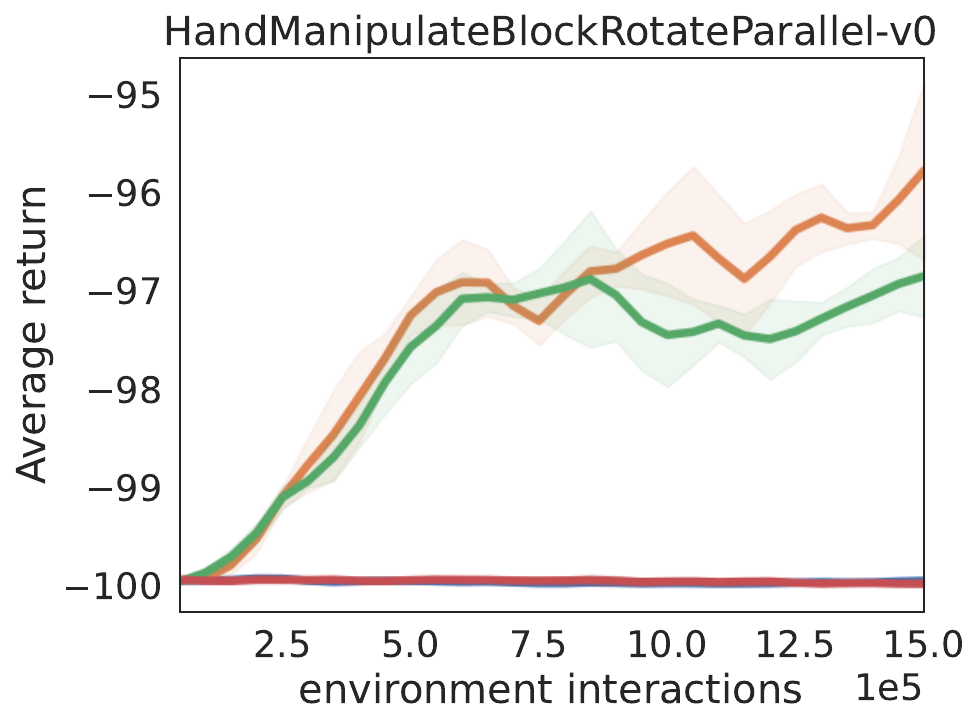}
\end{minipage}
\vspace{-0.7\baselineskip}
\caption{
Return improvement in each of 12 Robotics tasks. 
Figures show that HER alone significantly contributes to the return improvement in HandManipulate tasks, whereas both HER and BQ significantly contribute to the improvement in the Fetch tasks (except for FetchReach).
}
\label{fig:experiment-return}
\vspace{-0.5\baselineskip}
\end{figure*}

\textbf{Experiment for Q1: Both HER and BQ contribute to the performance improvement.} 
The experiment results (the left-hand side figure of Figs.~\ref{fig:experiment-return-sr-summary}) shows that both HER and BQ effectively improve the performance (return) over the 12 Robotics tasks. 
We can see that REDQ+HER+BQ improves performance more than REDQ+HER, REDQ+BQ, and REDQ. 
In addition, results for each task (Fig.~\ref{fig:experiment-return}) show that HER significantly improves the performance in HandManipulate tasks, while both HER and BQ significantly improve the performance in Fetch tasks (except for the FetchReach task). 
The main reason for these different trends in Fetch and HandManipulates tasks would be that the divergence of the Q-value estimation in Fetch tasks is more significant and greatly suppressed by BQ than that in HandManipulate tasks (Fig.~\ref{fig:app-method-boundq-qvals} in the appendix). 

\begin{figure*}[t!]
\begin{minipage}{1.0\hsize}
\includegraphics[clip, width=0.24\hsize]{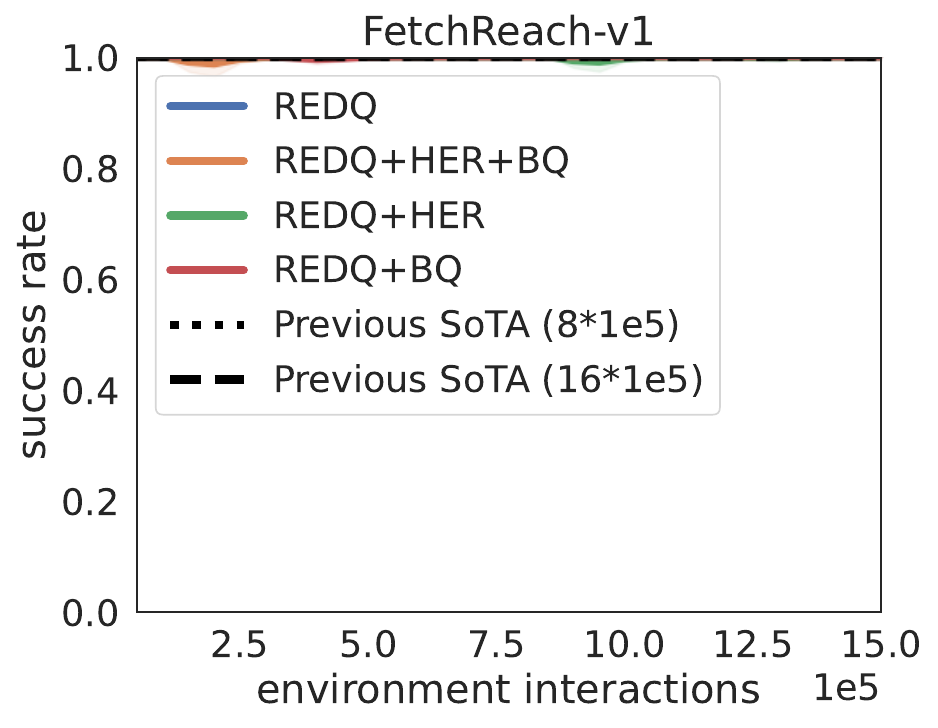}
\includegraphics[clip, width=0.24\hsize]{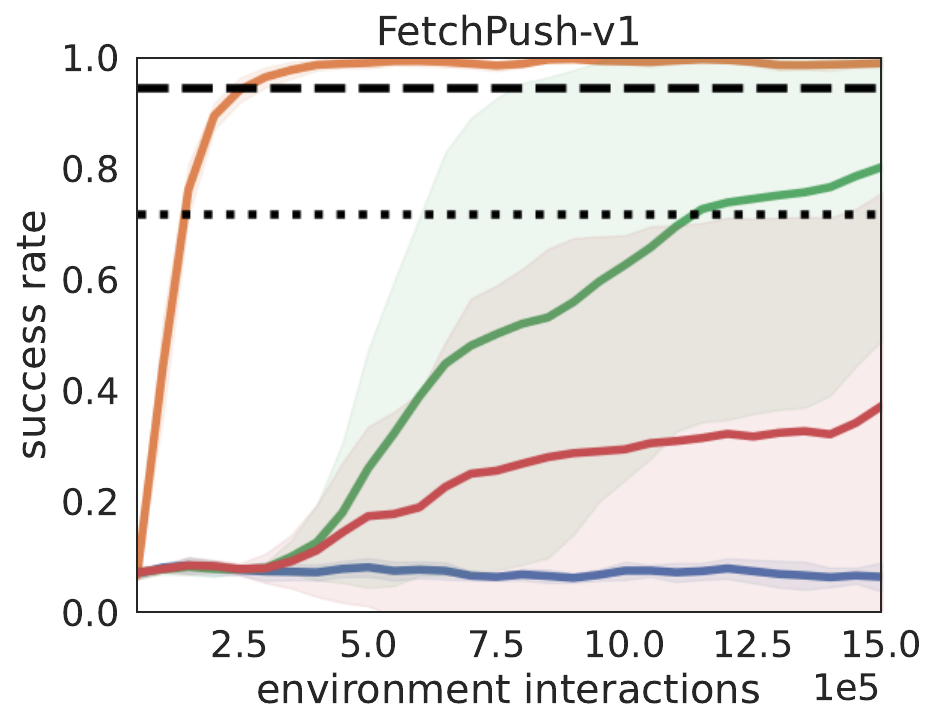}
\includegraphics[clip, width=0.24\hsize]{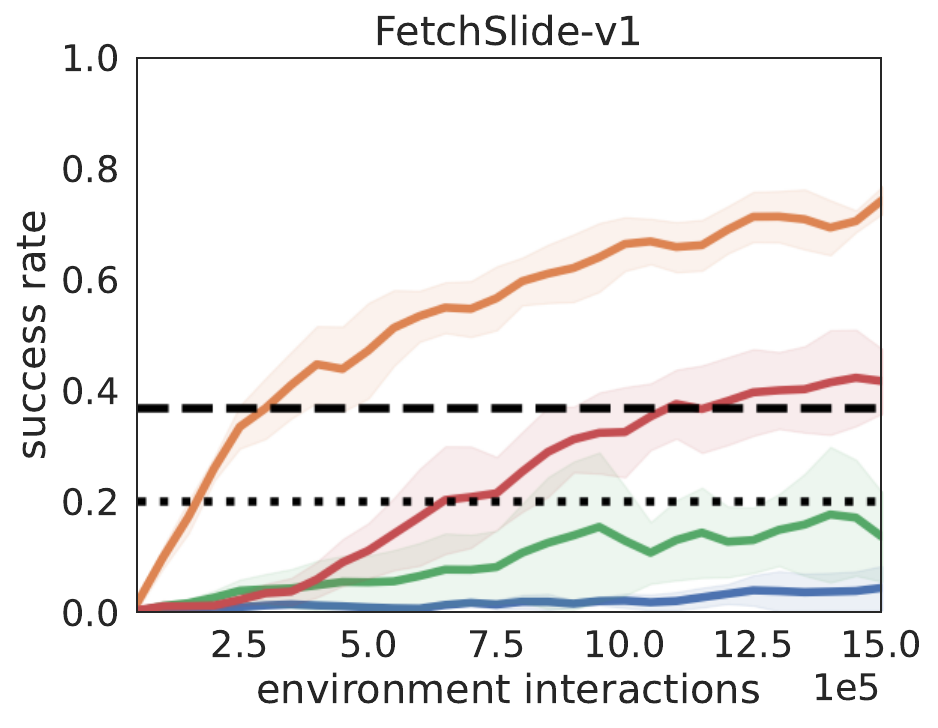}
\includegraphics[clip, width=0.24\hsize]{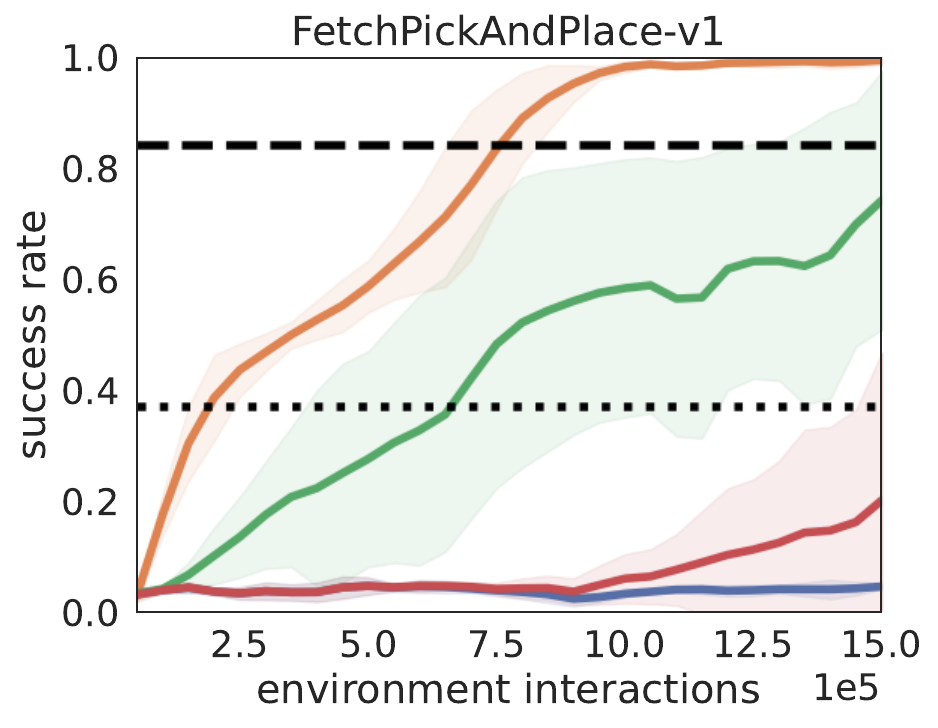}
\end{minipage}
\begin{minipage}{1.0\hsize}
\includegraphics[clip, width=0.24\hsize]{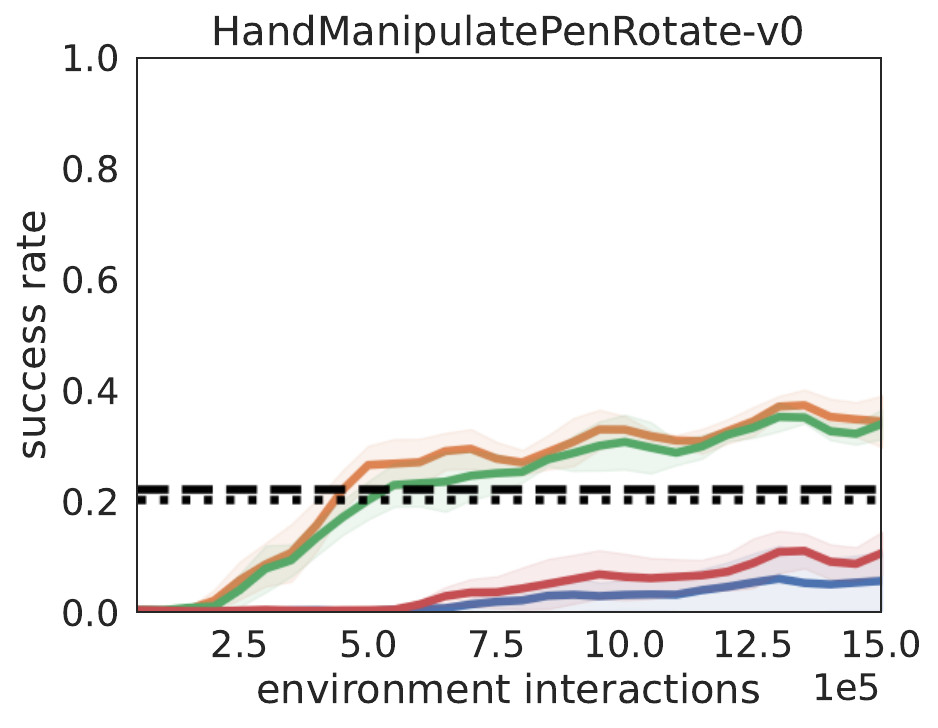}
\includegraphics[clip, width=0.24\hsize]{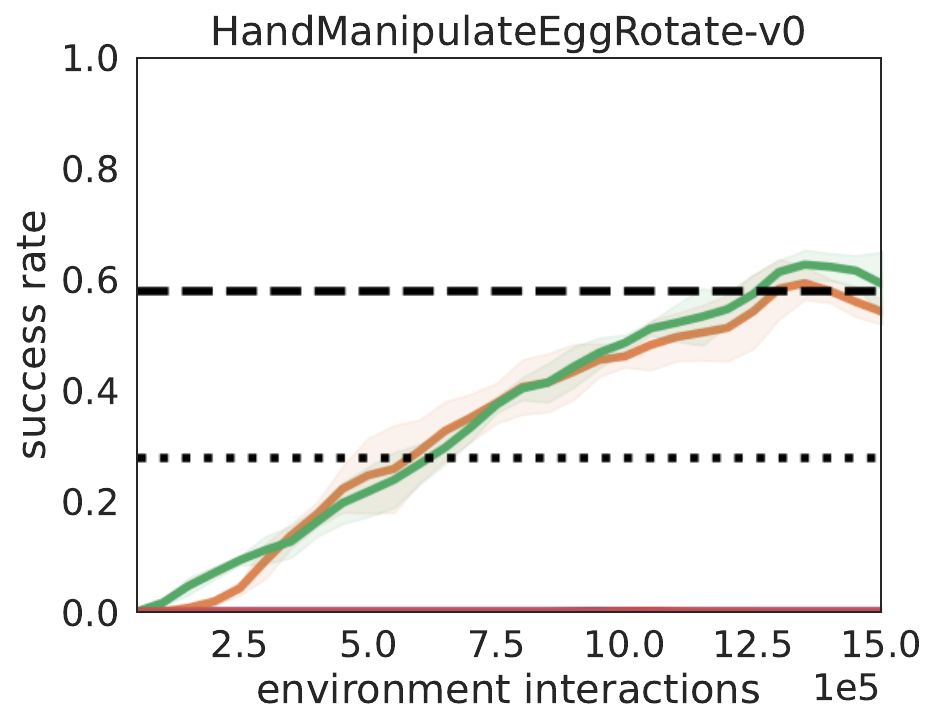}
\includegraphics[clip, width=0.24\hsize]{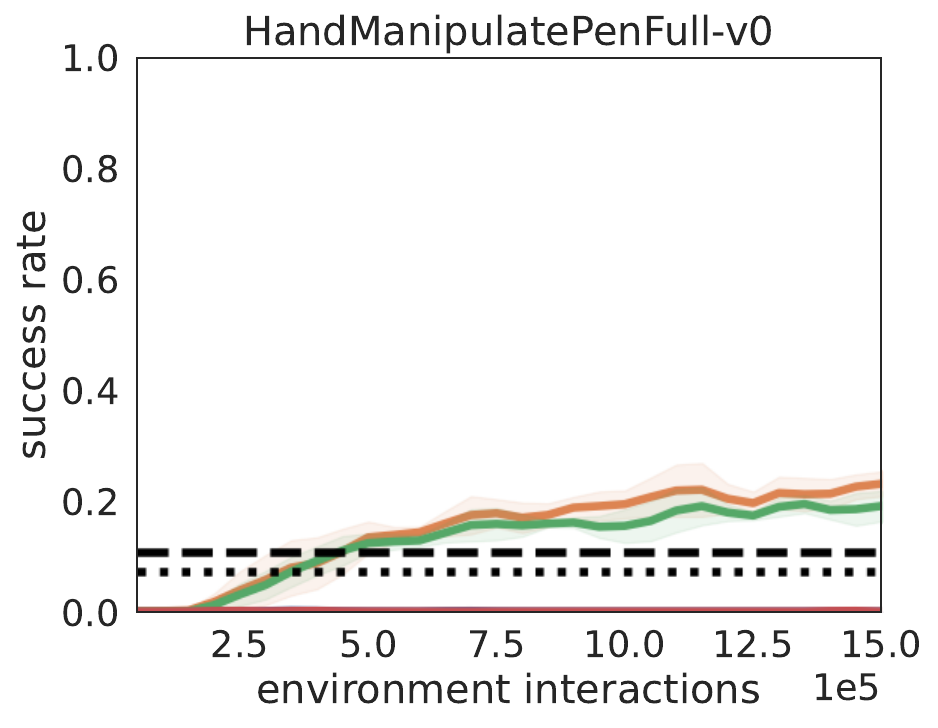}
\includegraphics[clip, width=0.24\hsize]{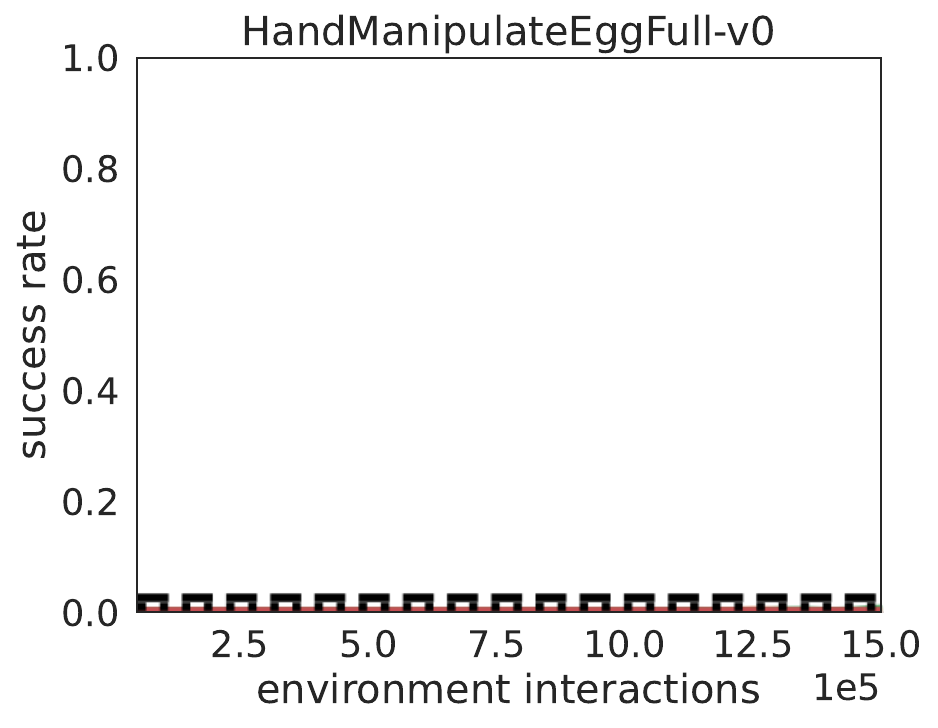}
\end{minipage}
\begin{minipage}{1.0\hsize}
\includegraphics[clip, width=0.24\hsize]{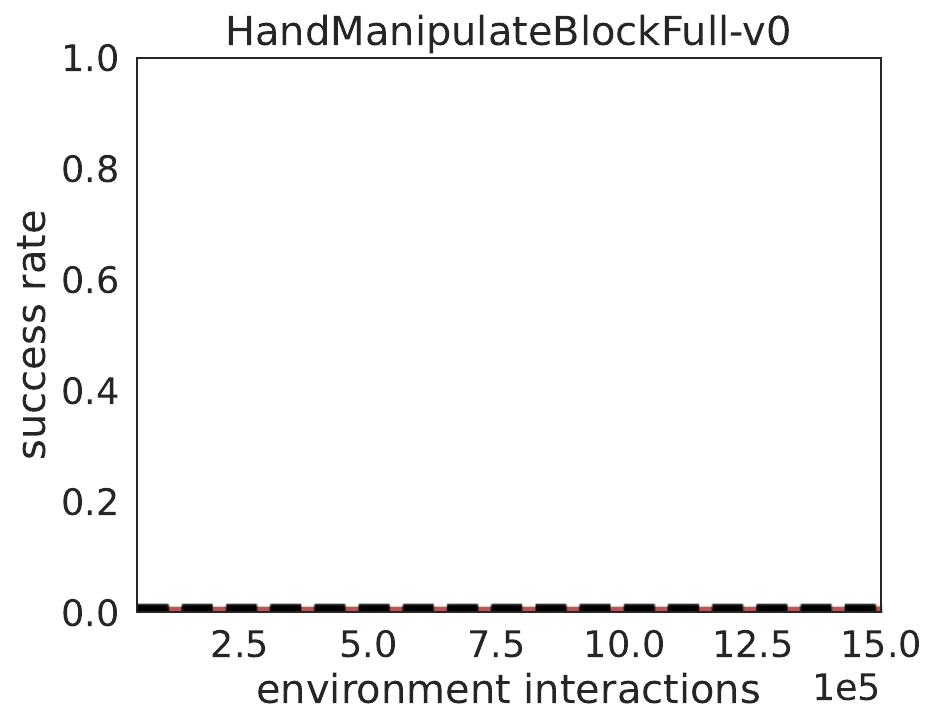}
\includegraphics[clip, width=0.24\hsize]{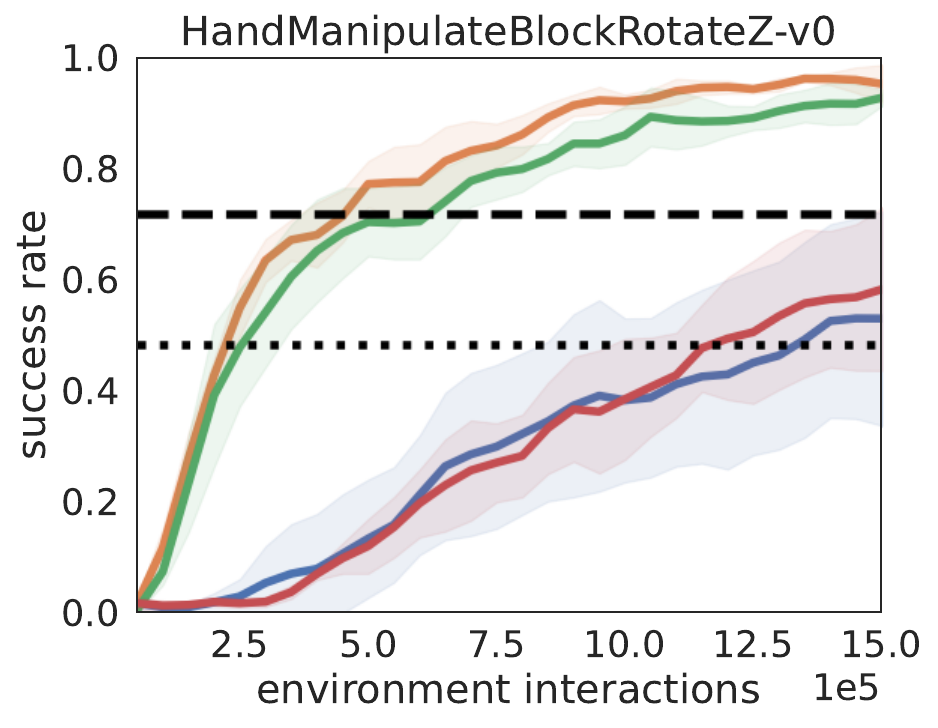}
\includegraphics[clip, width=0.24\hsize]{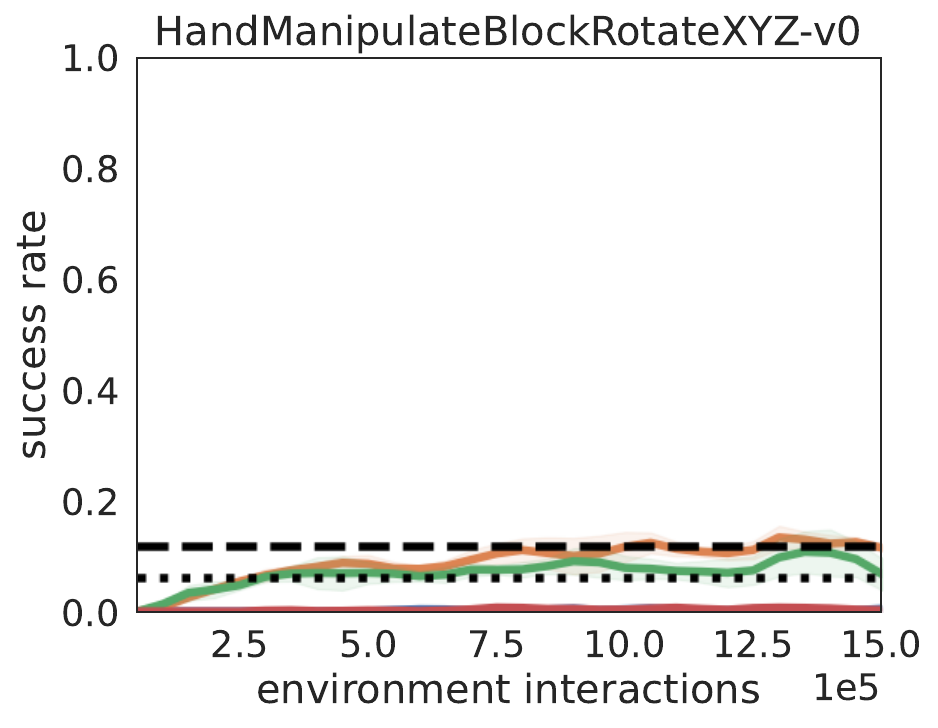}
\includegraphics[clip, width=0.24\hsize]{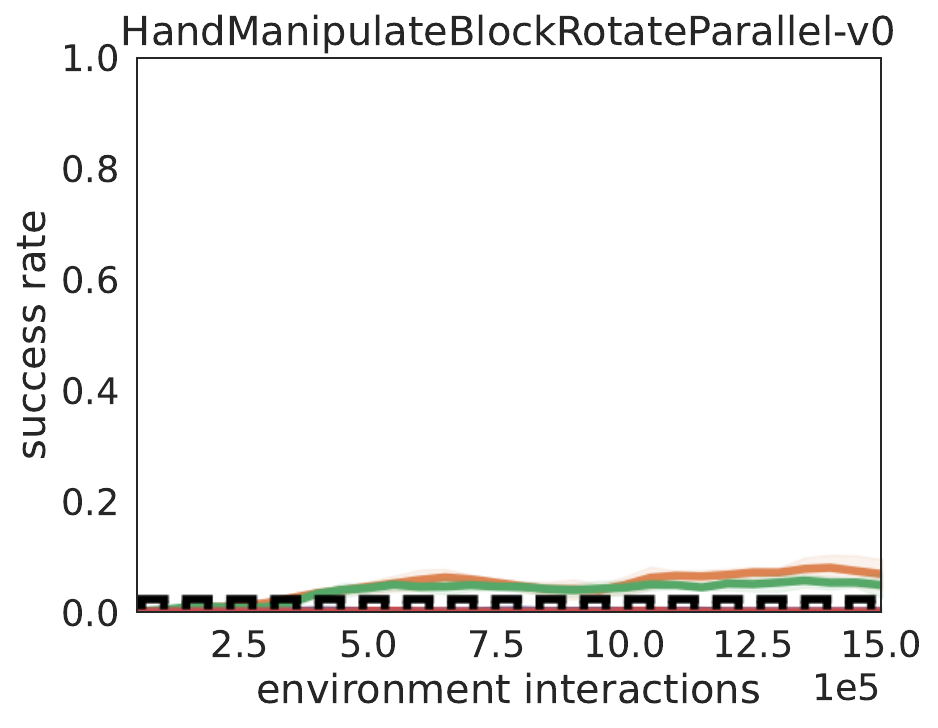}
\end{minipage}
\vspace{-0.7\baselineskip}
\caption{
The success rate in 12 Robotics tasks. 
The figures show that REDQ+HER+BQ exhibits particularly significant improvements compared with previous SoTA methods in the FetchSlide and HandManipulateBlockRotateZ tasks. 
}
\label{fig:experiment-sr}
\vspace{-0.5\baselineskip}
\end{figure*}

\textbf{Experiment for Q2: Our method performs better than previous SoTA RL methods.} 
We compare REDQ+HER+BQ with previous SoTA methods. 
For previous SoTA RL methods in Robotics tasks, we use HEREBP~\citep{zhao2018energy}, CHER~\citep{zhao2019curiosity}, DTGSH~\citep{dai2021diversity}, and VCPHER~\citep{xu2023efficient}. 
All these methods use a low RR ($ \leq 1$) and no regularization, unlike REDQ. 
We use the score of the best one among these previous methods, with $8 \cdot 10^5$ and $16 \cdot 16^5$ samples, for each task. 
As in previous works, we compare REDQ+HER+BQ and the previous SoTA on the basis of their sample efficiency in terms of task success rate. 
The experiment results (the right-hand side figure in Figs.~\ref{fig:experiment-return-sr-summary}) show that REDQ+HER+BQ achieves about $2 \times$ better sample efficiency than the previous SoTA. 
REDQ+HER+BQ with $8 \cdot 10^5$ samples performs comparably to the previous SoTA with $16 \cdot 10^5$ samples. 
In addition, REDQ+HER+BQ with $4 \cdot 10^5$ samples performs comparably to the previous SoTA with $8 \cdot 10^5$ samples. 
Looking at scores in each task (Fig.~\ref{fig:experiment-sr}), REDQ+HER+BQ makes particularly significant improvements against previous SoTA in, e.g., FetchSlide and HandManipulateBlockRotateZ tasks. 
On the other hand, the success rate of REDQ+HER+BQ is consistently close to 0, similar to the previous SoTA, in very difficult tasks such as HandManipulateEggFull and HandManipulateBlockFull. 

\section{Simplifying Our Method (REDQ+HER+BQ)}\label{sec:simplification}
In this section, we conduct experiments to gain insights for simplifying our method (REDQ+HER+BQ). 
For this, we assess the necessity of components of REDQ+HER+BQ. 
As we already assessed the necessity of HER and BQ in the previous section, we will focus on the components of REDQ in this section. 

\begin{figure}[t!]
\centering
\begin{minipage}{.5\hsize}
\includegraphics[clip, width=0.49\hsize]{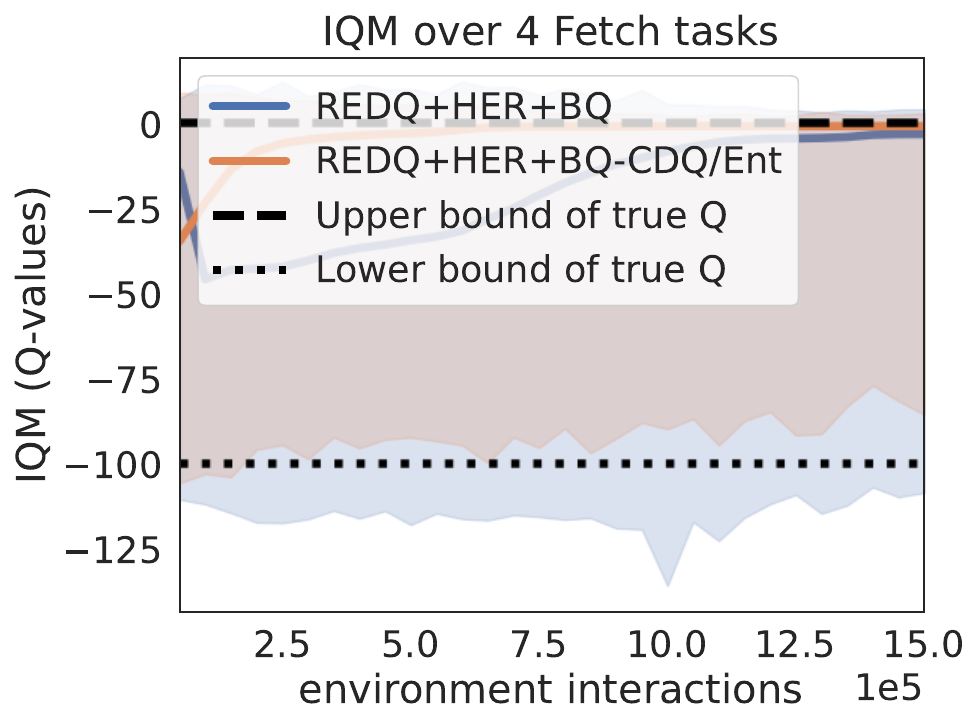}
\includegraphics[clip, width=0.49\hsize]{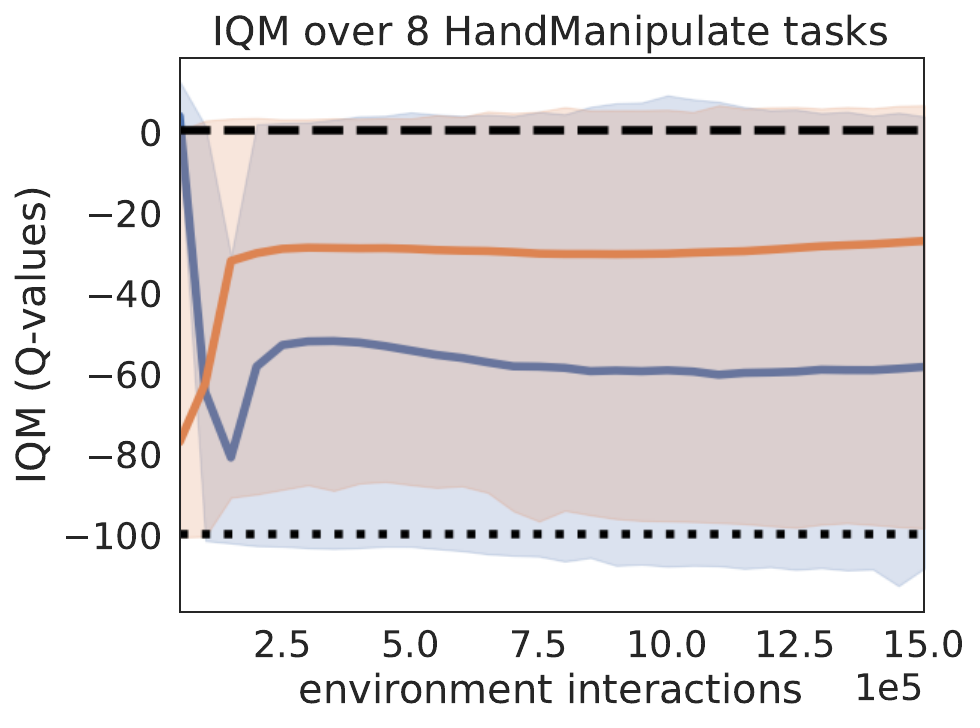}
\end{minipage}
\vspace{-0.7\baselineskip}
\caption{
The effect of removing CDQ and the entropy term on Q-value divergence. 
The figure shows that the method simplified by removing them (REDQ+HER+BQ-CDQ/Ent) can suppress the divergence of the Q-value to a similar extent as the method without the simplification (REDQ+HER+BQ). 
The results for all tasks are shown in Fig.~\ref{fig:app-simplication-qvals} in the appendix. 
}
\label{fig:simplication-qvals}
\vspace{-0.5\baselineskip}
\end{figure}

\begin{figure}[t!]
\centering
\begin{minipage}{0.5\hsize}
\includegraphics[clip, width=0.49\hsize]{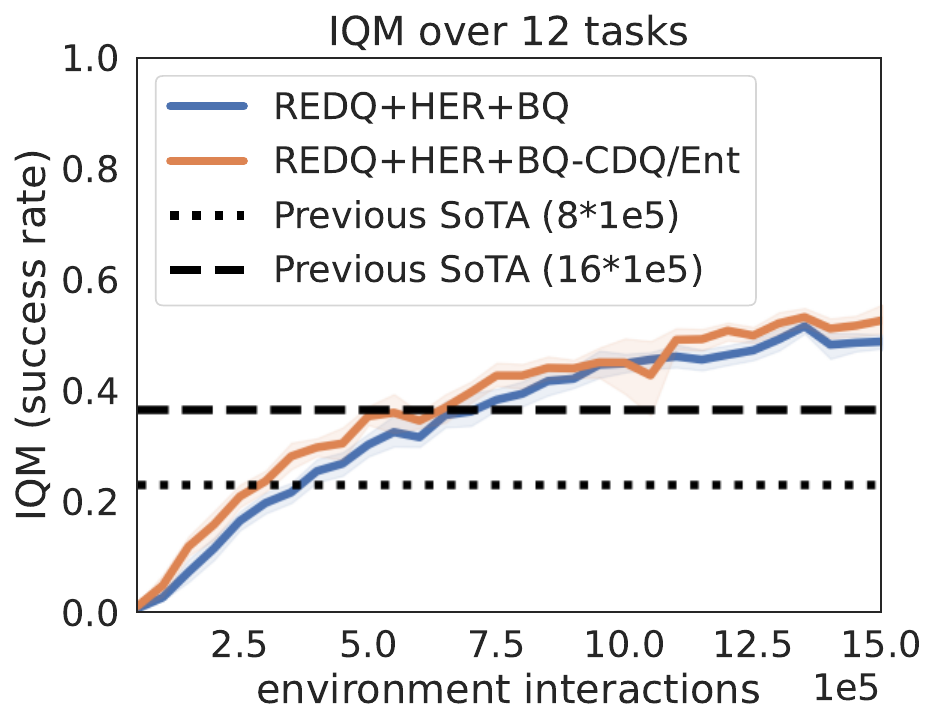}
\includegraphics[clip, width=0.49\hsize]{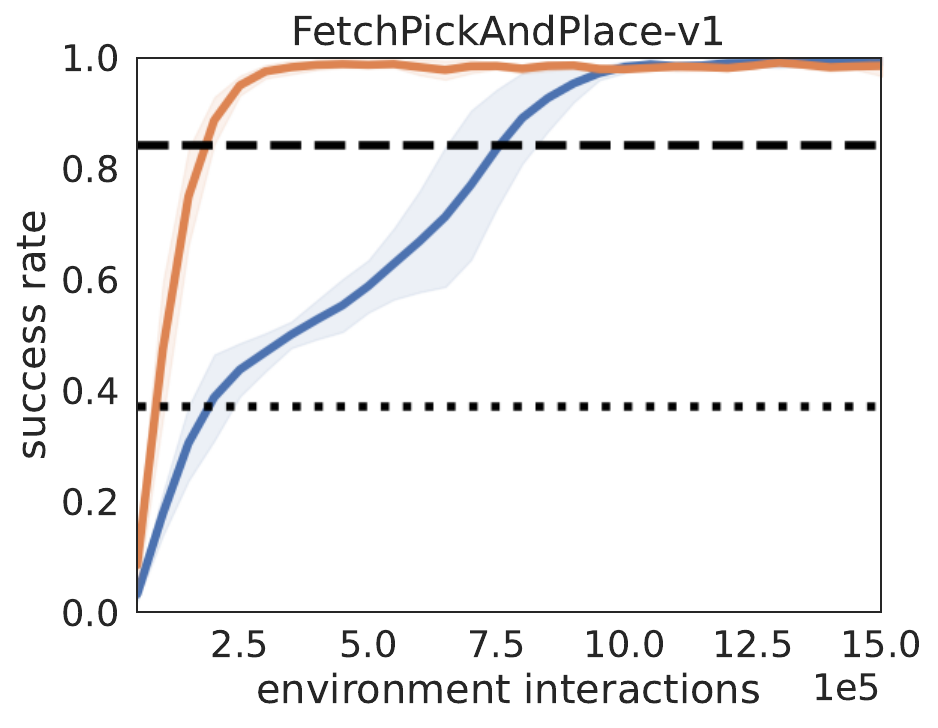}
\end{minipage}
\vspace{-0.7\baselineskip}
\caption{
The effect of removing CDQ and the entropy term on performance. 
The left-hand side figure: the IQM scores over 12 Robotics tasks. 
The right-hand side figure: the average return and success rate in FetchPickAndPlace. 
The left-hand side figure shows that the method not using CDQ and entropy term (REDQ+HER+BQ-CDQ/Ent) achieves an overall performance comparable to that of the original method (REDQ+HER+BQ). 
The right-hand side figure shows that REDQ+HER+BQ-CDQ/Ent achieves $\sim 8 \times$ better sample efficiency than the previous SoTA in the FetchPickAndPlace task. 
The results for all tasks are shown in Figs.~\ref{fig:app-simplification-return} and \ref{fig:app-simplification-sr} in the appendix. 
}
\label{fig:simplification-return-sr-summary}
\vspace{-0.5\baselineskip}
\end{figure}

\textbf{Are clipped double Q-learning and an entropy term removable? Yes.} 
REDQ calculates the target Q-value (Eq.~\ref{eq:bounded_target}) with clipped double Q-learning (CDQ)~\citep{fujimoto2018addressing} and an entropy term: 
(i) CDQ $\min_{i \in \mathcal{M}} Q_{\bar{\phi}_i}(s', a', g)$, and 
(ii) the entropy term $\alpha \log \pi_\theta(a' | s', g)$. 
The effectiveness of these components often depends heavily on the specific task~\citep{ball2023efficient}. 
Thus, we investigate whether they are necessary in our task or not. 
For this, we remove CDQ and the entropy term as: 
\begin{dmath}
    y = r + \gamma \min \left( \max \left( \frac{1}{|\mathcal{M}|}\sum_{i \in \mathcal{M}} Q_{\bar{\phi}_i}(s', a', g), Q_\min \right), Q_\max \right).
\end{dmath}
Here, the average operator $\frac{1}{|\mathcal{M}|}\sum_{i \in \mathcal{M}}$ is used instead of the minimum operator $\min_{i \in \mathcal{M}}$. 
The method simplified in this way (REDQ+HER+BQ-CDQ/Ent) can suppress Q-value divergence to a similar extent to the original method (REDQ+HER+BQ) (Fig.~\ref{fig:simplication-qvals}). 
In addition, REDQ+HER+BQ-CDQ/Ent can achieve almost the same overall (IQM) performance as REDQ+HER+BQ (the left-hand side figure in Figs.~\ref{fig:simplification-return-sr-summary}). 
Furthermore, REDQ+HER+BQ-CDQ/Ent achieves $\sim 8 \times$ better sample efficiency than the previous SoTA in the FetchPickAndPlace task (the right-hand side figure in Figs.~\ref{fig:simplification-return-sr-summary}). 
Given these results, we conclude that CDQ and the entropy term are removable in our tasks. 

\begin{figure}[t!]
\centering
\begin{minipage}{.5\hsize}
\includegraphics[clip, width=0.49\hsize]{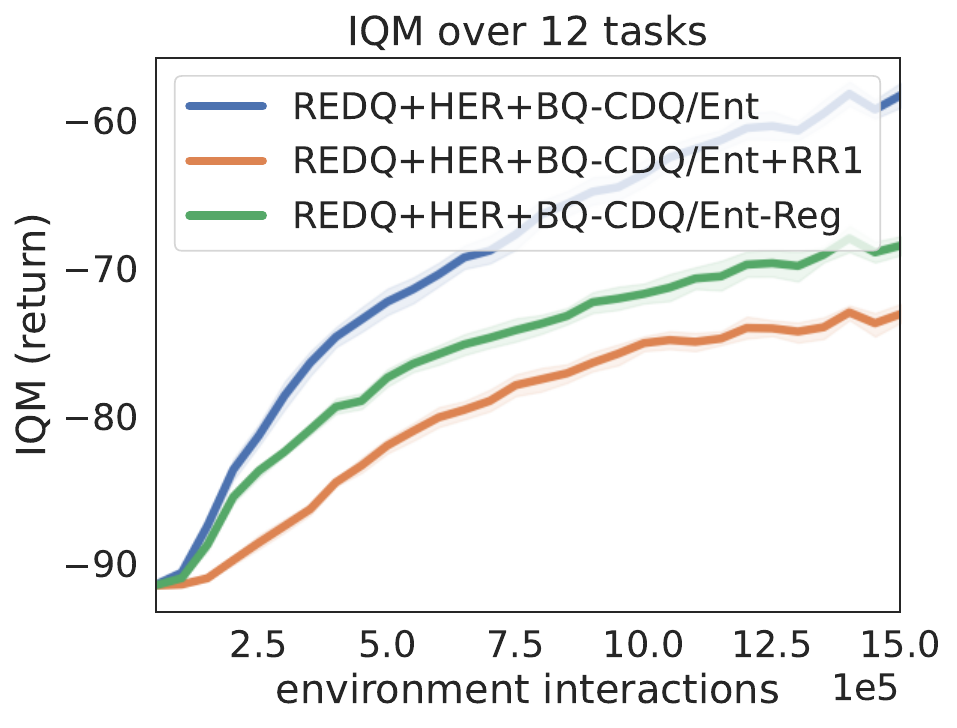}
\includegraphics[clip, width=0.49\hsize]{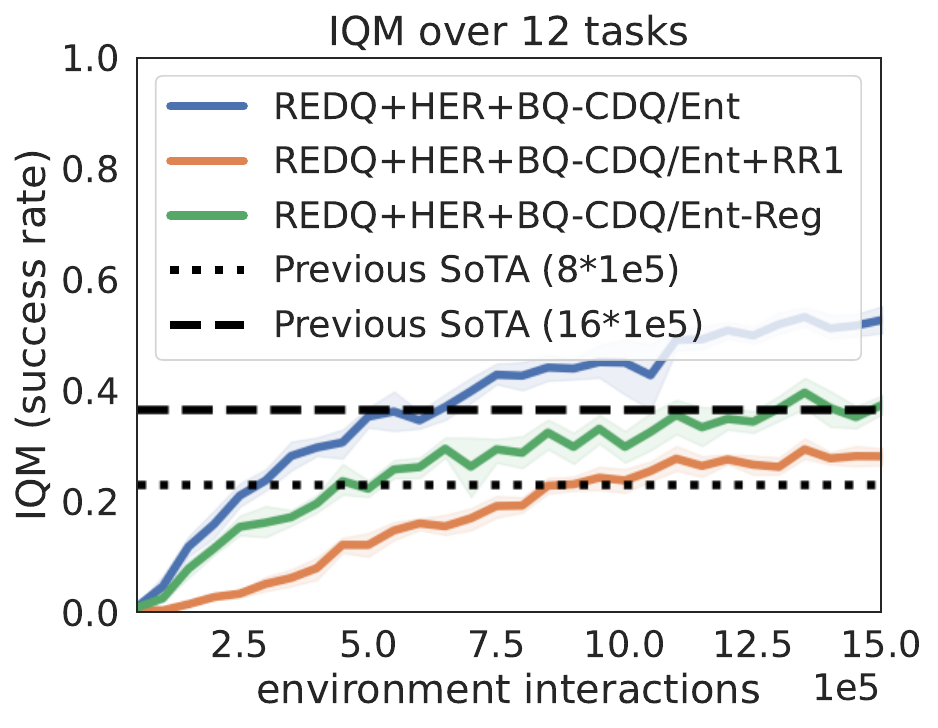}
\end{minipage}
\vspace{-0.7\baselineskip}
\caption{
The effect of removing a high RR and regularization on performance. 
The figure (IQM scores over 12 tasks) shows that both a high RR and regularization are necessary. 
The results for all tasks are shown in Figs.~\ref{fig:app-simplification2-return} and \ref{fig:app-simplification2-sr} in the appendix. 
}
\label{fig:simplification2-return-sr-summary}
\vspace{-0.5\baselineskip}
\end{figure}

\textbf{Are a high RR and regularization removable? No.} 
So far, we have introduced and reconsidered several design choices. 
Even after this consideration, are the core components of REDQ (i.e., a high RR and regularization) (Section~\ref{sec:base_method}) still necessary for our method? 
To answer this question, we evaluate two variants of REDQ+HER+BQ-CDQ/Ent that do not use a high RR and regularization:\\
1. REDQ+HER+BQ-CDQ/Ent+RR1: The method without a high RR. It uses a low RR of 1.\\ 
2. REDQ+HER+BQ-CDQ/Ent-Reg: The method without regularization. It uses a small ensemble (i.e., two Q-functions) and no layer normalization.\\
The evaluation results (Fig.~\ref{fig:simplification2-return-sr-summary}) show that both a high RR and regularization are still necessary for our method. 
We can see that REDQ+HER+BQ-CDQ/Ent achieves better sample efficiency than REDQ+HER+BQ-CDQ/Ent+RR1 and REDQ+HER+BQ-CDQ/Ent-Reg. 

\textbf{Can REDQ be replaced with a simpler method (Reset~\citep{pmlr-v162-nikishin22a})? No.} 
There are RL methods other than REDQ that have a high RR and regularization (see ``RL methods with a high RR and regularization'' in Section~\ref{sec:related_work}). 
Can we use these other methods, especially a simple one, instead of REDQ for our base RL method? 
To answer this, we compare our REDQ-based methods with methods based on Reset~\citep{pmlr-v162-nikishin22a}. 
Reset realizes regularization simply by periodically initializing the parameters of the agent's components (policy and Q-functions). 
Despite its simplicity, it performs equally to or better than REDQ in some dense-reward continuous-control tasks~\citep{d'oro2023sampleefficient}. We use four Reset-based methods for our comparison:\\ 
1. Reset([the number of resets]): Reset~\citep{pmlr-v162-nikishin22a} itself. ``[number of resets]'' means the total number of resets during training. In our experiments, we use Reset(1), Reset(4), and Reset(9). In addition, we use an RR of 20 as with our REDQ.\\
2. Reset([the number of resets])+HER: Reset with HER.\\
3. Reset([the number of resets])+BQ: Reset with BQ.\\
4. Reset([the number of resets])+HER+BQ: Reset with HER and BQ.\\
The algorithmic description of these Reset-based methods is summarized in Algorithm~\ref{alg1:ResetwithOurDesigneDecision} in the appendix. 
The comparison results (Fig.~\ref{fig:reset-sr-summary}) show that REDQ is more suitable for our base RL method than Reset \textit{in our setting}. 
We can see that REDQ+HER+BQ  performs better than other Reset-based methods.\\
\textbf{Complementary analysis: Does Reset also benefit from HER and BQ? Yes.} 
Interestingly, Reset also benefits from HER and BQ (Fig.~\ref{fig:reset-sr-summary}). 
We can see that Reset(1, 4, 9)+HER achieves better sample efficiency than Reset(1, 4, 9). 
In addition, Reset(1, 4, 9)+HER+BQ achieves the same or better sample efficiency than Reset(1, 4, 9)+HER. 
Especially, Reset(1)+HER+BQ significantly improves its sample efficiency compared with Reset(1)+HER. 
This may be because Reset(1)+HER has significant Q-estimation divergence (and BQ suppresses it) (Fig.~\ref{fig:method-reset-qvals}). 
We can see that Q-value estimates of Reset(1)+HER more significantly surpass the theoretical upper bound than Reset(4, 9)+HER, especially at the first-half stage of training ($\sim 7.5 \cdot 10^5$ samples). 
\begin{figure}[t!]
\begin{minipage}{1.0\hsize}
\includegraphics[clip, width=0.245\hsize]{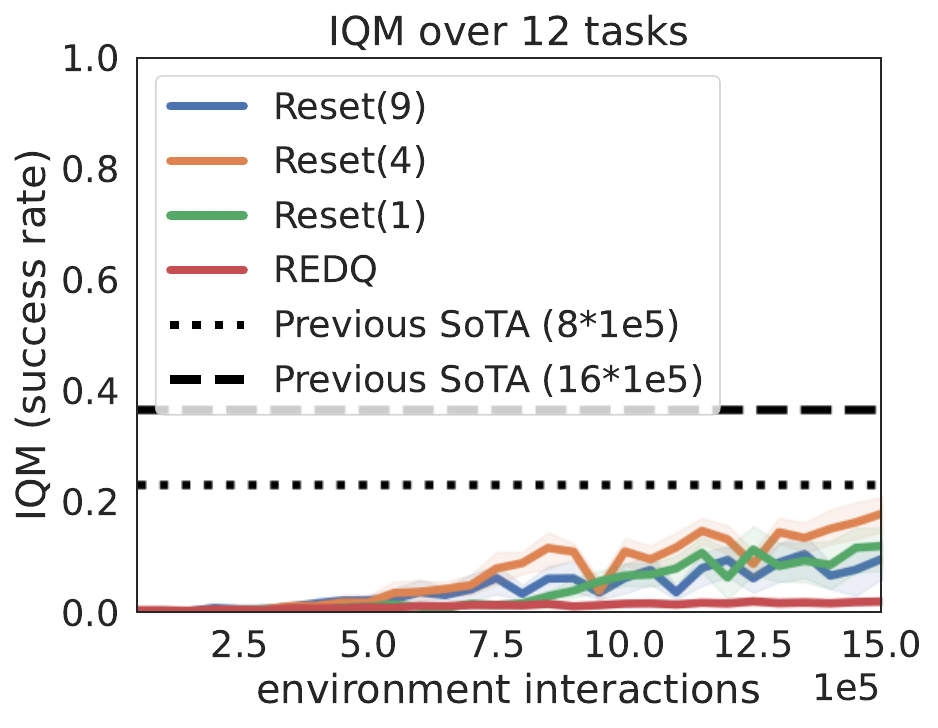}
\includegraphics[clip, width=0.245\hsize]{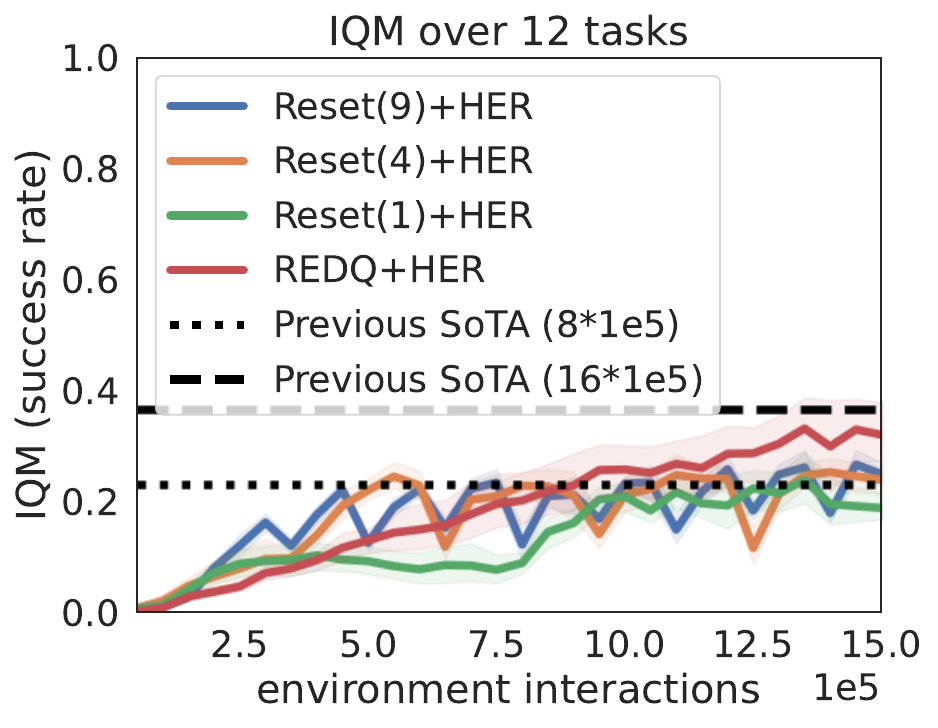}
\includegraphics[clip, width=0.245\hsize]{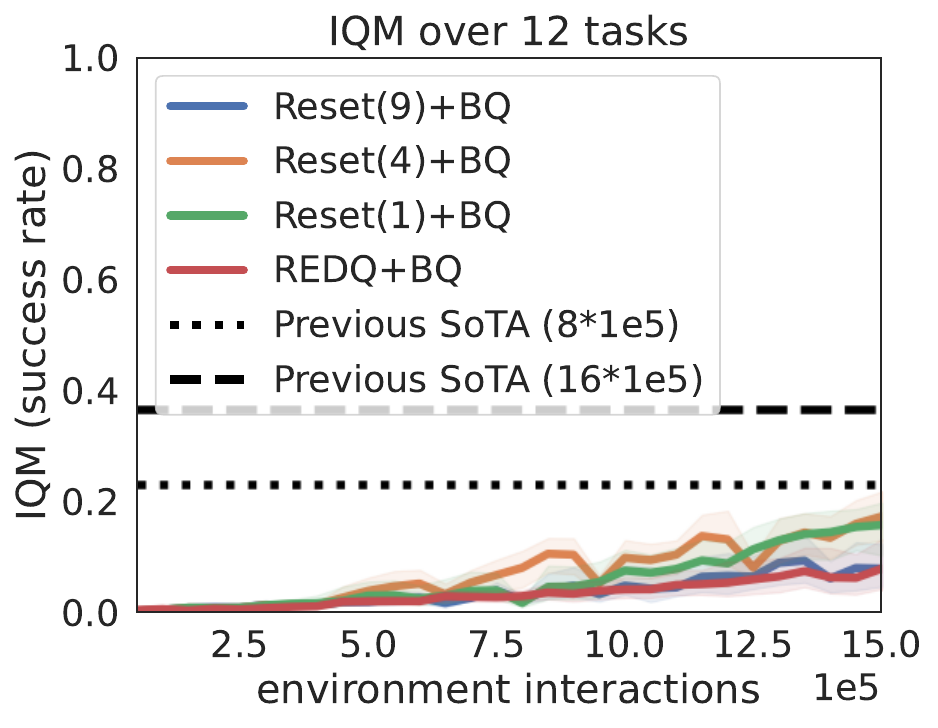}
\includegraphics[clip, width=0.245\hsize]{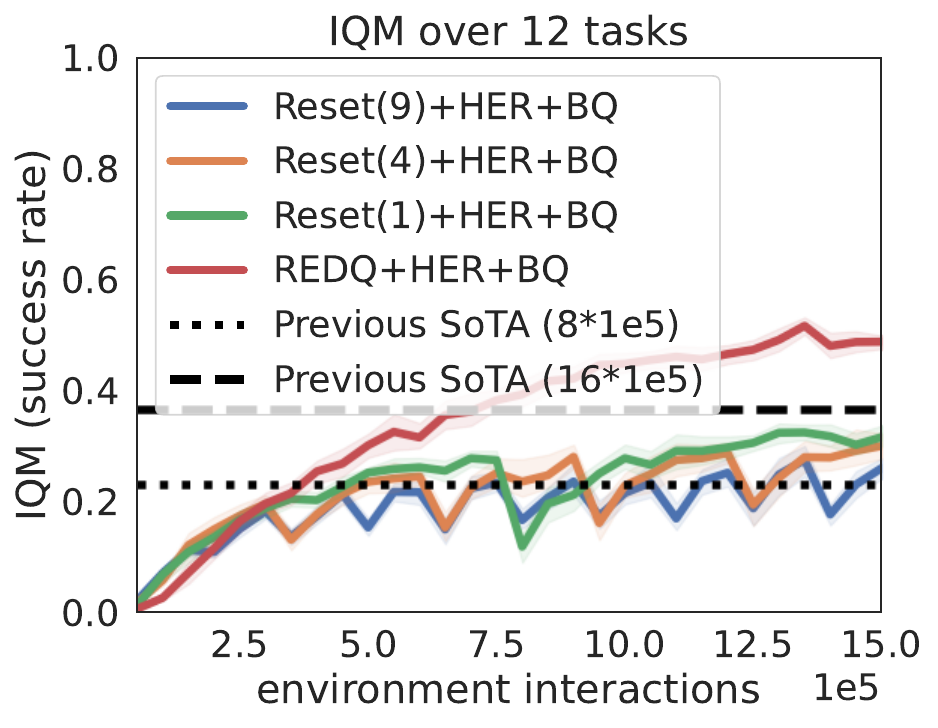}
\end{minipage}
\vspace{-0.7\baselineskip}
\caption{
The effect of replacing REDQ with Reset~\citep{pmlr-v162-nikishin22a} on performance (success rate). 
The figure shows that REDQ is more suitable for our base RL method than Reset. 
We can see that REDQ+HER+BQ performs better than other Reset-based methods. 
The results for all tasks are shown in Figs.~\ref{fig:app-reset-sr}, \ref{fig:app-reset-her-sr}, \ref{fig:app-reset-bq-sr}, and \ref{fig:app-reset-her-bq-sr} in the appendix. 
}
\label{fig:reset-sr-summary}
\vspace{-0.5\baselineskip}
\end{figure}
\begin{figure}[t!]
\begin{minipage}{1.0\hsize}
\includegraphics[clip, width=0.245\hsize]{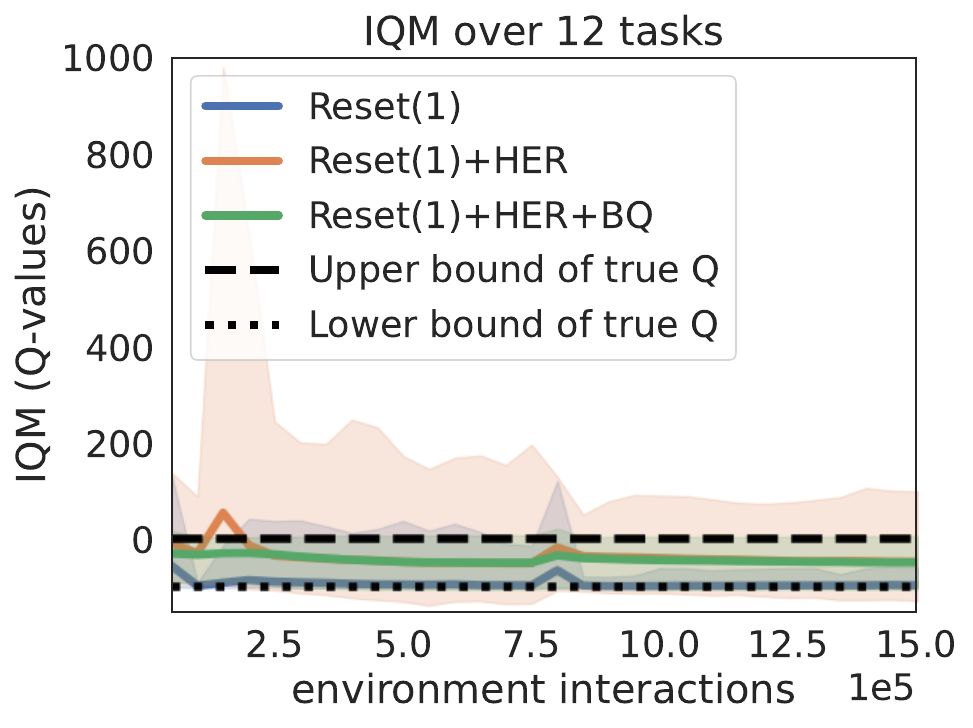}
\includegraphics[clip, width=0.245\hsize]{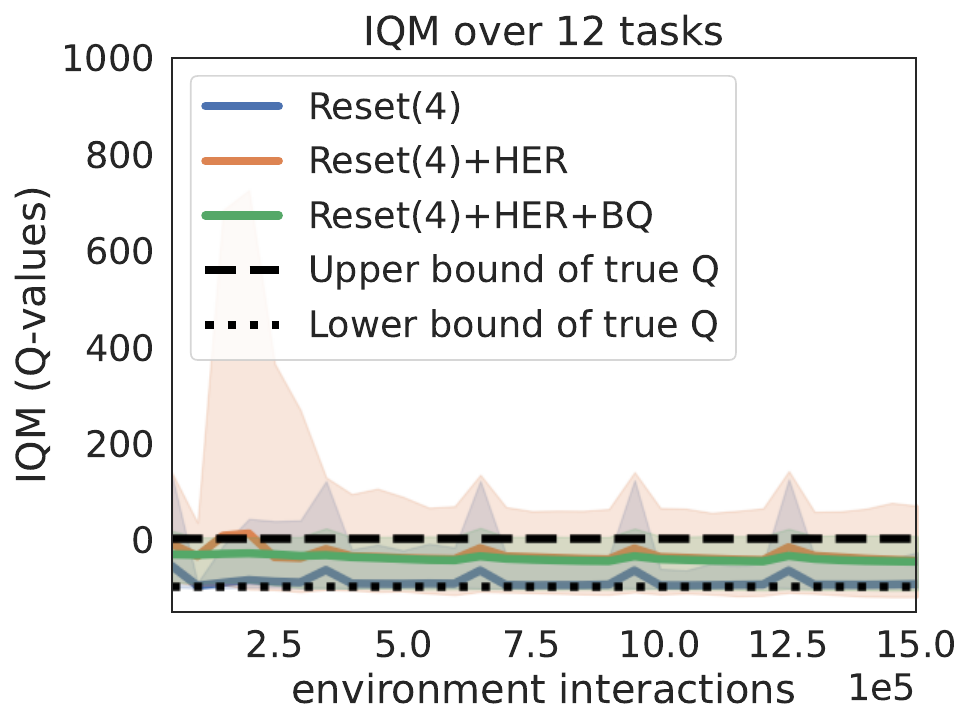}
\centering
\includegraphics[clip, width=0.245\hsize]{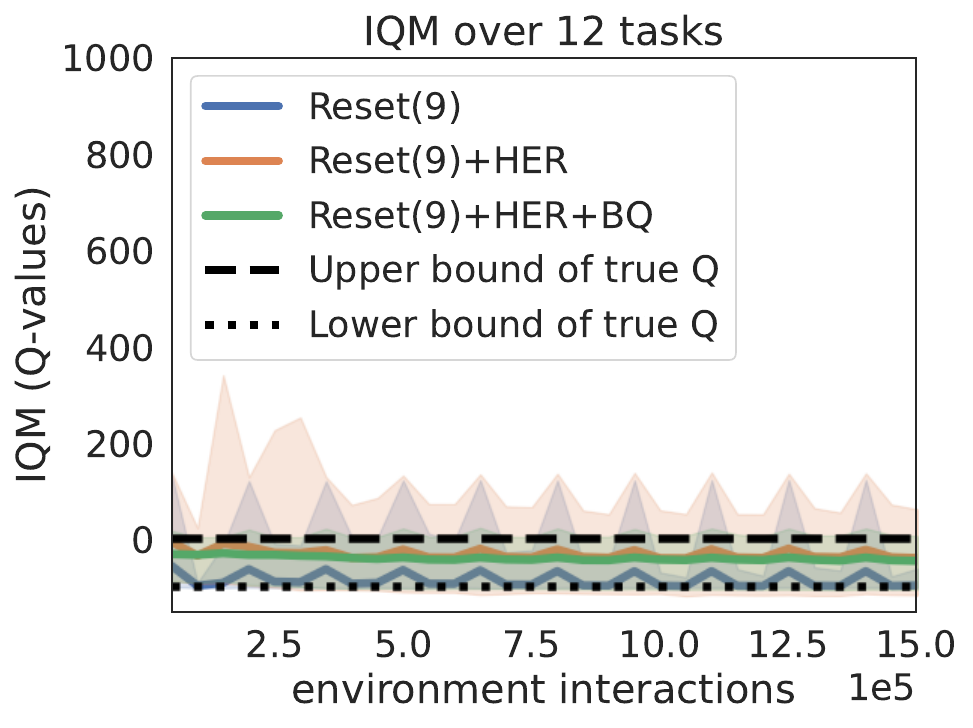}
\end{minipage}
\vspace{-0.7\baselineskip}
\caption{
The effect of replacing REDQ with Reset on Q-value divergence. 
The figures show that HER causes Q-estimation divergence more significantly for the Reset method with a smaller reset number. 
We can see that Q-estimates of Reset(1)+HER exceed the bound range more significantly than Reset(4, 9)+HER. 
The results for all tasks are shown in Figs.~\ref{fig:app-reset1-qvals}, \ref{fig:app-reset4-qvals}, and \ref{fig:app-reset9-qvals} in the appendix.
}
\label{fig:method-reset-qvals}
\vspace{-0.5\baselineskip}
\end{figure}

\section{Related Works}\label{sec:related_work}
\textbf{RL methods with a high RR and regularization.} 
We have applied RL methods with a high RR and regularization to sparse-reward tasks (Sections~\ref{sec:method} and \ref{sec:experiment}). 
Most previous works on RL methods with a high RR and regularization have focused primarily on dense-reward tasks~\citep{janner2019trust,kumar2021implicit,chen2021randomized,hiraoka2022dropout,smith2022walk,pmlr-v162-nikishin22a,li2022efficient,d'oro2023sampleefficient,smith2023learning,sokar2023dormant,pmlr-v202-schwarzer23a,lee2023plastic}. 
Some works~\citep{vecerik2017leveraging,sharma2023self,ball2023efficient,nakamoto2023cal,li2023accelerating} have considered sparse-reward tasks, but they assume situations where prior data (e.g., expert demonstrations) are available. 
On the other hand, we assume sparse-reward tasks where such prior data are unavailable. 
Our work is orthogonal to the above previous works, and some of our modifications may be useful in them. 
For example, we bound the target Q-value to deal with the instability of Q-value estimation (Section~\ref{sec:boundq}), and a similar instability problem also appears in the above works (see Fig. 2 in \citet{ball2023efficient} for example). 

\textbf{Sparse-reward goal-conditioned RL.} 
There are previous works on sparse-reward goal-conditioned RL. 
These works have used a deep deterministic policy gradient~\citep{lillicrap2015continuous} (or soft actor-critic~\citep{haarnoja2018soft}) with HER~\citep{andrychowicz2017hindsight} as a base RL method and improved its sampling prioritization scheme~\citep{zhao2018energy,zhao2019maximum, zhao2019curiosity,dai2021diversity,xu2023efficient} and new-goal-selection strategies~\citep{meng2019cher,pitis2020maximum,ren2019exploration,chanesane2021goal,luo2022relay}. 
In these works, the base RL method uses low RR ($\leq 1$) and no regularization, which is contrary to our base RL method (Section~\ref{sec:base_method}). 
We showed that the RL method with a high RR and regularization can achieve a better sample efficiency than methods with a low RR and no regularization (the right-hand side figure in Figs.~\ref{fig:experiment-return-sr-summary} in Section~\ref{sec:experiment}). 

\textbf{Bounding Q-value.} 
We bounded the target Q-value (Section~\ref{sec:boundq}). 
Previous works have proposed to bound Q-values in online RL settings~\citep{blundell2016model,he2017learning,oh2018self,lin2018episodic,tang2020self,fujita2020distributed,hoppe2020qgraph,zhao2023faster,fujimoto2023sale}. 
These previous works have verified a positive effect of the bounding on RL methods with a low RR and no regularization in dense-reward tasks. 
In contrast, our work has verified a positive effect of the bounding on the RL method with a high RR and regularization in sparse-reward tasks. 
The technical modification that improves certain RL methods in certain tasks could impair other methods in other tasks~\citep{NEURIPS2021_517f24c0}. 
Thus, our verification would be valuable in clarifying the generality of a positive effect of Q-values-bound techniques. 

\section{Conclusion, Limitations, and Future Works}\label{sec:conclusion}
\textbf{Conclusion.} We applied a reinforcement learning (RL) method (Randomized Ensemble Double Q-learning; REDQ) with a high replay ratio (RR) and regularization to sparse-reward goal-conditioned tasks. 
We introduced hindsight experience replay (HER) and bounding target Q-value (BQ) to REDQ and showed that REDQ with them achieves about $2 \times$ better sample efficiency than previous state-of-the-art (SoTA) methods in 12 Robotics tasks. 
We also showed that REDQ with HER and BQ can be simplified by removing clipped double Q-learning (CDQ) and entropy terms. 
The simplified REDQ with our modifications achieved $\sim 8 \times$ better sample efficiency than the SoTA methods in the 4 Fetch tasks of Robotics. 
We hope that these findings will push the boundaries of the application of RL methods with a high RR and regularization from dense-reward tasks to sparse-reward tasks. 

\textbf{Limitations and future work.} 
Our work leaves limitations and future work. 
First, our RL method did not significantly improve the sampling efficiency in extremely hard tasks (e.g., HandManipulateBlockFull). 
Improving the efficiency in these tasks is an interesting future work. 
Second, our experiments are conducted in simulated environments, not real ones. 
Our primary interest lies more in investigating decision choices for RL methods rather than in demonstration in real environments. 
Nevertheless, demonstration in real environments would be one of the natural future steps for our work.

\bibliographystyle{tmlr}
\bibliography{icml2023}
\clearpage
\onecolumn
\appendix

\section{Detailed Experimental Results}

\subsection{Modification 2: Bounding Target Q-Value}
\begin{figure*}[h!]
\begin{minipage}{1.0\hsize}
\includegraphics[clip, width=0.24\hsize]{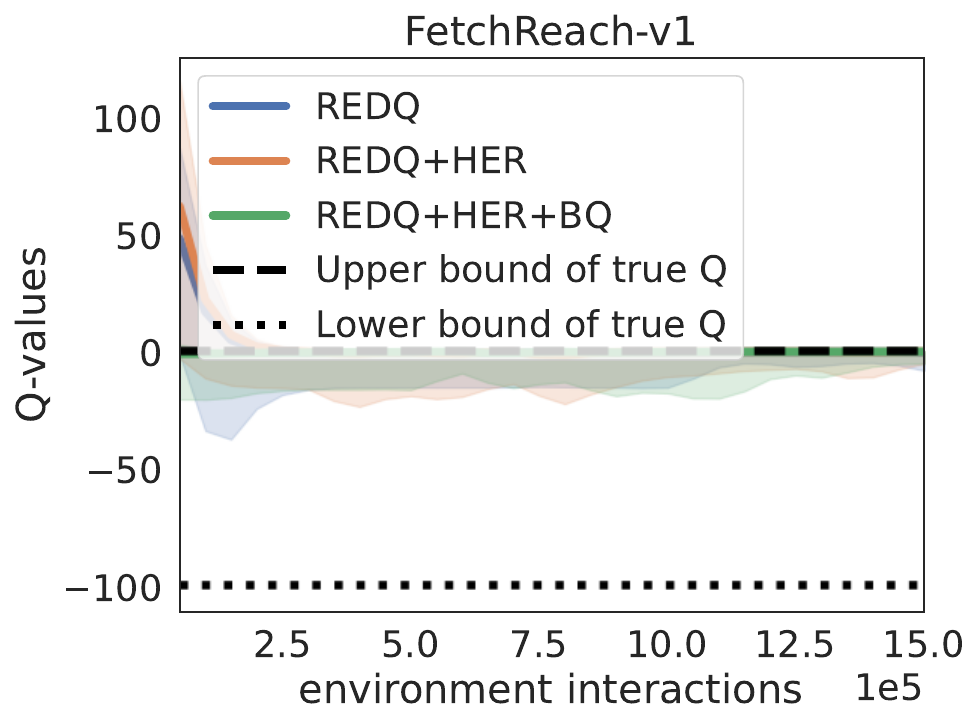}
\includegraphics[clip, width=0.24\hsize]{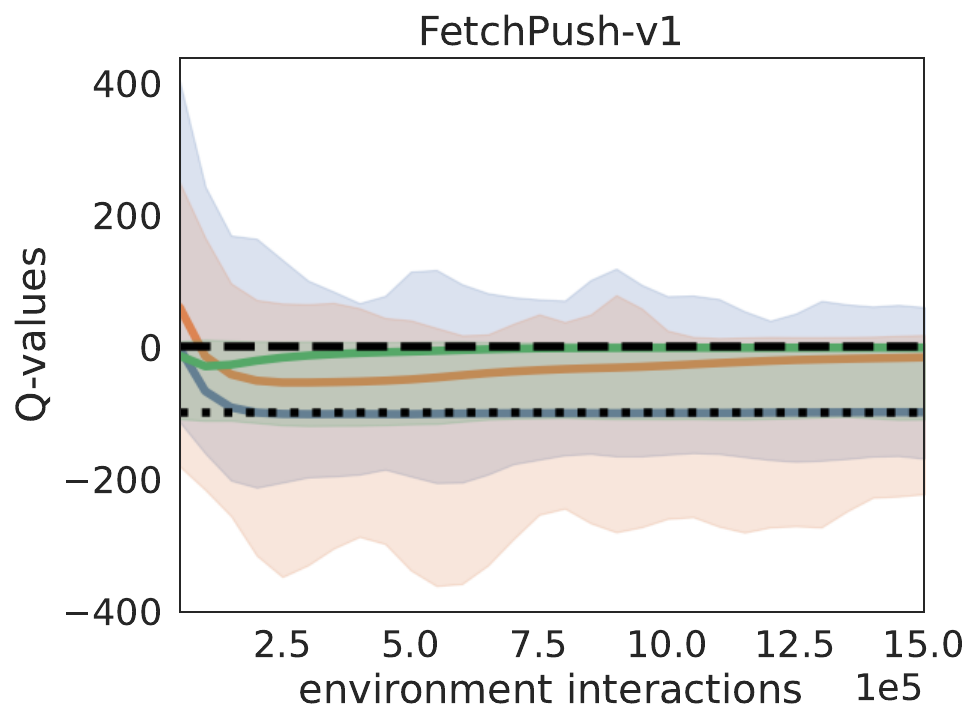}
\includegraphics[clip, width=0.24\hsize]{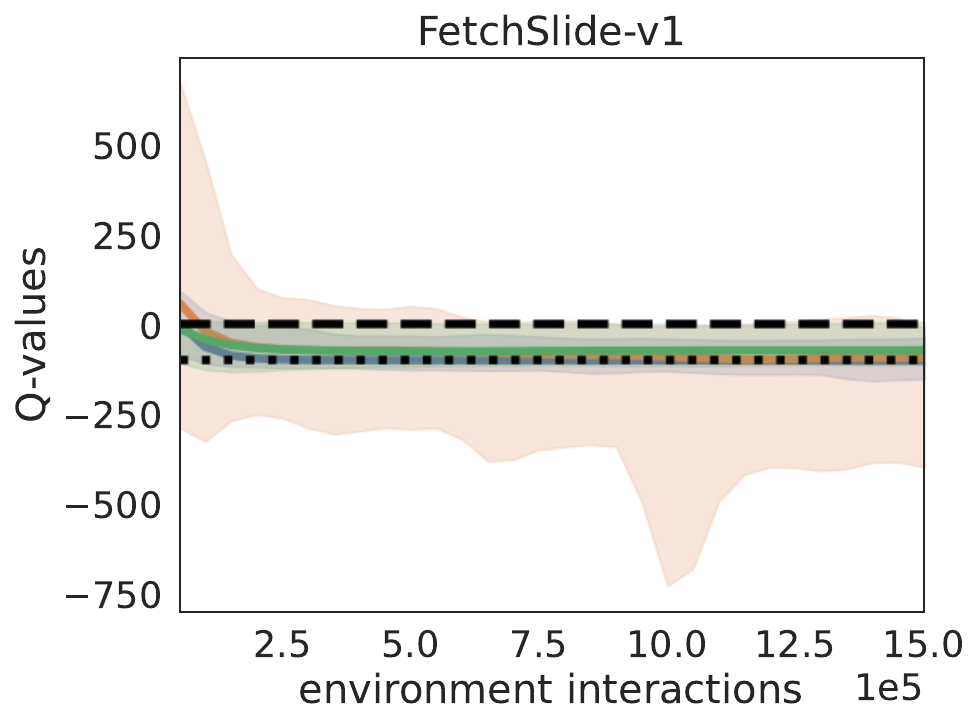}
\includegraphics[clip, width=0.24\hsize]{Figures/Method-BoundQ/FetchPickAndPlace-v1_0.pdf}
\end{minipage}
\begin{minipage}{1.0\hsize}
\includegraphics[clip, width=0.24\hsize]{Figures/Method-BoundQ/HandManipulatePenRotate-v0_0.pdf}
\includegraphics[clip, width=0.24\hsize]{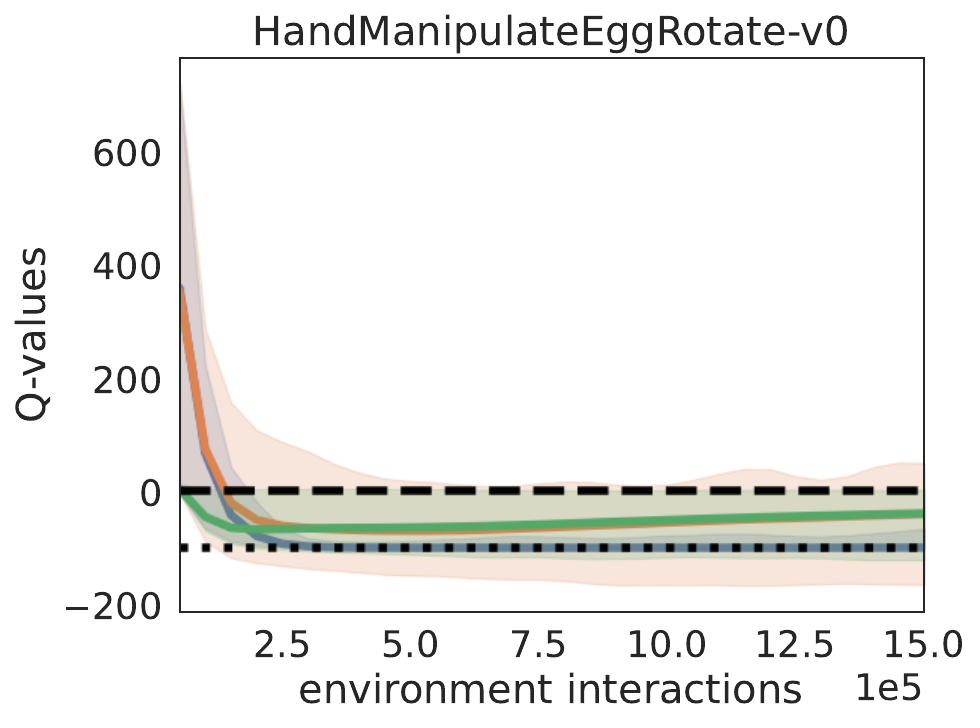}
\includegraphics[clip, width=0.24\hsize]{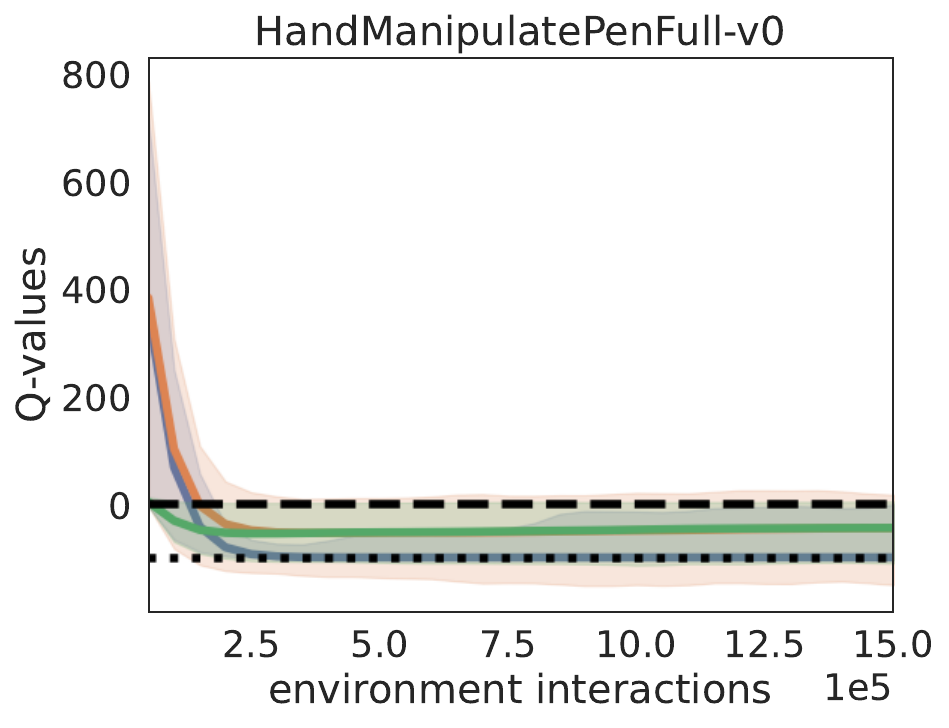}
\includegraphics[clip, width=0.24\hsize]{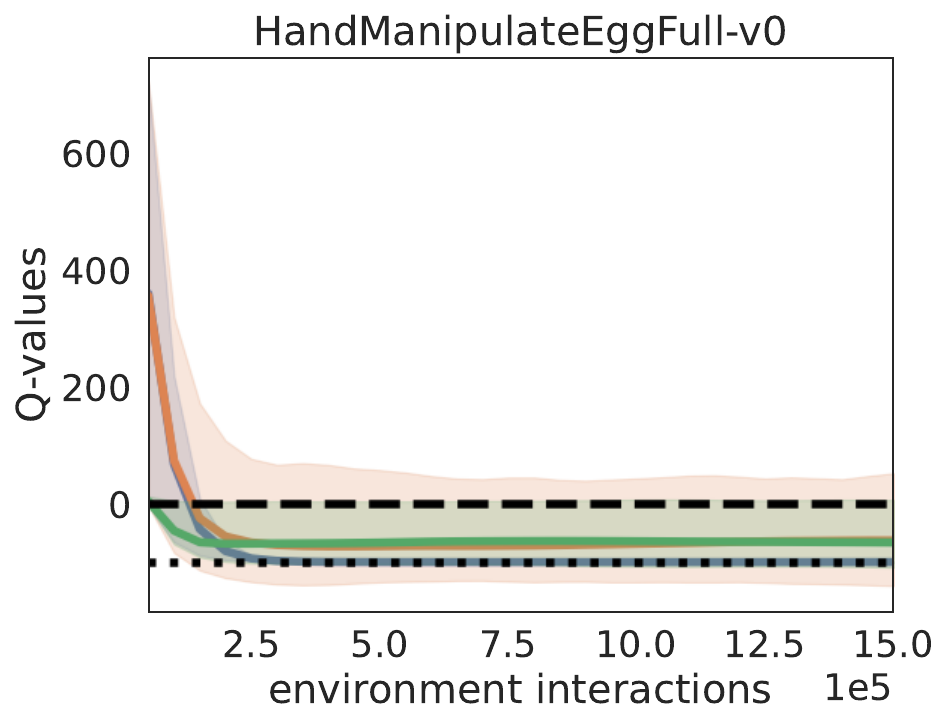}
\end{minipage}
\begin{minipage}{1.0\hsize}
\includegraphics[clip, width=0.24\hsize]{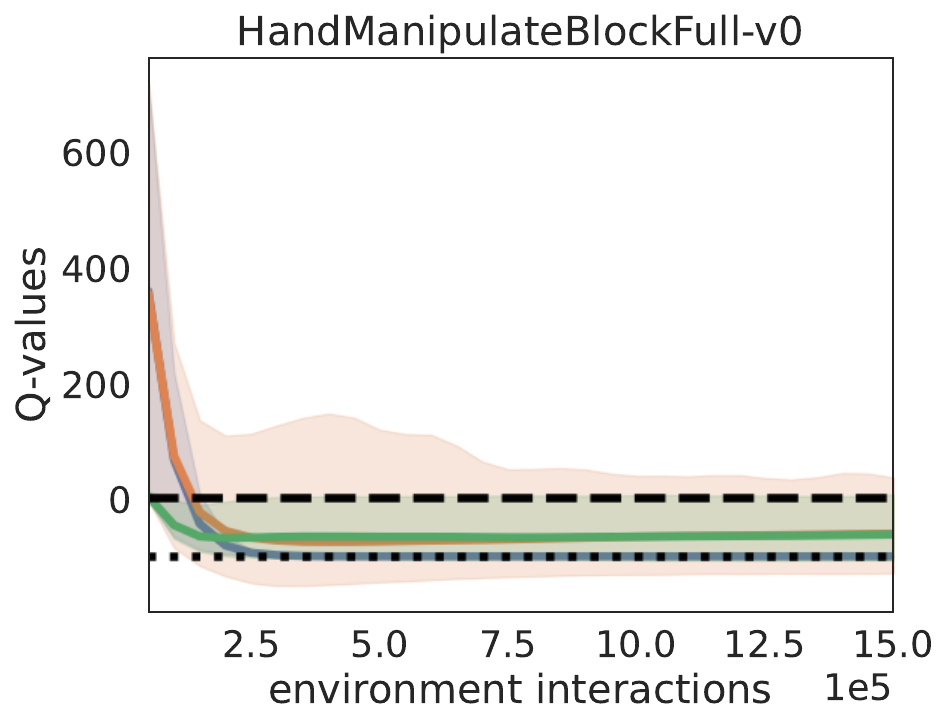}
\includegraphics[clip, width=0.24\hsize]{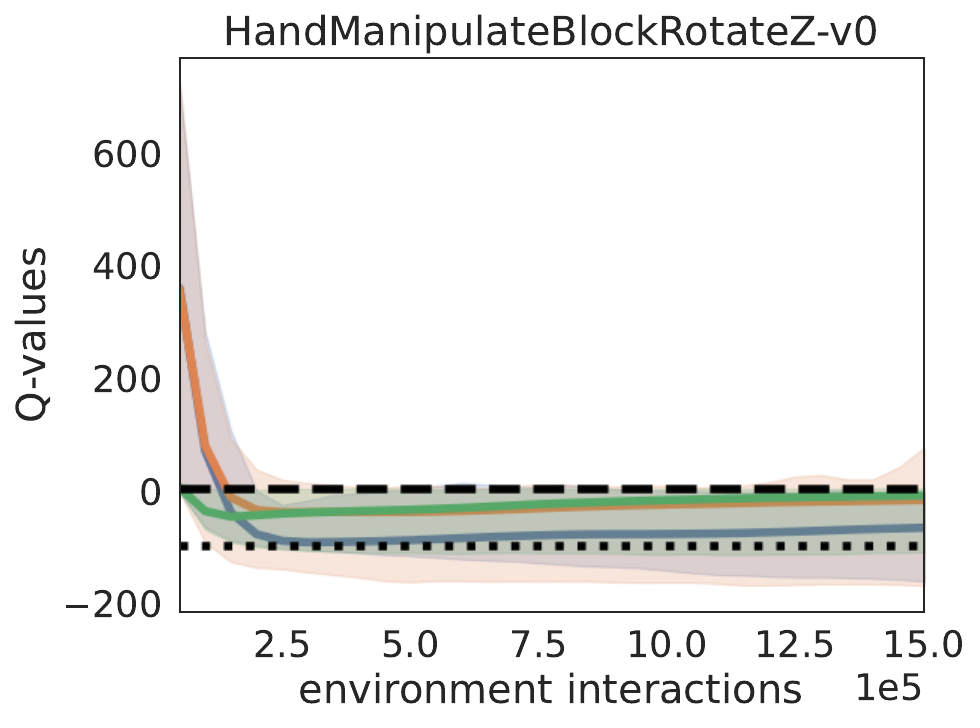}
\includegraphics[clip, width=0.24\hsize]{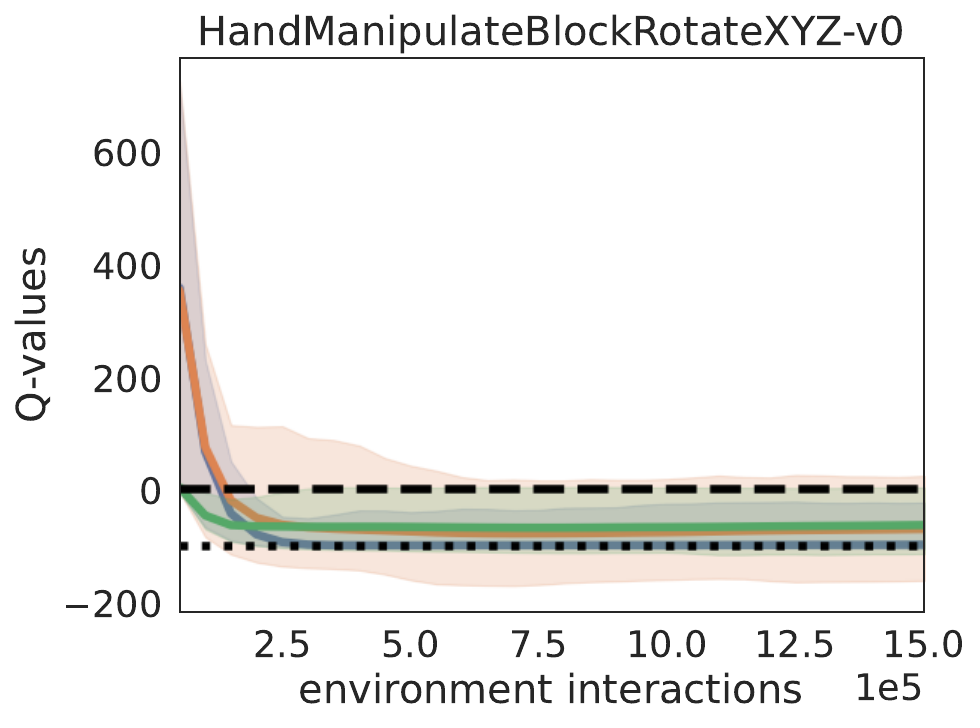}
\includegraphics[clip, width=0.24\hsize]{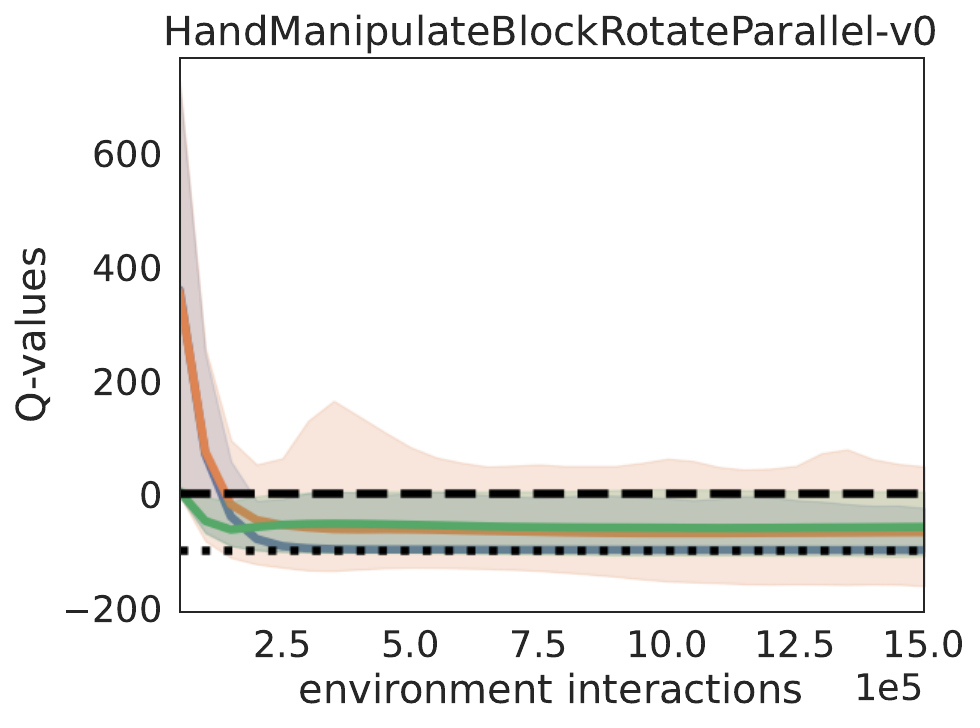}
\end{minipage}
\vspace{-0.7\baselineskip}
\caption{
The effect of BQ on Q-value divergence. 
}
\label{fig:app-method-boundq-qvals}
\vspace{-0.5\baselineskip}
\end{figure*}

\clearpage
\subsection{Simplifying Our Method (REDQ+HER+BQ)}
\begin{figure*}[h!]
\begin{minipage}{1.0\hsize}
\includegraphics[clip, width=0.24\hsize]{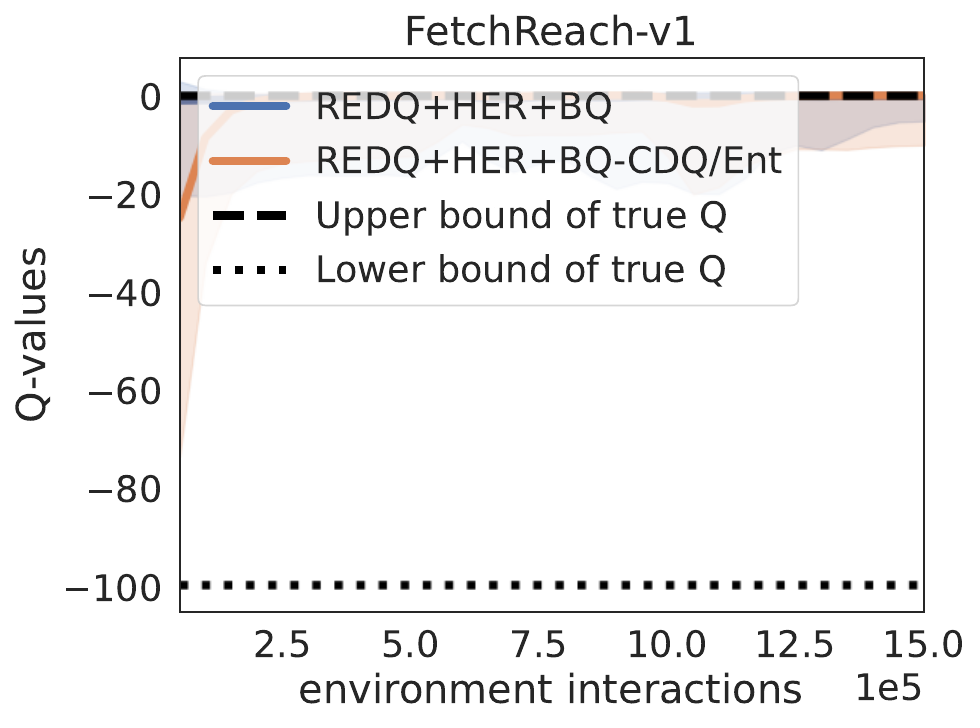}
\includegraphics[clip, width=0.24\hsize]{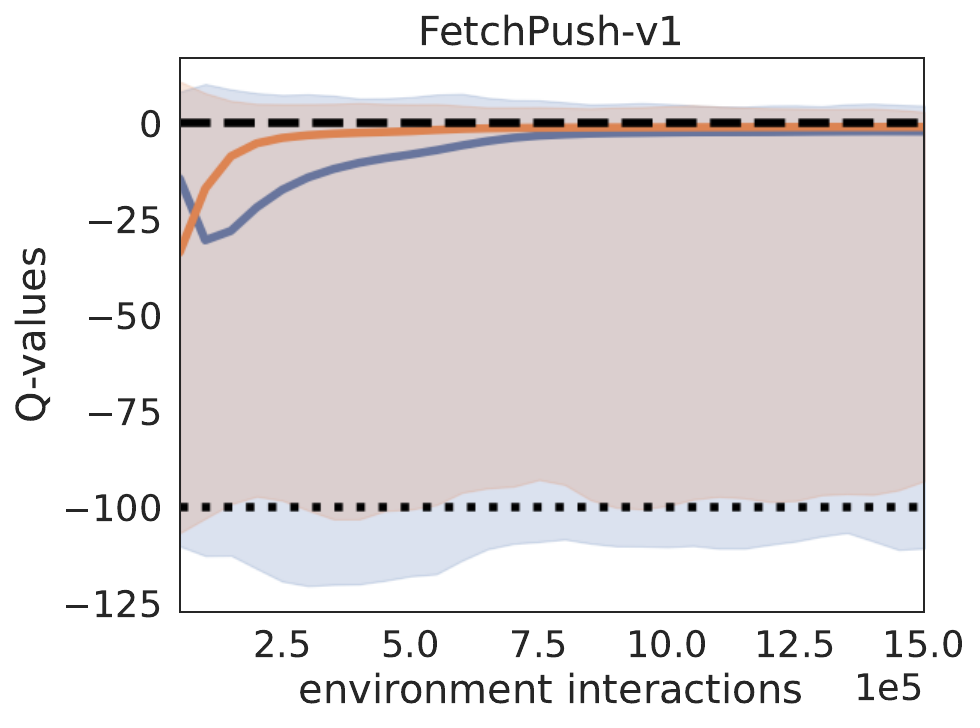}
\includegraphics[clip, width=0.24\hsize]{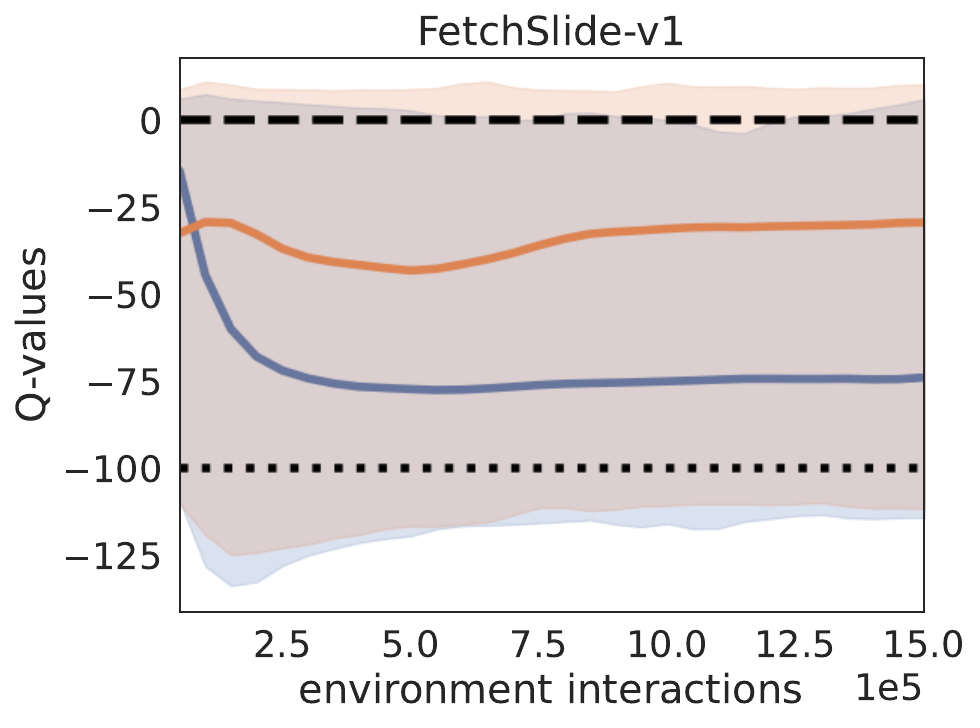}
\includegraphics[clip, width=0.24\hsize]{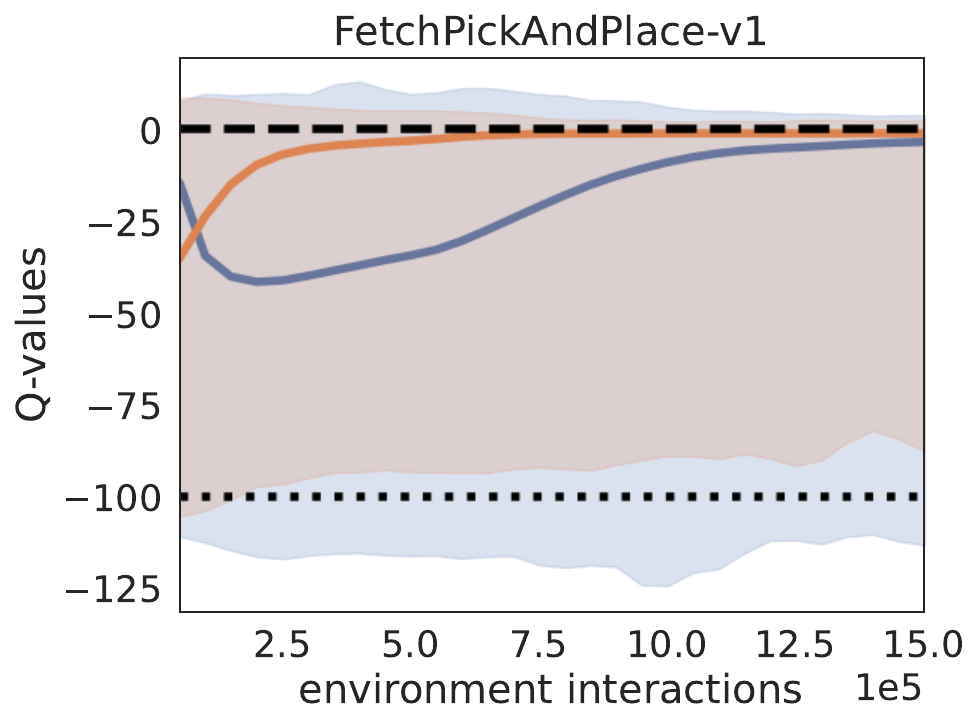}
\end{minipage}
\begin{minipage}{1.0\hsize}
\includegraphics[clip, width=0.24\hsize]{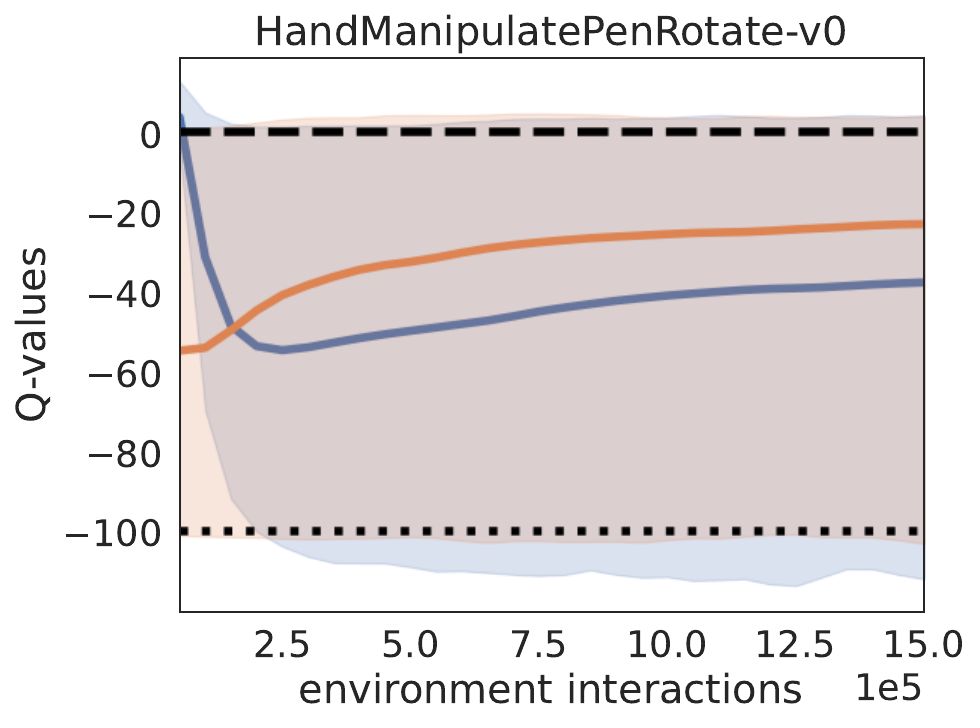}
\includegraphics[clip, width=0.24\hsize]{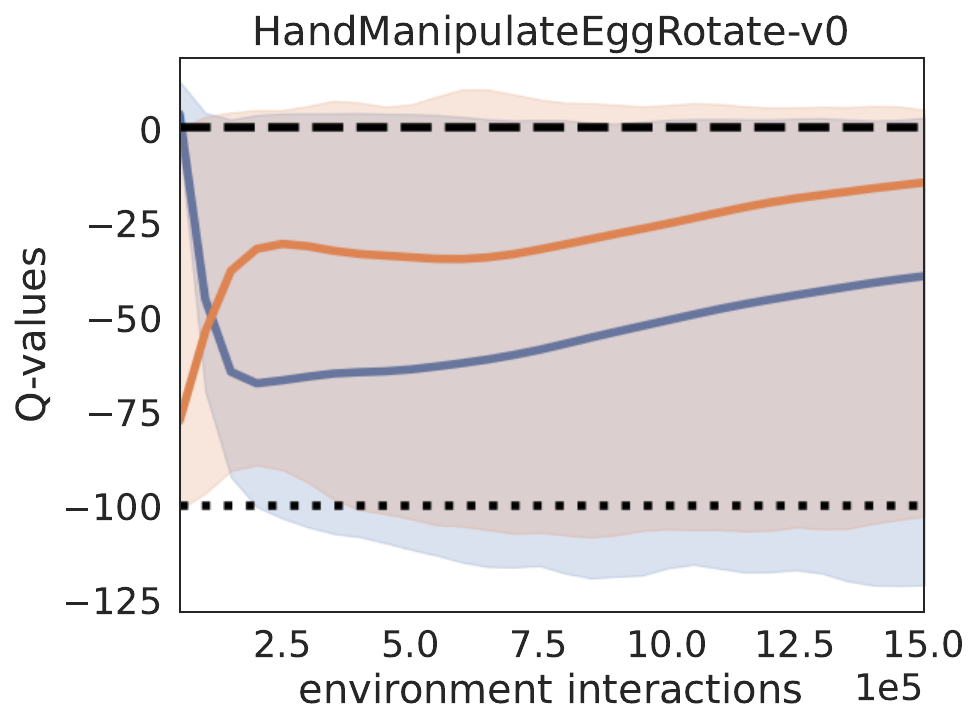}
\includegraphics[clip, width=0.24\hsize]{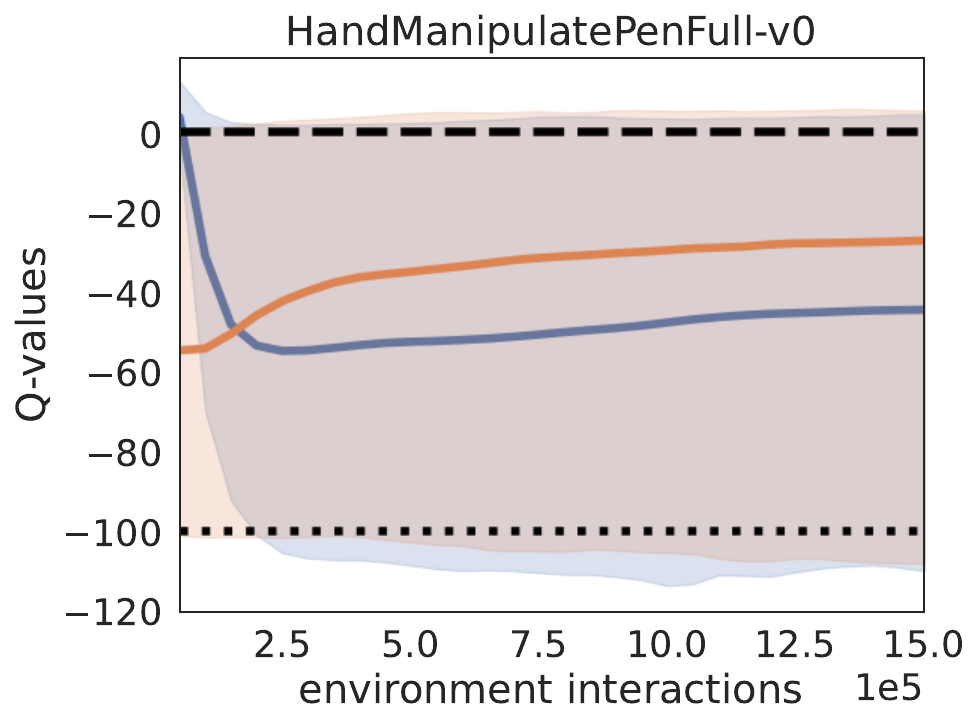}
\includegraphics[clip, width=0.24\hsize]{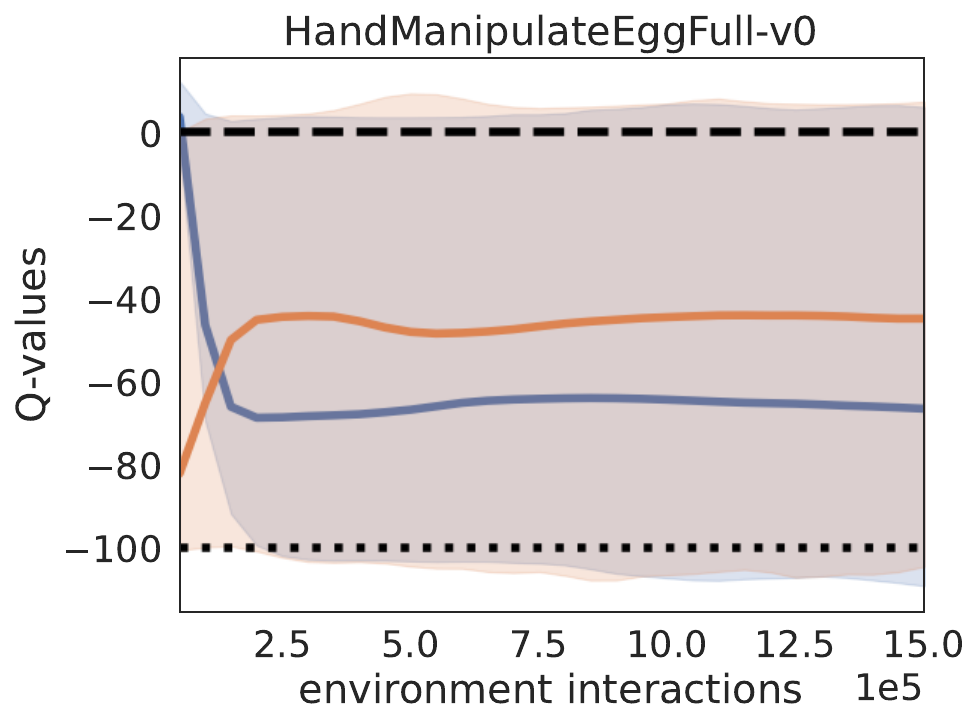}
\end{minipage}
\begin{minipage}{1.0\hsize}
\includegraphics[clip, width=0.24\hsize]{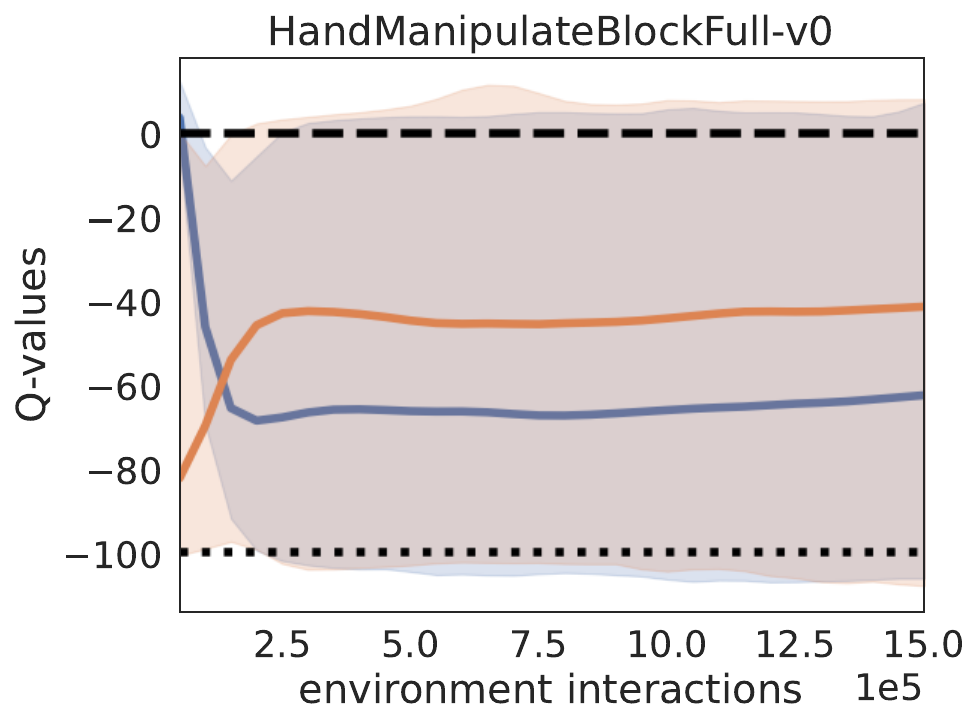}
\includegraphics[clip, width=0.24\hsize]{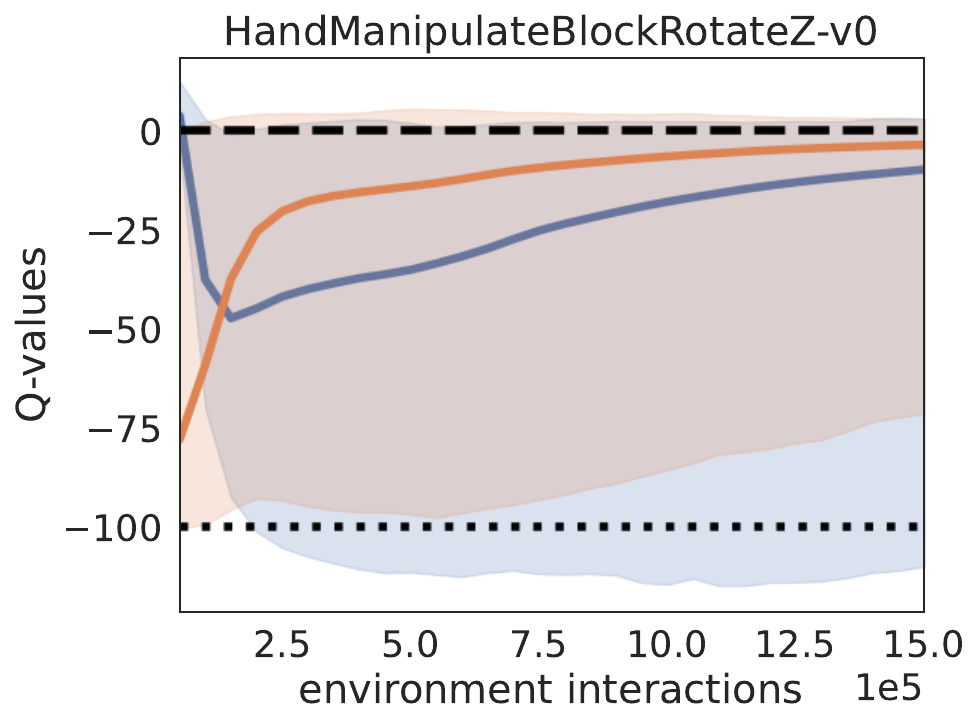}
\includegraphics[clip, width=0.24\hsize]{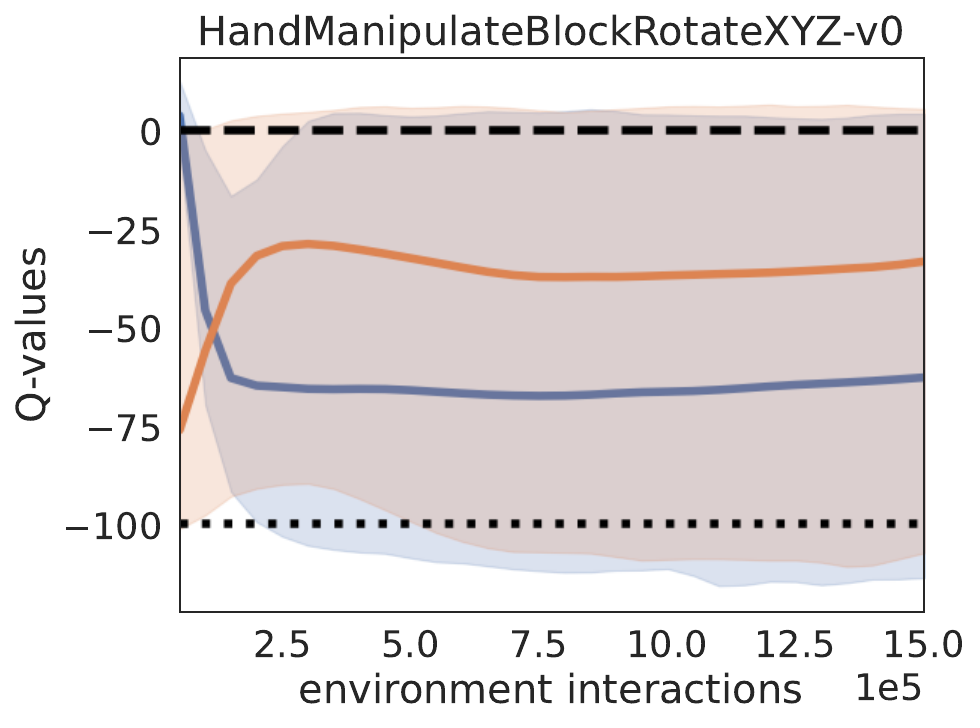}
\includegraphics[clip, width=0.24\hsize]{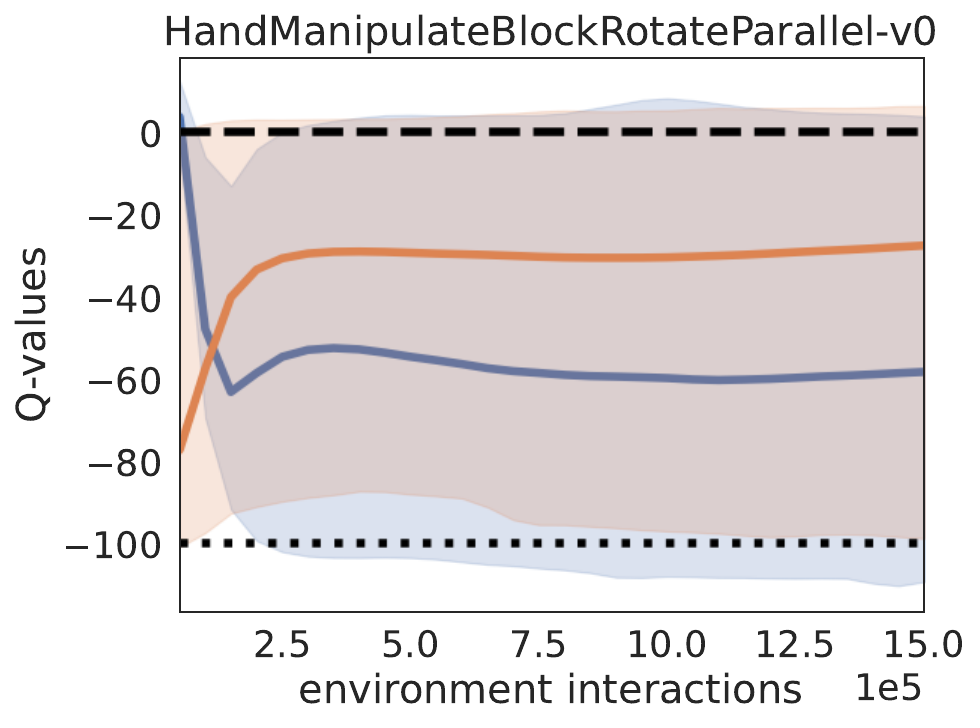}
\end{minipage}
\vspace{-0.7\baselineskip}
\caption{
The effect of removing CDQ and entropy term on Q-value divergence. 
}
\label{fig:app-simplication-qvals}
\vspace{-0.5\baselineskip}
\end{figure*}

\begin{figure}[h!]
\begin{minipage}{1.0\hsize}
\includegraphics[clip, width=0.24\hsize]{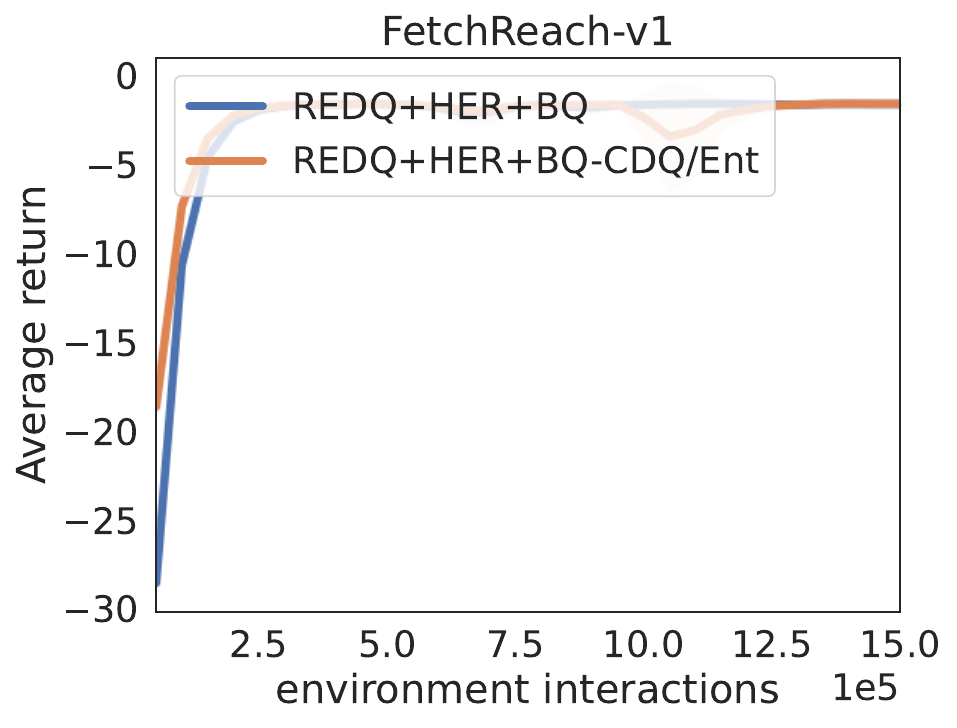}
\includegraphics[clip, width=0.24\hsize]{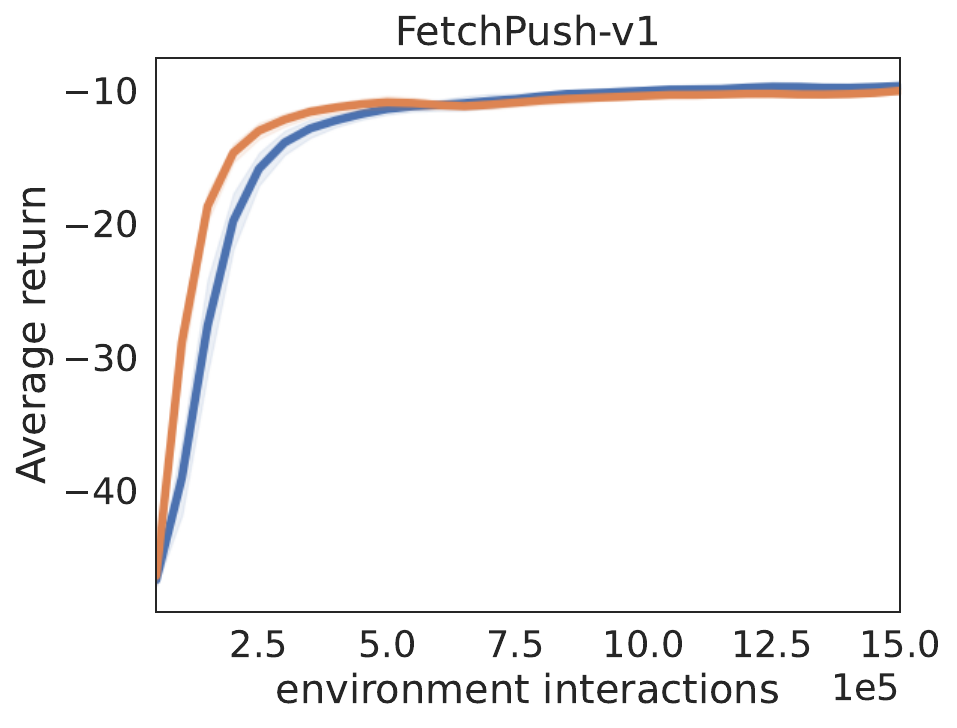}
\includegraphics[clip, width=0.24\hsize]{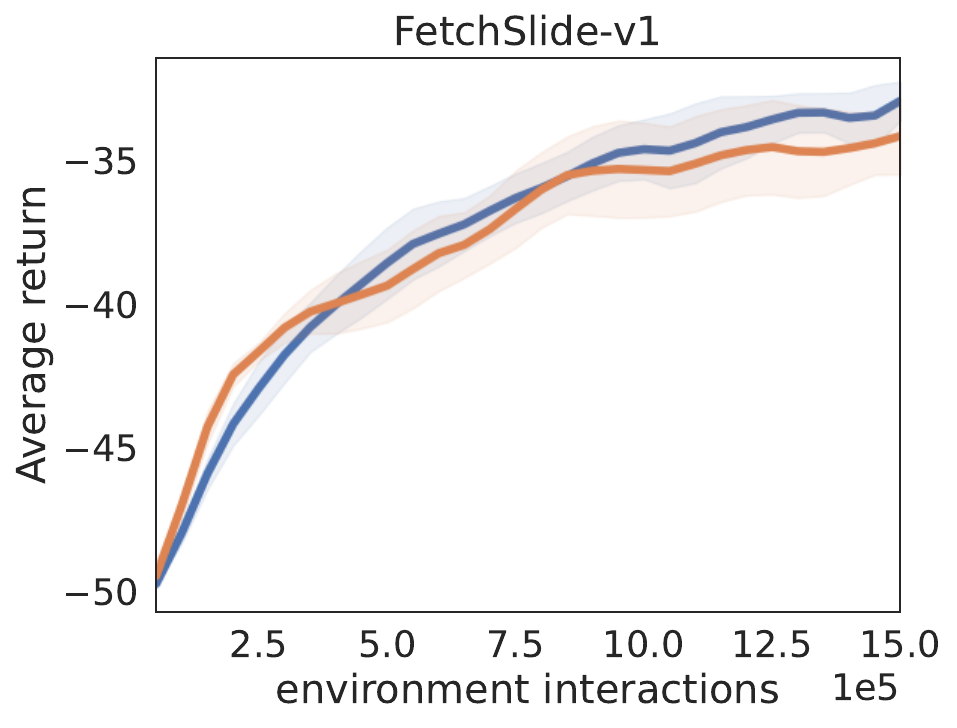}
\includegraphics[clip, width=0.24\hsize]{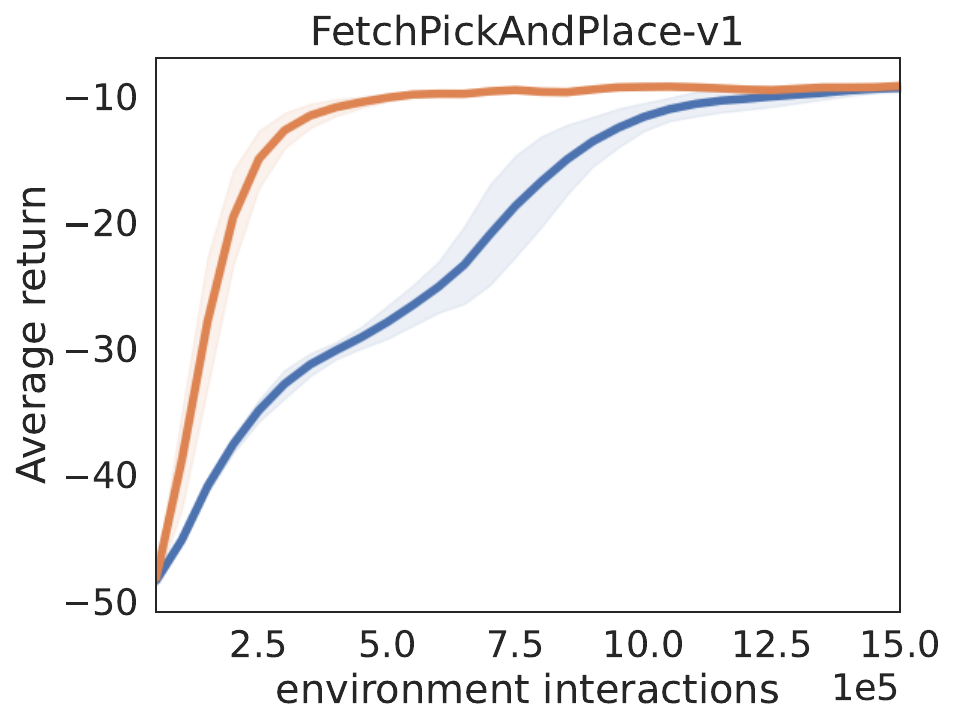}
\end{minipage}
\begin{minipage}{1.0\hsize}
\includegraphics[clip, width=0.24\hsize]{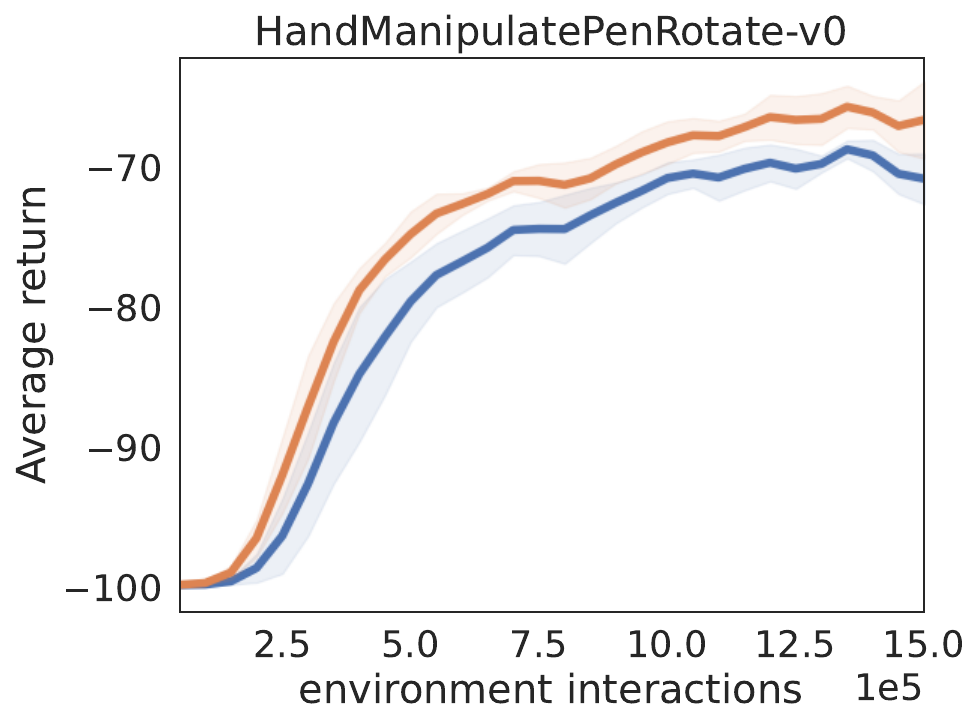}
\includegraphics[clip, width=0.24\hsize]{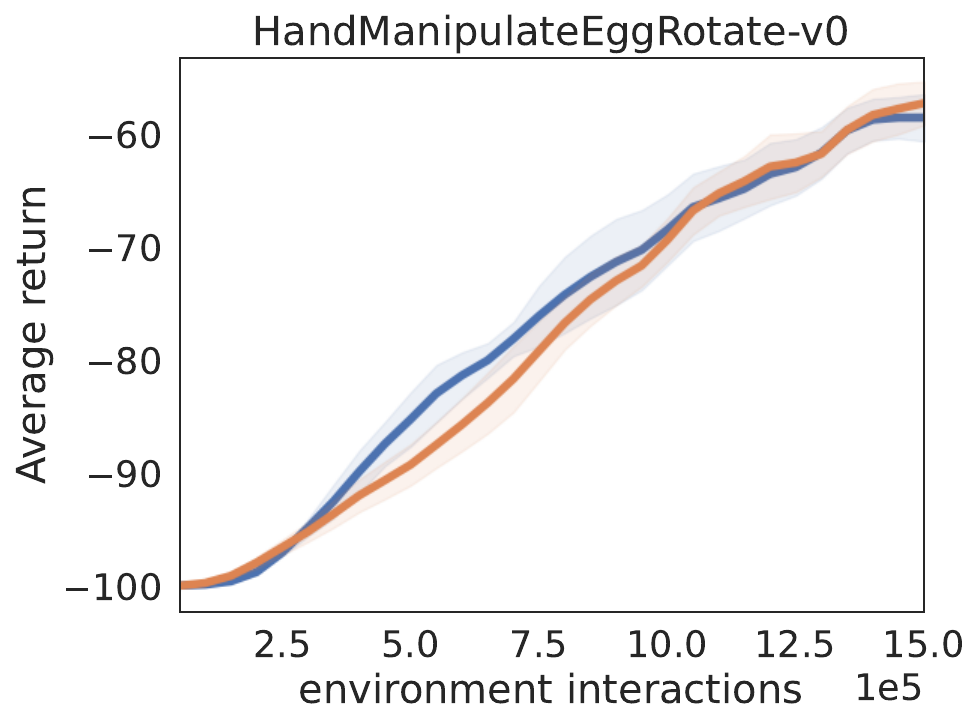}
\includegraphics[clip, width=0.24\hsize]{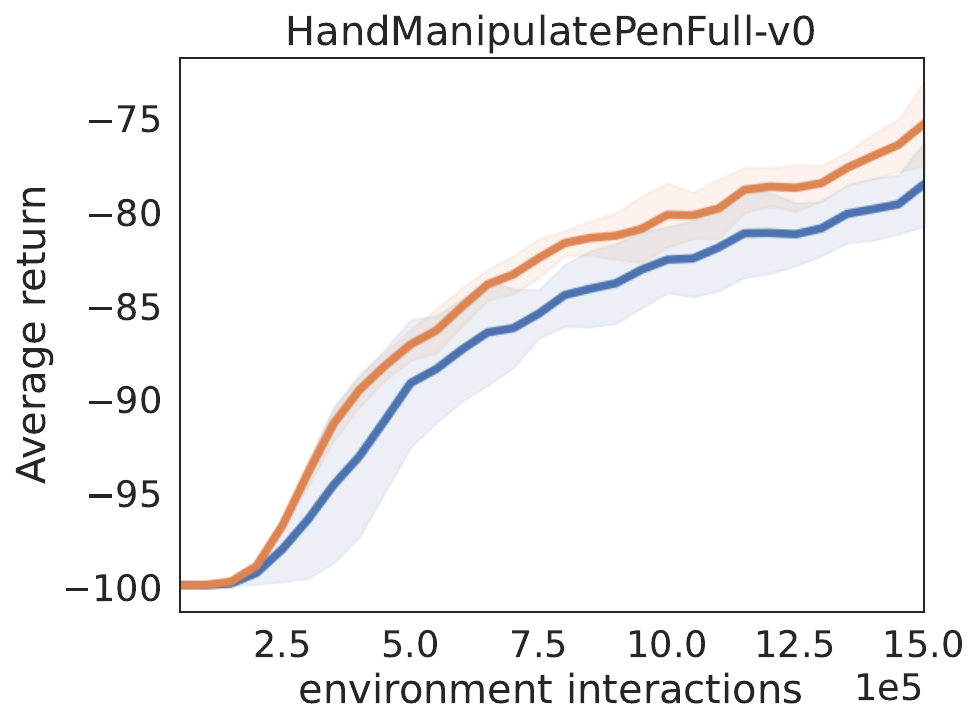}
\includegraphics[clip, width=0.24\hsize]{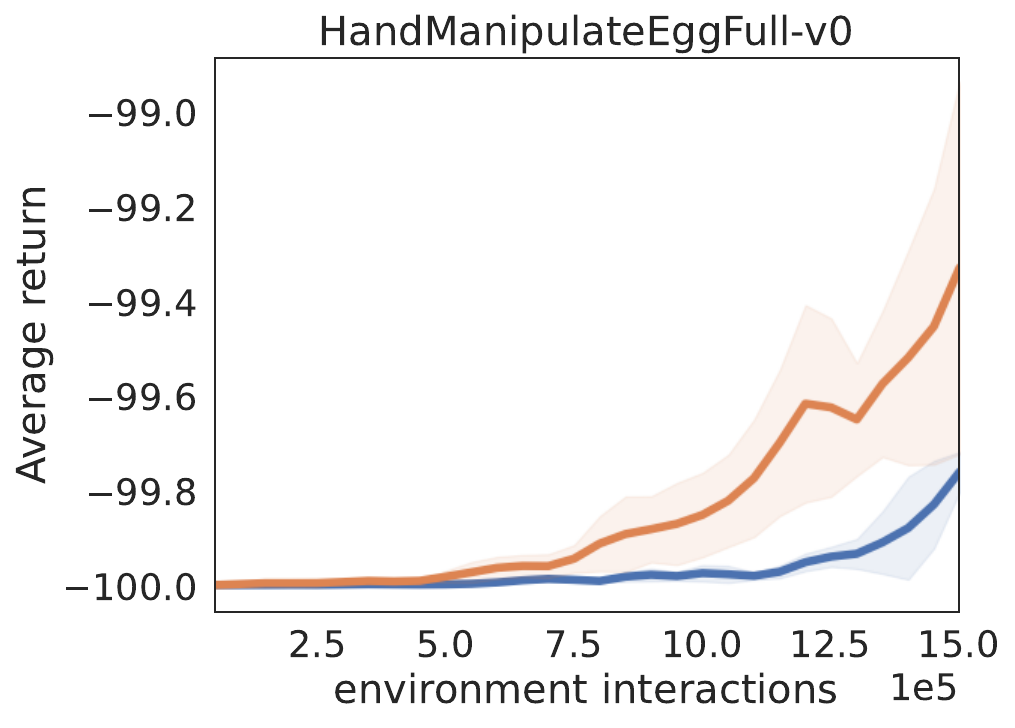}
\end{minipage}
\begin{minipage}{1.0\hsize}
\includegraphics[clip, width=0.24\hsize]{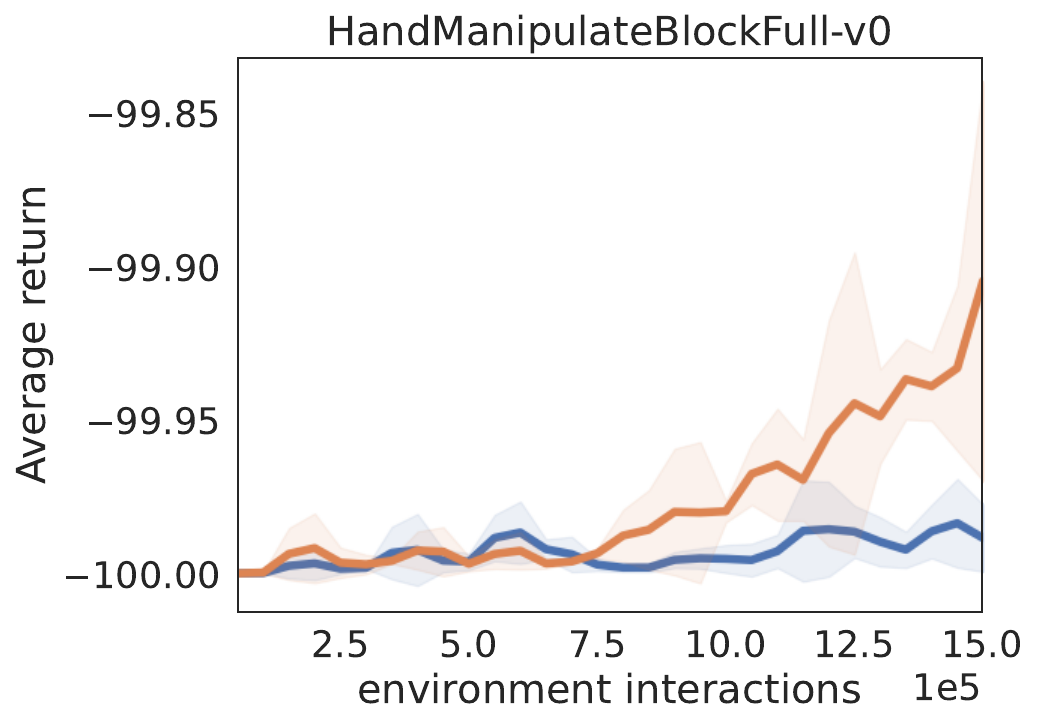}
\includegraphics[clip, width=0.24\hsize]{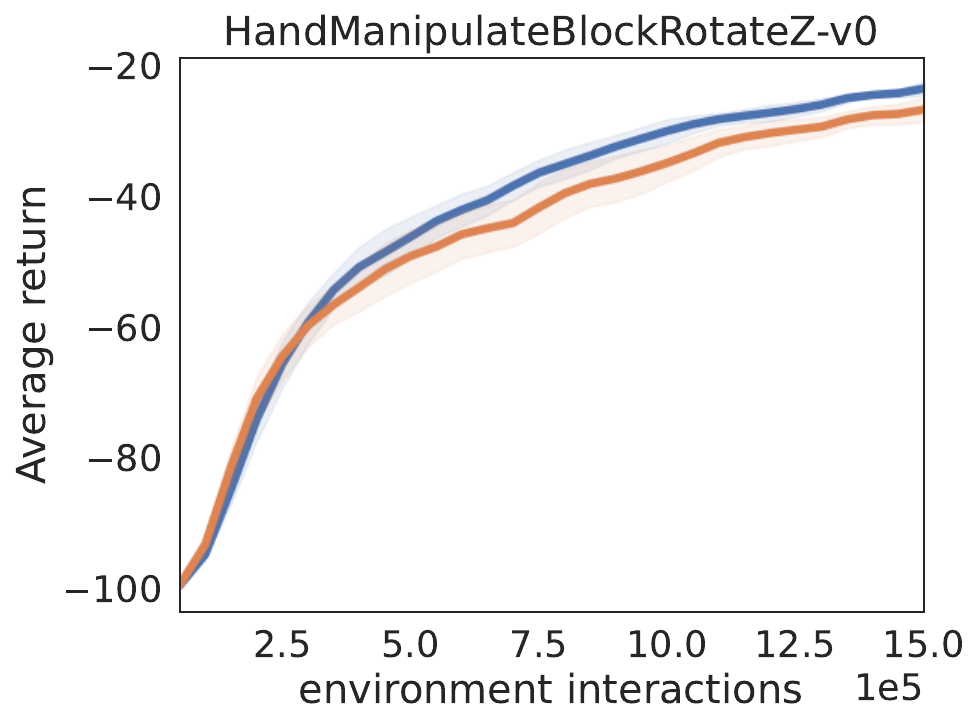}
\includegraphics[clip, width=0.24\hsize]{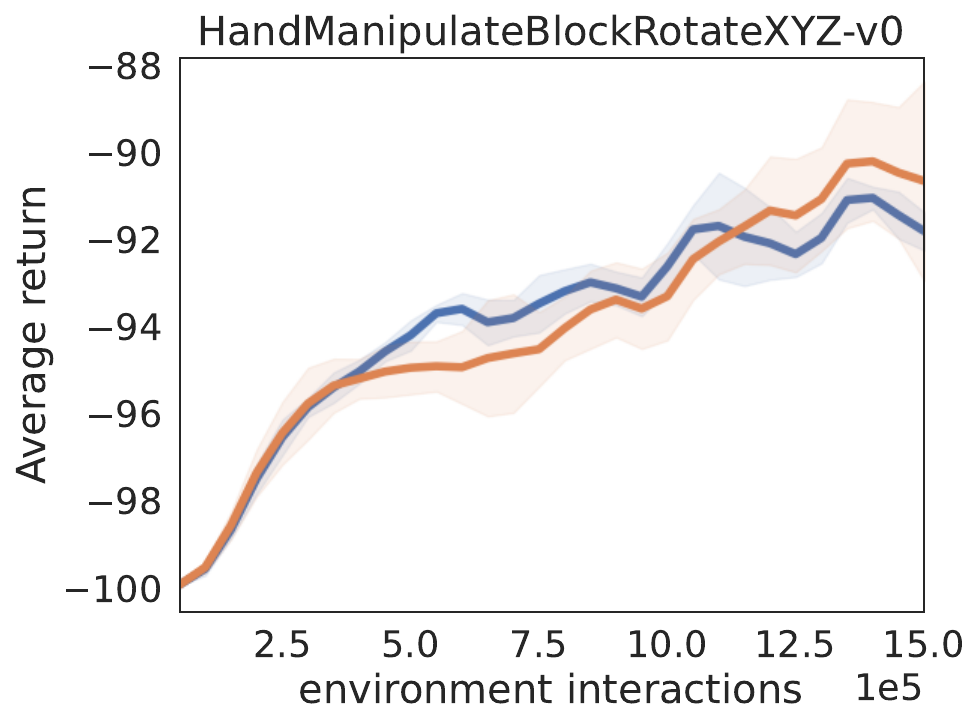}
\includegraphics[clip, width=0.24\hsize]{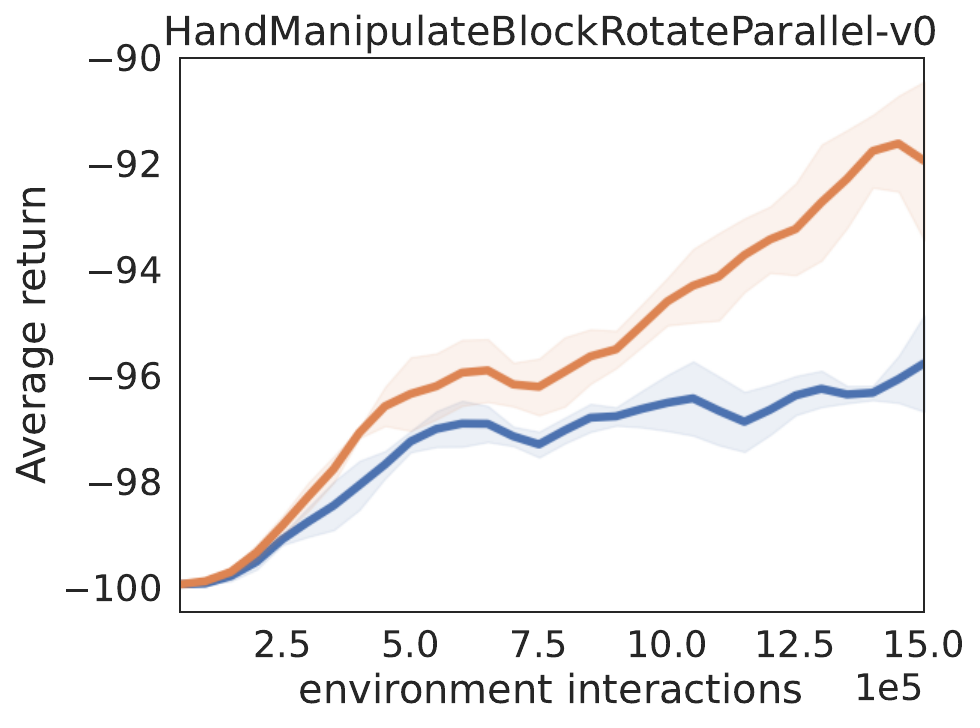}
\end{minipage}
\vspace{-0.7\baselineskip}
\caption{
The effect of removing CDQ and the entropy term on performance (return). 
}
\label{fig:app-simplification-return}
\vspace{-0.5\baselineskip}
\end{figure}

\begin{figure}[h!]
\begin{minipage}{1.0\hsize}
\includegraphics[clip, width=0.24\hsize]{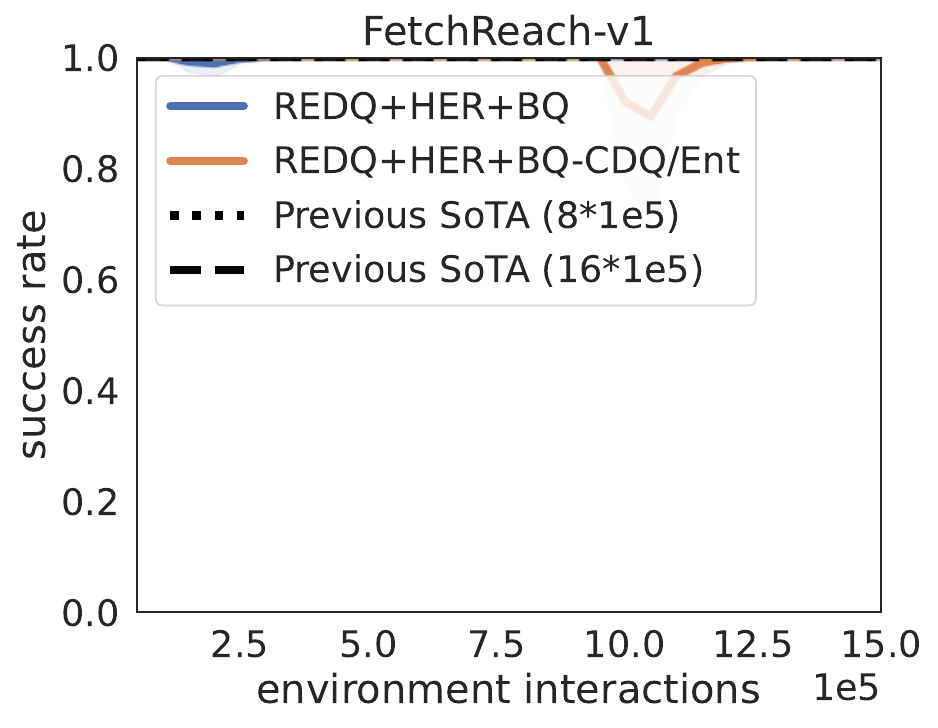}
\includegraphics[clip, width=0.24\hsize]{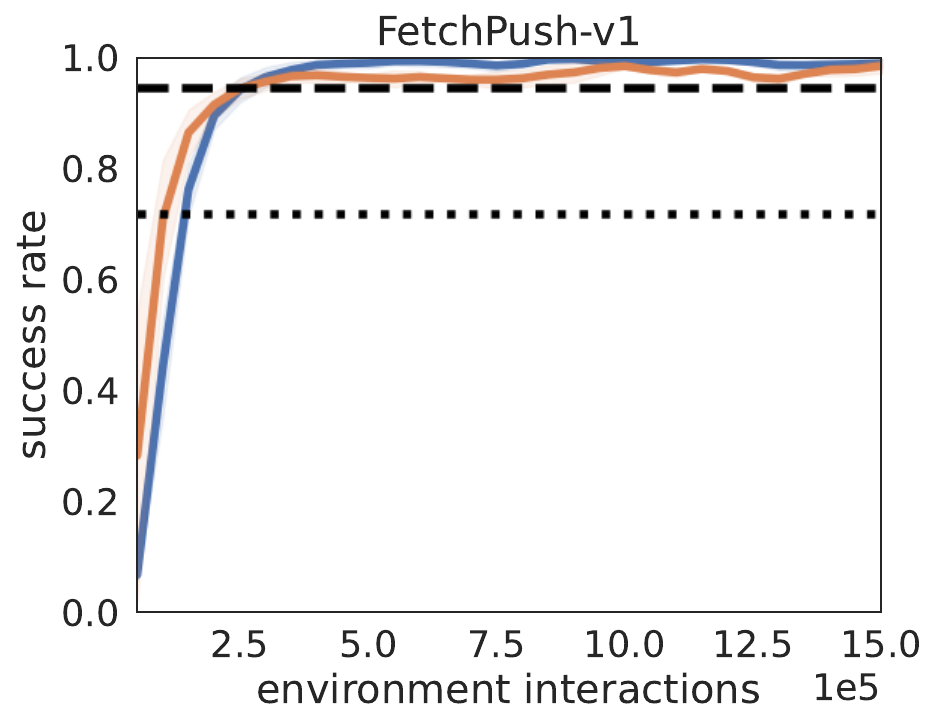}
\includegraphics[clip, width=0.24\hsize]{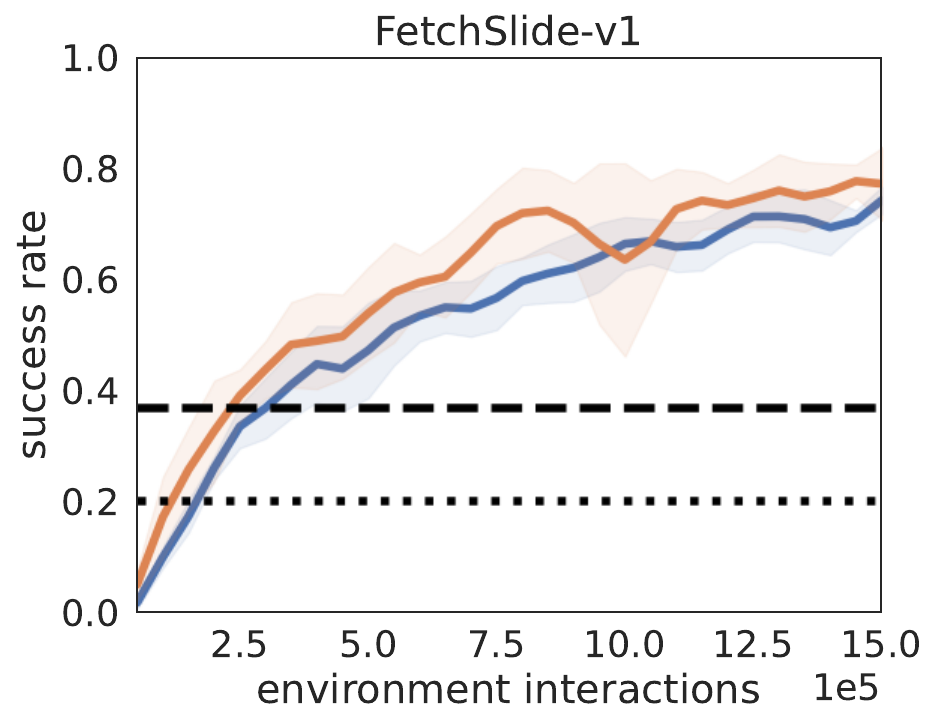}
\includegraphics[clip, width=0.24\hsize]{Figures/Simplification/SuccessRate/FetchPickAndPlace-v1_0.pdf}
\end{minipage}
\begin{minipage}{1.0\hsize}
\includegraphics[clip, width=0.24\hsize]{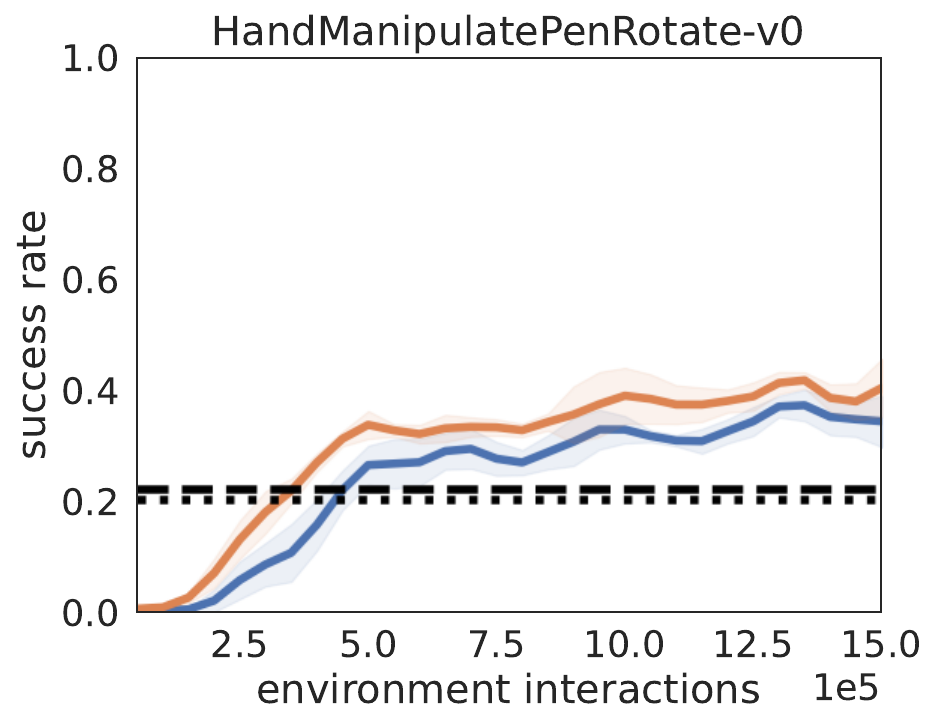}
\includegraphics[clip, width=0.24\hsize]{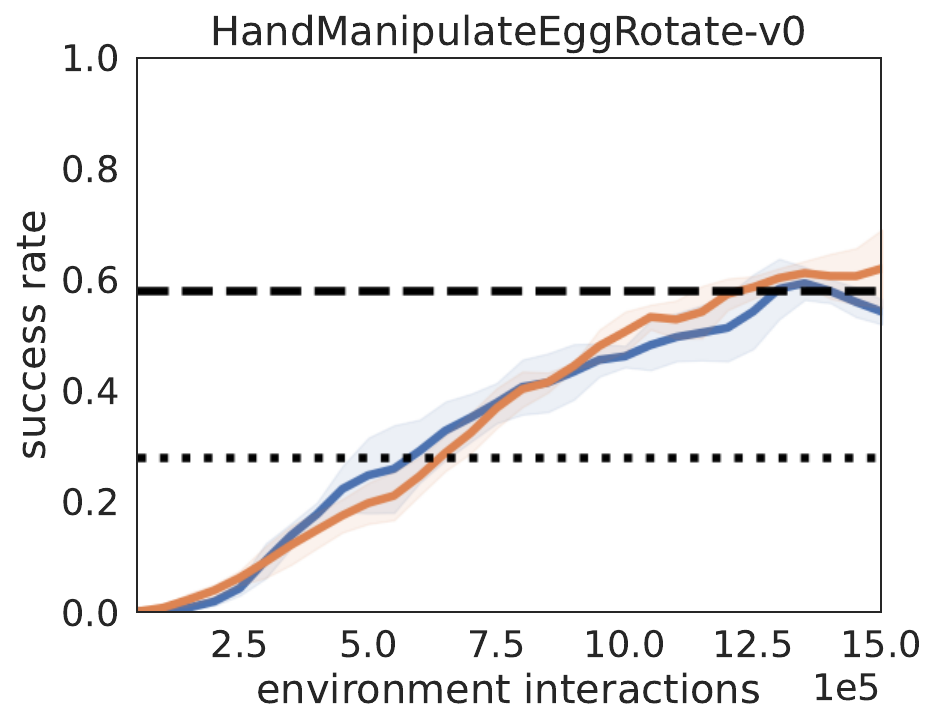}
\includegraphics[clip, width=0.24\hsize]{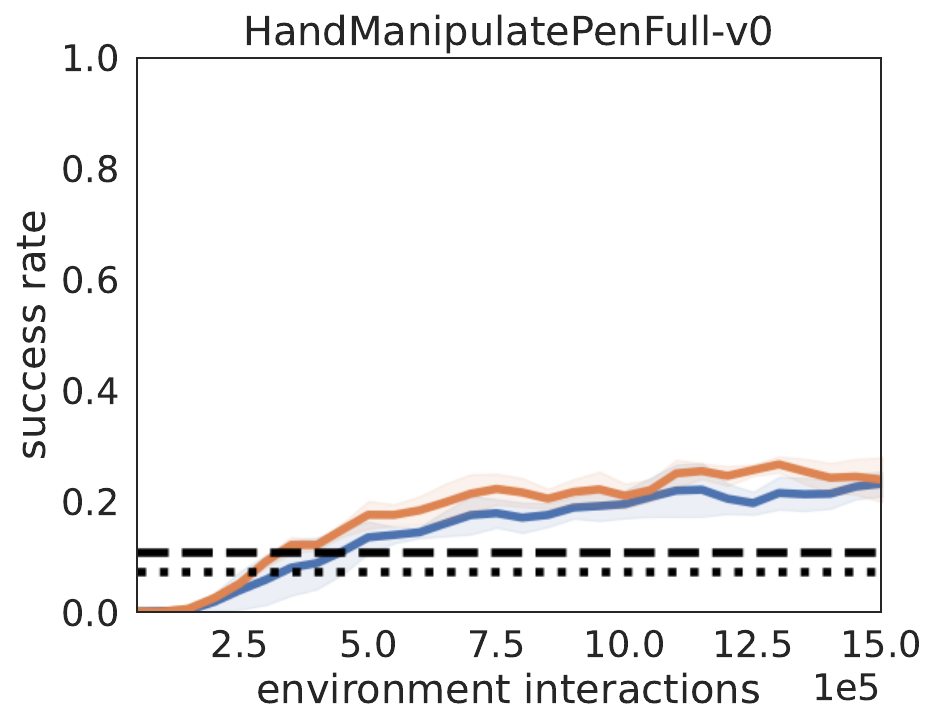}
\includegraphics[clip, width=0.24\hsize]{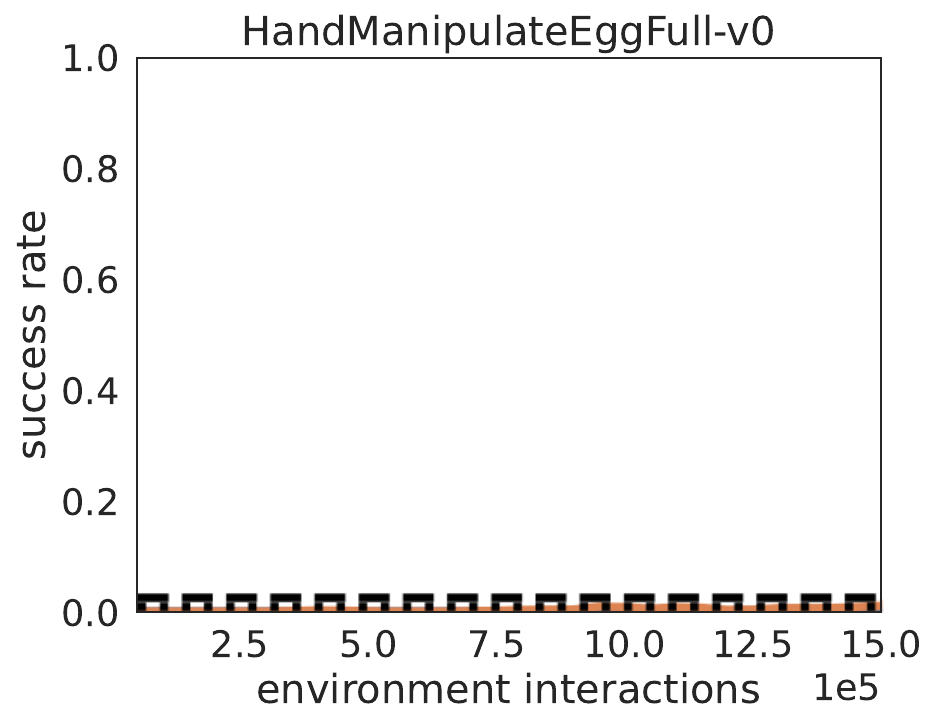}
\end{minipage}
\begin{minipage}{1.0\hsize}
\includegraphics[clip, width=0.24\hsize]{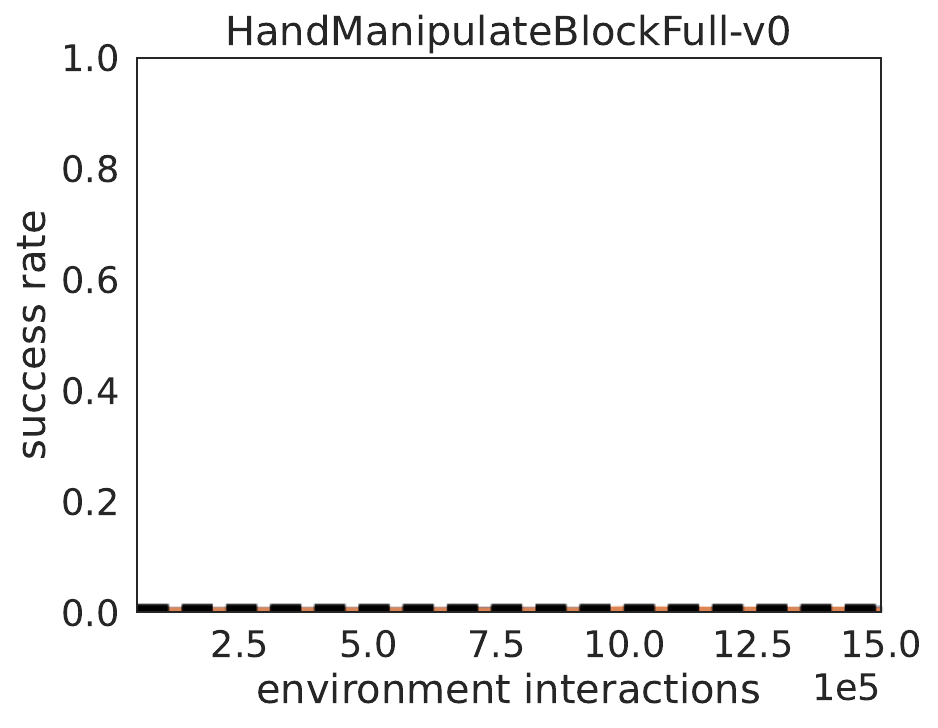}
\includegraphics[clip, width=0.24\hsize]{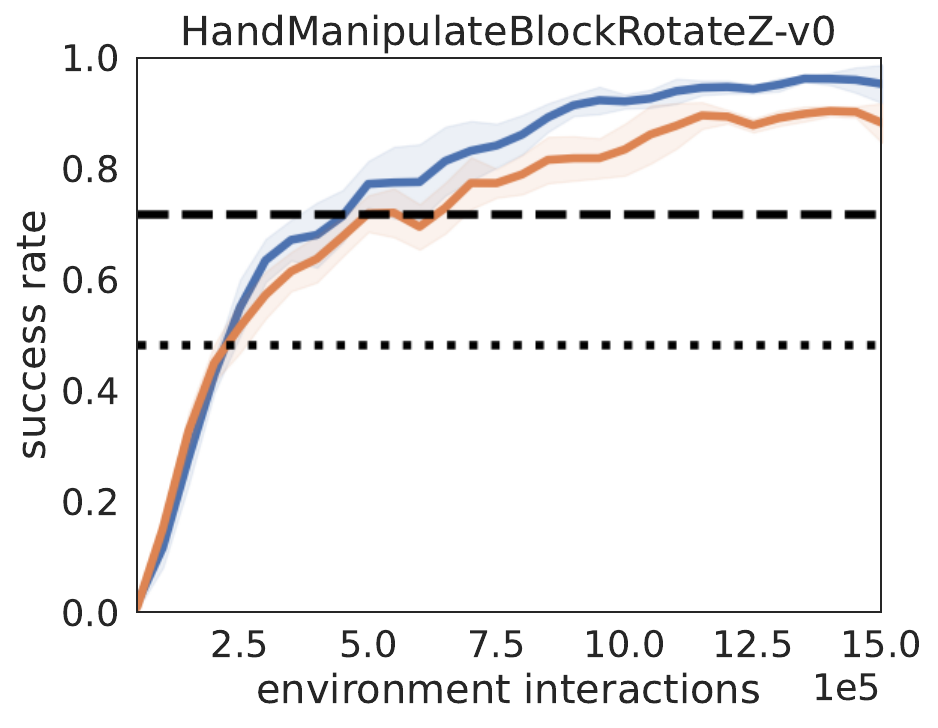}
\includegraphics[clip, width=0.24\hsize]{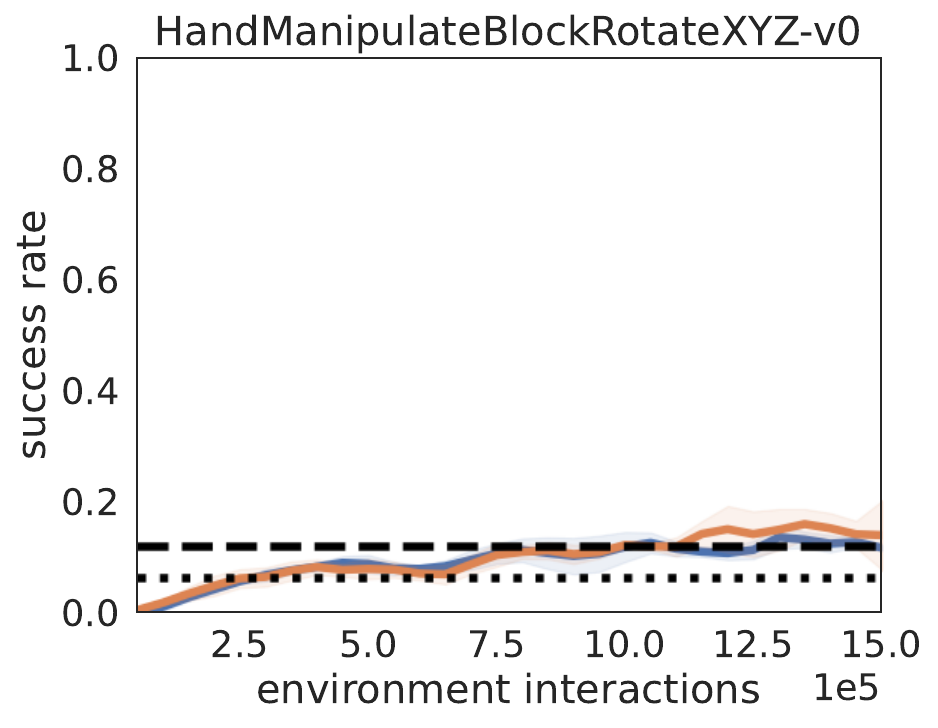}
\includegraphics[clip, width=0.24\hsize]{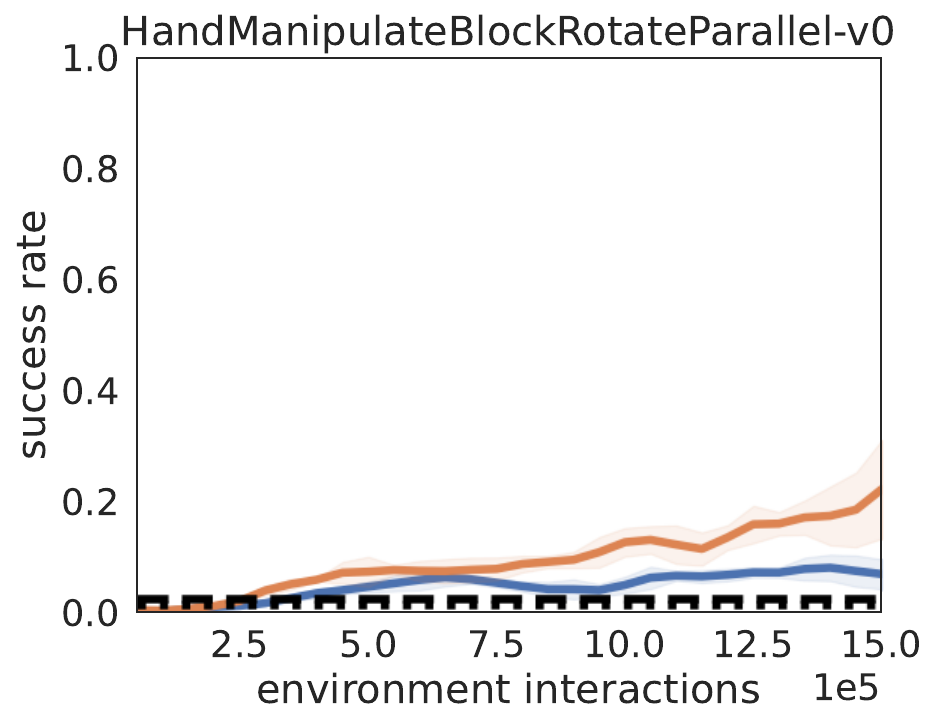}
\end{minipage}
\vspace{-0.7\baselineskip}
\caption{
The effect of removing CDQ and the entropy term on performance (success rate). 
}
\label{fig:app-simplification-sr}
\vspace{-0.5\baselineskip}
\end{figure}

\begin{figure}[h!]
\begin{minipage}{1.0\hsize}
\includegraphics[clip, width=0.24\hsize]{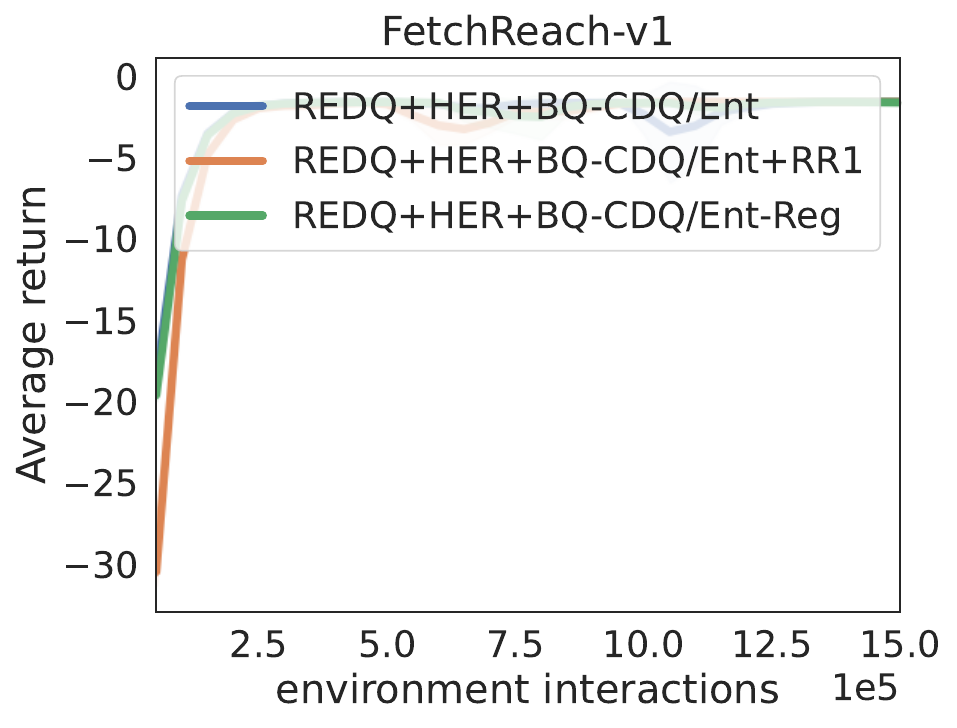}
\includegraphics[clip, width=0.24\hsize]{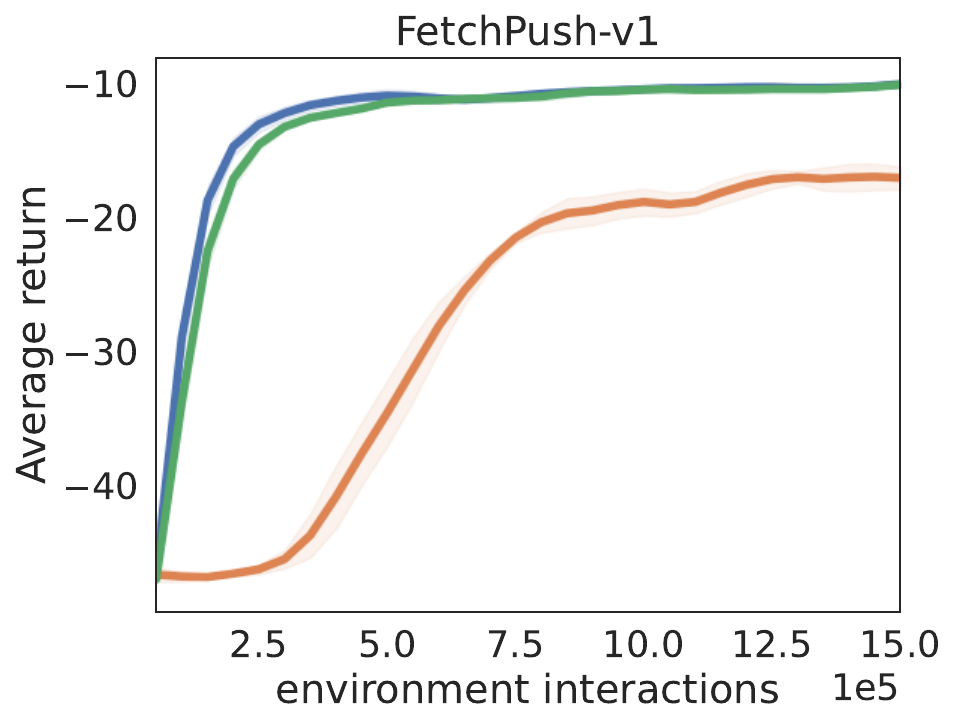}
\includegraphics[clip, width=0.24\hsize]{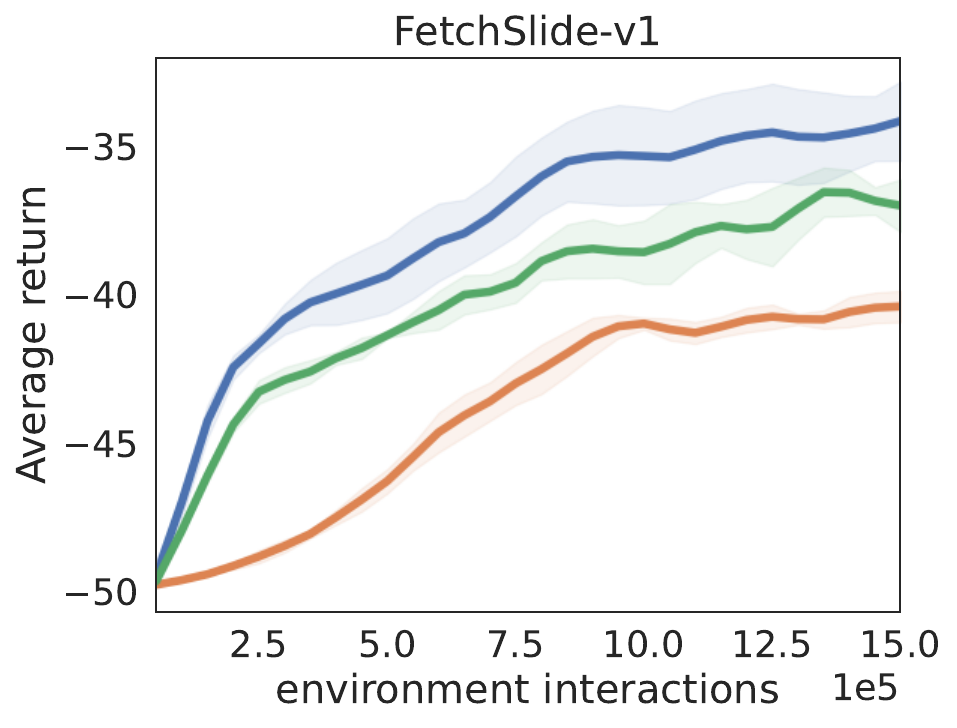}
\includegraphics[clip, width=0.24\hsize]{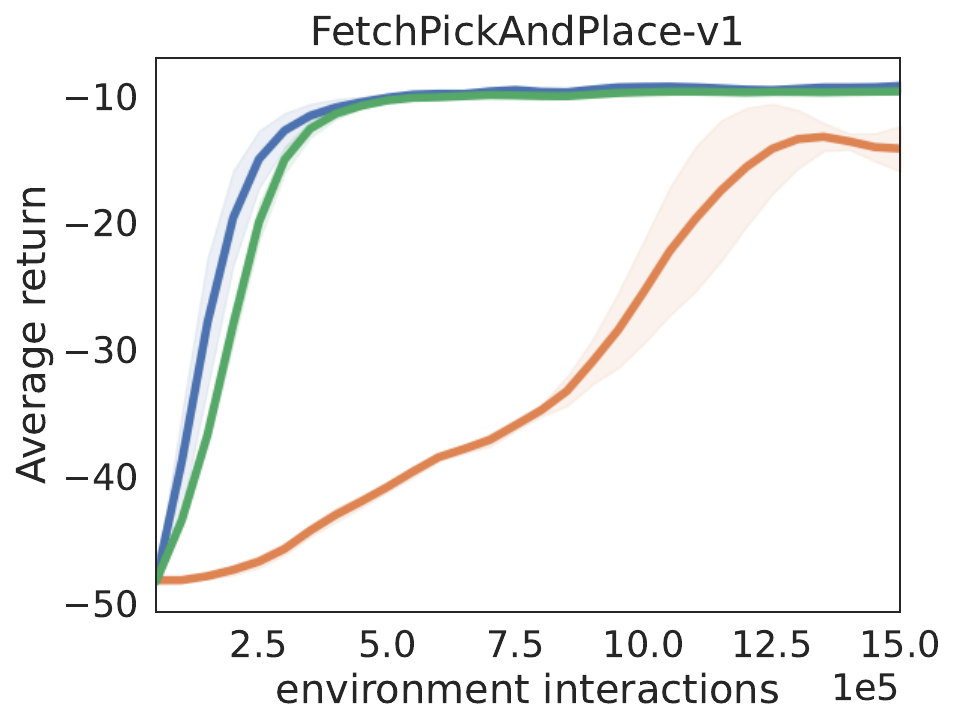}
\end{minipage}
\begin{minipage}{1.0\hsize}
\includegraphics[clip, width=0.24\hsize]{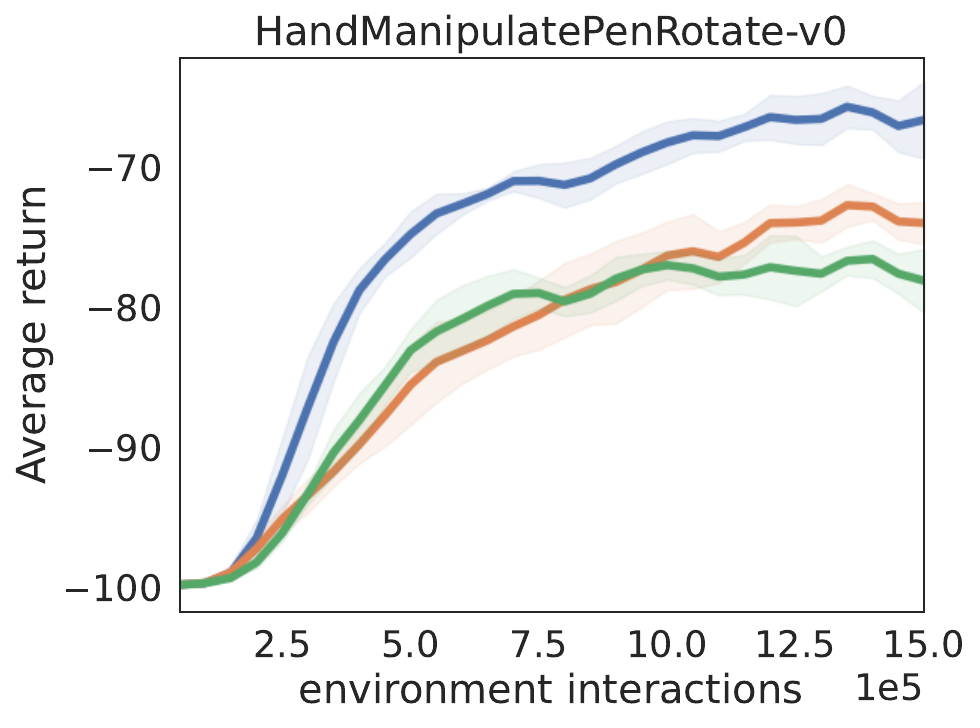}
\includegraphics[clip, width=0.24\hsize]{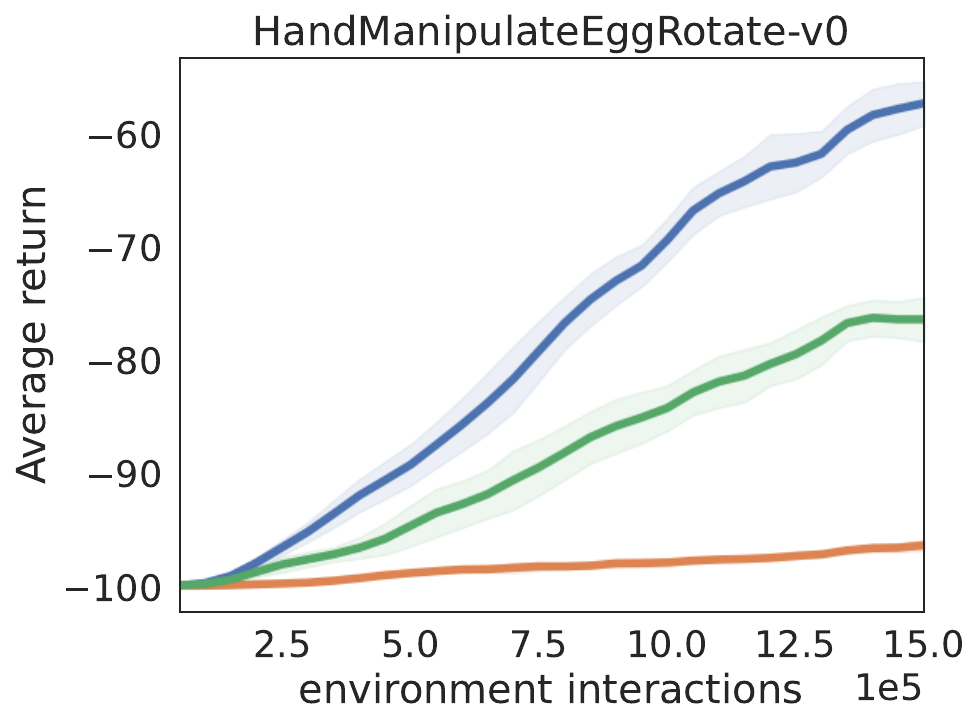}
\includegraphics[clip, width=0.24\hsize]{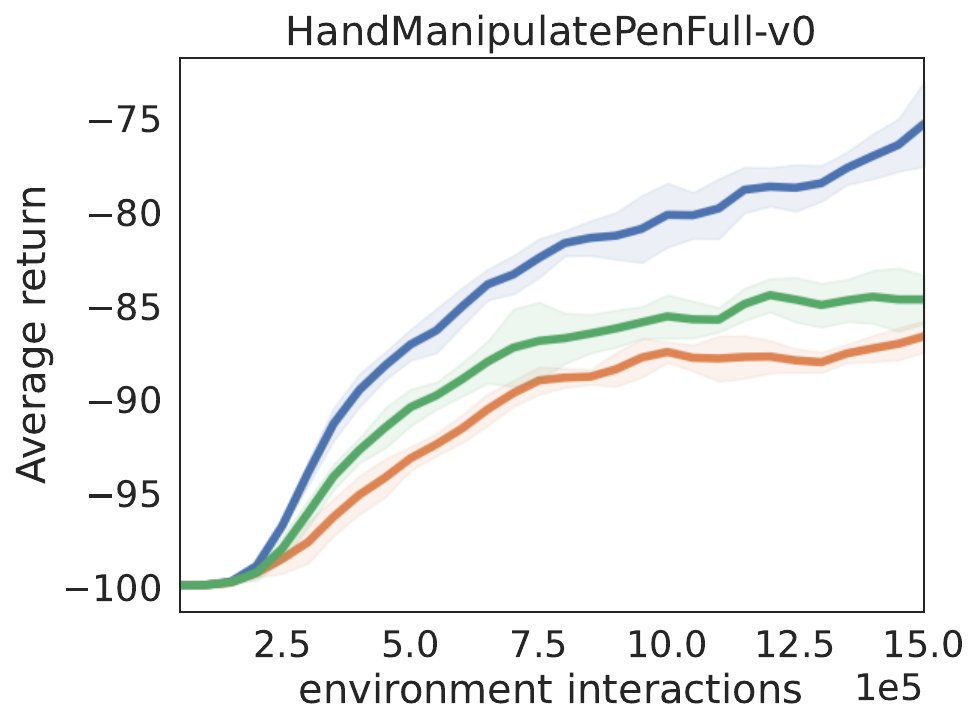}
\includegraphics[clip, width=0.24\hsize]{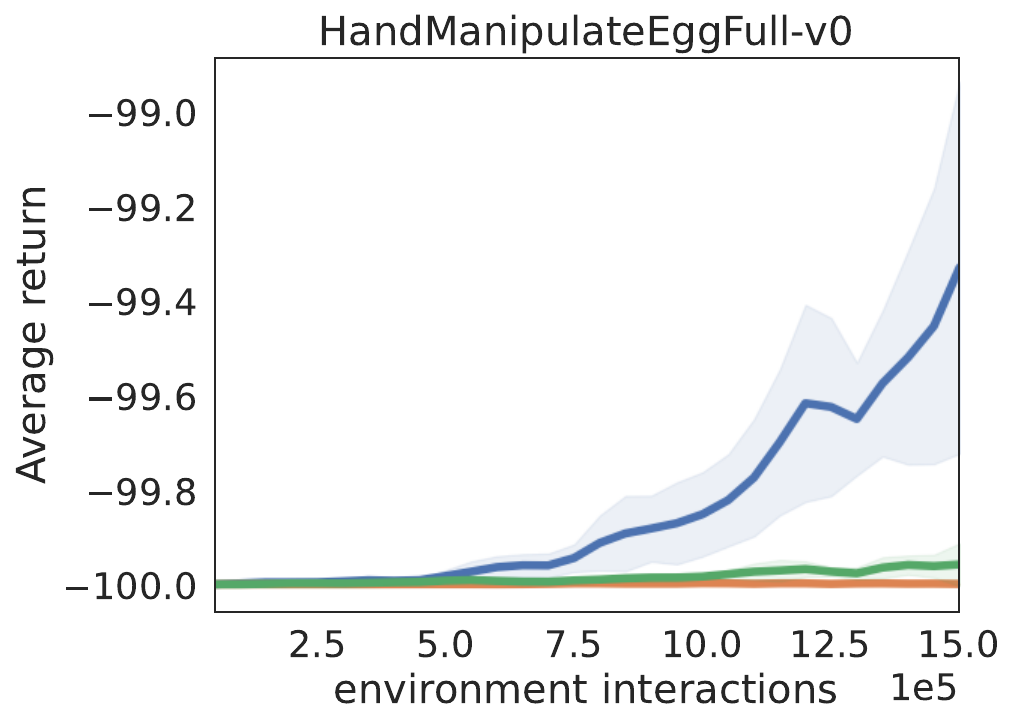}
\end{minipage}
\begin{minipage}{1.0\hsize}
\includegraphics[clip, width=0.24\hsize]{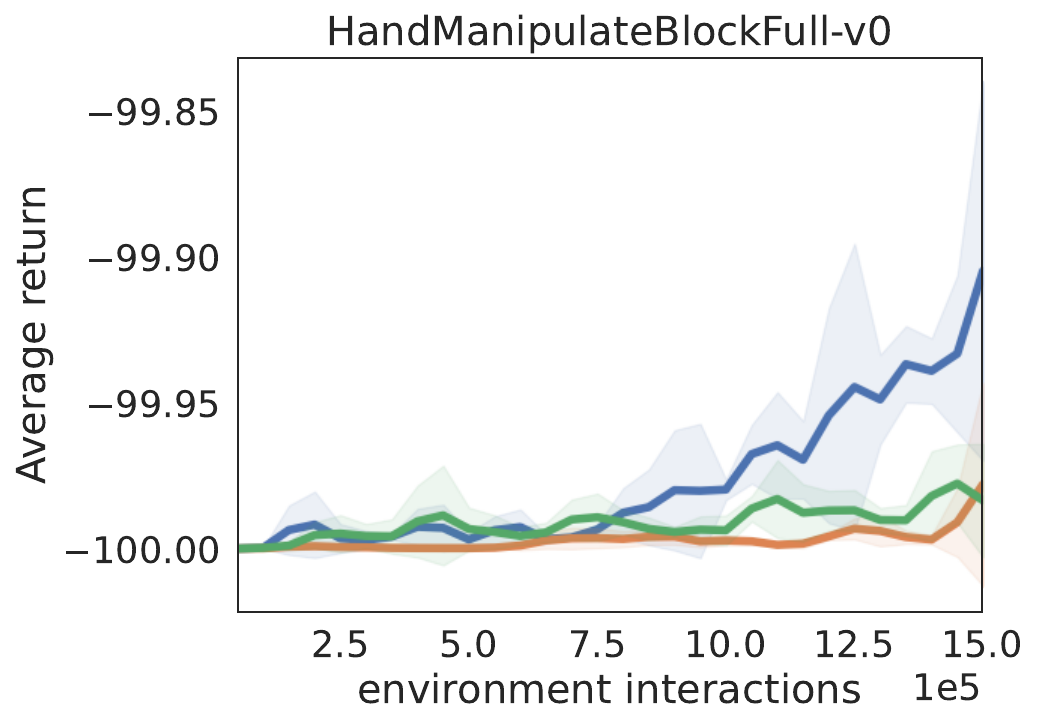}
\includegraphics[clip, width=0.24\hsize]{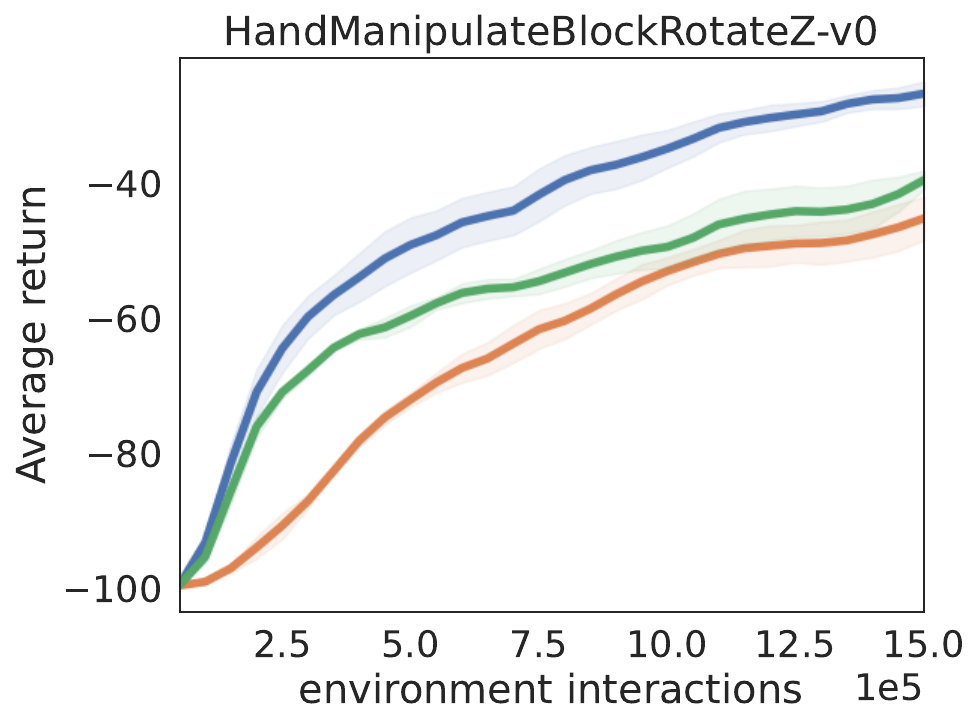}
\includegraphics[clip, width=0.24\hsize]{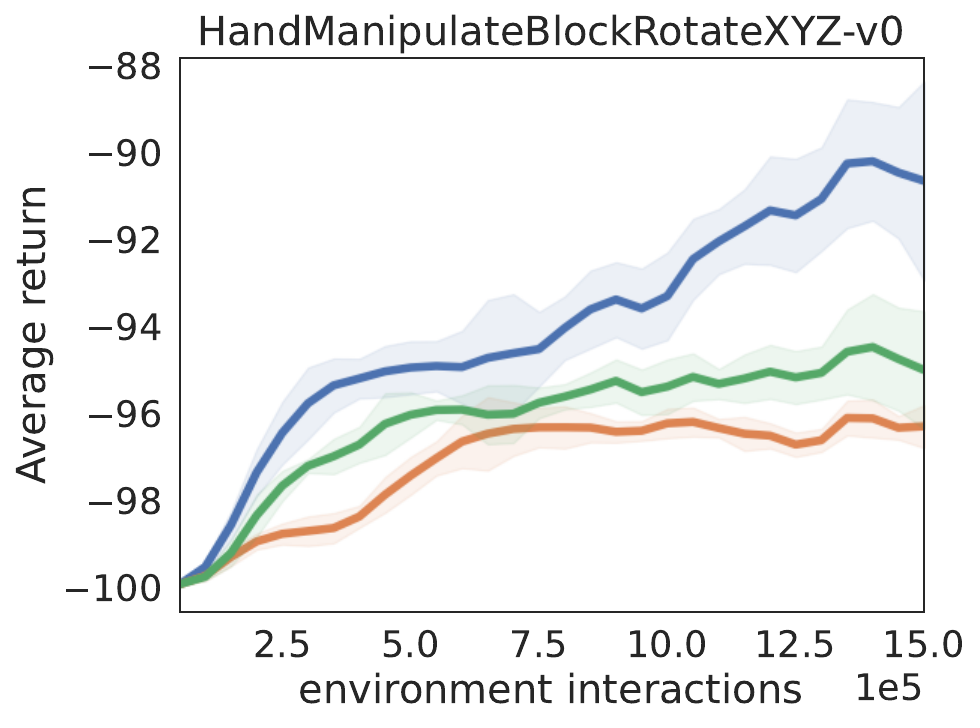}
\includegraphics[clip, width=0.24\hsize]{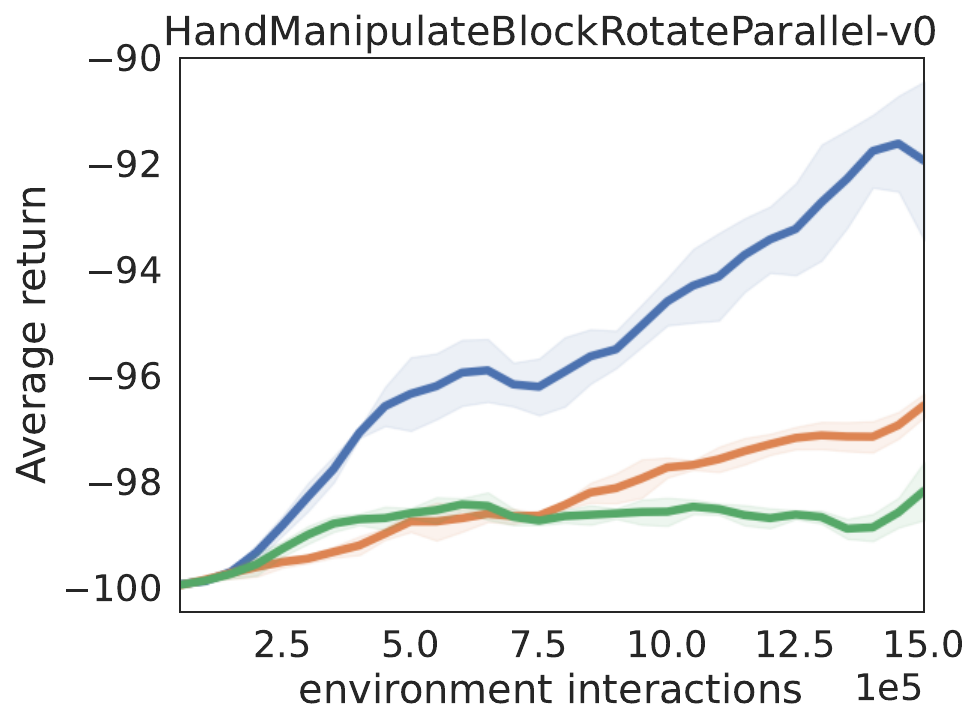}
\end{minipage}
\vspace{-0.7\baselineskip}
\caption{
The effect of removing a high RR and regularization on performance (return). 
}
\label{fig:app-simplification2-return}
\vspace{-0.5\baselineskip}
\end{figure}

\begin{figure}[h!]
\begin{minipage}{1.0\hsize}
\includegraphics[clip, width=0.24\hsize]{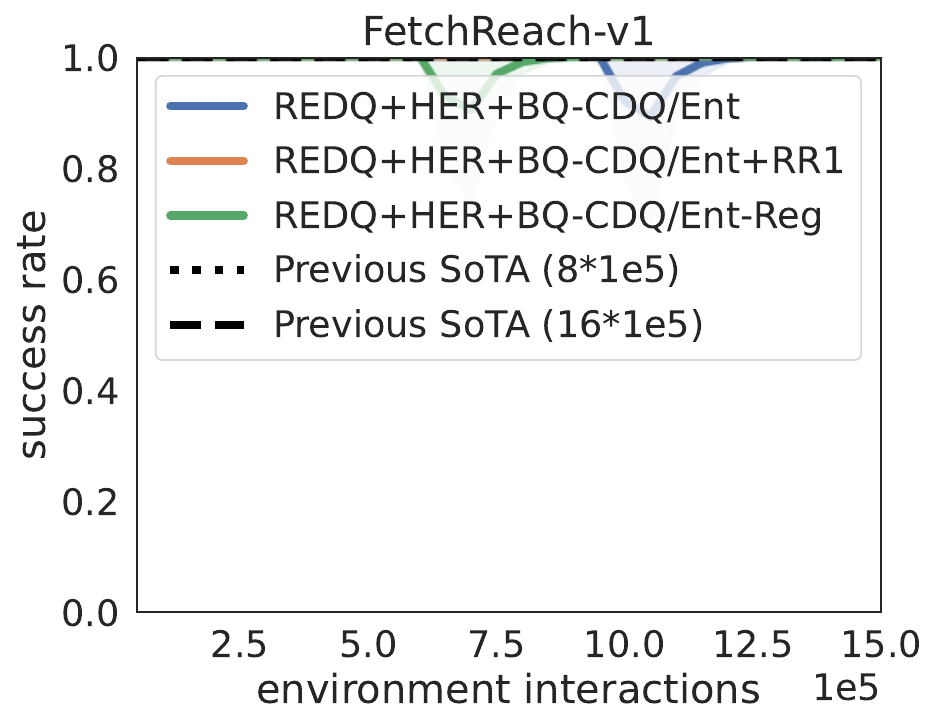}
\includegraphics[clip, width=0.24\hsize]{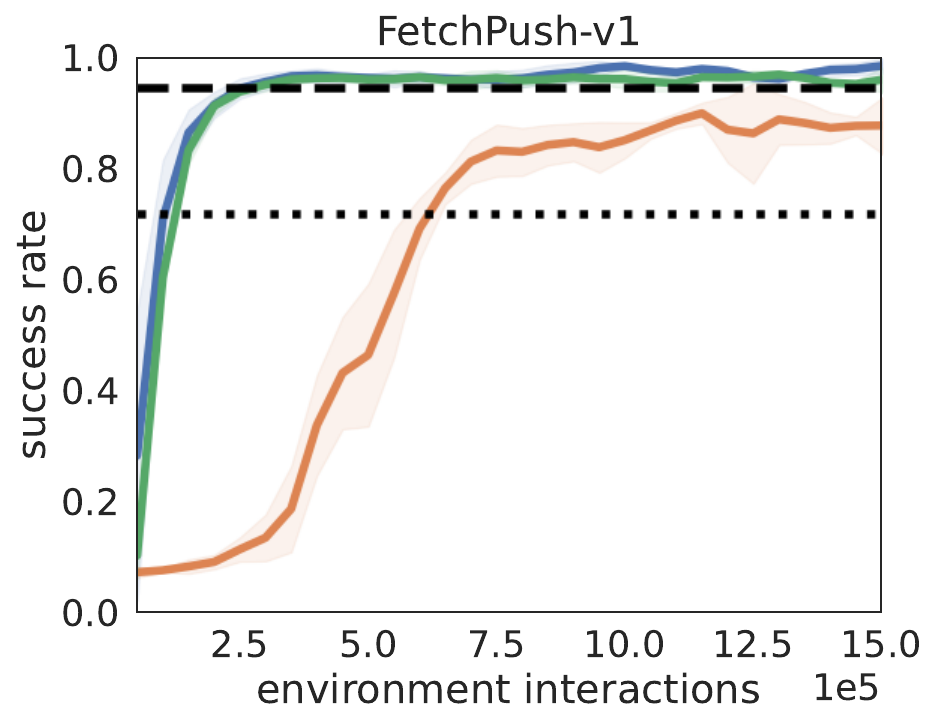}
\includegraphics[clip, width=0.24\hsize]{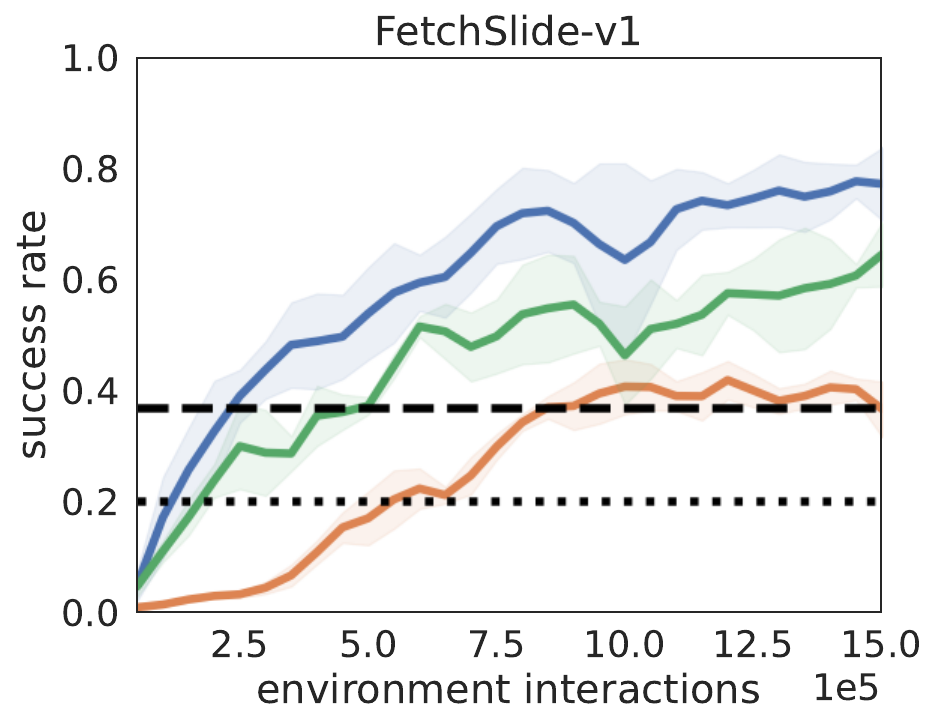}
\includegraphics[clip, width=0.24\hsize]{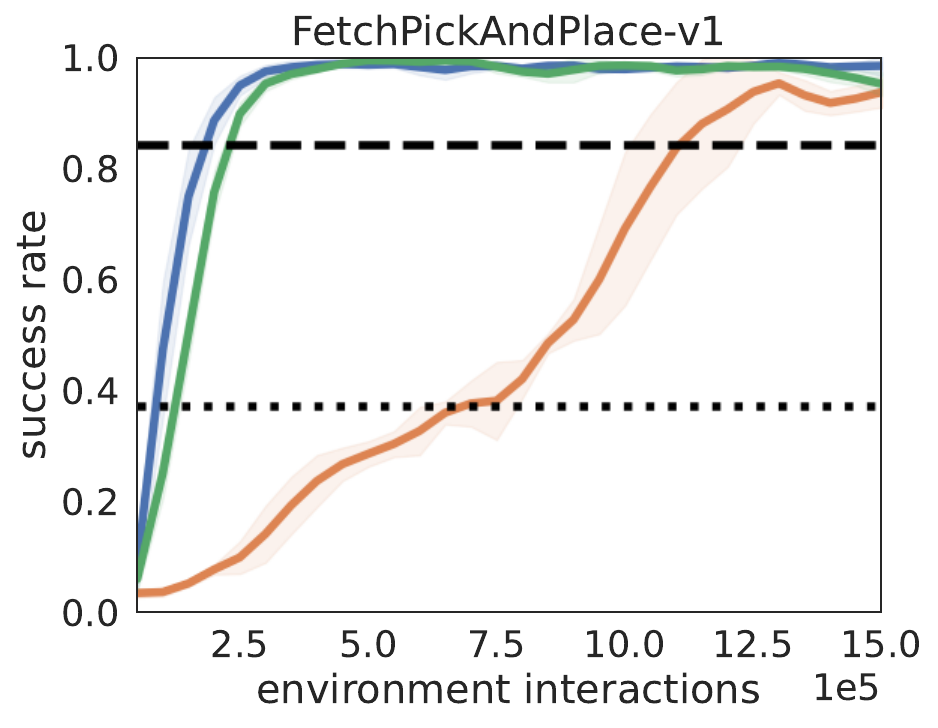}
\end{minipage}
\begin{minipage}{1.0\hsize}
\includegraphics[clip, width=0.24\hsize]{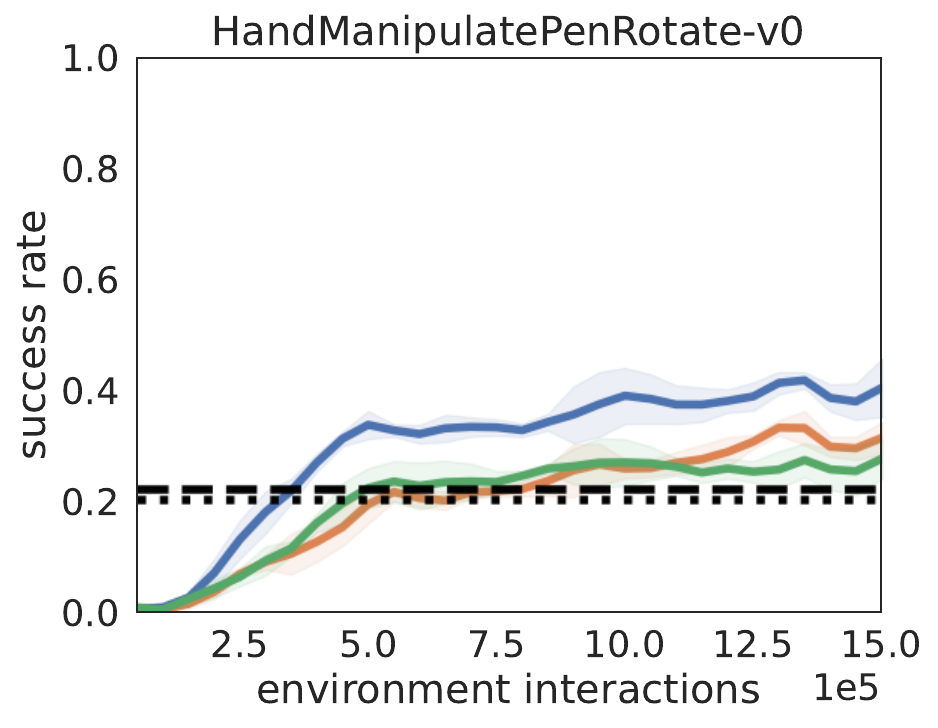}
\includegraphics[clip, width=0.24\hsize]{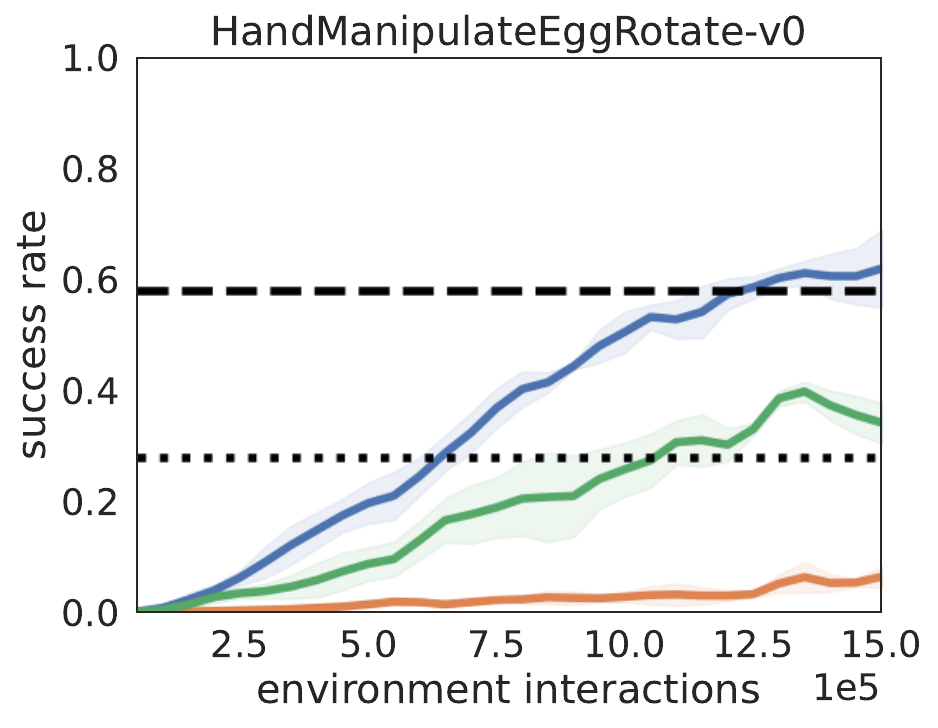}
\includegraphics[clip, width=0.24\hsize]{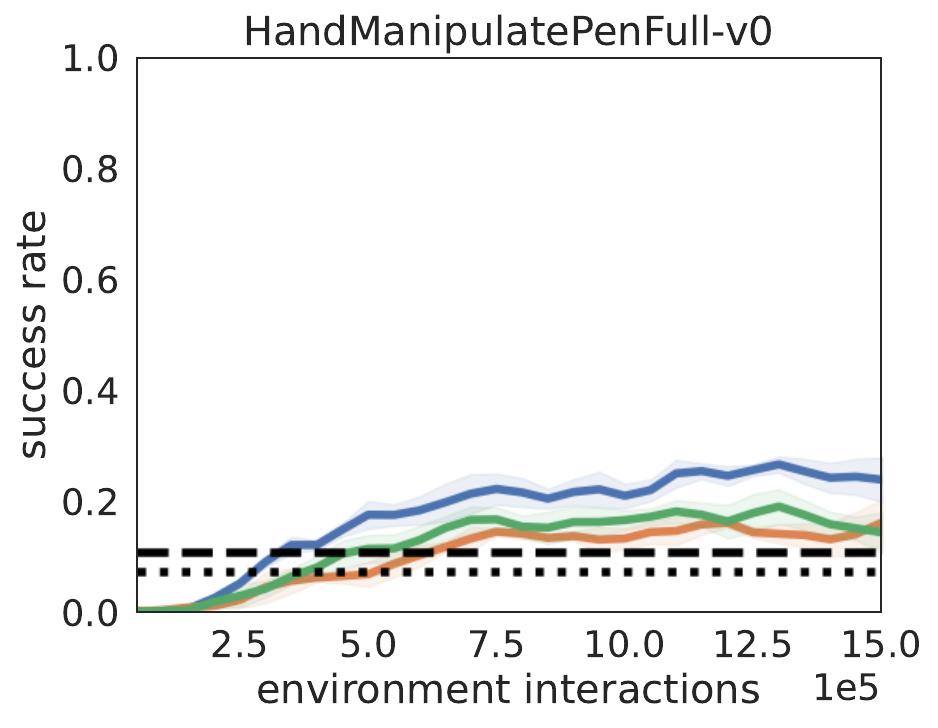}
\includegraphics[clip, width=0.24\hsize]{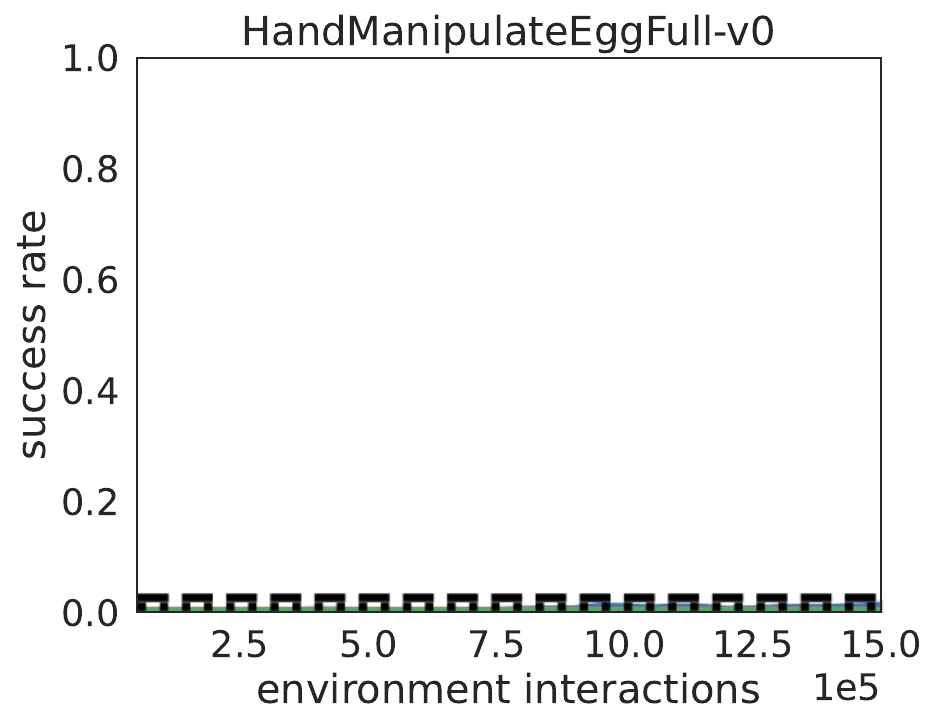}
\end{minipage}
\begin{minipage}{1.0\hsize}
\includegraphics[clip, width=0.24\hsize]{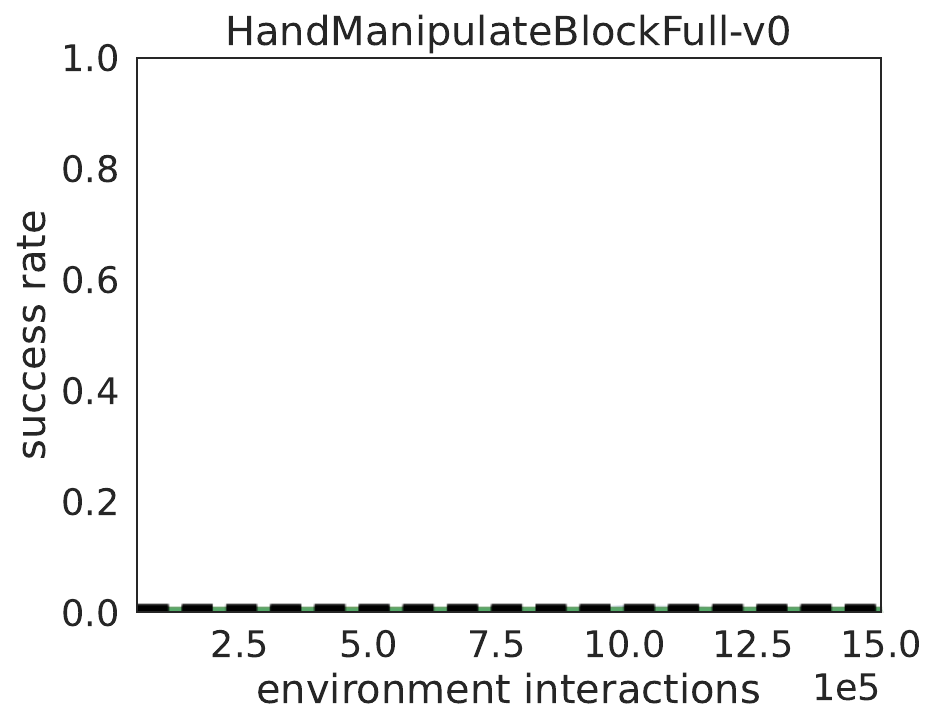}
\includegraphics[clip, width=0.24\hsize]{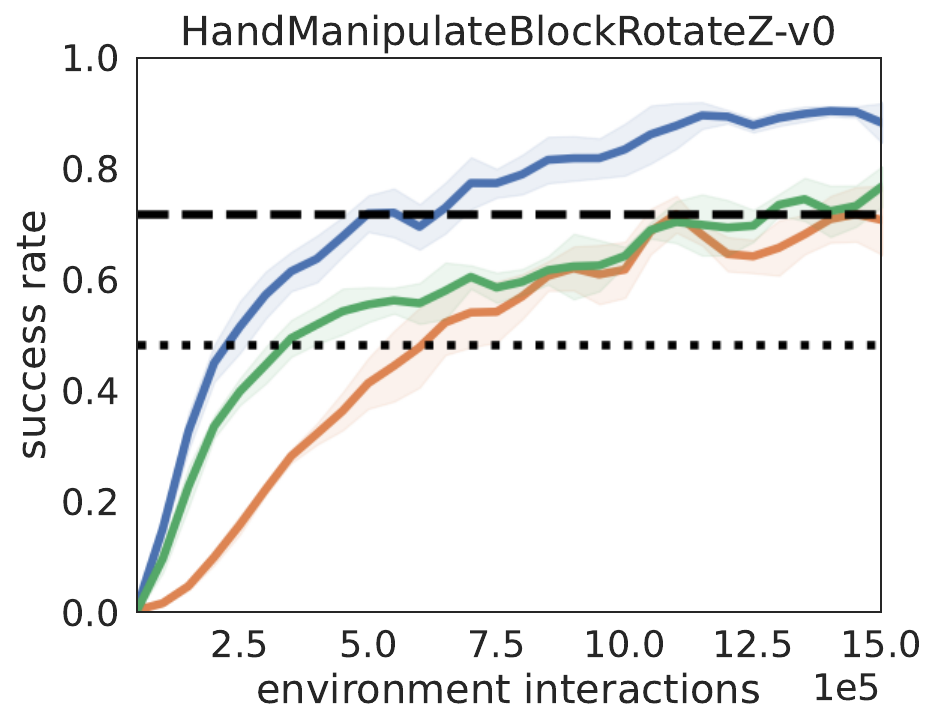}
\includegraphics[clip, width=0.24\hsize]{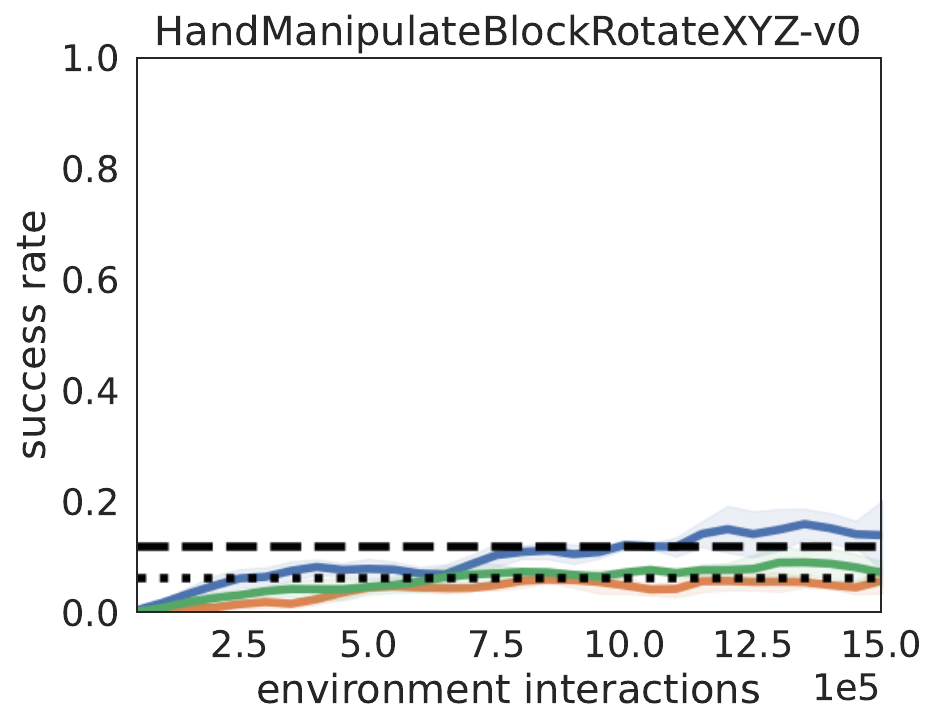}
\includegraphics[clip, width=0.24\hsize]{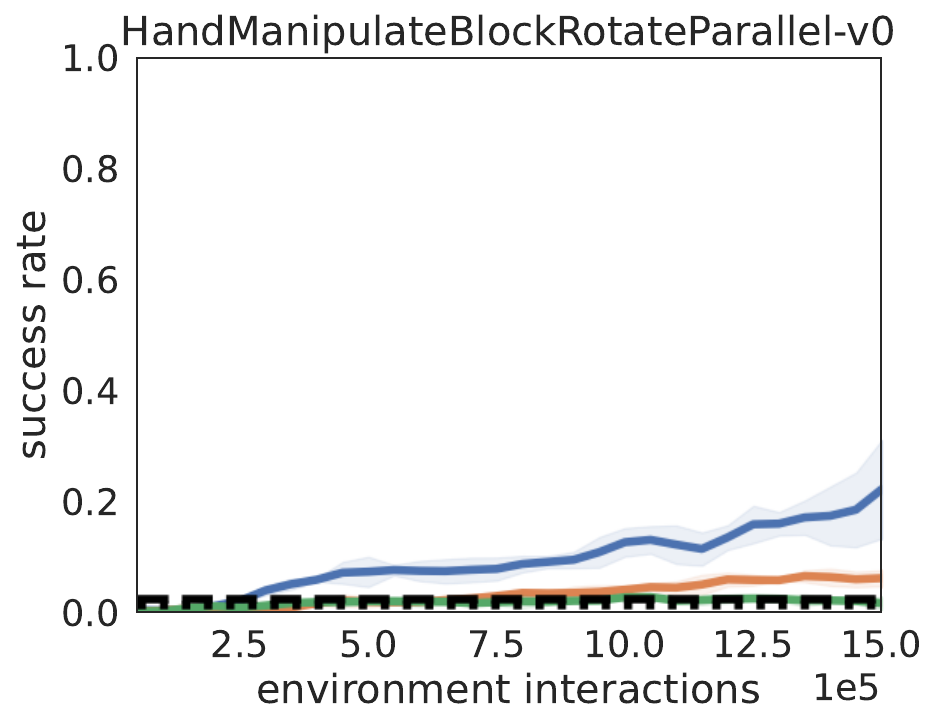}
\end{minipage}
\vspace{-0.7\baselineskip}
\caption{
The effect of removing a high RR and regularization on performance (success rate). 
}
\label{fig:app-simplification2-sr}
\vspace{-0.5\baselineskip}
\end{figure}

\begin{figure}[h!]
\begin{minipage}{1.0\hsize}
\includegraphics[clip, width=0.24\hsize]{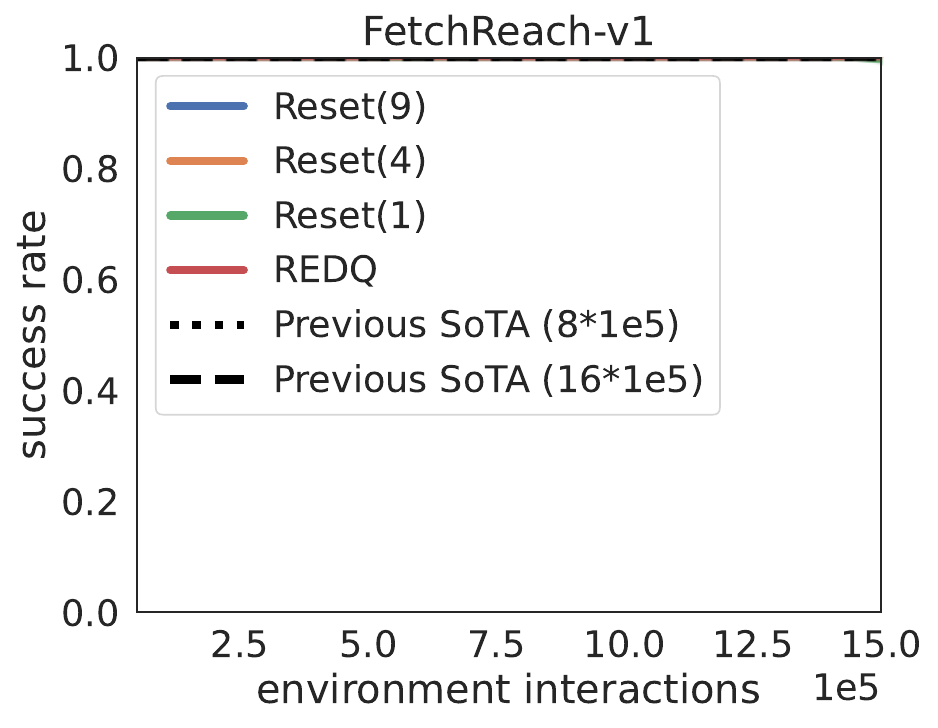}
\includegraphics[clip, width=0.24\hsize]{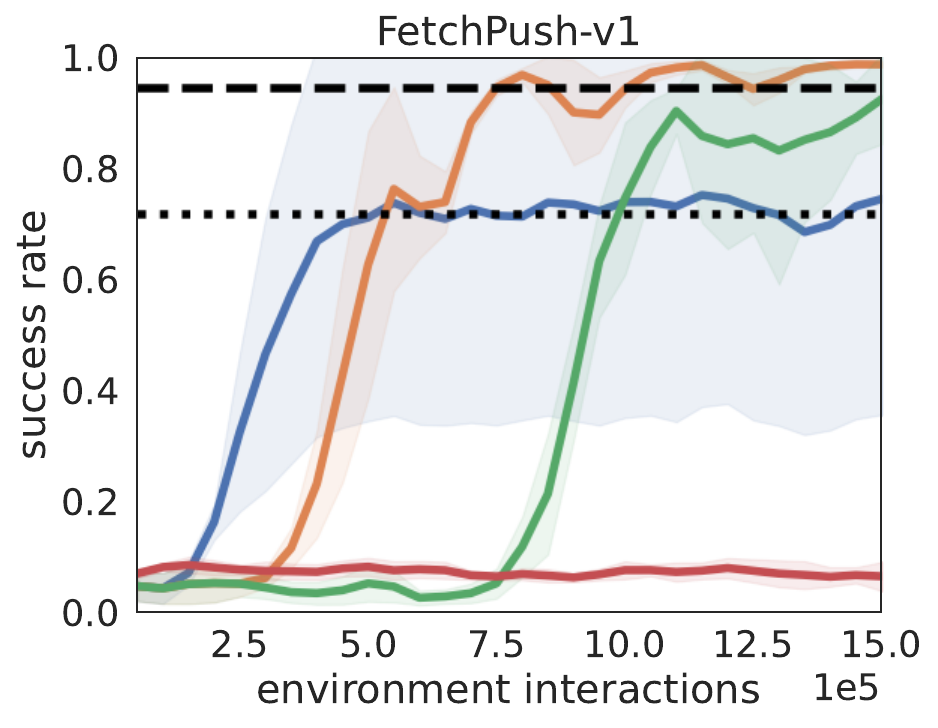}
\includegraphics[clip, width=0.24\hsize]{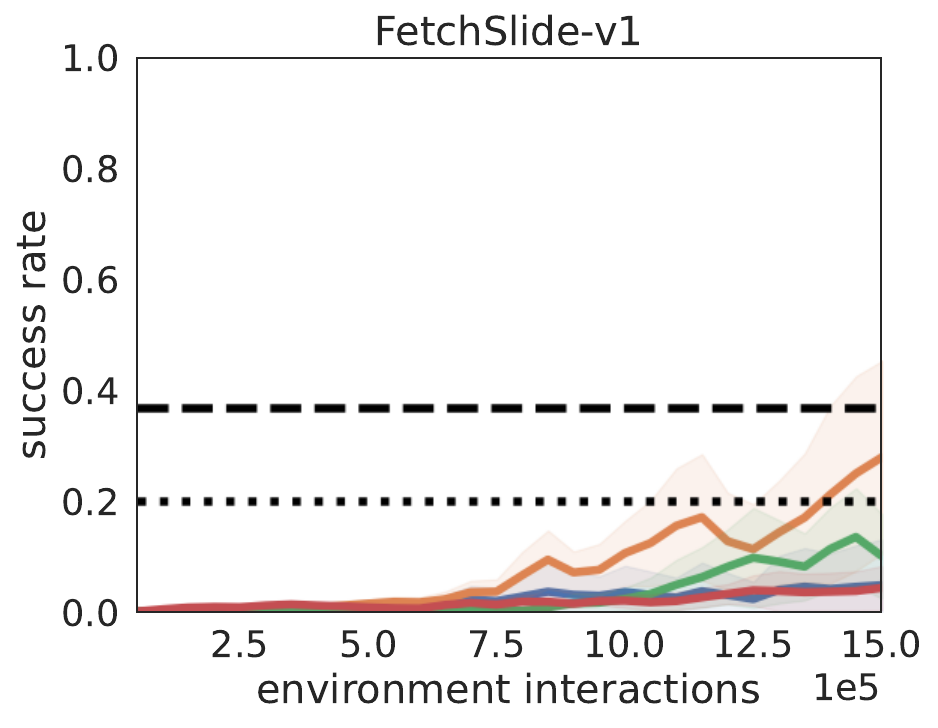}
\includegraphics[clip, width=0.24\hsize]{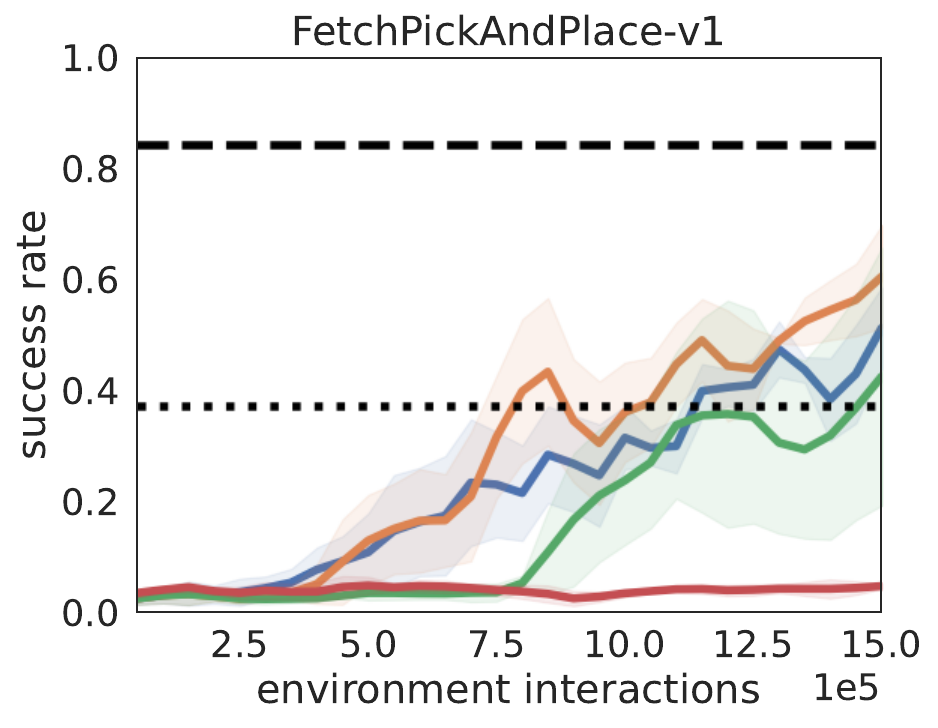}
\end{minipage}
\begin{minipage}{1.0\hsize}
\includegraphics[clip, width=0.24\hsize]{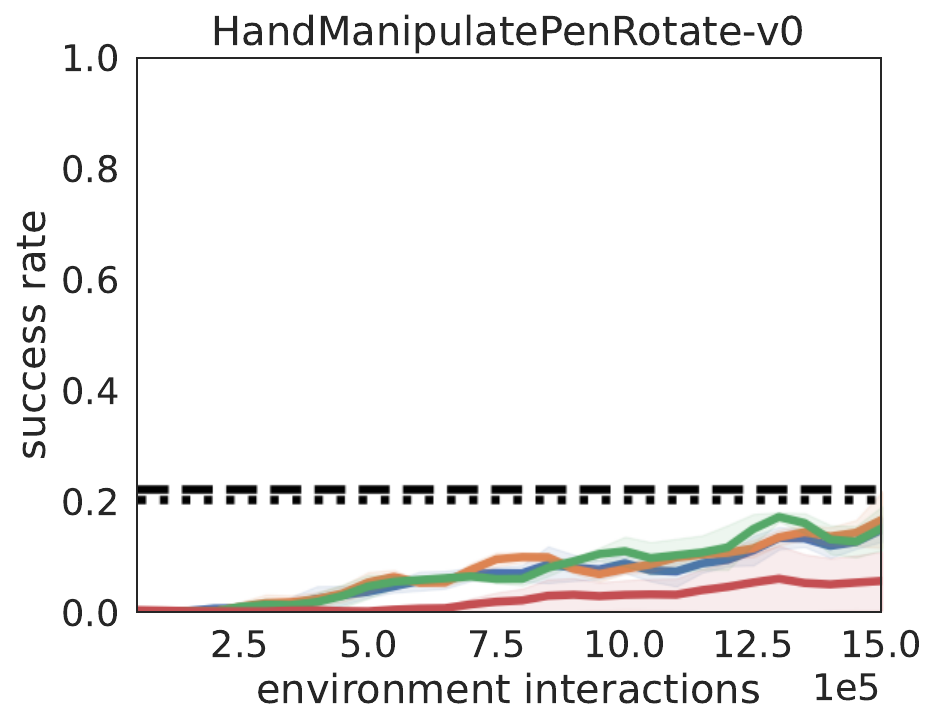}
\includegraphics[clip, width=0.24\hsize]{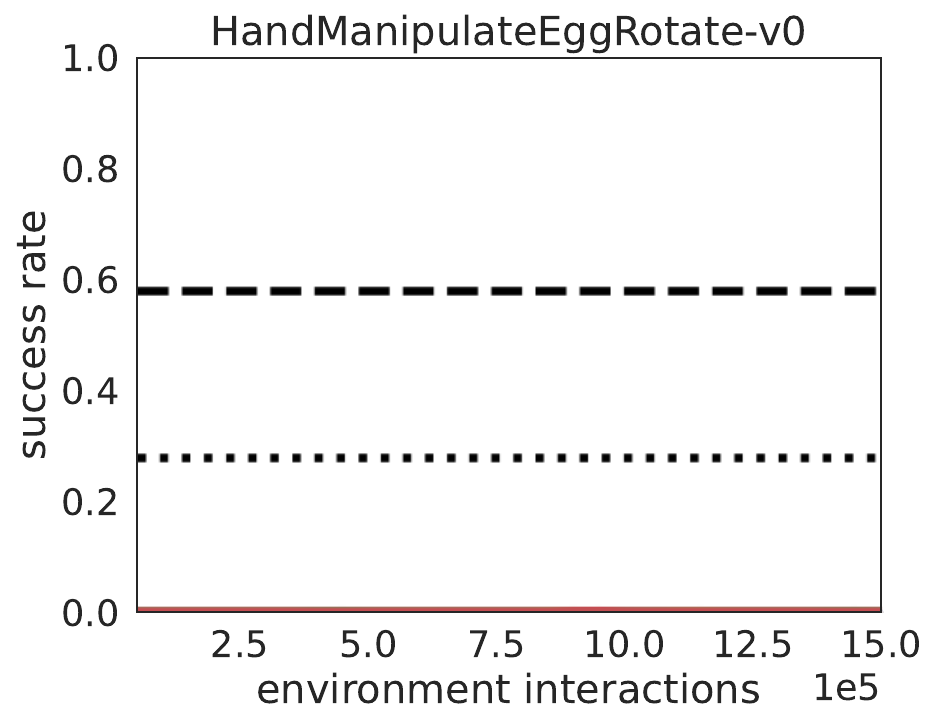}
\includegraphics[clip, width=0.24\hsize]{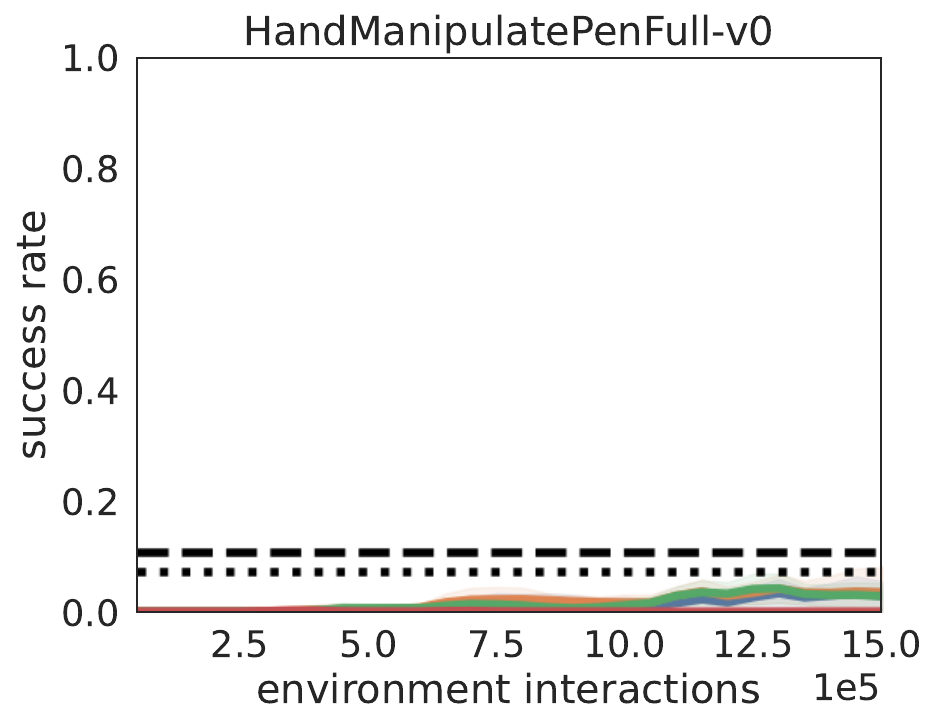}
\includegraphics[clip, width=0.24\hsize]{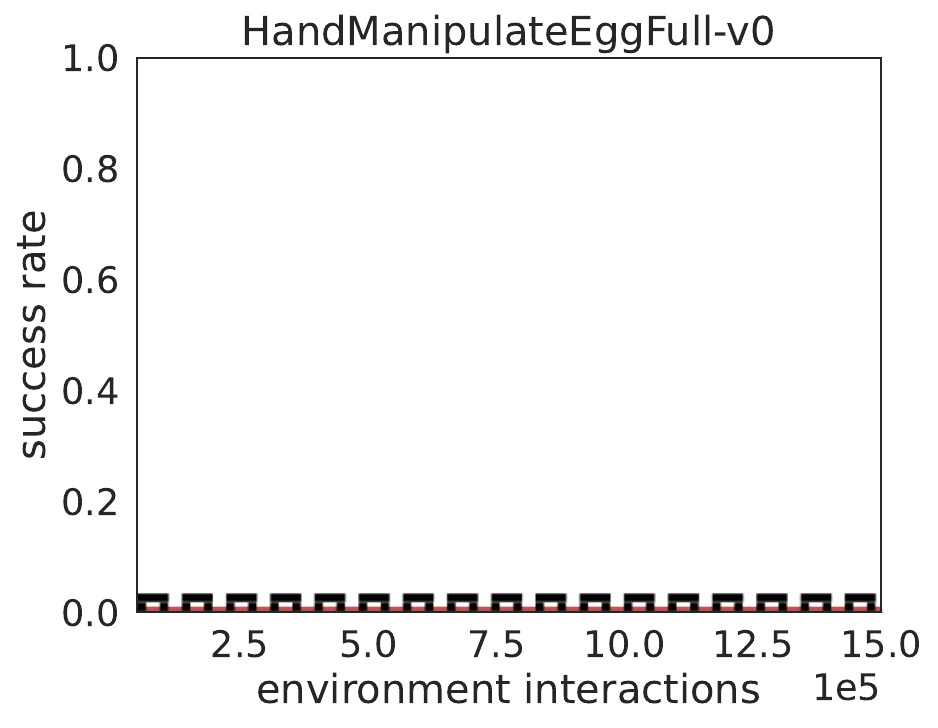}
\end{minipage}
\begin{minipage}{1.0\hsize}
\includegraphics[clip, width=0.24\hsize]{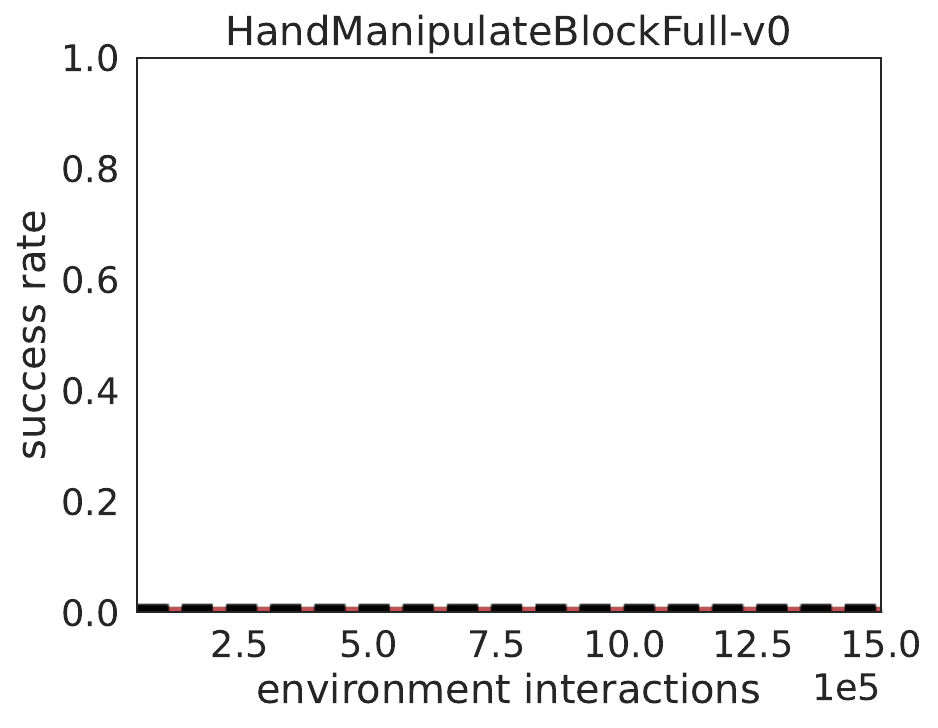}
\includegraphics[clip, width=0.24\hsize]{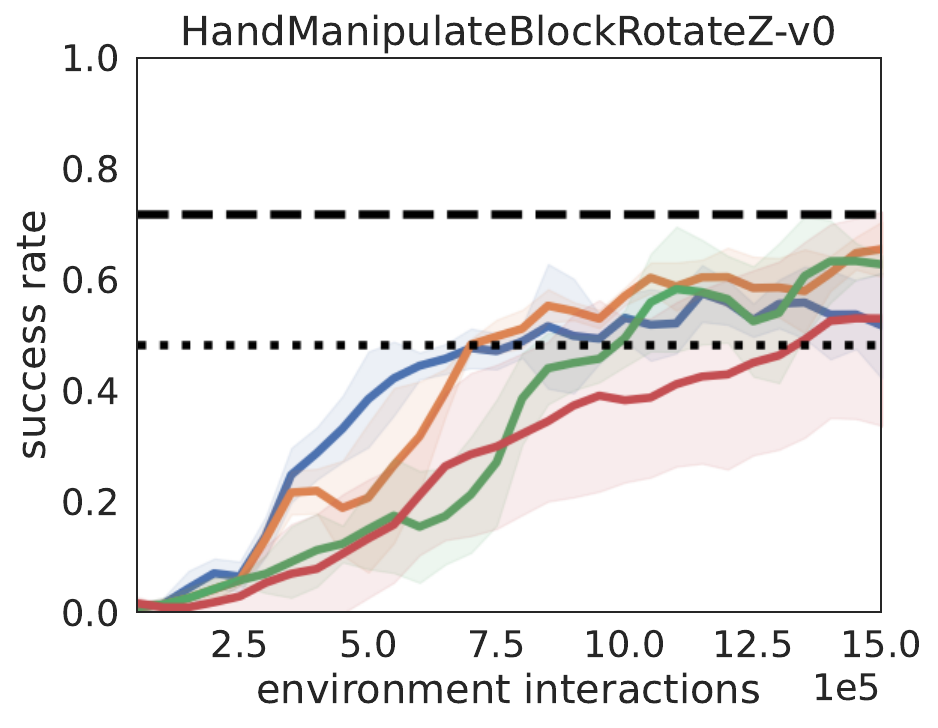}
\includegraphics[clip, width=0.24\hsize]{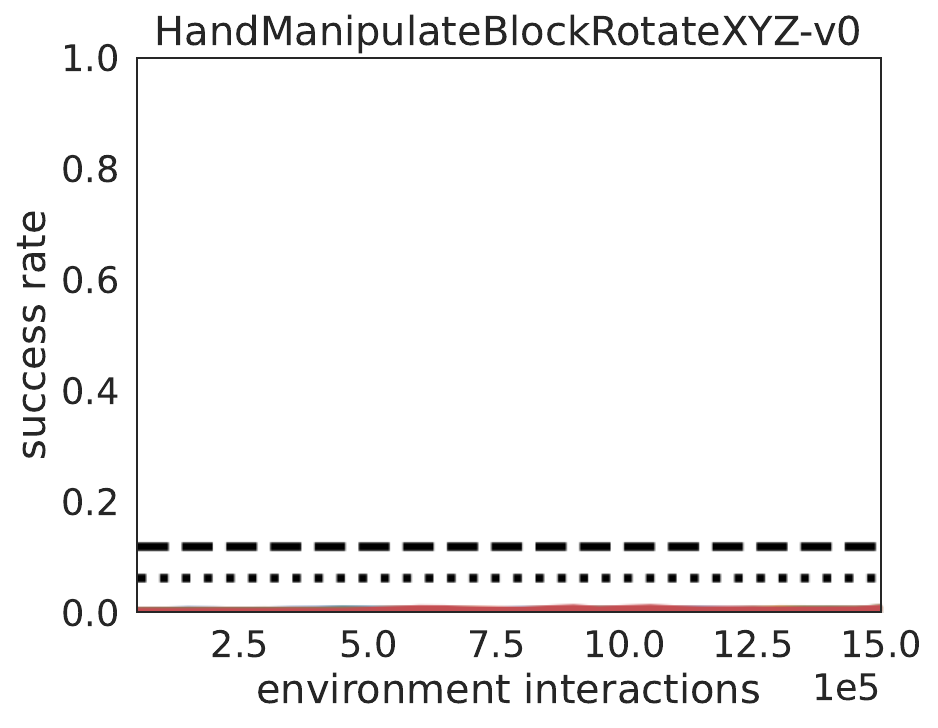}
\includegraphics[clip, width=0.24\hsize]{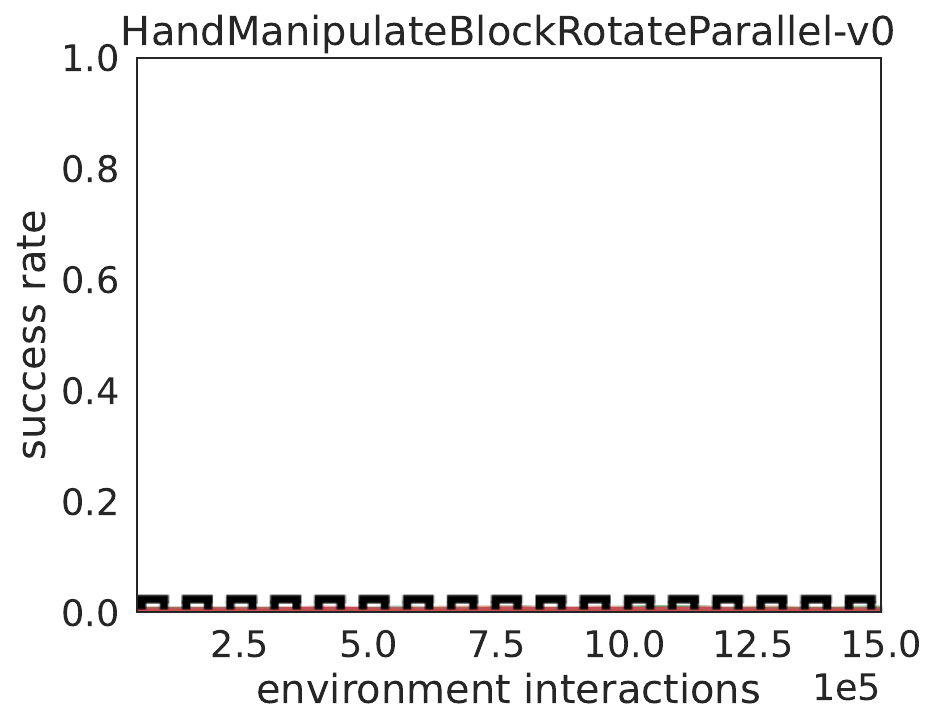}
\end{minipage}
\vspace{-0.7\baselineskip}
\caption{
The effect of replacing REDQ with Reset on performance (success rate). 
}
\label{fig:app-reset-sr}
\vspace{-0.5\baselineskip}
\end{figure}

%
\begin{figure}[h!]
\begin{minipage}{1.0\hsize}
\includegraphics[clip, width=0.24\hsize]{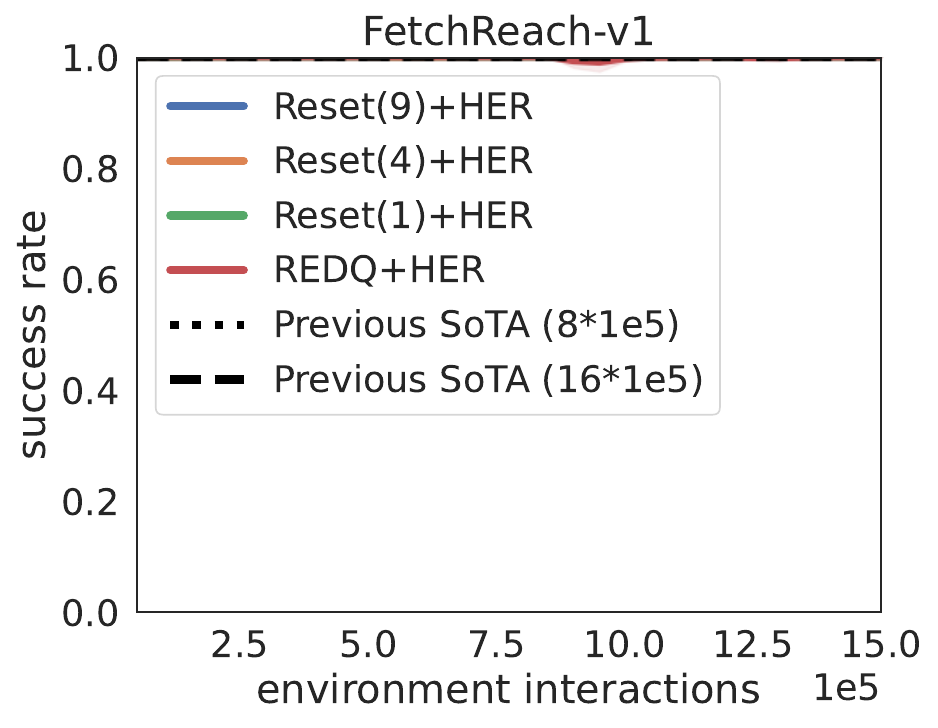}
\includegraphics[clip, width=0.24\hsize]{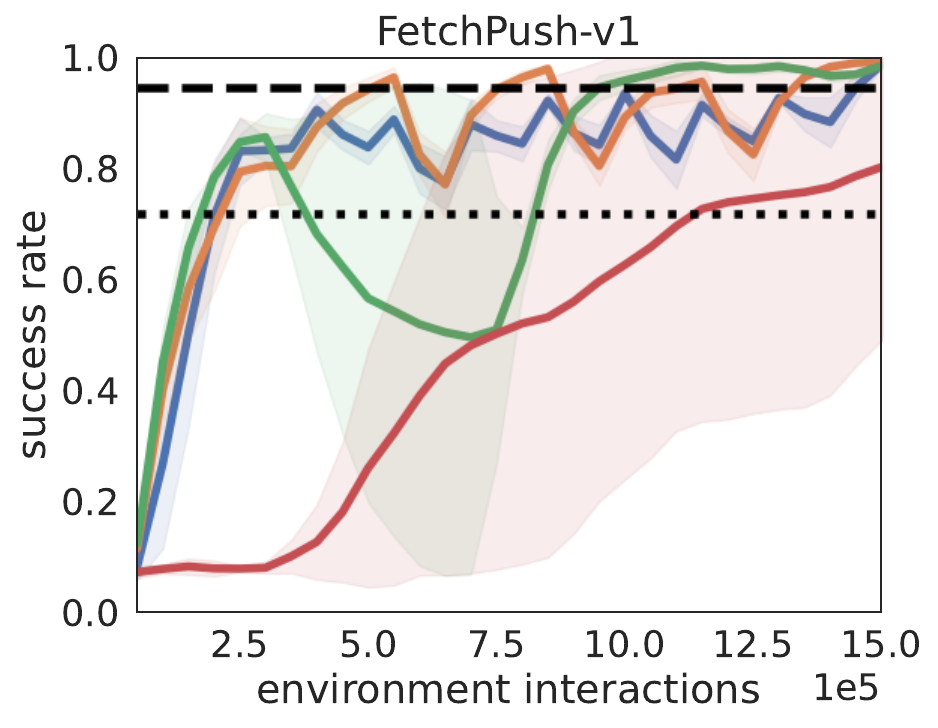}
\includegraphics[clip, width=0.24\hsize]{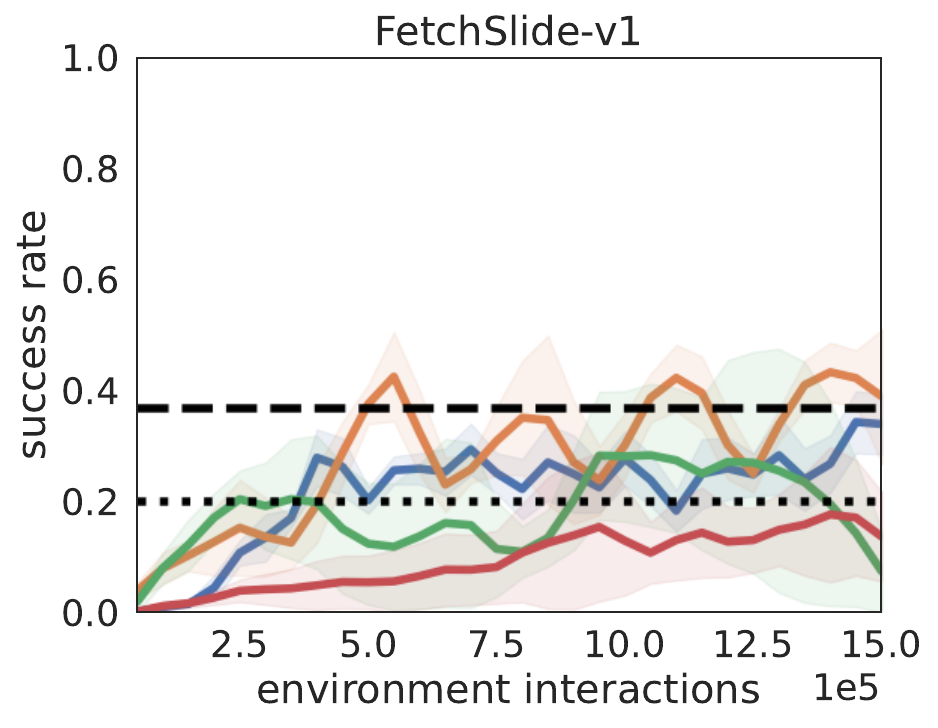}
\includegraphics[clip, width=0.24\hsize]{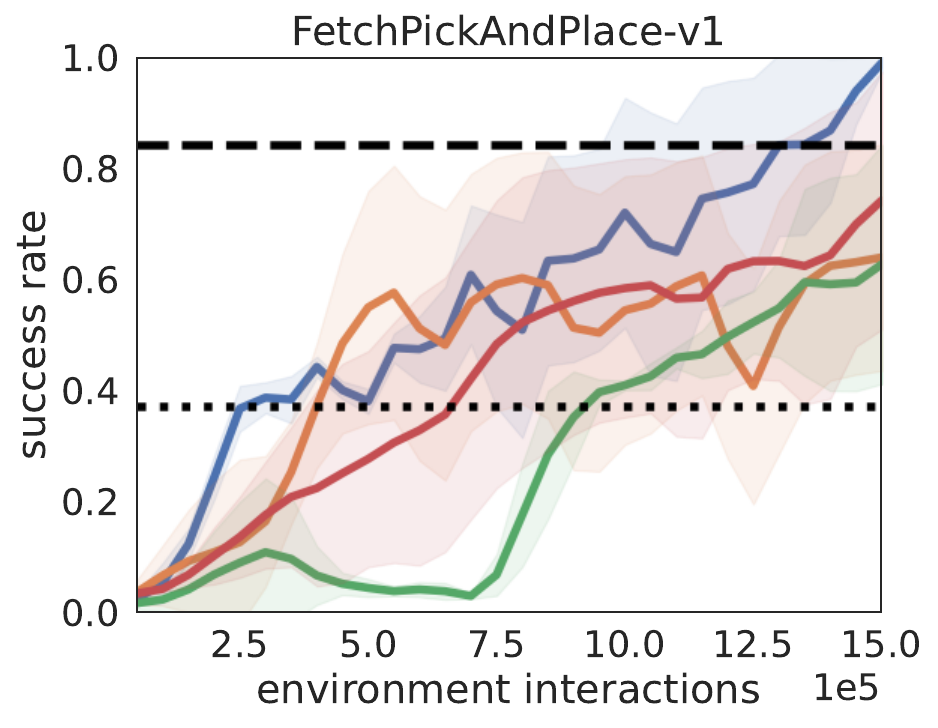}
\end{minipage}
\begin{minipage}{1.0\hsize}
\includegraphics[clip, width=0.24\hsize]{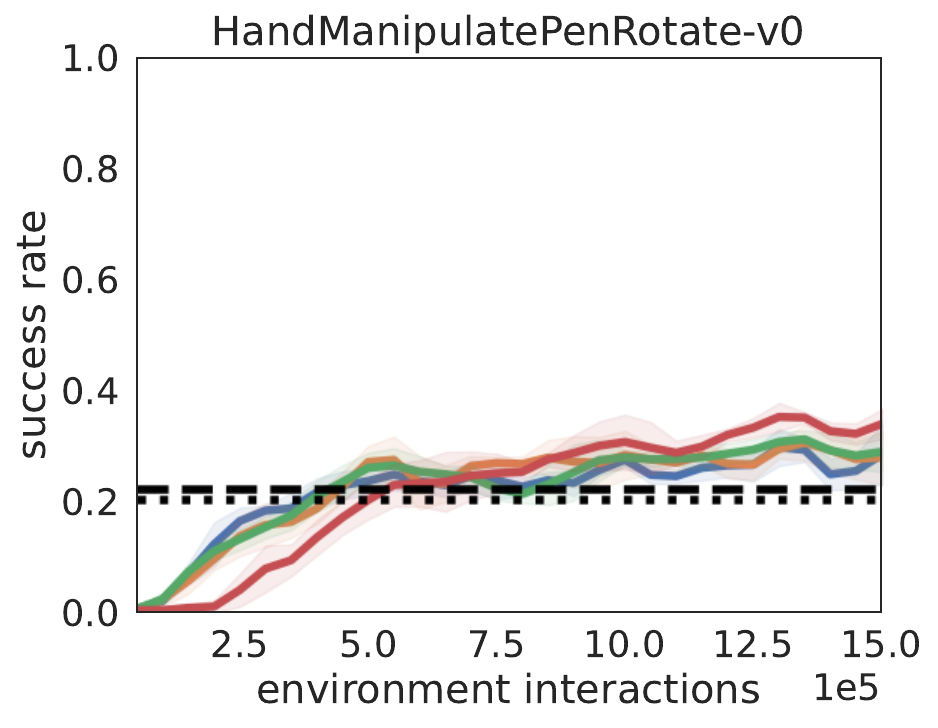}
\includegraphics[clip, width=0.24\hsize]{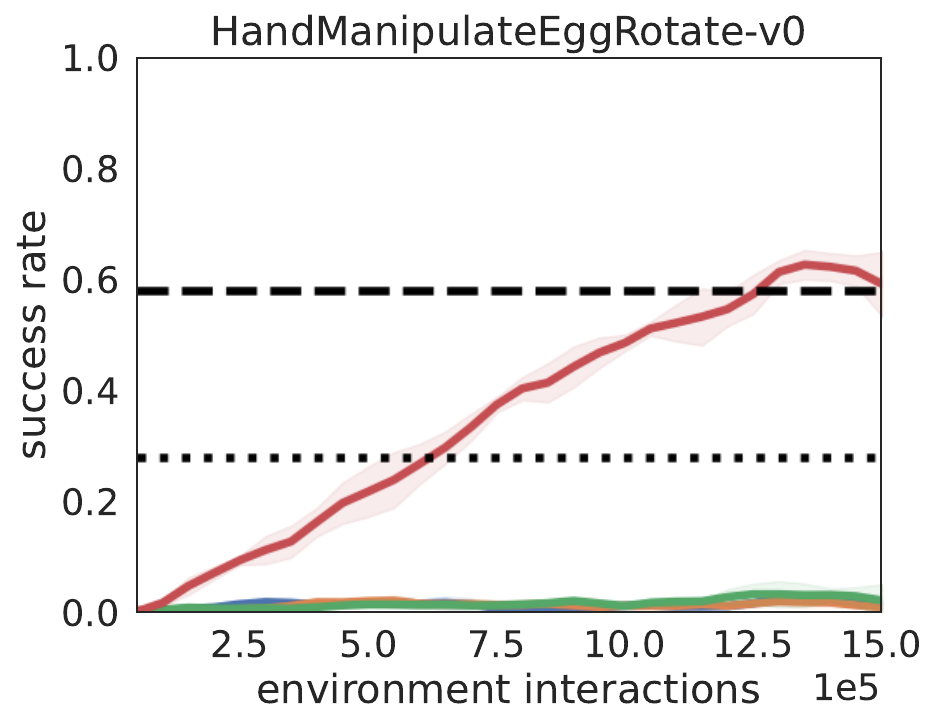}
\includegraphics[clip, width=0.24\hsize]{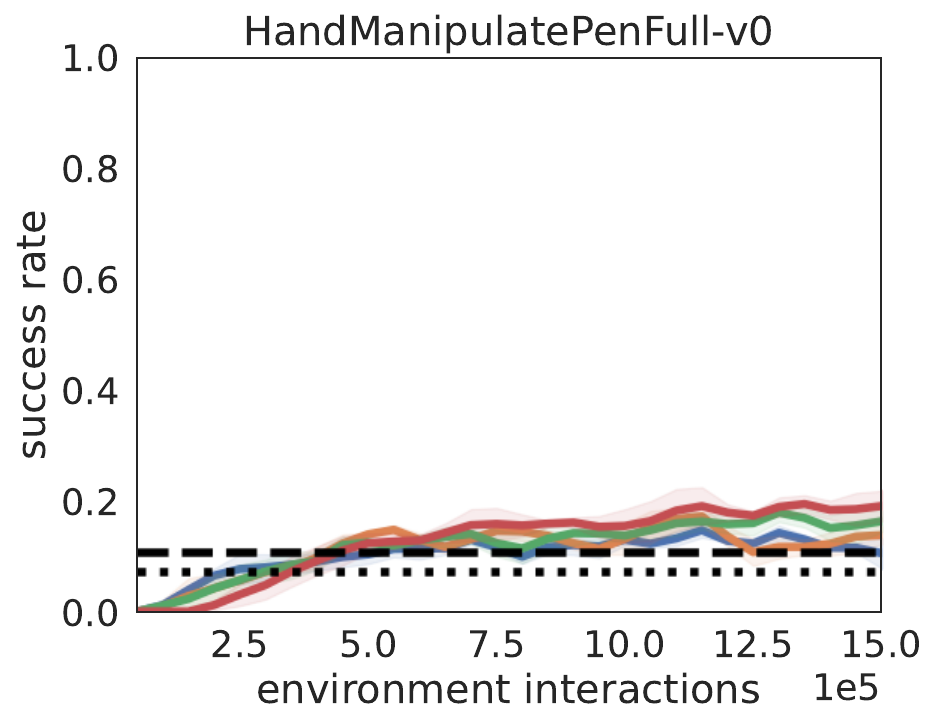}
\includegraphics[clip, width=0.24\hsize]{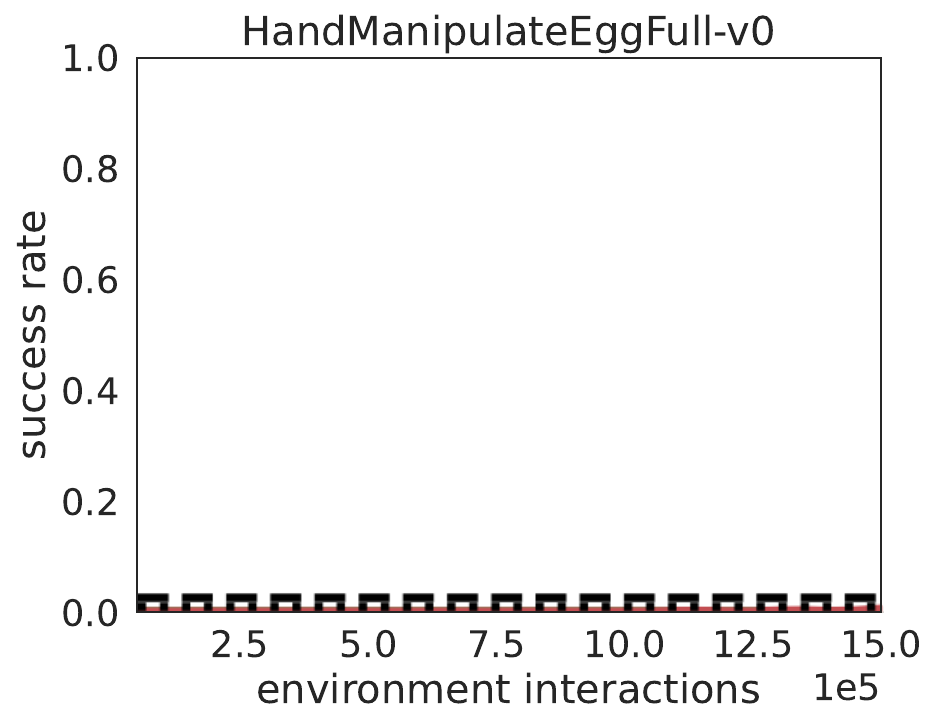}
\end{minipage}
\begin{minipage}{1.0\hsize}
\includegraphics[clip, width=0.24\hsize]{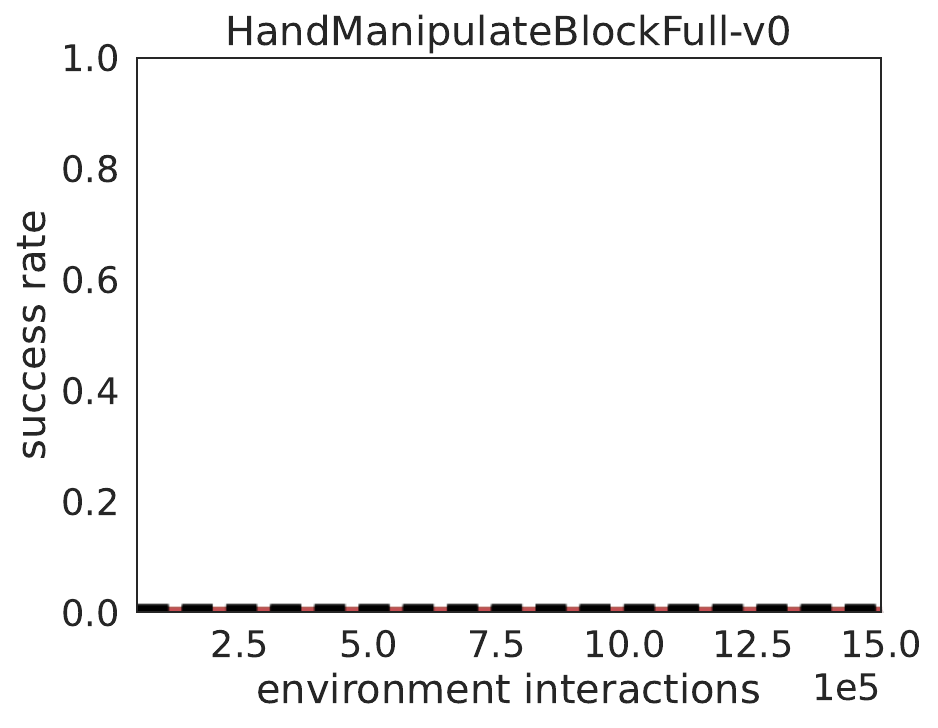}
\includegraphics[clip, width=0.24\hsize]{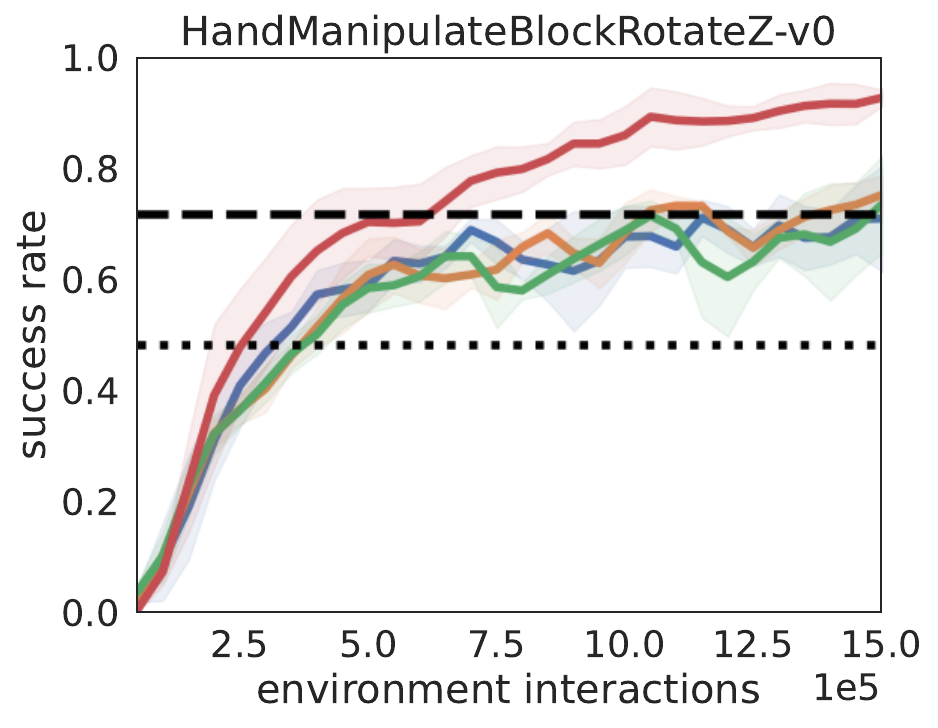}
\includegraphics[clip, width=0.24\hsize]{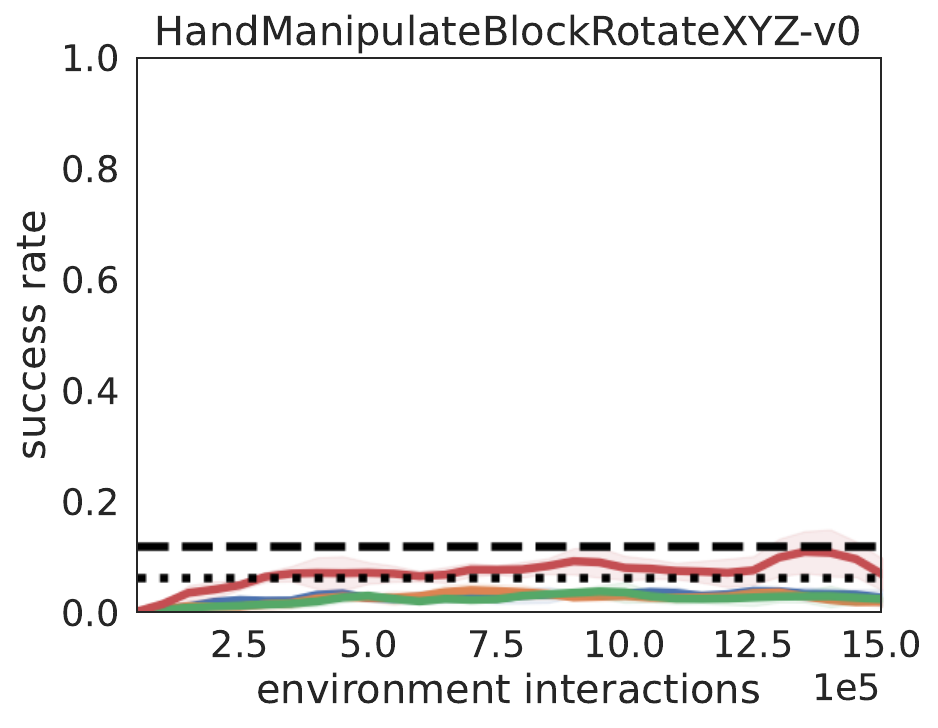}
\includegraphics[clip, width=0.24\hsize]{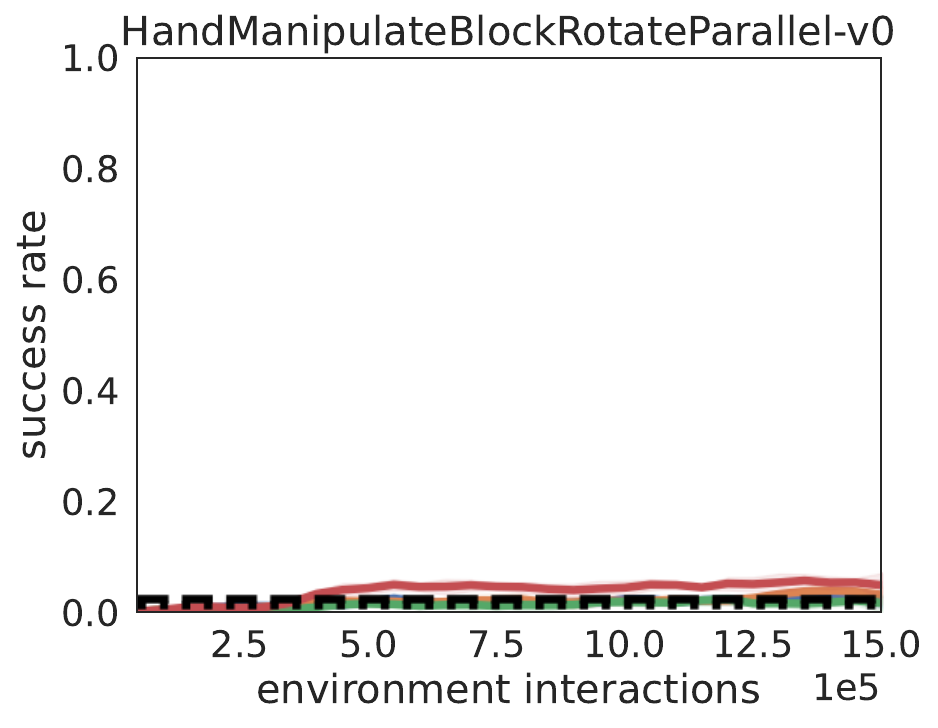}
\end{minipage}
\vspace{-0.7\baselineskip}
\caption{
The effect of replacing REDQ+HER with Reset+HER on performance (success rate). 
}
\label{fig:app-reset-her-sr}
\vspace{-0.5\baselineskip}
\end{figure}

\begin{figure}[h!]
\begin{minipage}{1.0\hsize}
\includegraphics[clip, width=0.24\hsize]{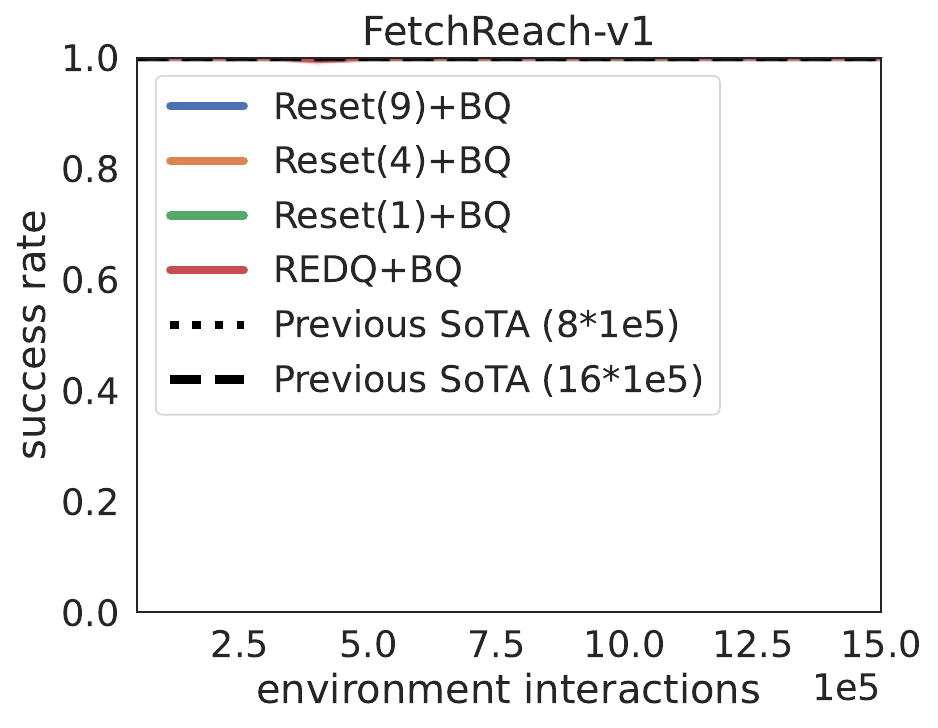}
\includegraphics[clip, width=0.24\hsize]{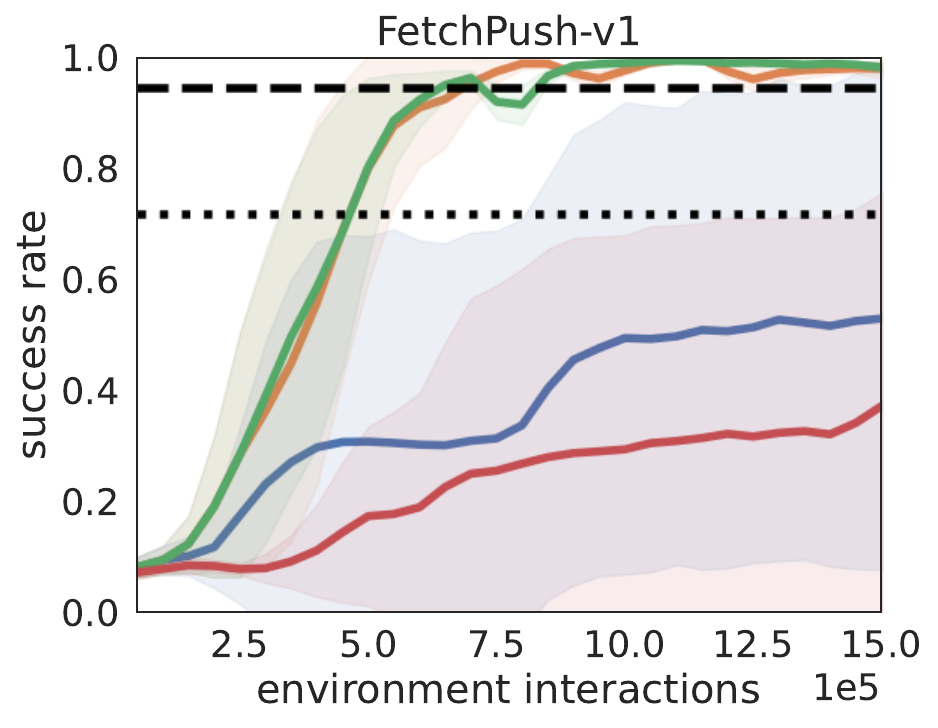}
\includegraphics[clip, width=0.24\hsize]{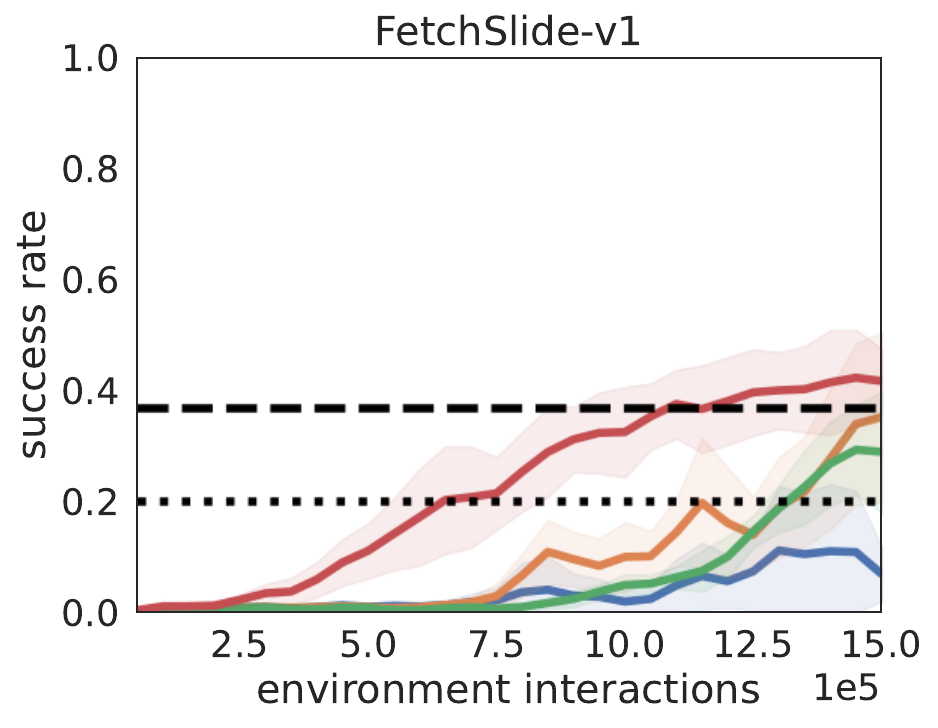}
\includegraphics[clip, width=0.24\hsize]{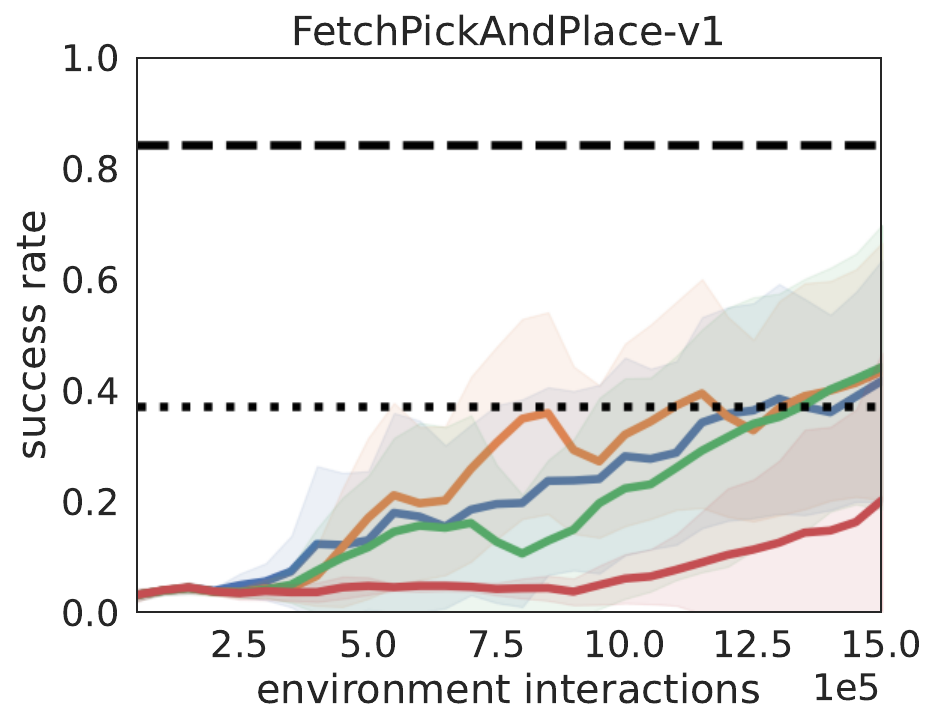}
\end{minipage}
\begin{minipage}{1.0\hsize}
\includegraphics[clip, width=0.24\hsize]{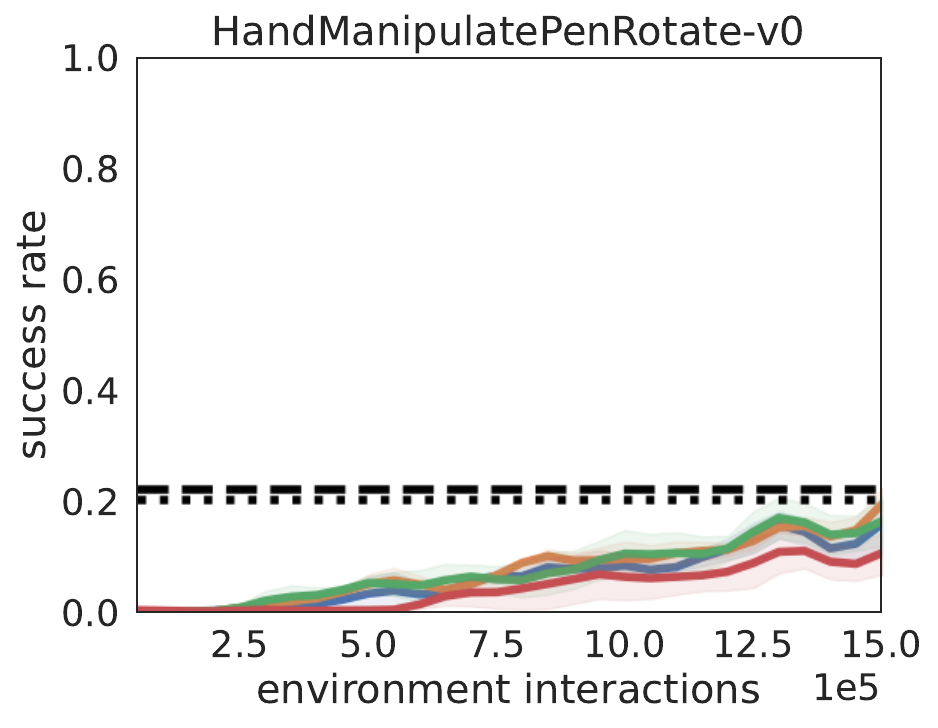}
\includegraphics[clip, width=0.24\hsize]{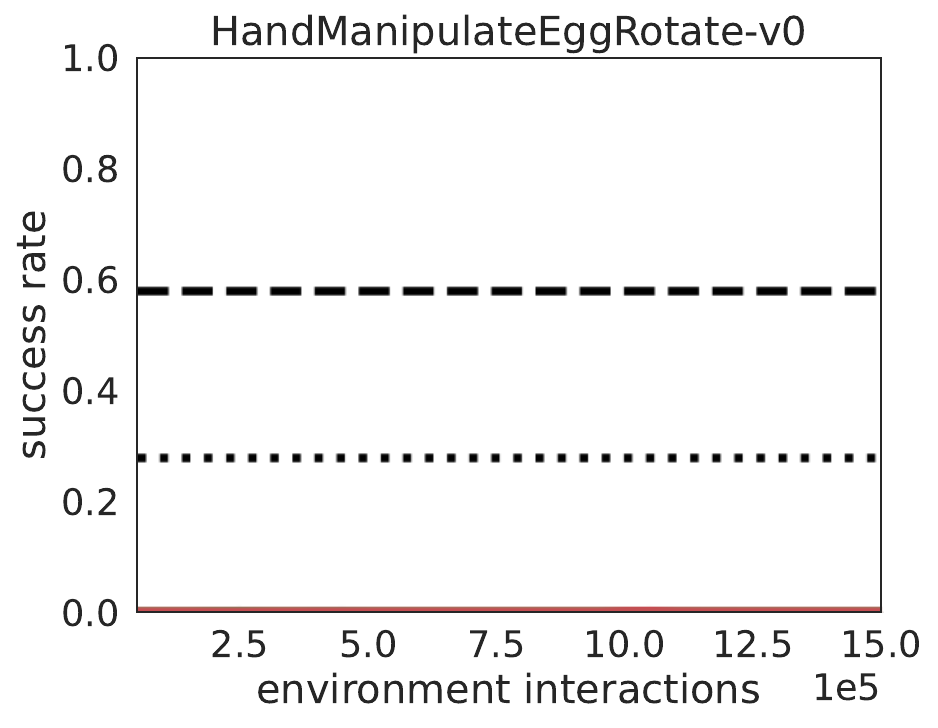}
\includegraphics[clip, width=0.24\hsize]{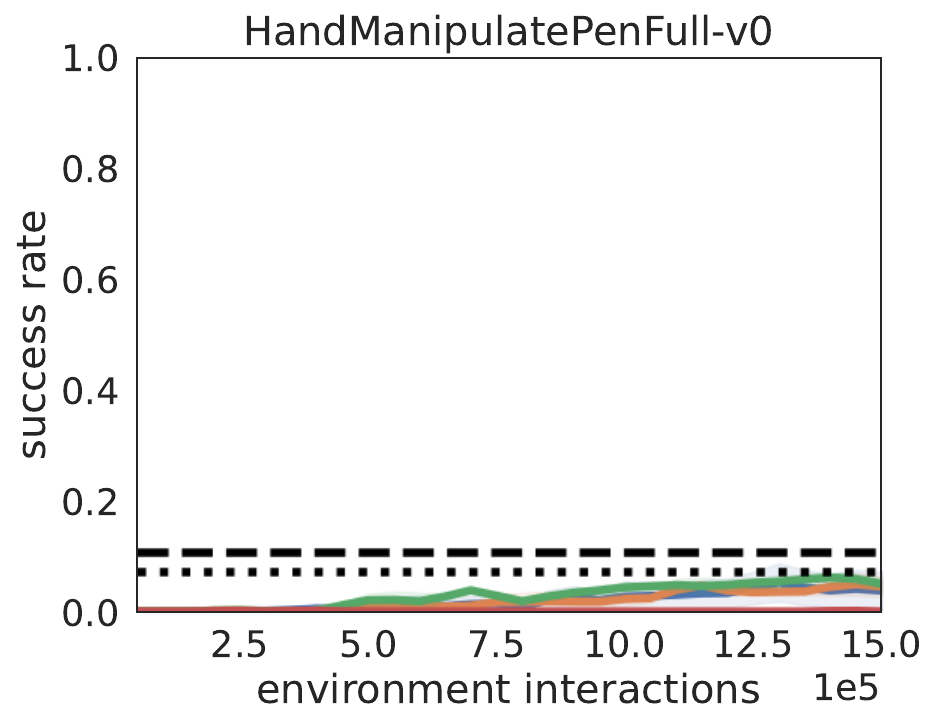}
\includegraphics[clip, width=0.24\hsize]{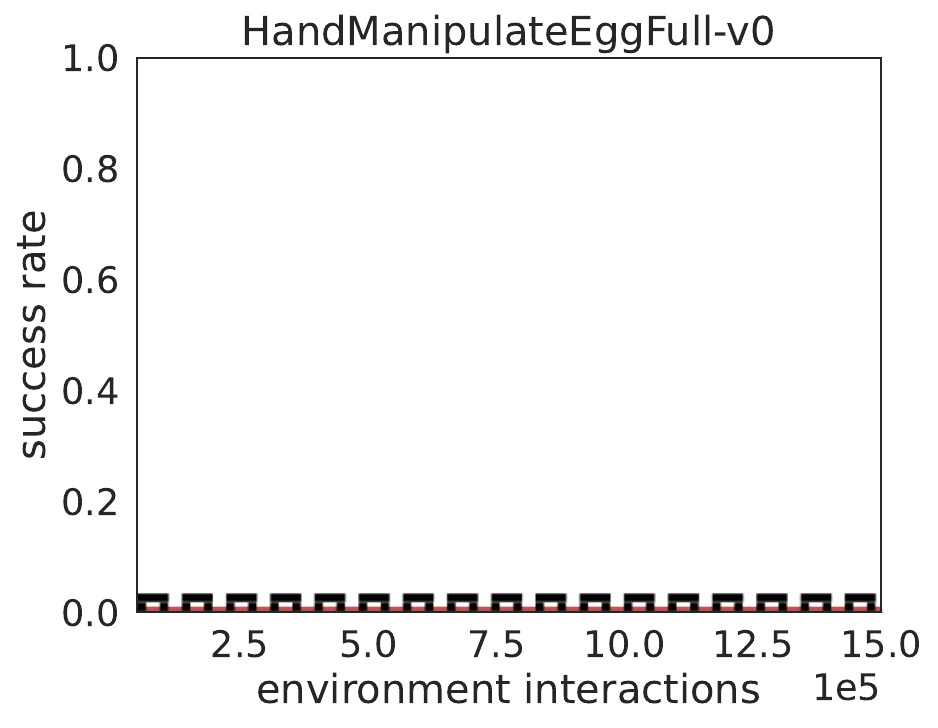}
\end{minipage}
\begin{minipage}{1.0\hsize}
\includegraphics[clip, width=0.24\hsize]{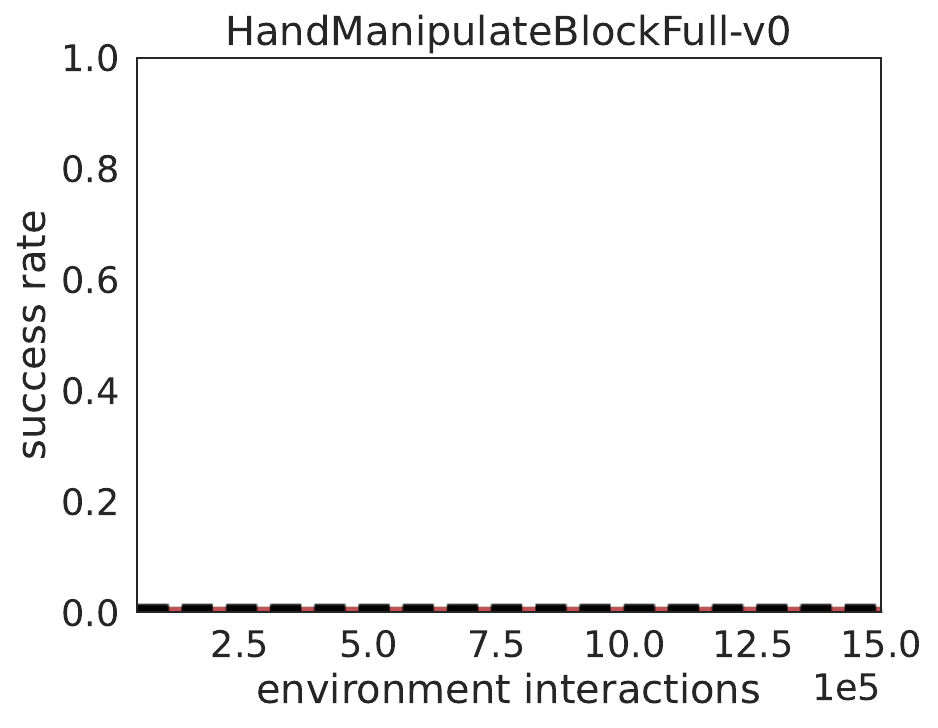}
\includegraphics[clip, width=0.24\hsize]{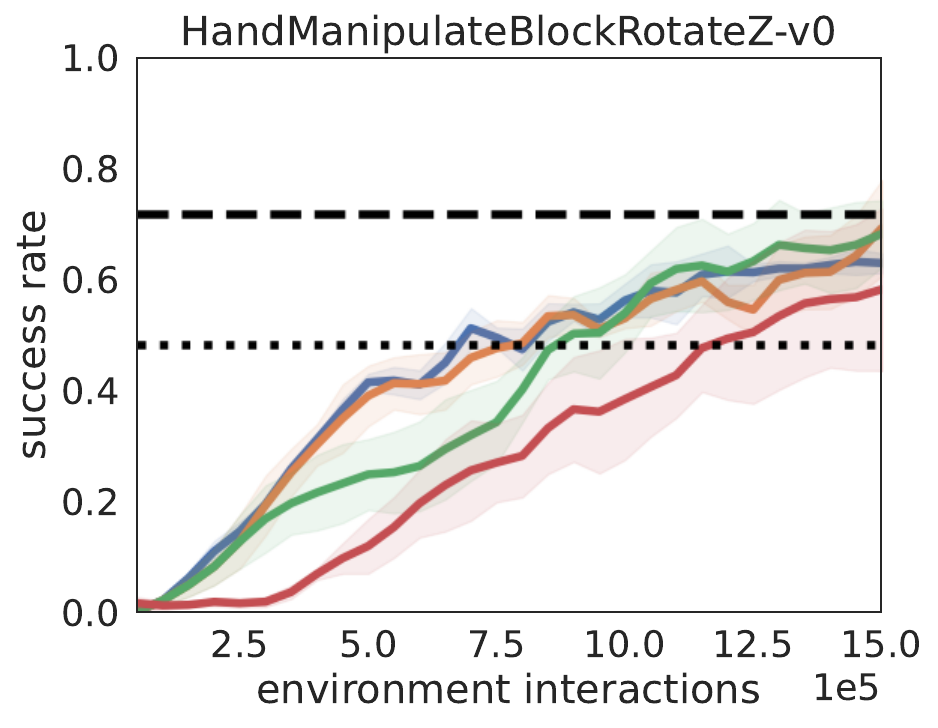}
\includegraphics[clip, width=0.24\hsize]{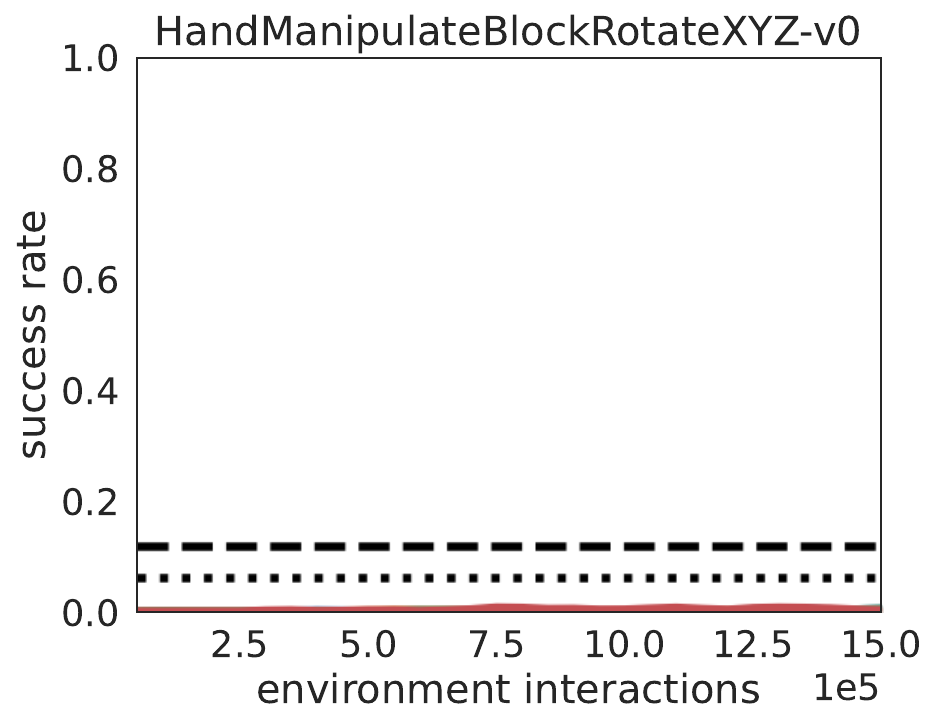}
\includegraphics[clip, width=0.24\hsize]{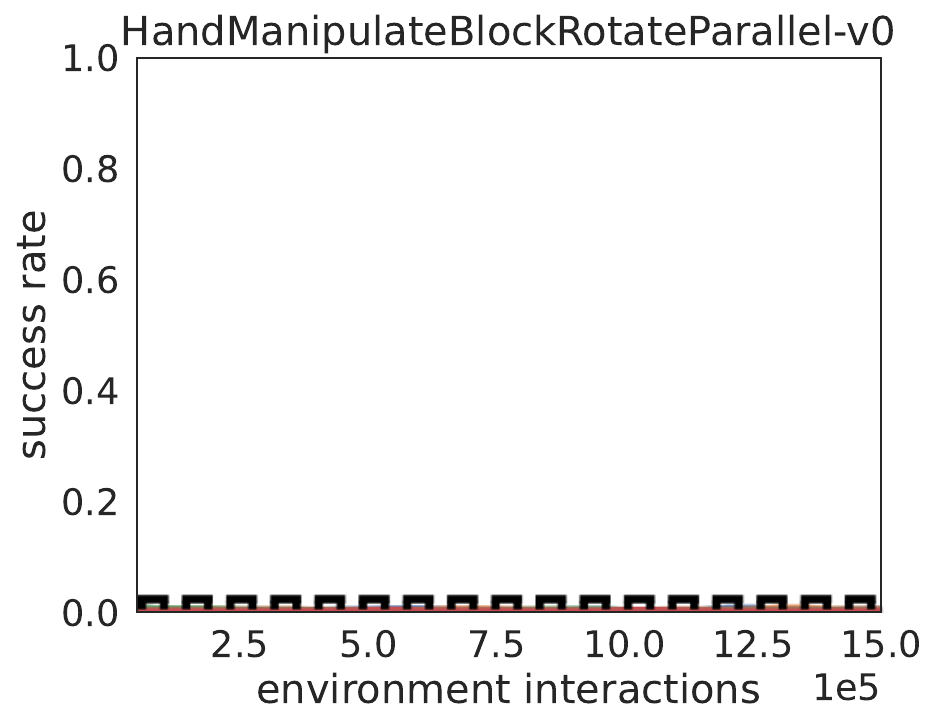}
\end{minipage}
\vspace{-0.7\baselineskip}
\caption{
The effect of replacing REDQ+BQ with Reset+BQ on performance (success rate). 
}
\label{fig:app-reset-bq-sr}
\vspace{-0.5\baselineskip}
\end{figure}

\begin{figure}[h!]
\begin{minipage}{1.0\hsize}
\includegraphics[clip, width=0.24\hsize]{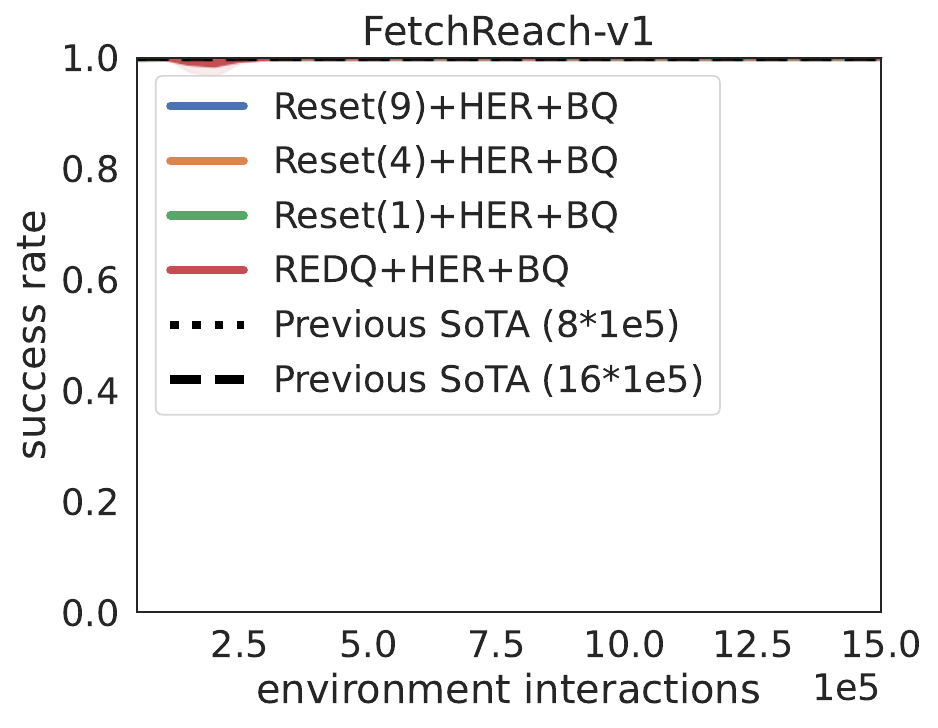}
\includegraphics[clip, width=0.24\hsize]{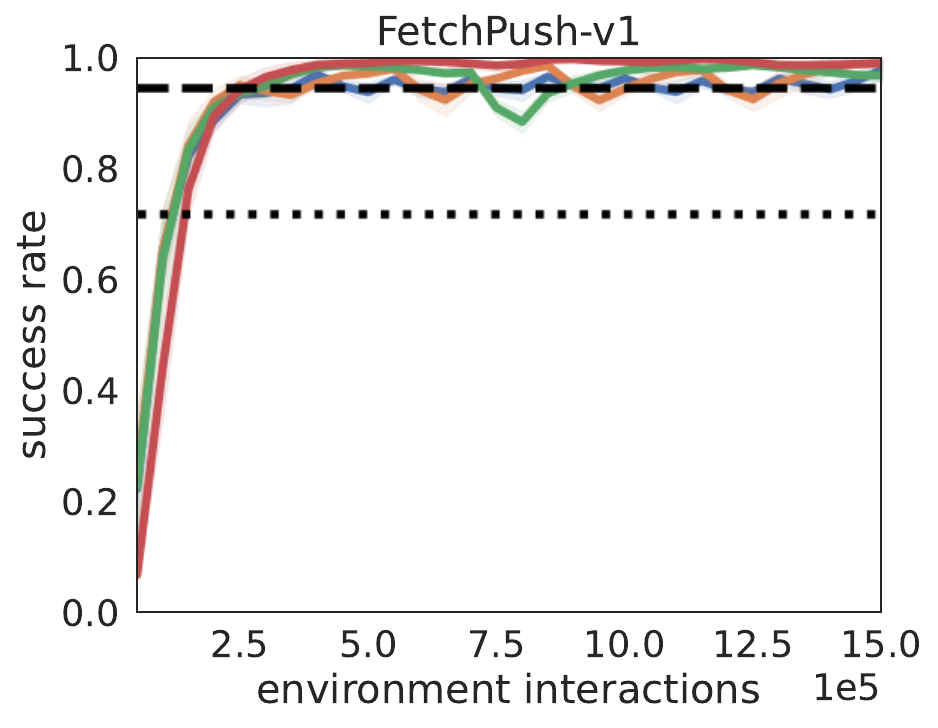}
\includegraphics[clip, width=0.24\hsize]{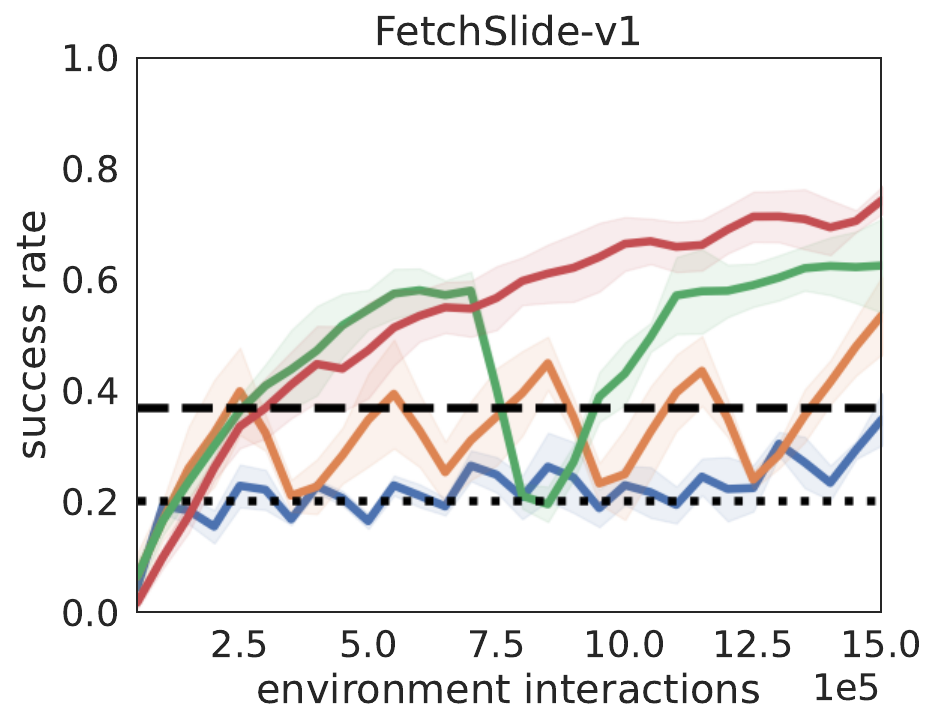}
\includegraphics[clip, width=0.24\hsize]{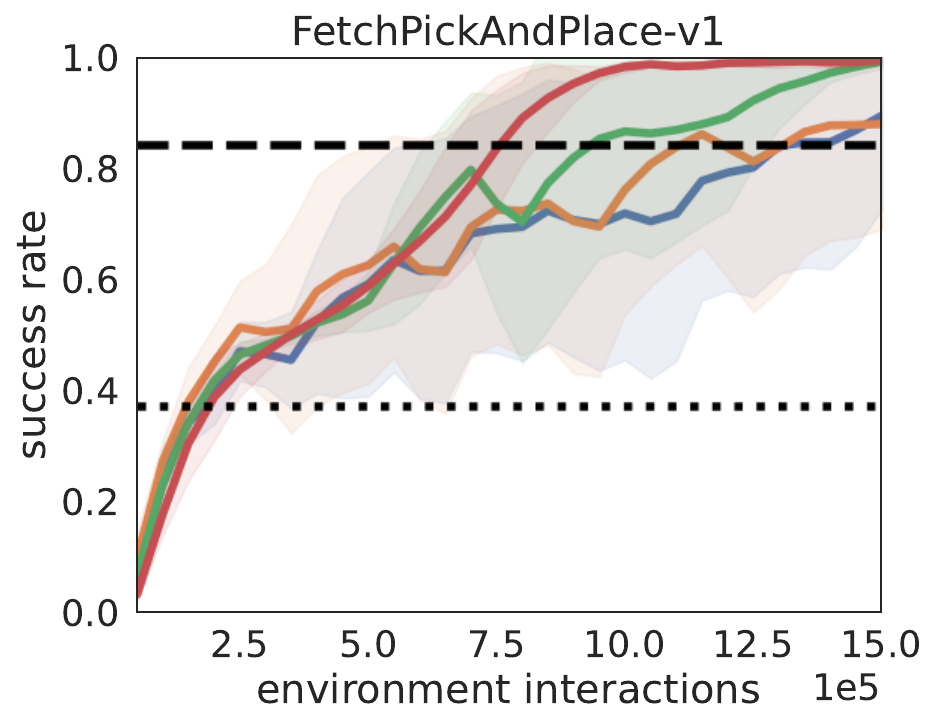}
\end{minipage}
\begin{minipage}{1.0\hsize}
\includegraphics[clip, width=0.24\hsize]{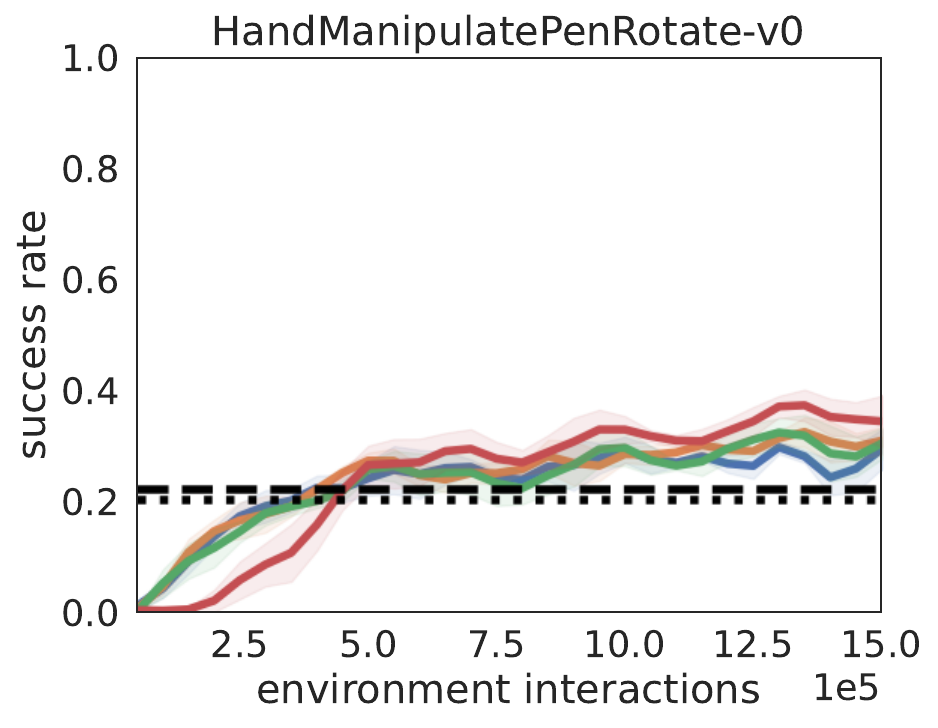}
\includegraphics[clip, width=0.24\hsize]{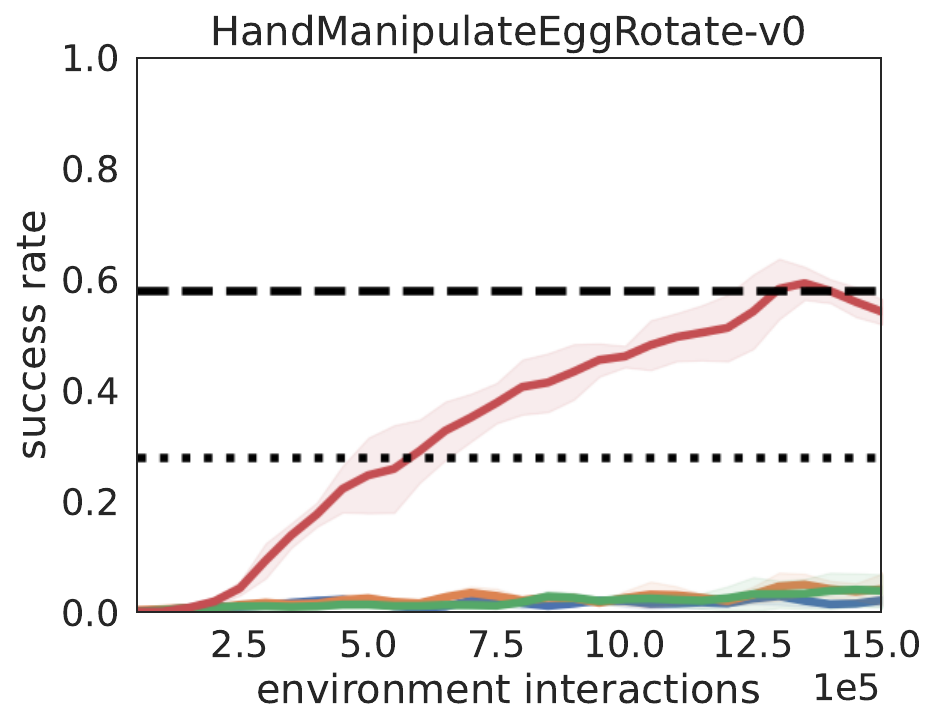}
\includegraphics[clip, width=0.24\hsize]{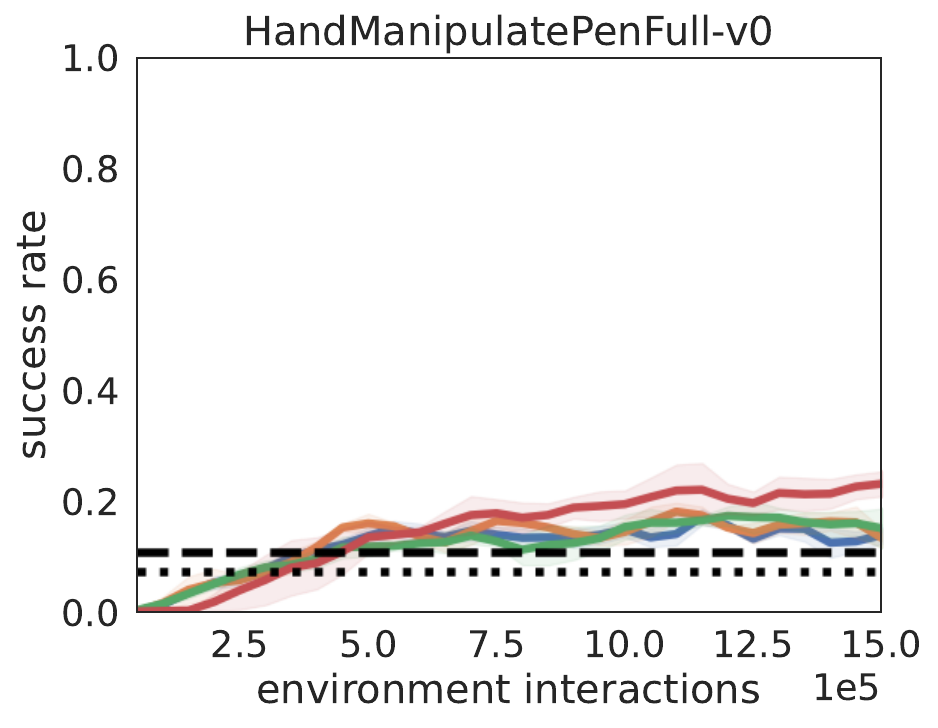}
\includegraphics[clip, width=0.24\hsize]{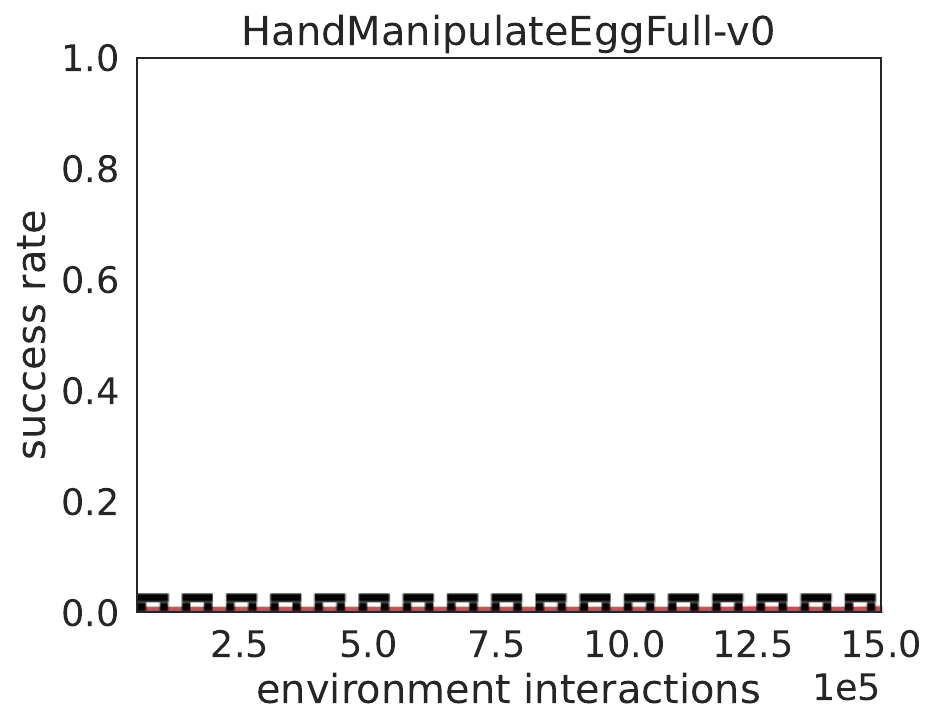}
\end{minipage}
\begin{minipage}{1.0\hsize}
\includegraphics[clip, width=0.24\hsize]{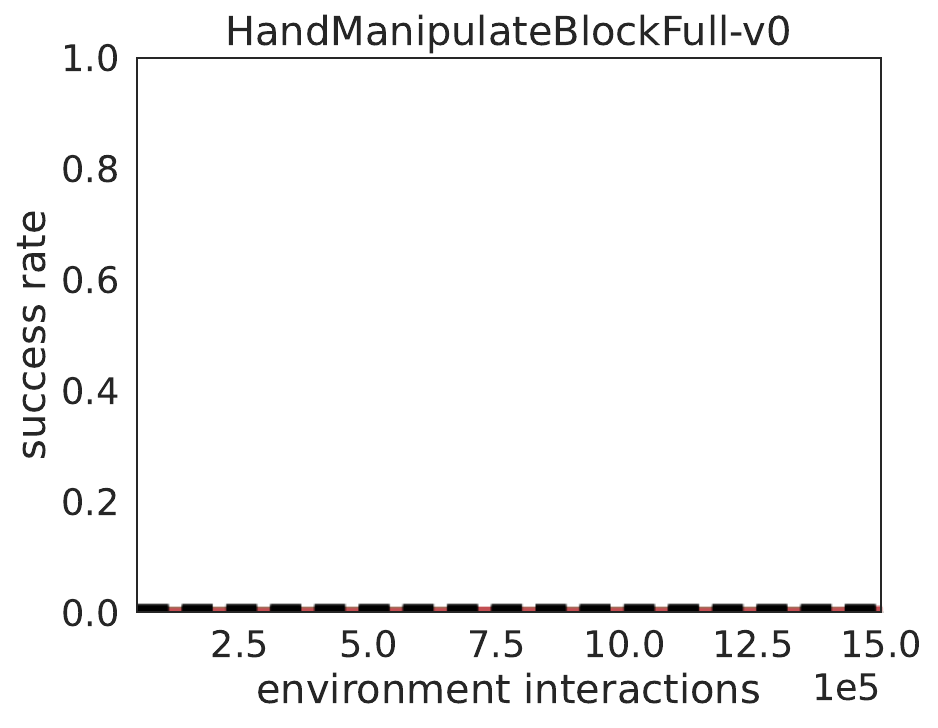}
\includegraphics[clip, width=0.24\hsize]{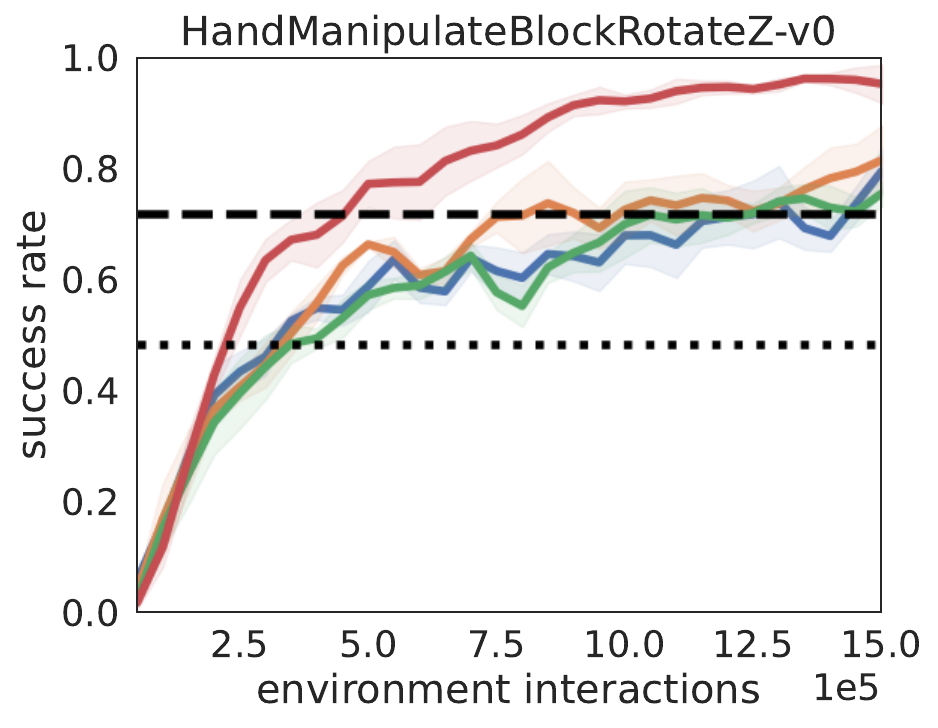}
\includegraphics[clip, width=0.24\hsize]{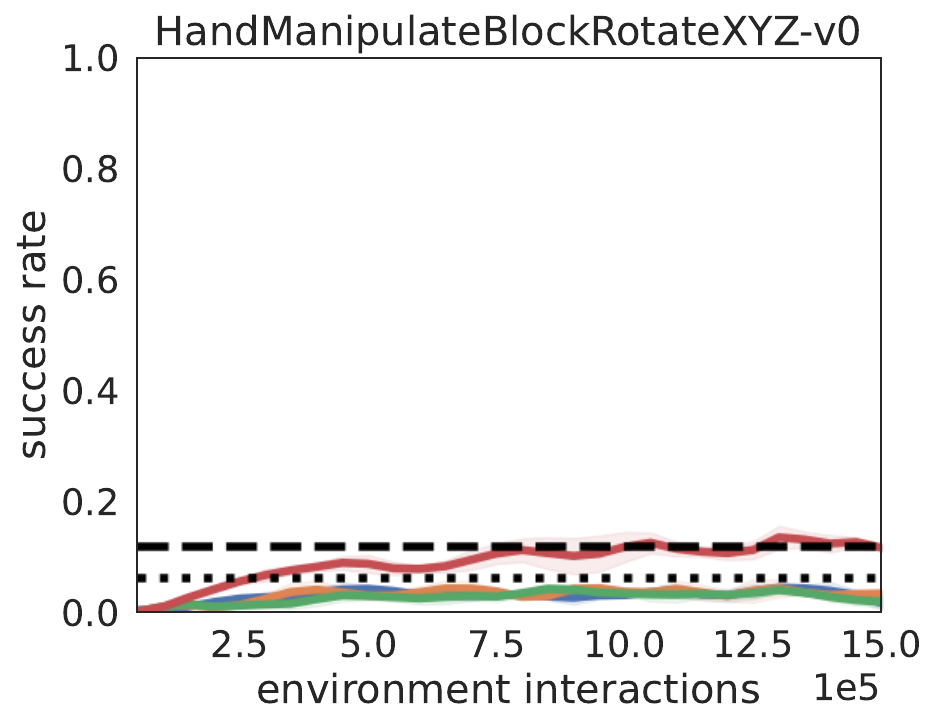}
\includegraphics[clip, width=0.24\hsize]{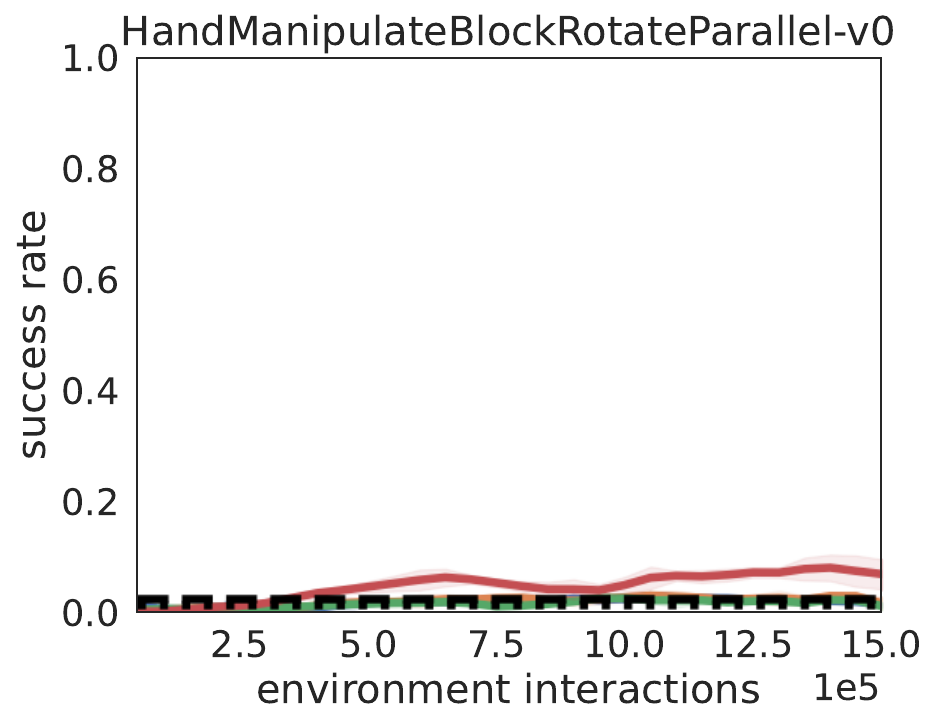}
\end{minipage}
\vspace{-0.7\baselineskip}
\caption{
The effect of replacing REDQ+HER with Reset+HER on performance (success rate). 
}
\label{fig:app-reset-her-bq-sr}
\vspace{-0.5\baselineskip}
\end{figure}

\begin{figure*}[h!]
\begin{minipage}{1.0\hsize}
\includegraphics[clip, width=0.24\hsize]{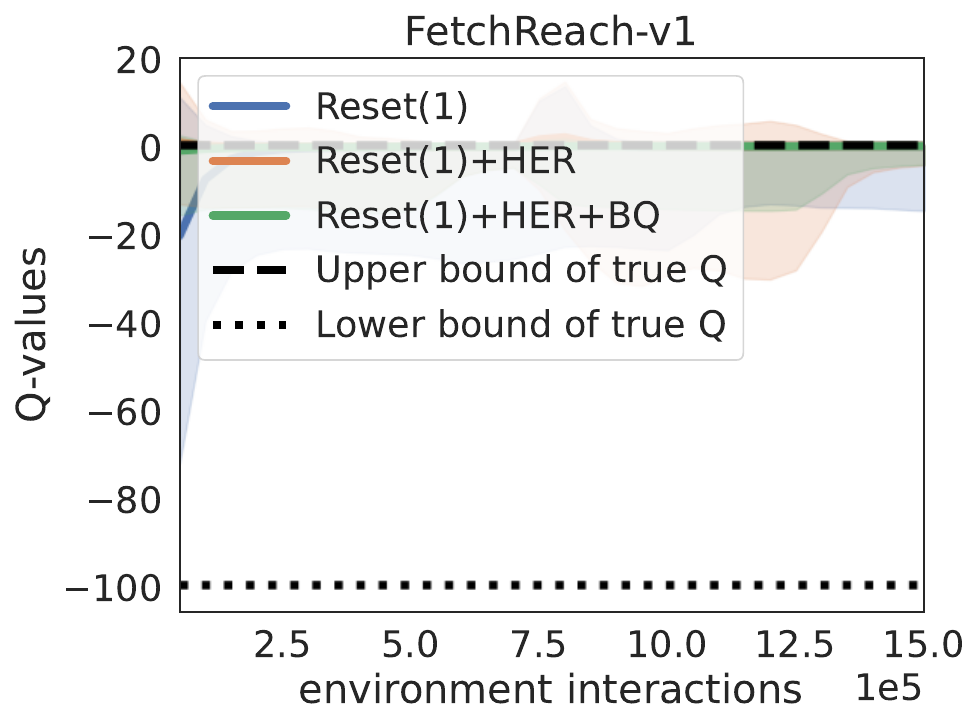}
\includegraphics[clip, width=0.24\hsize]{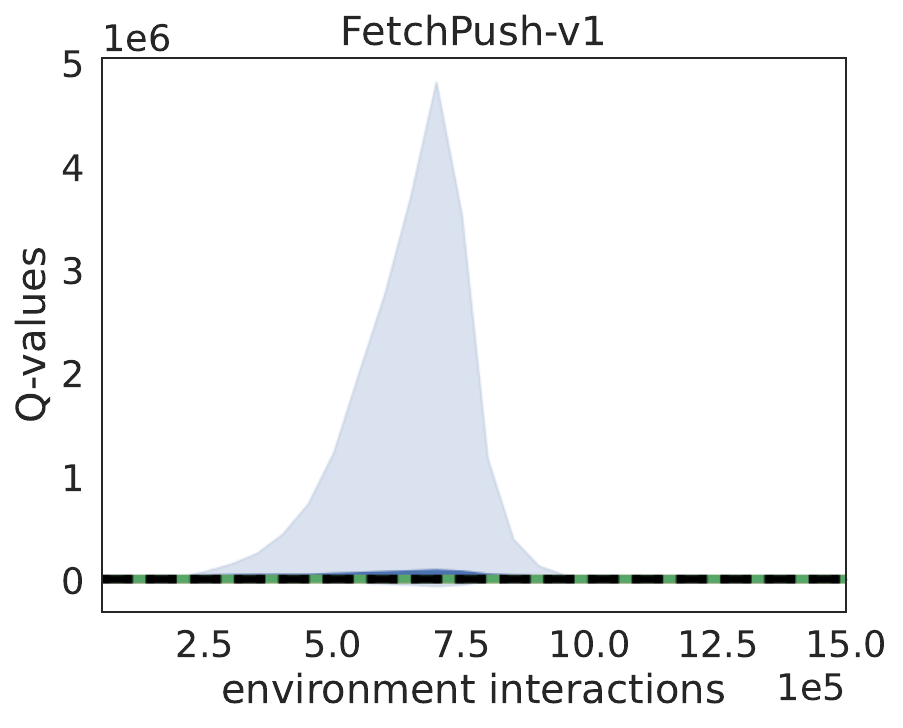}
\includegraphics[clip, width=0.24\hsize]{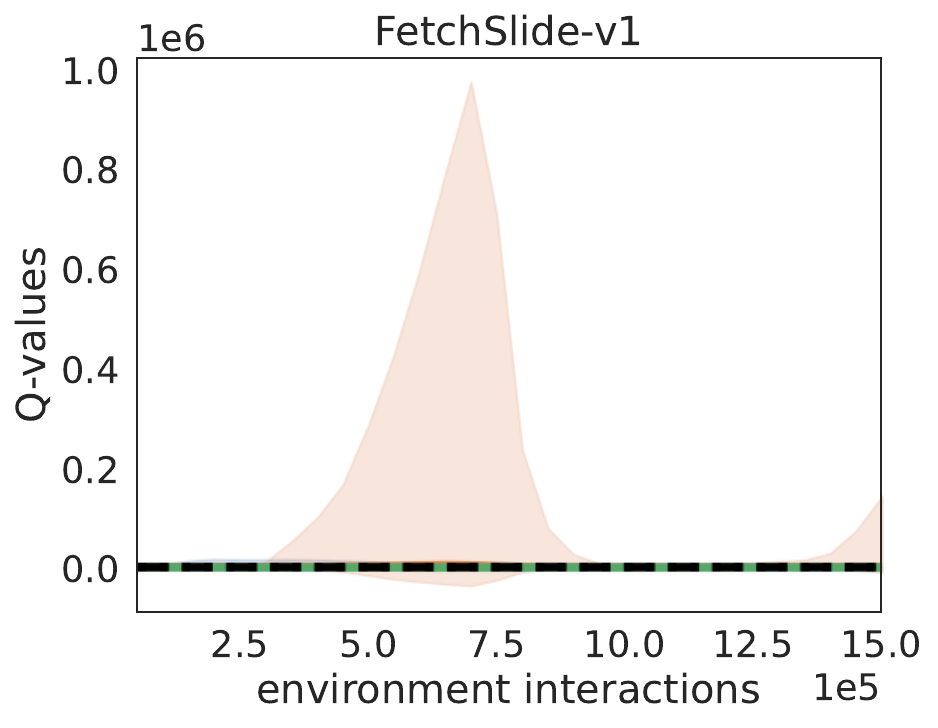}
\includegraphics[clip, width=0.24\hsize]{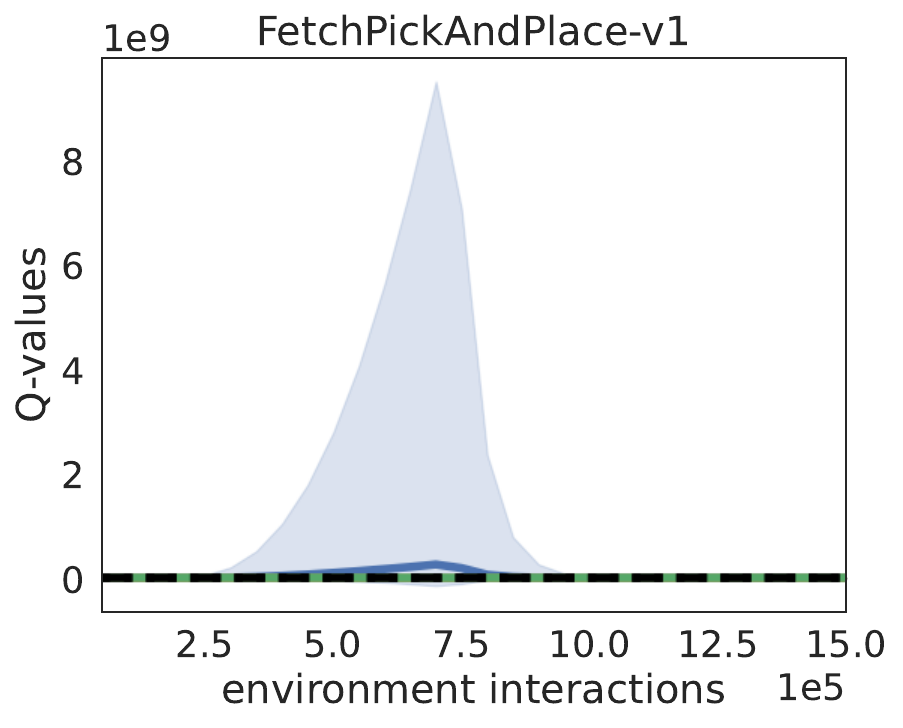}
\end{minipage}
\begin{minipage}{1.0\hsize}
\includegraphics[clip, width=0.24\hsize]{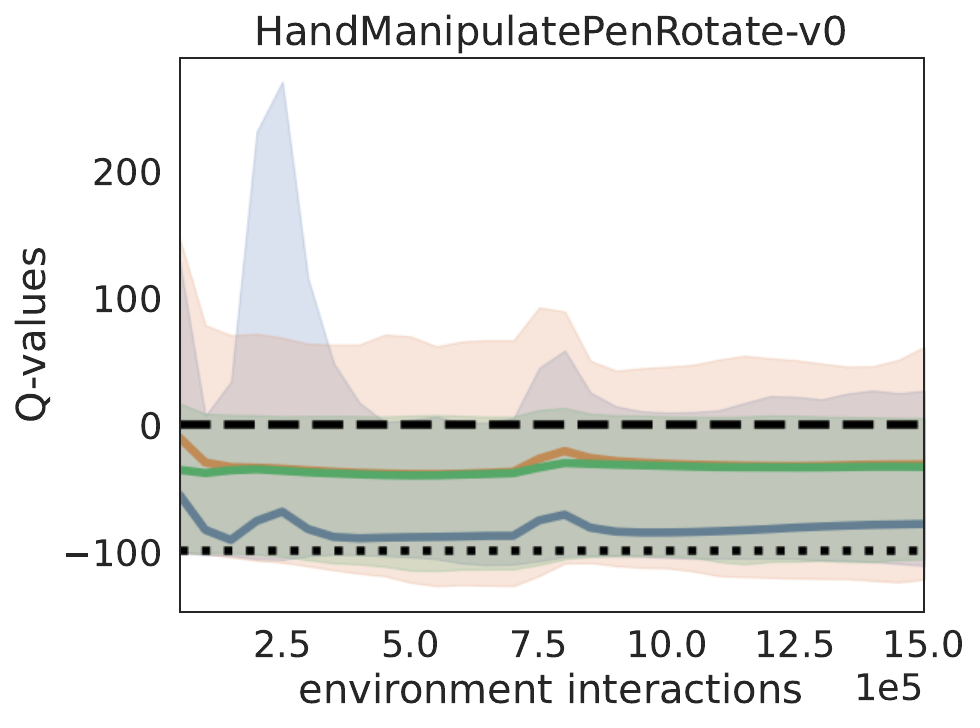}
\includegraphics[clip, width=0.24\hsize]{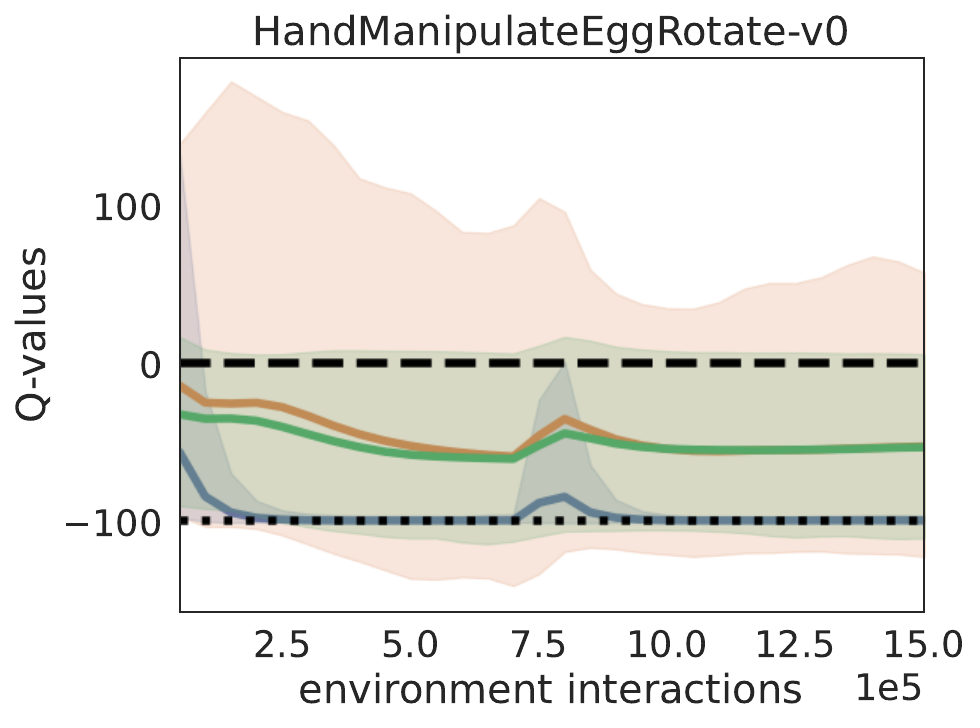}
\includegraphics[clip, width=0.24\hsize]{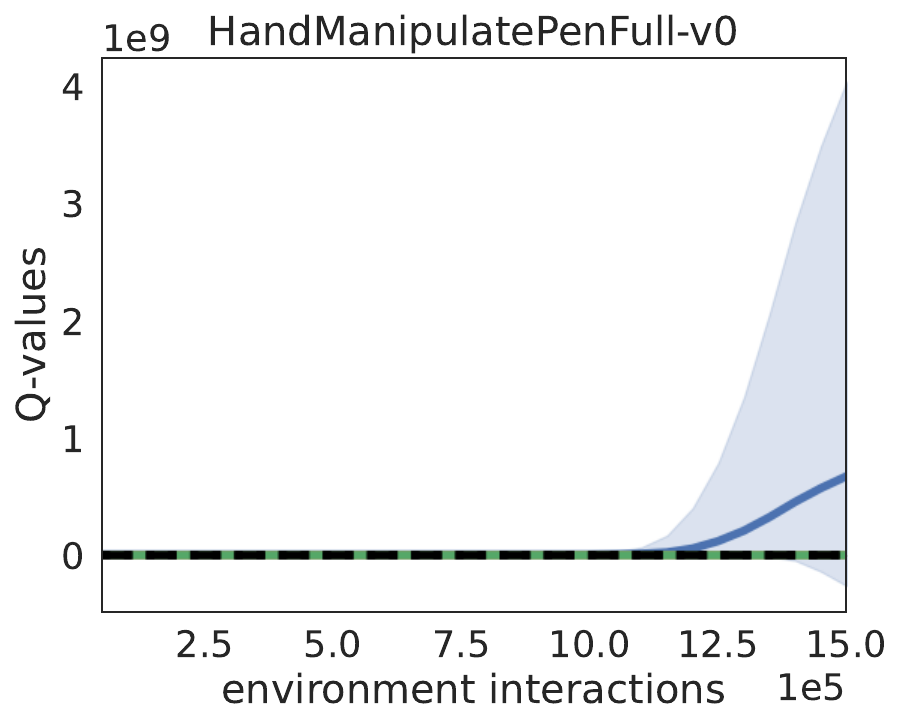}
\includegraphics[clip, width=0.24\hsize]{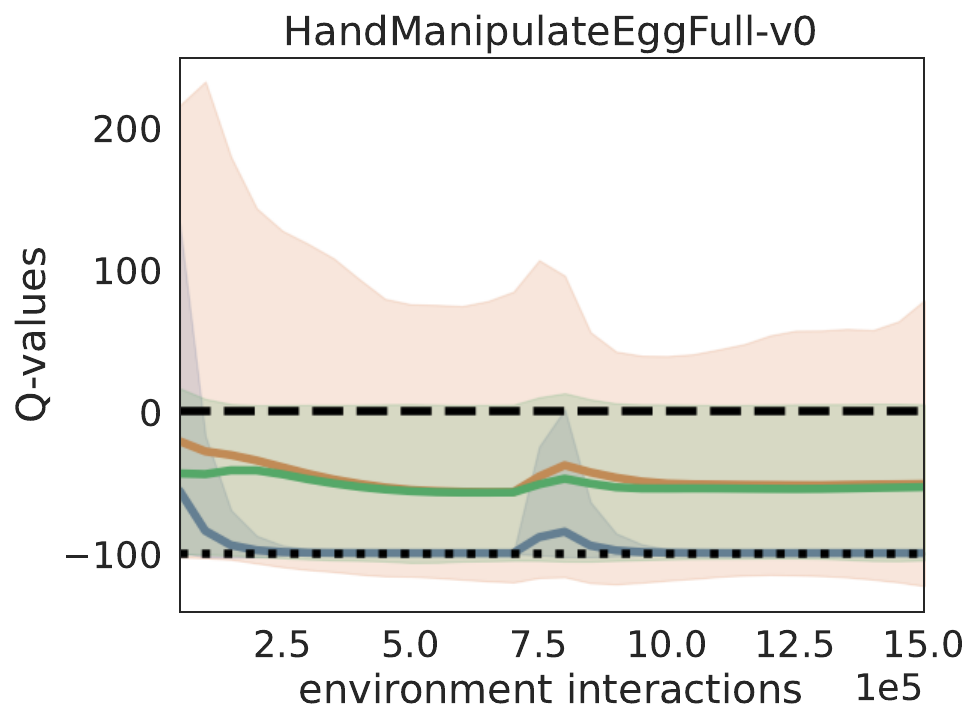}
\end{minipage}
\begin{minipage}{1.0\hsize}
\includegraphics[clip, width=0.24\hsize]{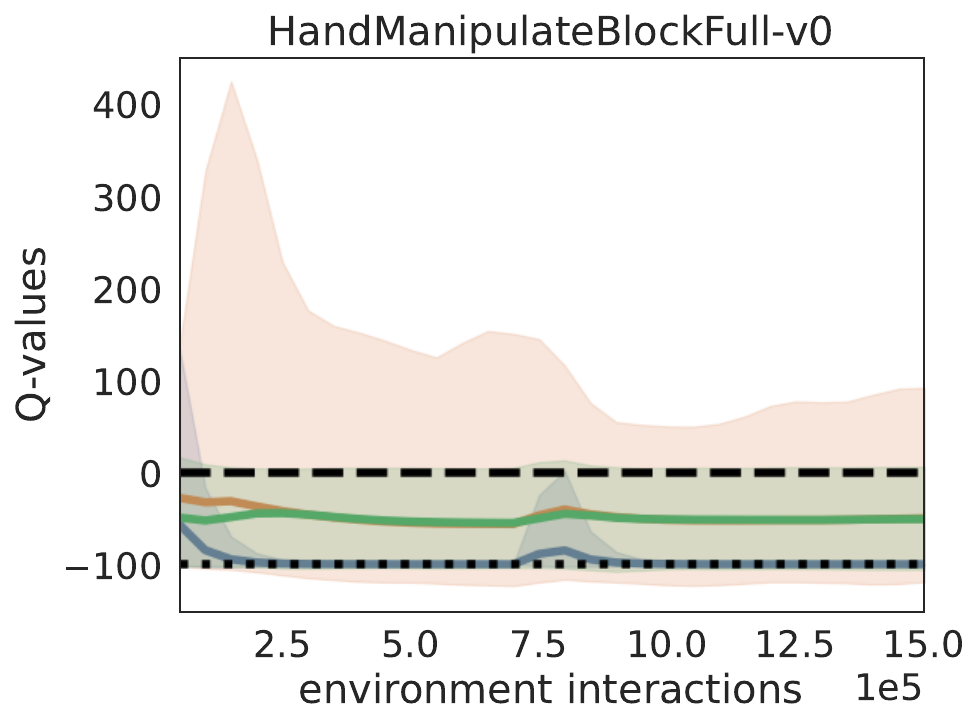}
\includegraphics[clip, width=0.24\hsize]{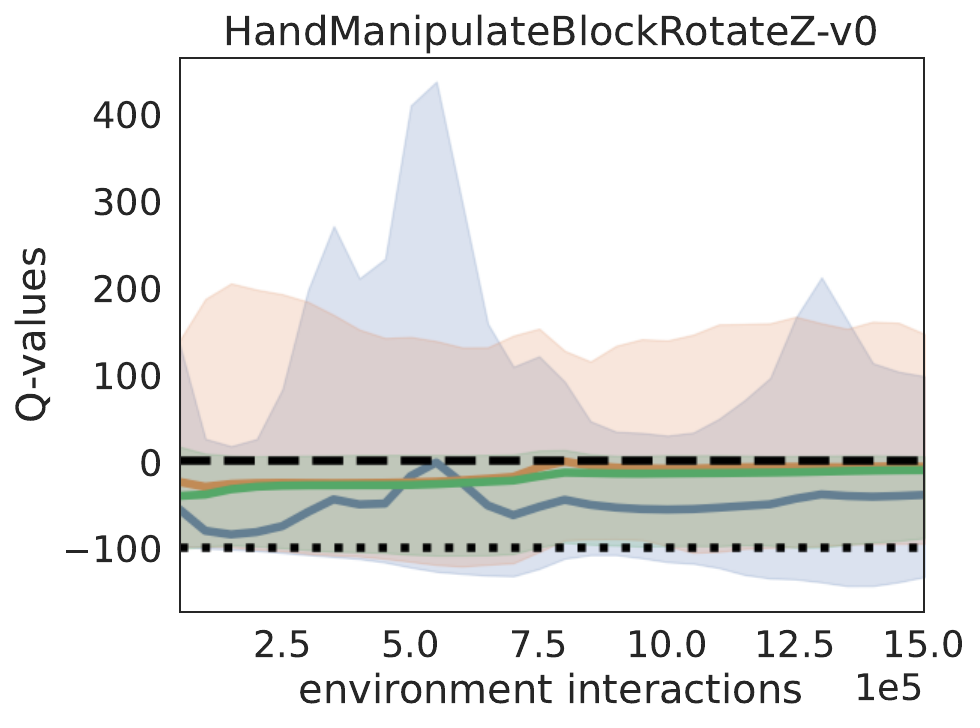}
\includegraphics[clip, width=0.24\hsize]{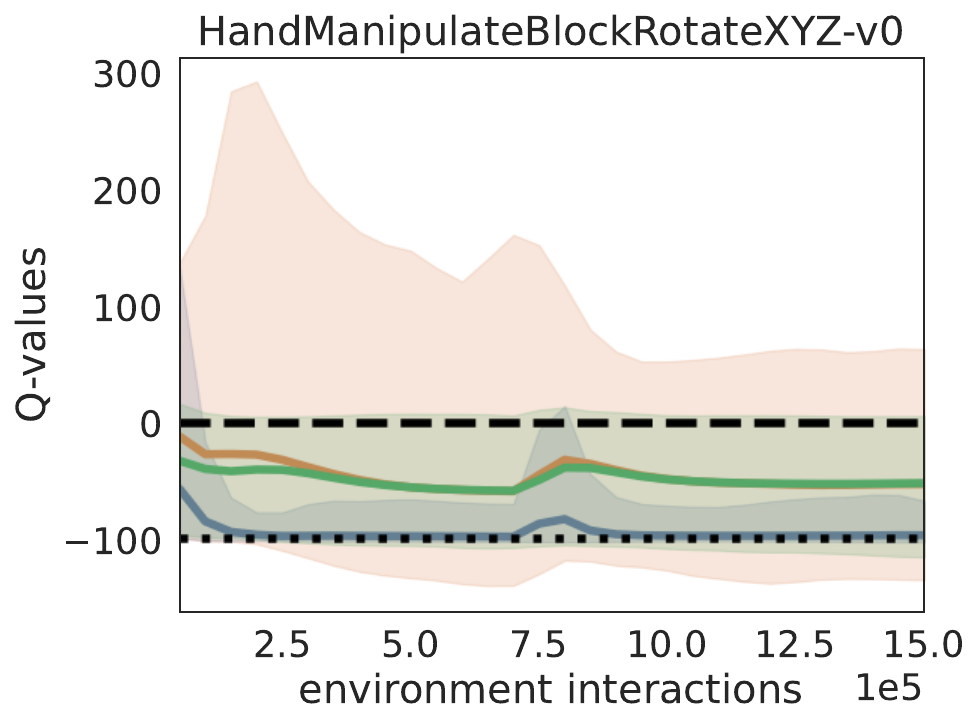}
\includegraphics[clip, width=0.24\hsize]{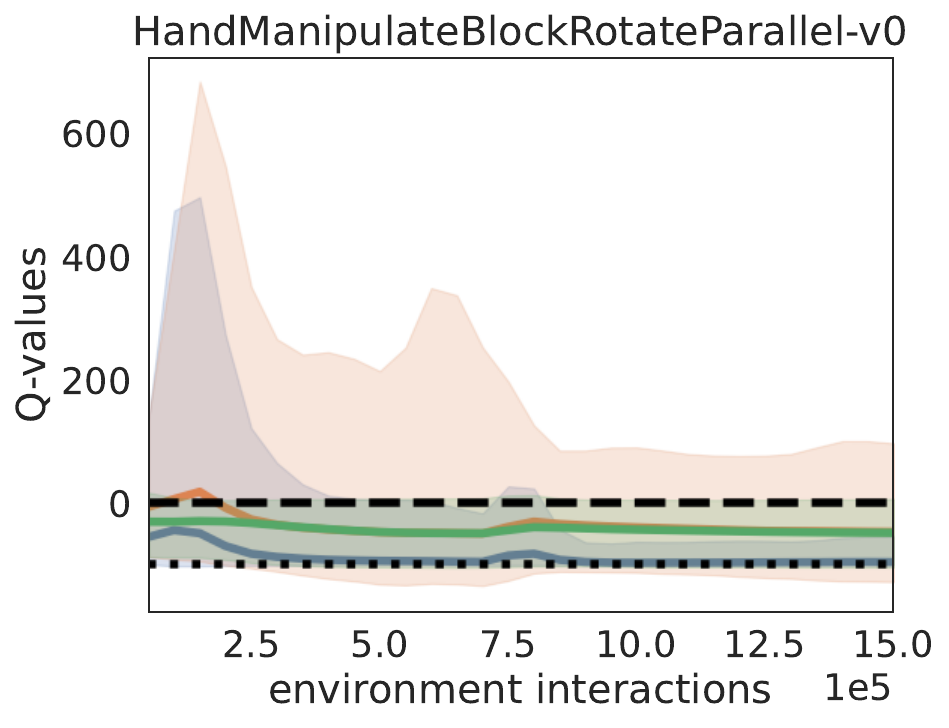}
\end{minipage}
\vspace{-0.7\baselineskip}
\caption{
The effect of replacing REDQ with Reset(1) on Q-value divergence. 
}
\label{fig:app-reset1-qvals}
\vspace{-0.5\baselineskip}
\end{figure*}

\begin{figure*}[h!]
\begin{minipage}{1.0\hsize}
\includegraphics[clip, width=0.24\hsize]{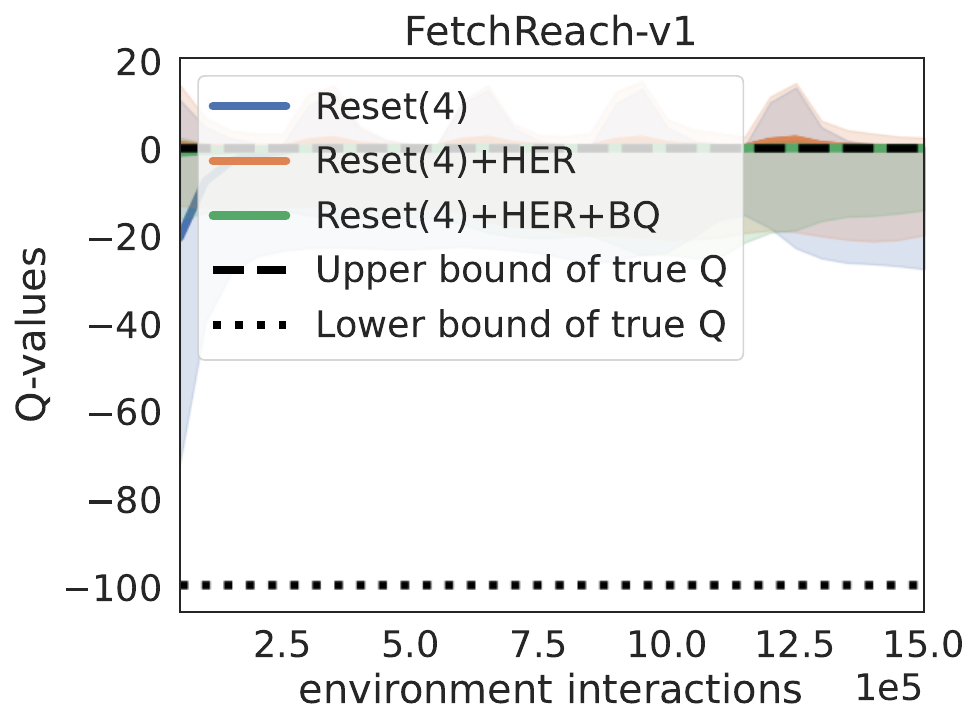}
\includegraphics[clip, width=0.24\hsize]{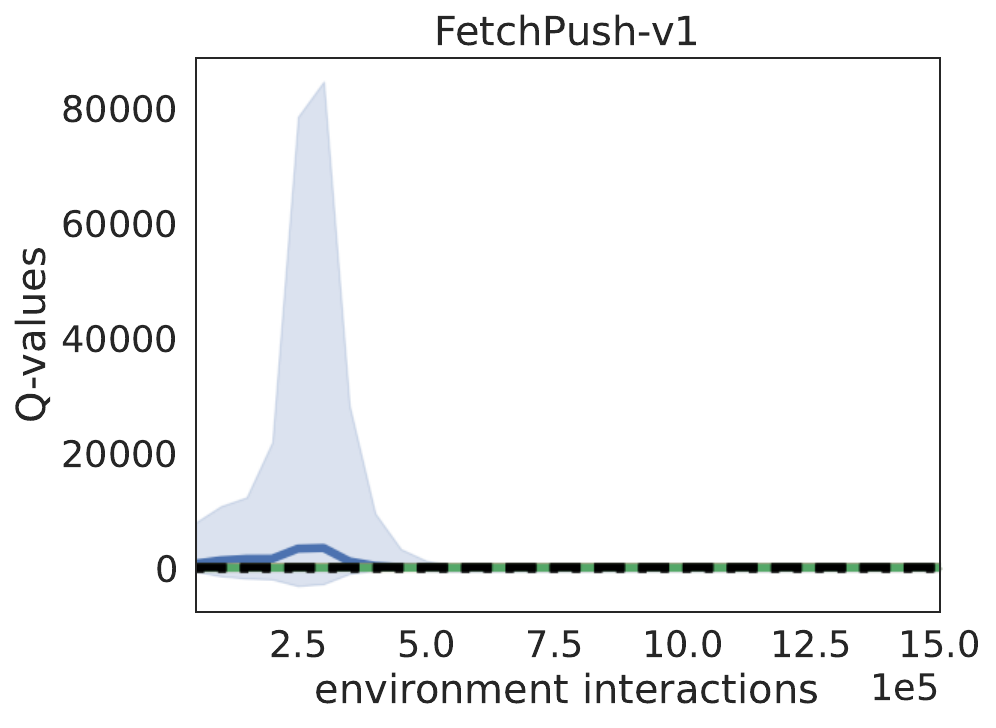}
\includegraphics[clip, width=0.24\hsize]{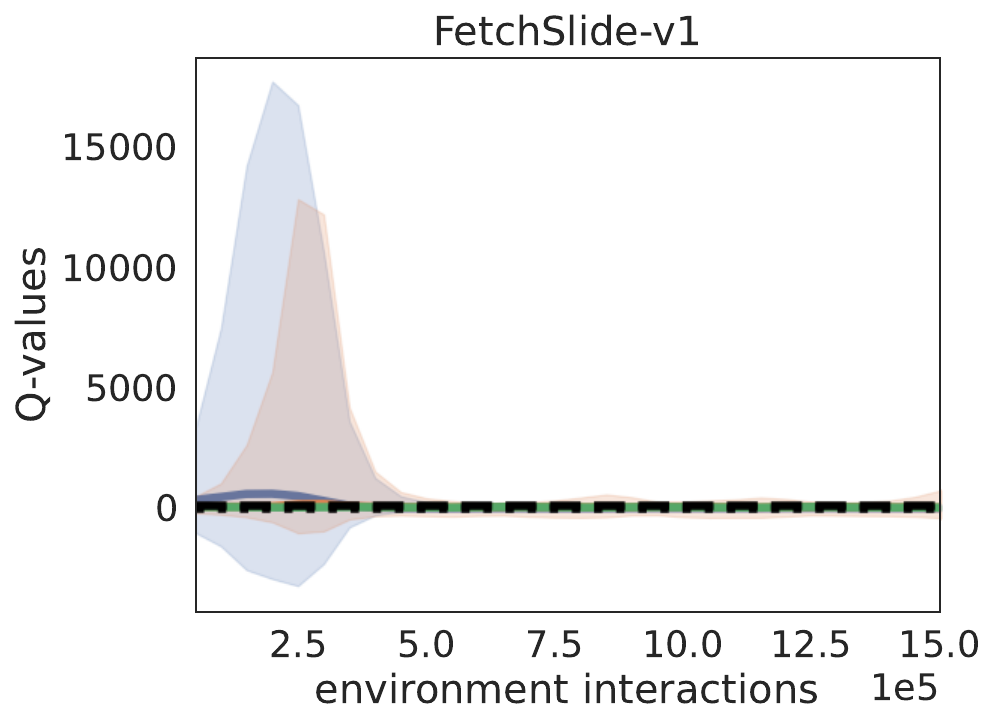}
\includegraphics[clip, width=0.24\hsize]{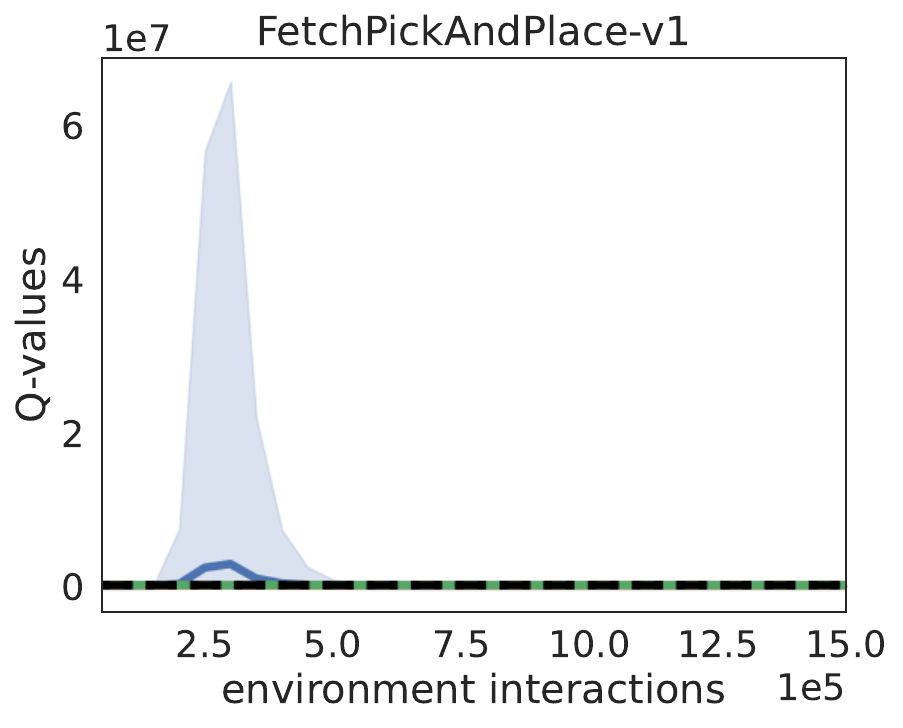}
\end{minipage}
\begin{minipage}{1.0\hsize}
\includegraphics[clip, width=0.24\hsize]{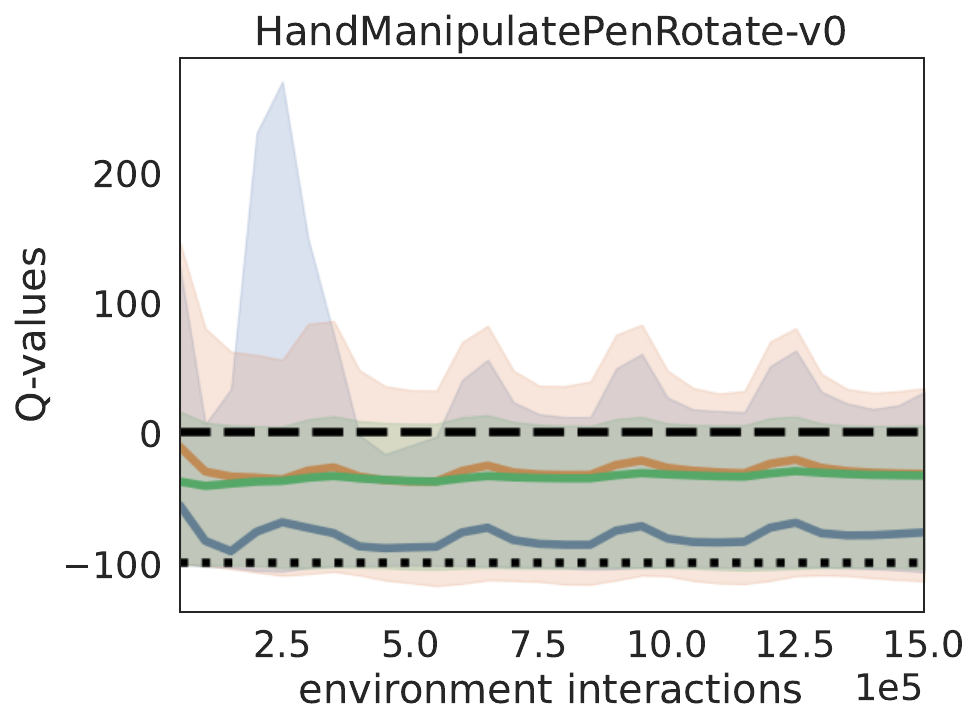}
\includegraphics[clip, width=0.24\hsize]{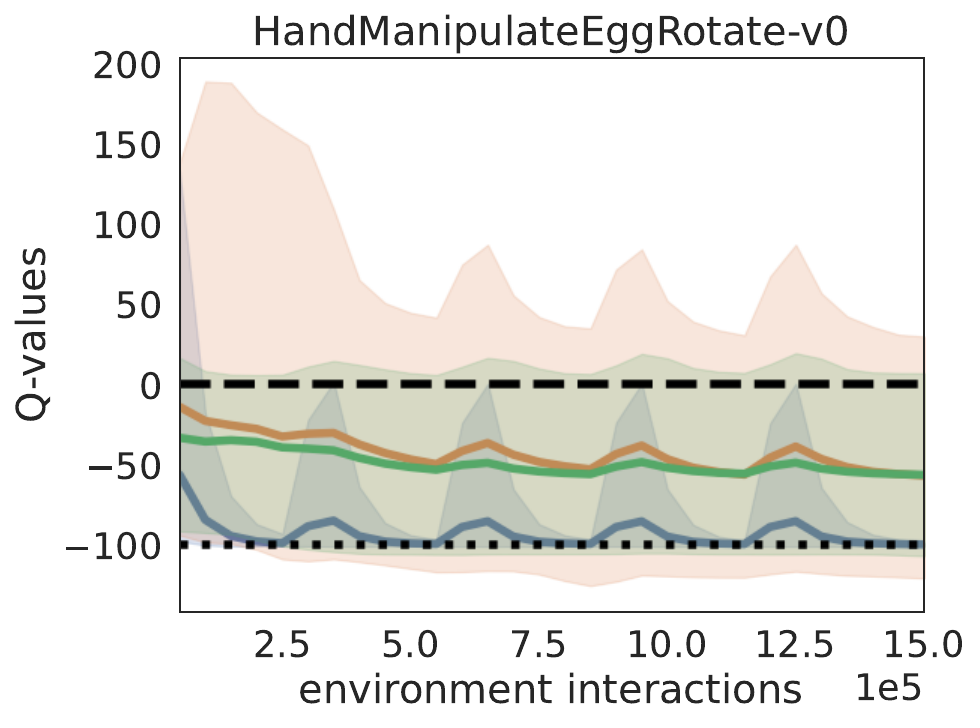}
\includegraphics[clip, width=0.24\hsize]{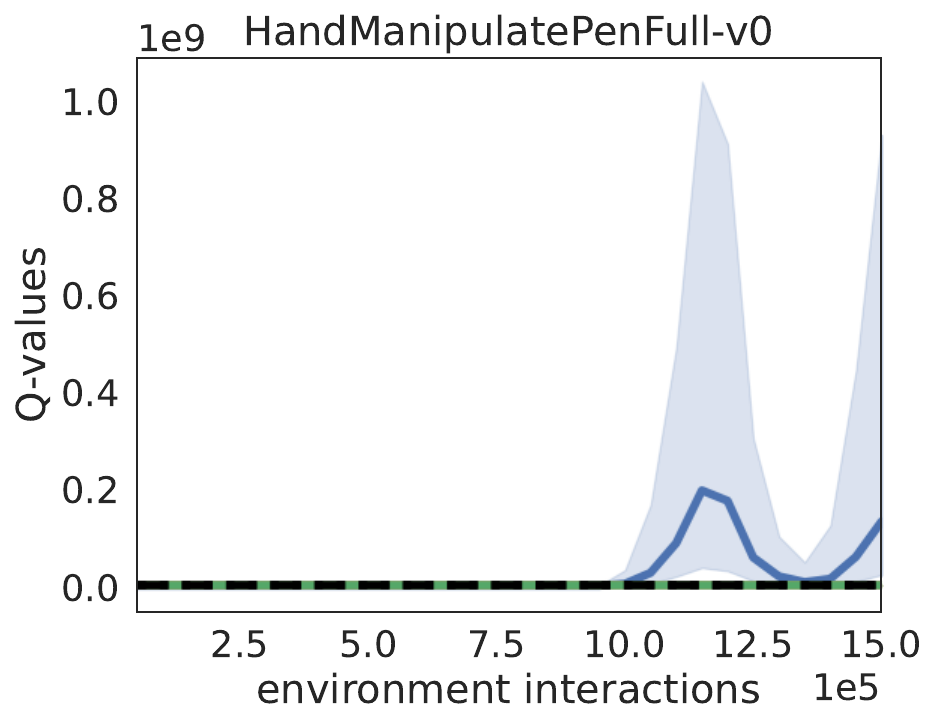}
\includegraphics[clip, width=0.24\hsize]{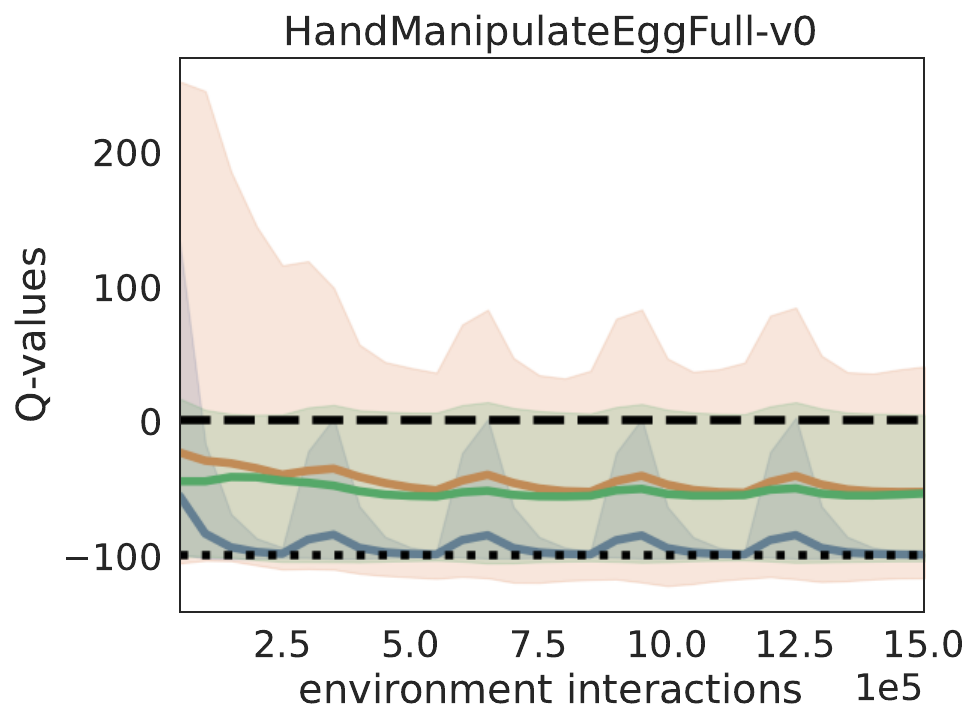}
\end{minipage}
\begin{minipage}{1.0\hsize}
\includegraphics[clip, width=0.24\hsize]{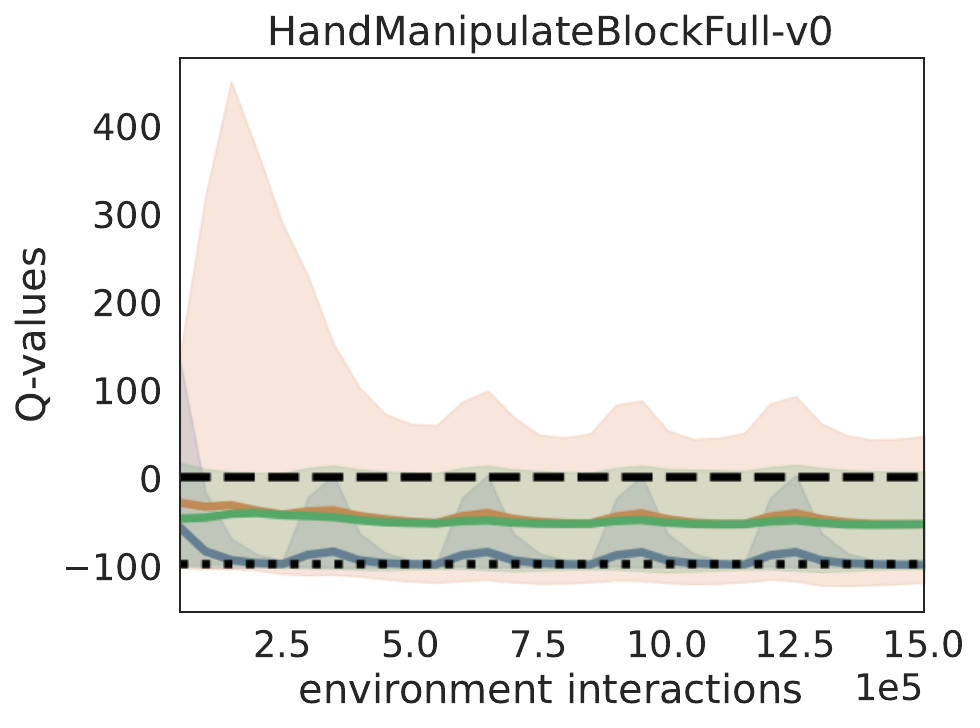}
\includegraphics[clip, width=0.24\hsize]{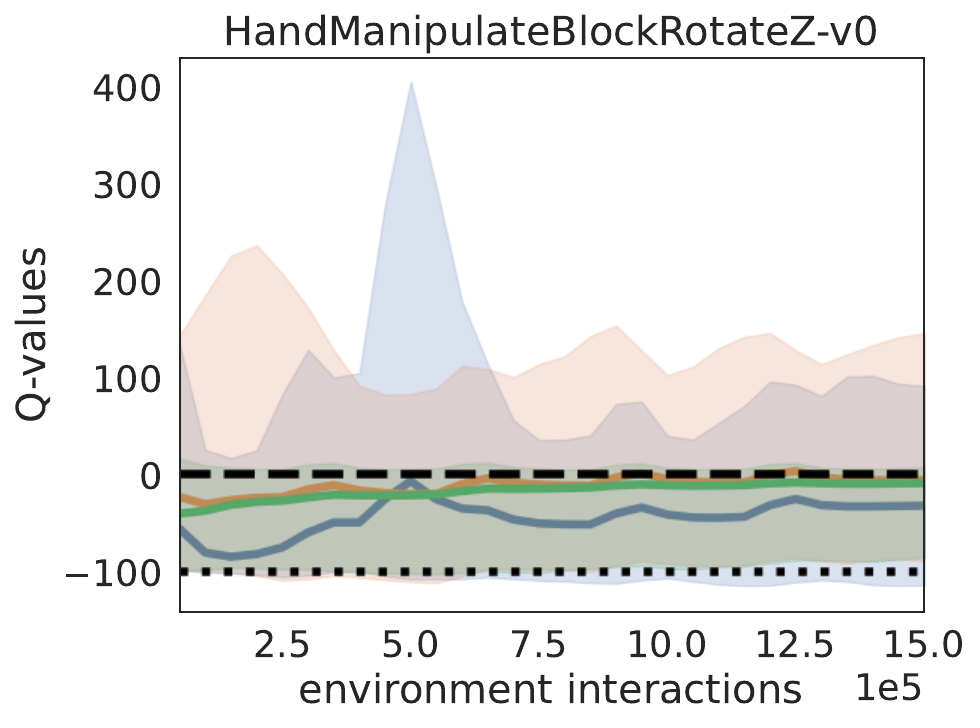}
\includegraphics[clip, width=0.24\hsize]{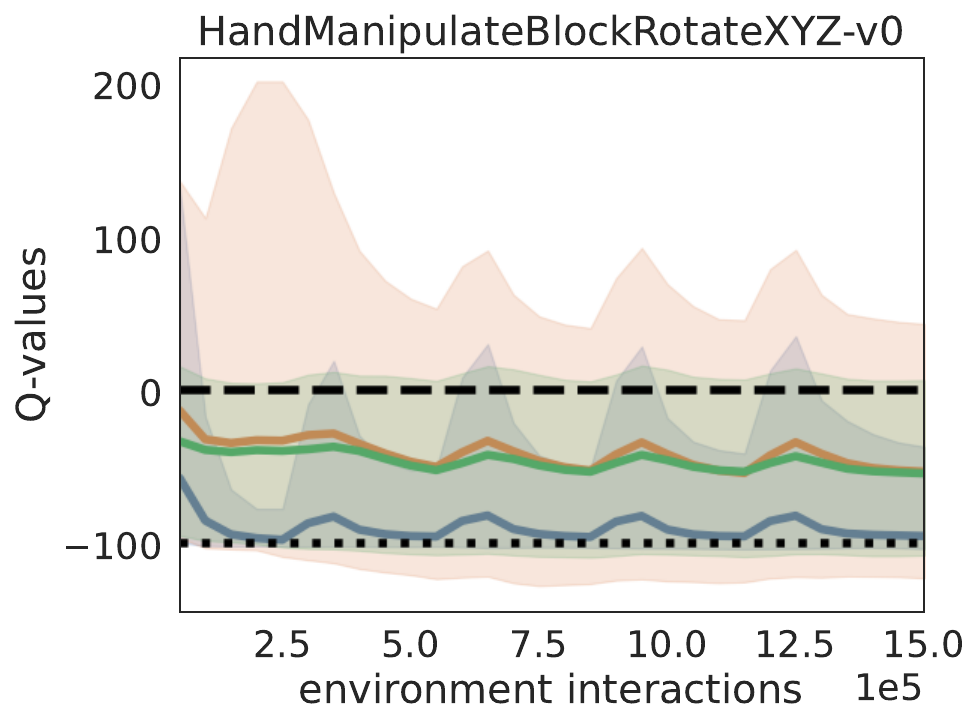}
\includegraphics[clip, width=0.24\hsize]{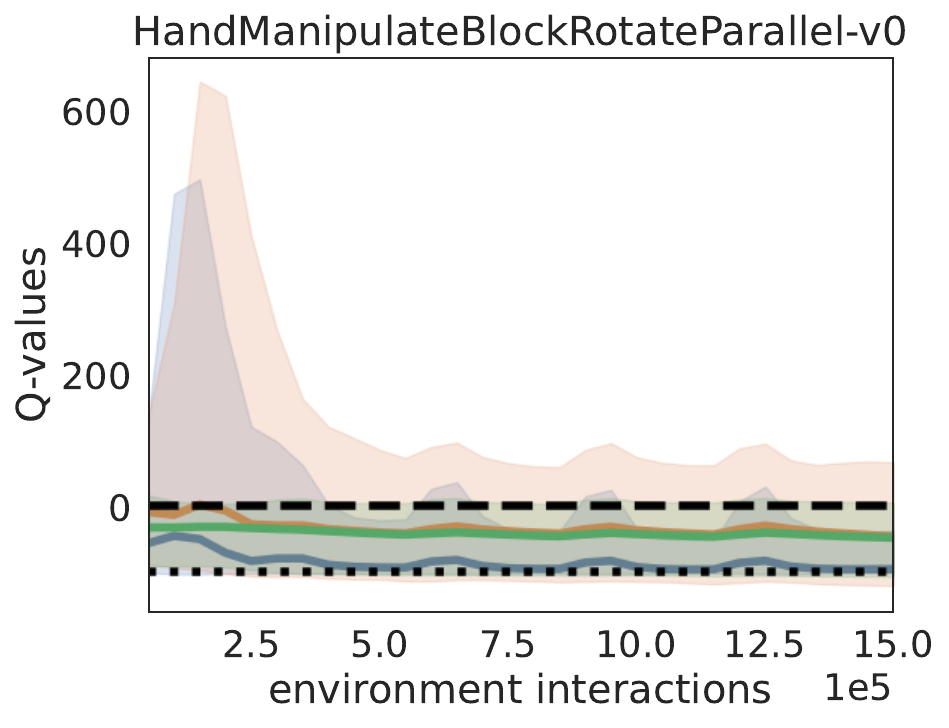}
\end{minipage}
\vspace{-0.7\baselineskip}
\caption{
The effect of replacing REDQ with Reset(4) on Q-value divergence. 
}
\label{fig:app-reset4-qvals}
\vspace{-0.5\baselineskip}
\end{figure*}

\begin{figure*}[h!]
\begin{minipage}{1.0\hsize}
\includegraphics[clip, width=0.24\hsize]{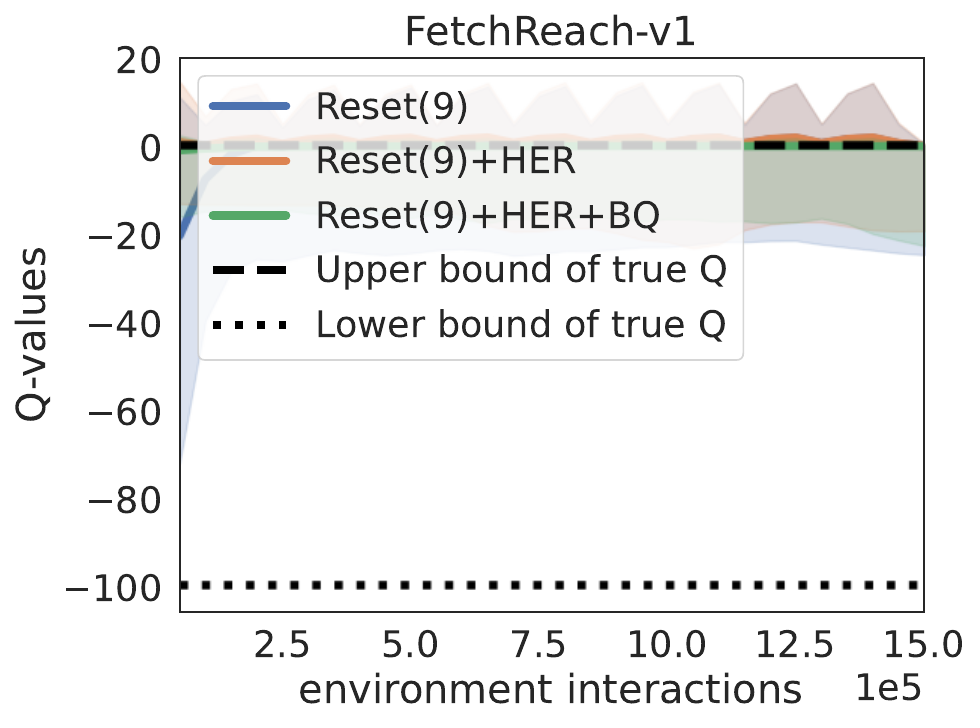}
\includegraphics[clip, width=0.24\hsize]{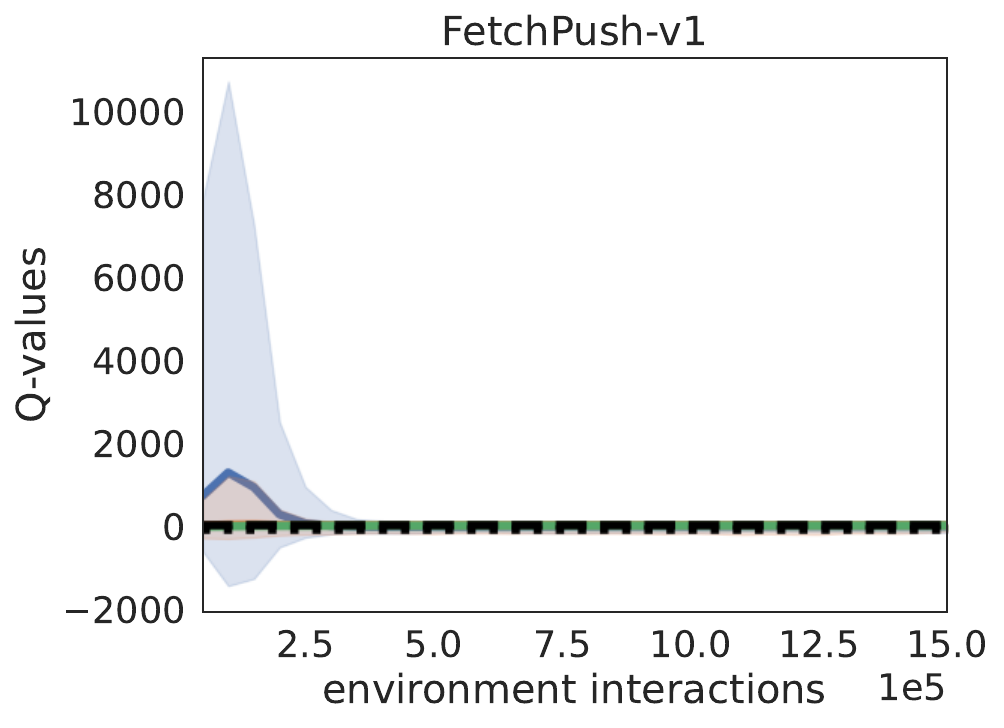}
\includegraphics[clip, width=0.24\hsize]{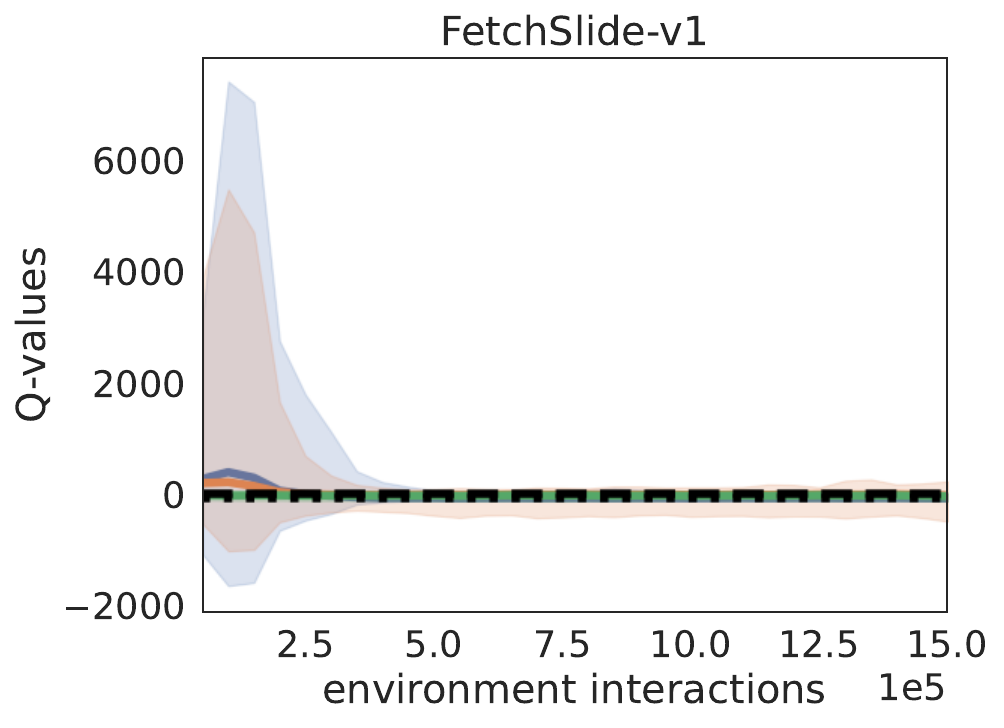}
\includegraphics[clip, width=0.24\hsize]{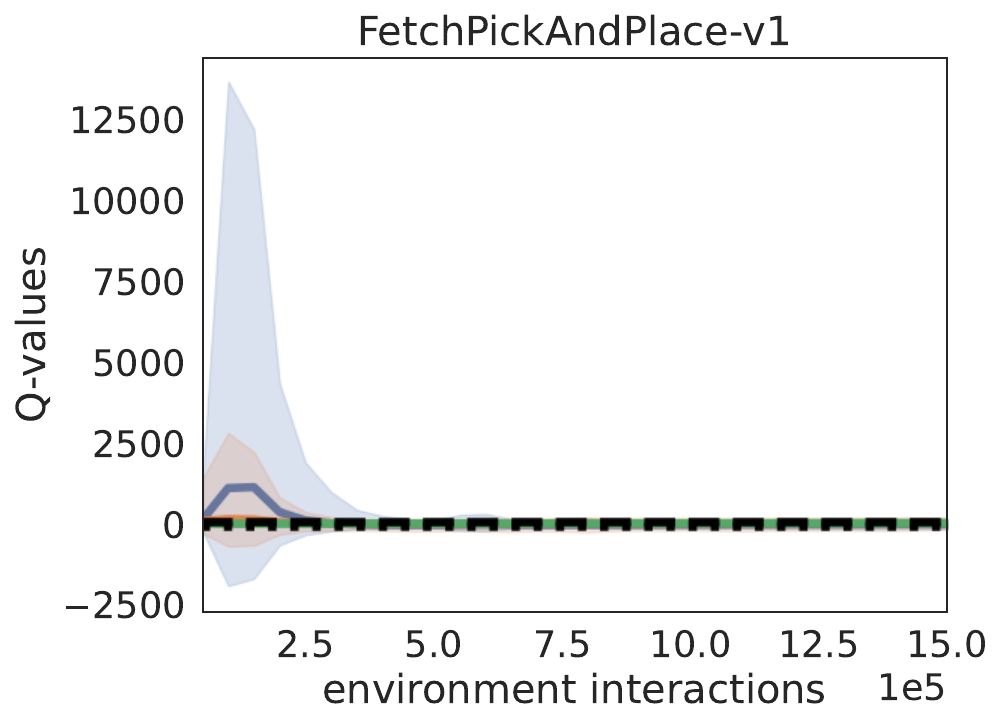}
\end{minipage}
\begin{minipage}{1.0\hsize}
\includegraphics[clip, width=0.24\hsize]{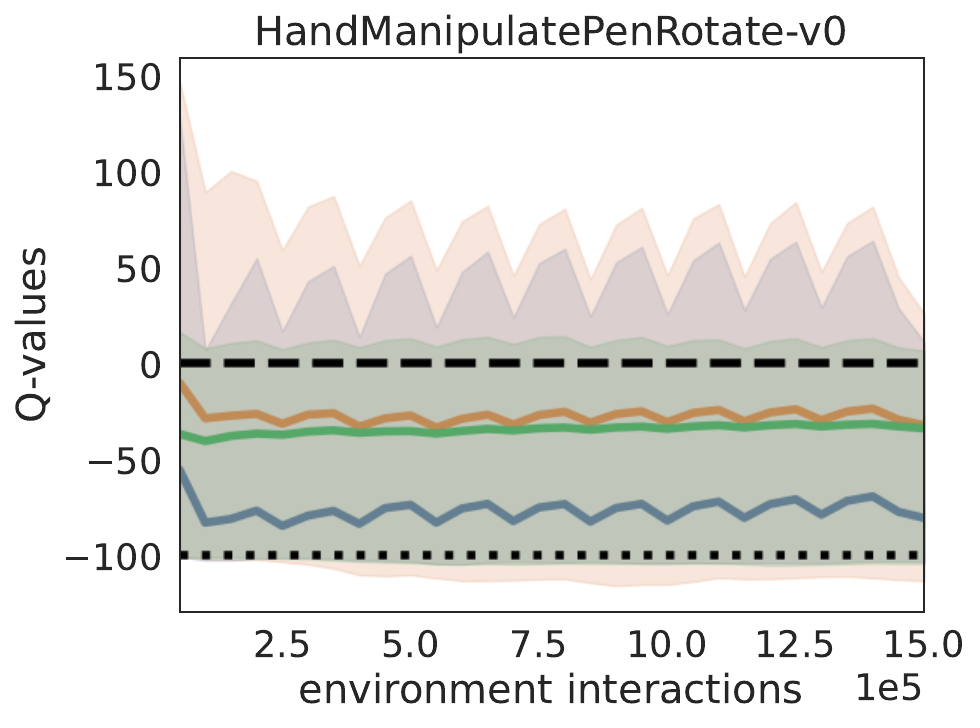}
\includegraphics[clip, width=0.24\hsize]{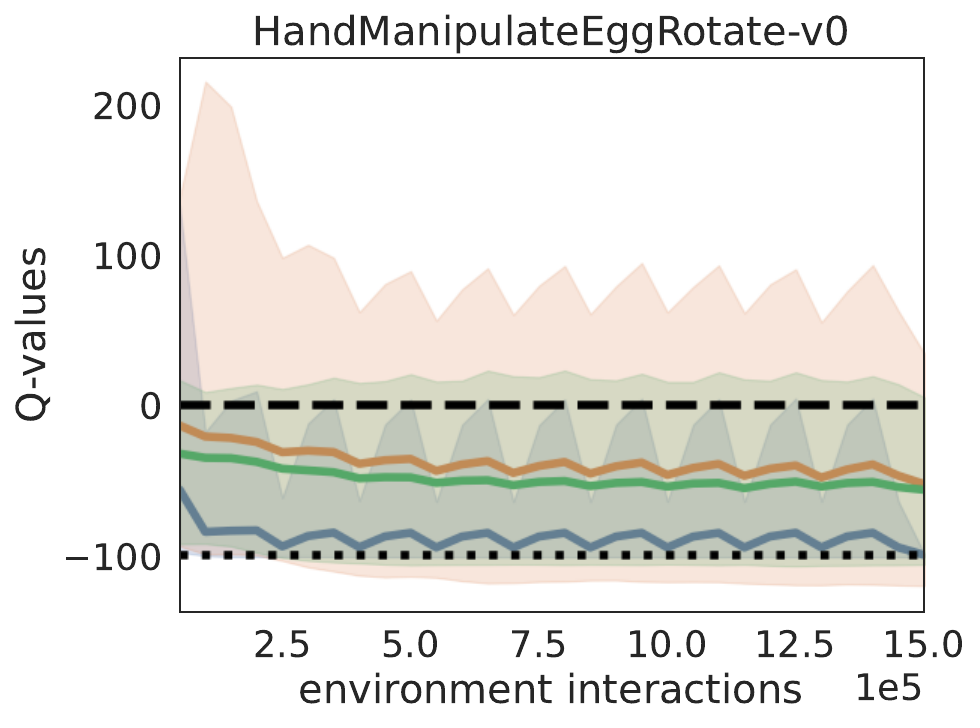}
\includegraphics[clip, width=0.24\hsize]{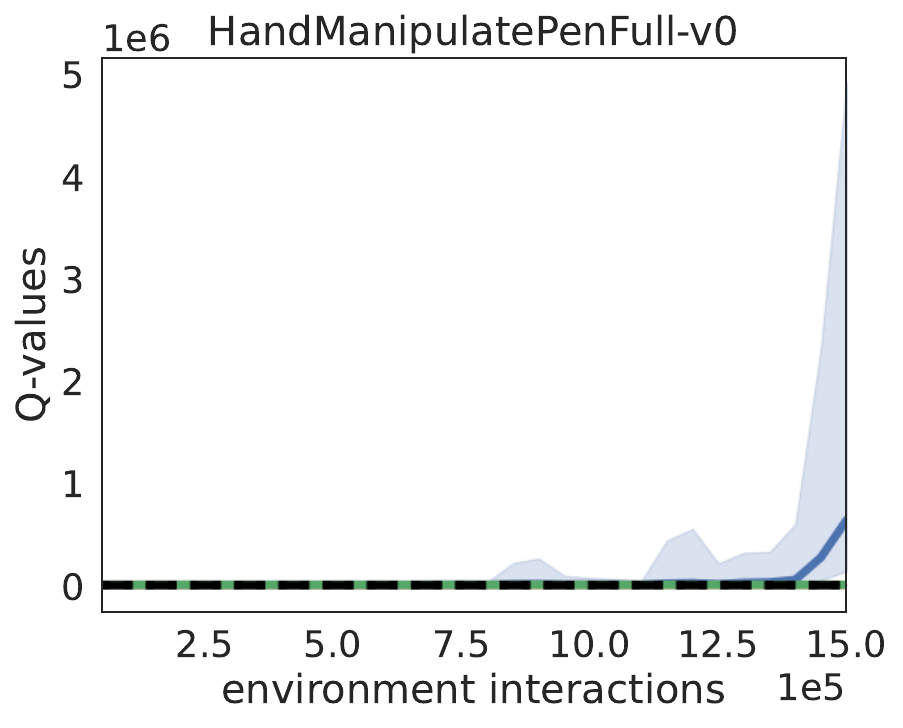}
\includegraphics[clip, width=0.24\hsize]{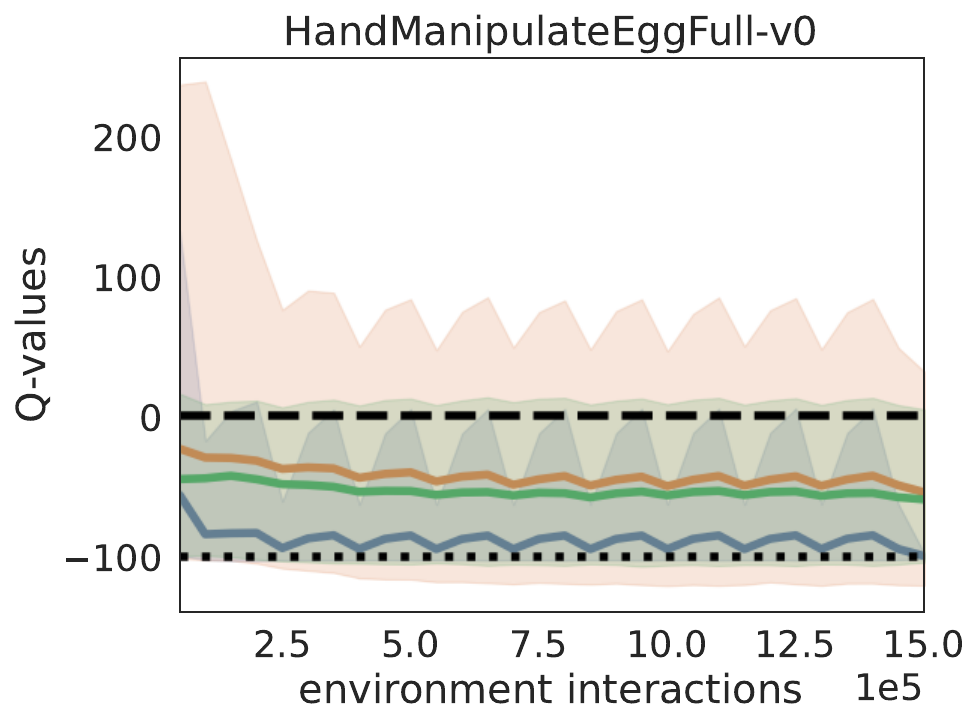}
\end{minipage}
\begin{minipage}{1.0\hsize}
\includegraphics[clip, width=0.24\hsize]{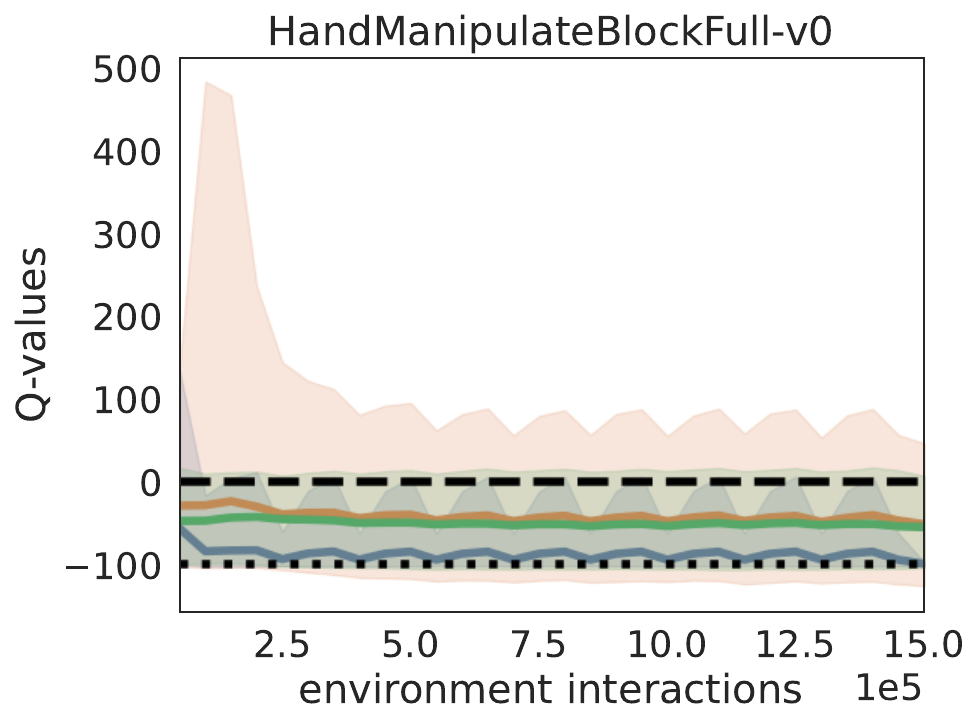}
\includegraphics[clip, width=0.24\hsize]{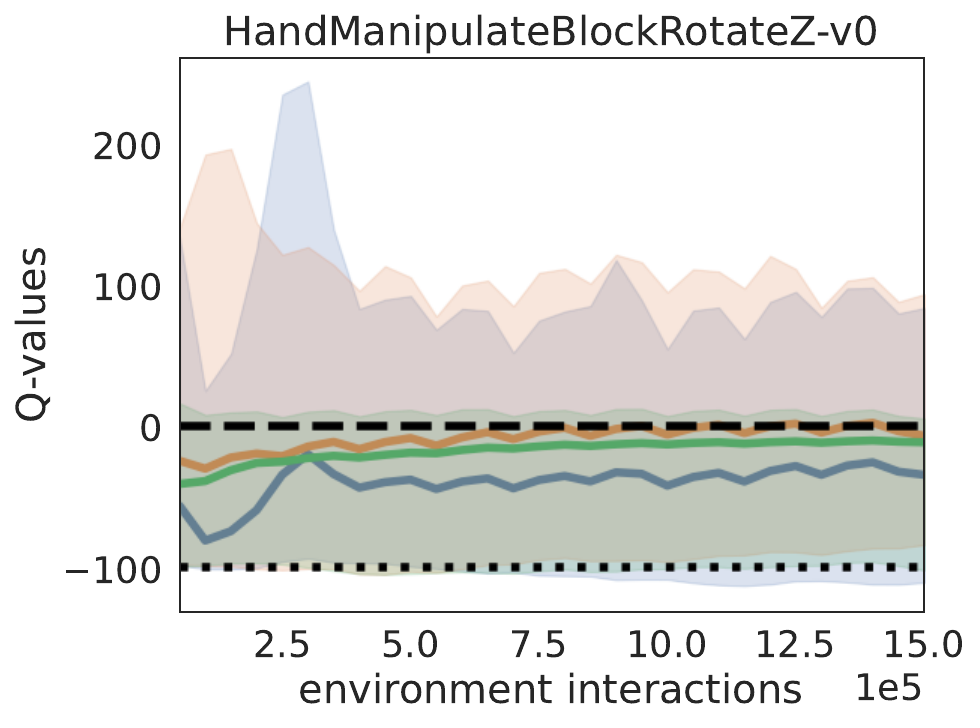}
\includegraphics[clip, width=0.24\hsize]{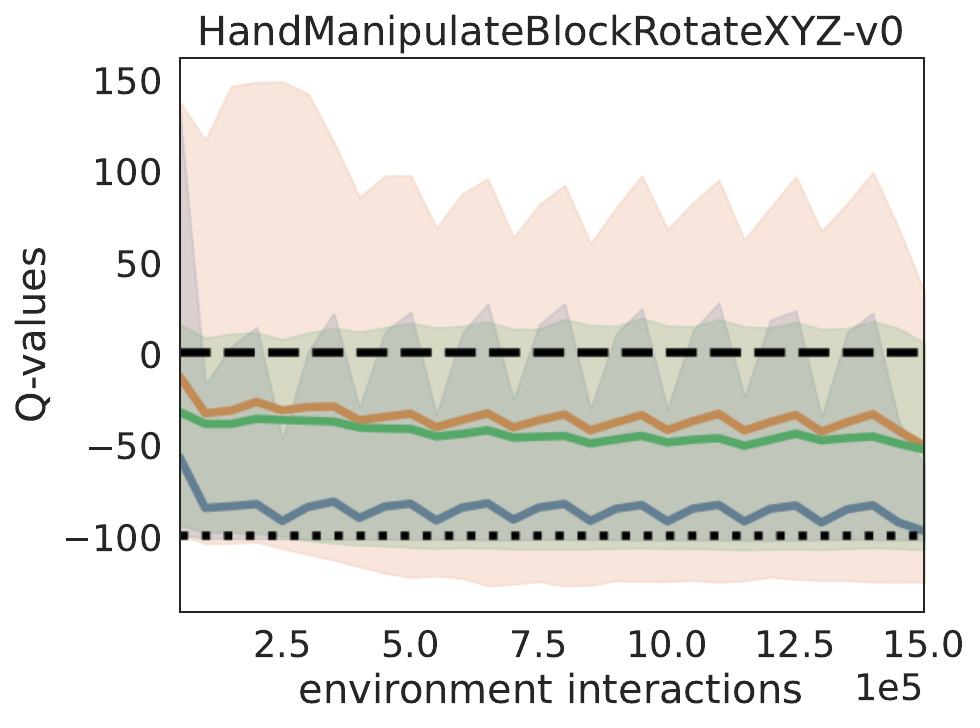}
\includegraphics[clip, width=0.24\hsize]{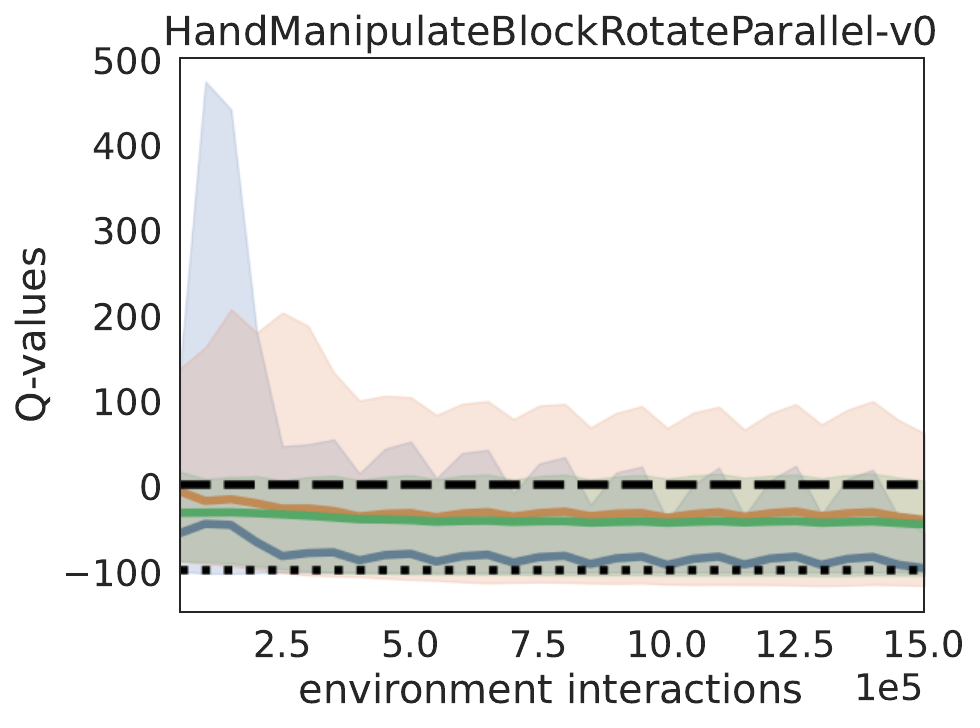}
\end{minipage}
\vspace{-0.7\baselineskip}
\caption{
The effect of replacing REDQ with Reset(9) on Q-value divergence. 
}
\label{fig:app-reset9-qvals}
\vspace{-0.5\baselineskip}
\end{figure*}


\clearpage
\section{Algorithmic Description of Reset~\citep{pmlr-v162-nikishin22a} with Our Modifications}\label{app:reset}
\begin{algorithm*}[h!]
\caption{Reset with our modifications (\textcolor{uscarlet}{HER} and \textcolor{uviolet}{BQ}) }
\label{alg1:ResetwithOurDesigneDecision}
\begin{algorithmic}[1]
\Statex Initialize policy parameters $\theta$, two Q-function parameters $\phi_i$, empty replay buffer $\mathcal{D}$, and episode length $T$. Set target parameters $\bar{\phi}_i \leftarrow \phi_i$, for $i = 1, 2$.
\State Sample goal $g \sim p_g(\cdot)$ and initial state $s_0 \sim p_{s_0}(\cdot)$
\For{$t=0, .., T$}
\State Take action $a_t \sim \pi_\theta(\cdot | s_t)$; Observe reward $r_t$ and next state $s_{t+1}$.
\If{$t = T$}
    \State $\mathcal{D} \leftarrow \mathcal{D} \bigcup \{(s_t, a_t, r_t, s_{t+1}, g)\}_{t=0}^{T}$; \textcolor{uscarlet}{Select new goal $g_t'$; Calculate new reward $r_t' \leftarrow \mathcal{R}(s_t, a_t, g_t')$; $\mathcal{D} \leftarrow \mathcal{D} \bigcup \{(s_t, a_t, r_t', s_{t+1}, g_t')\}_{t=0}^{T}$}
\EndIf
\For{$G$ updates}
    \State Sample a mini-batch $\mathcal{B} = \{ (s, a, r, s', g) \}$ from $\mathcal{D}$. 
    \State Compute the target Q-value $y$: 
    \begin{equation}
        y = r + \gamma \textcolor{uviolet}{\min \left( \max \left( \textcolor{black}{\min_{i \in \{1, 2\}} Q_{\bar{\phi}_i}(s', a', g) - \alpha \log \pi_\theta(a' | s', g)}, Q_\min \right), Q_\max \right)}, ~~ a' \sim \pi_\theta(\cdot | s', g) \nonumber
    \end{equation}
    \For{$i=1, 2$}
        \State Update $\phi_i$ with gradient descent using
        \begin{equation}
            \nabla_\phi \frac{1}{|B|} \sum_{(s, a, r, s', g) \in \mathcal{B}} \left( Q_{\phi_i}(s, a, g) - y \right)^2 \nonumber
        \end{equation}
    \State Update target networks with $\bar{\phi}_i \leftarrow \rho \bar{\phi}_i + (1-\rho) \phi_i $.
    \EndFor
    \State Update $\theta$ with gradient ascent using 
    \begin{equation}
           \nabla_\theta \frac{1}{|B|} \sum_{s \in \mathcal{B}} \left( \frac{1}{2} \sum_{i=1}^{2} Q_{\phi_i}(s, a, g) - \alpha \log \pi_\theta(a | s, g) \right), ~~~ a \sim \pi_\theta(\cdot | s, g) \nonumber
    \end{equation}
    \If{the number of environment interactions reaches a reset period}
         \State Reinitialize $\theta$ and $\phi_1, \phi_2$. 
    \EndIf
\EndFor
\EndFor
\end{algorithmic}
\end{algorithm*}

\clearpage
\section{Hyperparameter Settings}\label{app:hypara}

\begin{table}[h!]
\caption{Hyperparameter settings}
\label{tab:hyerparametsers}
\begin{center}
\scalebox{0.95}{
\begin{tabular}{l||l|l}\hline
Method                    & Parameter                               & Value          \\\hline\hline
REDQ                      & optimizer                               & Adam~\citep{kingma2014adam}           \\
Reset                     & learning rate                           & $3 \cdot 10^{-4}$ \\
                          & discount rate $\gamma$                  & 0.99           \\
                          & target-smoothing coefficient            & 0.005          \\
                          & replay buffer size                      & $10^6$         \\
                          & number of hidden layers for all networks & 2              \\
                          & number of hidden units per layer        & 256            \\
                          & mini-batch size                         & 256            \\
                          & random starting data                    & 10000 for HER-based methods and 5000 for the others           \\
                          & replay ratio $G$                           & 20             \\
                          & in-target minimization parameter $M$    & 2              \\
                          & ensemble size $N$                       & 5 for REDQ and 2 for Reset.           \\\hline
                          
HER                       & number of additional goals              & 1             \\\hline

BQ                    & upper bound of Q-value $Q_\max$              & 0 (i.e., $\sum_t^{\infty}\gamma^t \cdot 0$)             \\
                          & lower bound of Q-value $Q_\min$              & -100 (i.e., $\sum_t^{\infty}\gamma^t \cdot -1 = \frac{-1}{1 - \gamma}$ with $\gamma=0.99$)             \\\hline

\end{tabular}
}
\end{center}
\end{table}

\section{Our Source Code}
Our source code is available at \url{https://github.com/TakuyaHiraoka/Efficient-SRGC-RL-with-a-High-RR-and-Regularization}


\end{document}